\documentclass[twoside]{article}

\usepackage[accepted]{aistats2022}

\usepackage[round]{natbib}

\usepackage{amssymb}
\usepackage{graphicx}
\usepackage{subfigure}
\usepackage{algorithm}
\usepackage{algorithmic}
\usepackage{booktabs}
\usepackage{url}
\usepackage{enumerate}

\newcommand{\secref}[1]{Section~\ref{#1}}
\newcommand{\figref}[1]{Figure~\ref{#1}}
\newcommand{\tabref}[1]{Table~\ref{#1}}

\newcommand{\algref}[1]{Algorithm~\ref{#1}}

\newcommand{\argmax}{\operatornamewithlimits{\arg \max}}

\newcommand{\bstau}{\boldsymbol \tau}
\newcommand{\bseta}{\boldsymbol \eta}
\newcommand{\bsdelta}{\boldsymbol \delta}
\newcommand{\bstheta}{\boldsymbol \theta}

\newcommand{\bx}{\mathbf{x}}
\newcommand{\by}{\mathbf{y}}

\newcommand{\bB}{\mathbf{B}}

\newcommand{\bX}{\mathbf{X}}

\newcommand{\calL}{\mathcal{L}}
\newcommand{\calN}{\mathcal{N}}

\newcommand{\calT}{\mathcal{T}}

\newcommand{\bbE}{\mathbb{E}}

\newcommand{\bbR}{\mathbb{R}}

\usepackage{xr}
\makeatletter
\newcommand*{\addFileDependency}[1]{
  \typeout{(#1)}
  \@addtofilelist{#1}
  \IfFileExists{#1}{}{\typeout{No file #1.}}
}
\makeatother

\graphicspath{{./figures/}}

\begin{document}

\runningtitle{On Uncertainty Estimation by Tree-based Surrogate Models in SMO}

\twocolumn[

\aistatstitle{On Uncertainty Estimation by Tree-based Surrogate Models 
\\in Sequential Model-based Optimization}

\aistatsauthor{Jungtaek Kim \And Seungjin Choi}

\aistatsaddress{POSTECH \And BARO AI} ]

\begin{abstract}
Sequential model-based optimization sequentially selects a candidate point
by constructing a surrogate model with the history of evaluations,
to solve a black-box optimization problem.
Gaussian process (GP) regression is a popular choice as a surrogate model,
because of its capability of calculating prediction uncertainty analytically.
On the other hand, an ensemble of randomized trees is another option
and has practical merits over GPs due to its scalability and easiness of handling
continuous/discrete mixed variables.
In this paper we revisit various ensembles of randomized trees to investigate their
behavior in the perspective of prediction uncertainty estimation.
Then, we propose a new way of constructing an ensemble of randomized trees,
referred to as \emph{BwO forest}, where bagging with oversampling is employed to construct
bootstrapped samples that are used to build randomized trees with random splitting.
Experimental results demonstrate the validity and good performance of BwO forest over
existing tree-based models in various circumstances.
\end{abstract}

\section{INTRODUCTION\label{sec:introduction}}

Sequential model-based optimization (SMO)~\citep{BrochuE2010arxiv,HutterF2011lion}
constructs a statistical surrogate model in order to estimate
a function value and its uncertainty -- both estimates are employed
to balance a trade-off between exploitation and exploration.
To determine where next to evaluate carefully,
a surrogate model is one of key components in SMO~\citep{BodinE2020icml}.
In a common setting of SMO including the formulation of Bayesian optimization~\citep{SrinivasN2010icml,AzimiJ2010neurips,SnoekJ2012neurips},
Gaussian process (GP) regression~\citep{OHaganA1978jrssb,WilliamsCKI1995neurips}
is a popular choice as a surrogate model due to its flexibility and expressibility~\citep{RasmussenCE2006book}.
However, it requires the assumption on smoothness,
which can induce \emph{a mismatch on the smoothness degree} for an objective of interest~\citep{SchulzE2016neuripsw},
and moreover its exact complexity over
the number of the query points already evaluated scales \emph{cubically}~\citep{RasmussenCE2006book}.

Instead of GP regression, random forest regression~\citep{BreimanL2001ml}
is another option and has practical merits as a surrogate model;
sequential model-based algorithm configuration~\citep{HutterF2011lion} shows its strength
in various real-world applications such as
automated machine learning~\citep{FeurerM2015neurips},
neural architecture search~\citep{YingC2019icml}, and
water distribution system~\citep{CandelieriA2018jgo}.
In particular, it \emph{inherently deals with a categorical variable},
because a randomized tree -- a base estimator of random forest --
is capable of defining a split criterion for categorical variables
without any complex techniques.
Compared to GPs for categorical variables
(e.g., using the Aitchison and Aitken kernel~\citep{AitchisonJ1976biometrika}
or using a random embedding to lower-dimensional space~\citep{WangZ2016jair}),\footnote{Many GP-based approaches to solving this topic have been studied, but they are not the scope of this work.}
it provides an easy-to-use implementation as well as reliable performance.
Furthermore, a tree-based surrogate model tends to be robust in \emph{a high-dimensional search space},
in comparison with GP regression. These strengths mentioned above
let us pursue in-depth and thorough studies on randomized tree-based models.

\begin{figure*}[t]
	\centering
	\subfigure[GP]{
		\includegraphics[width=0.12\textwidth]{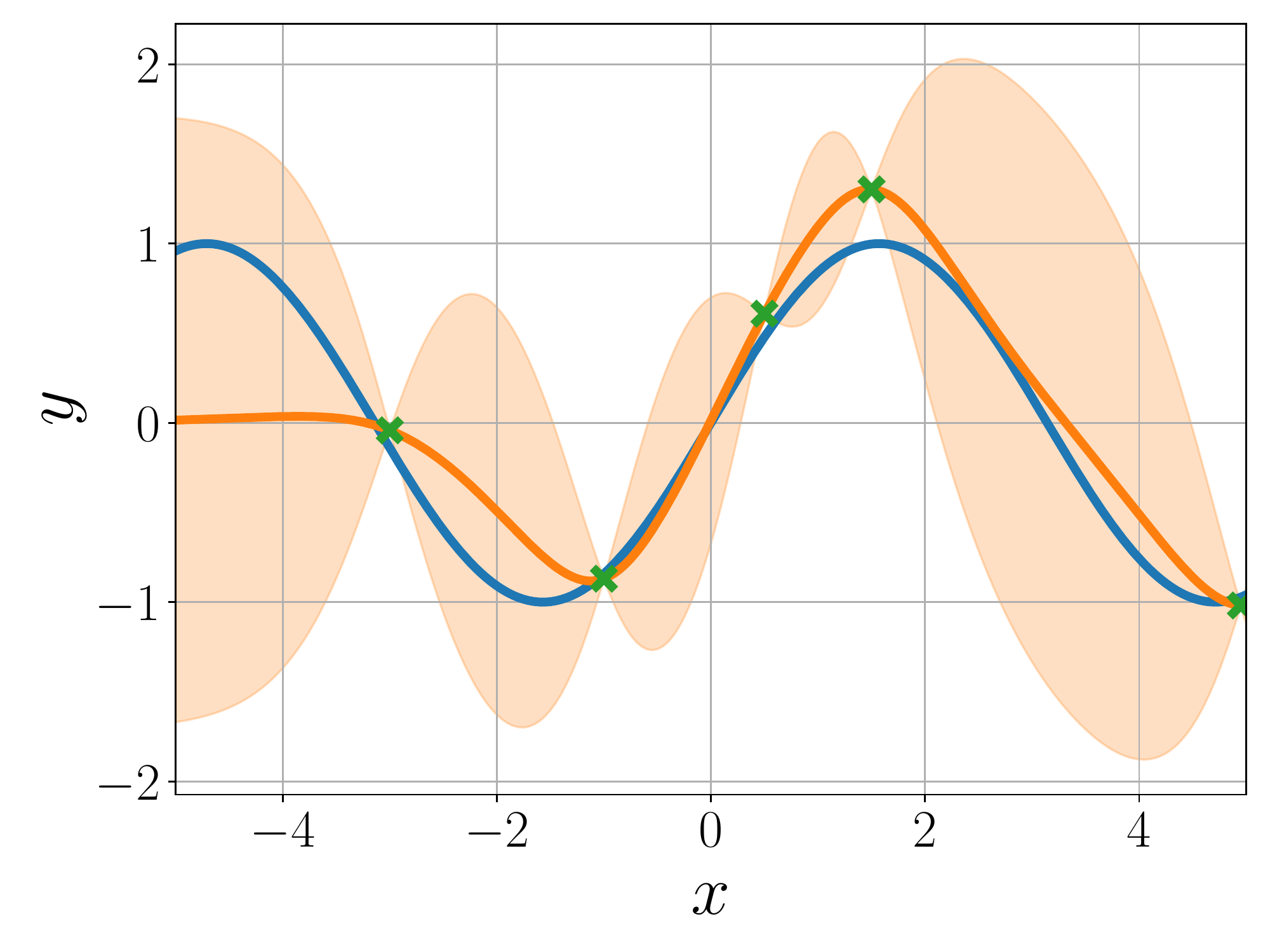}
		\label{fig:unc_1d_few_gp}
	}
	\subfigure[RF]{
		\includegraphics[width=0.12\textwidth]{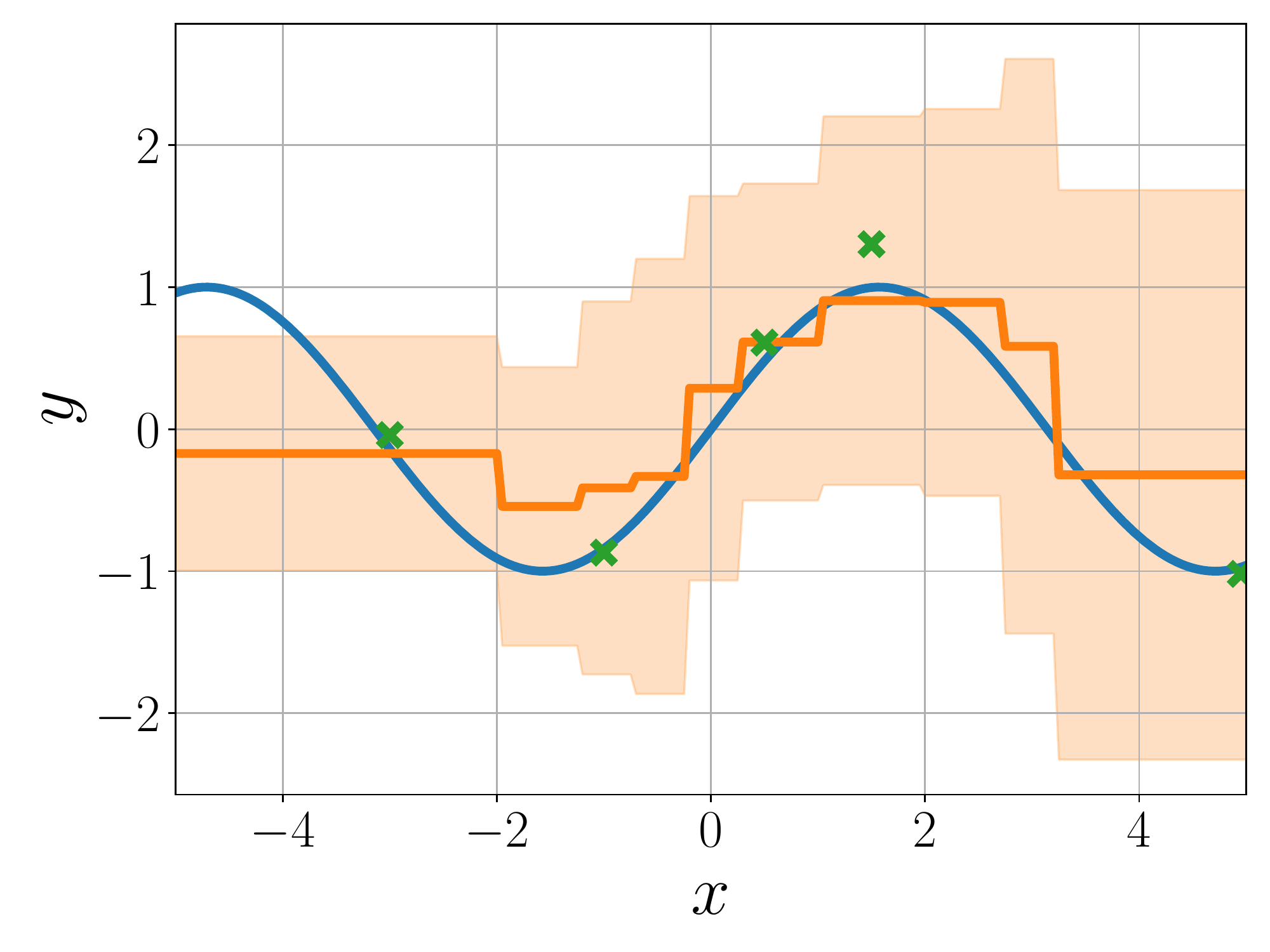}
		\label{fig:unc_1d_few_rf}
	}
	\subfigure[ERTs]{
		\includegraphics[width=0.12\textwidth]{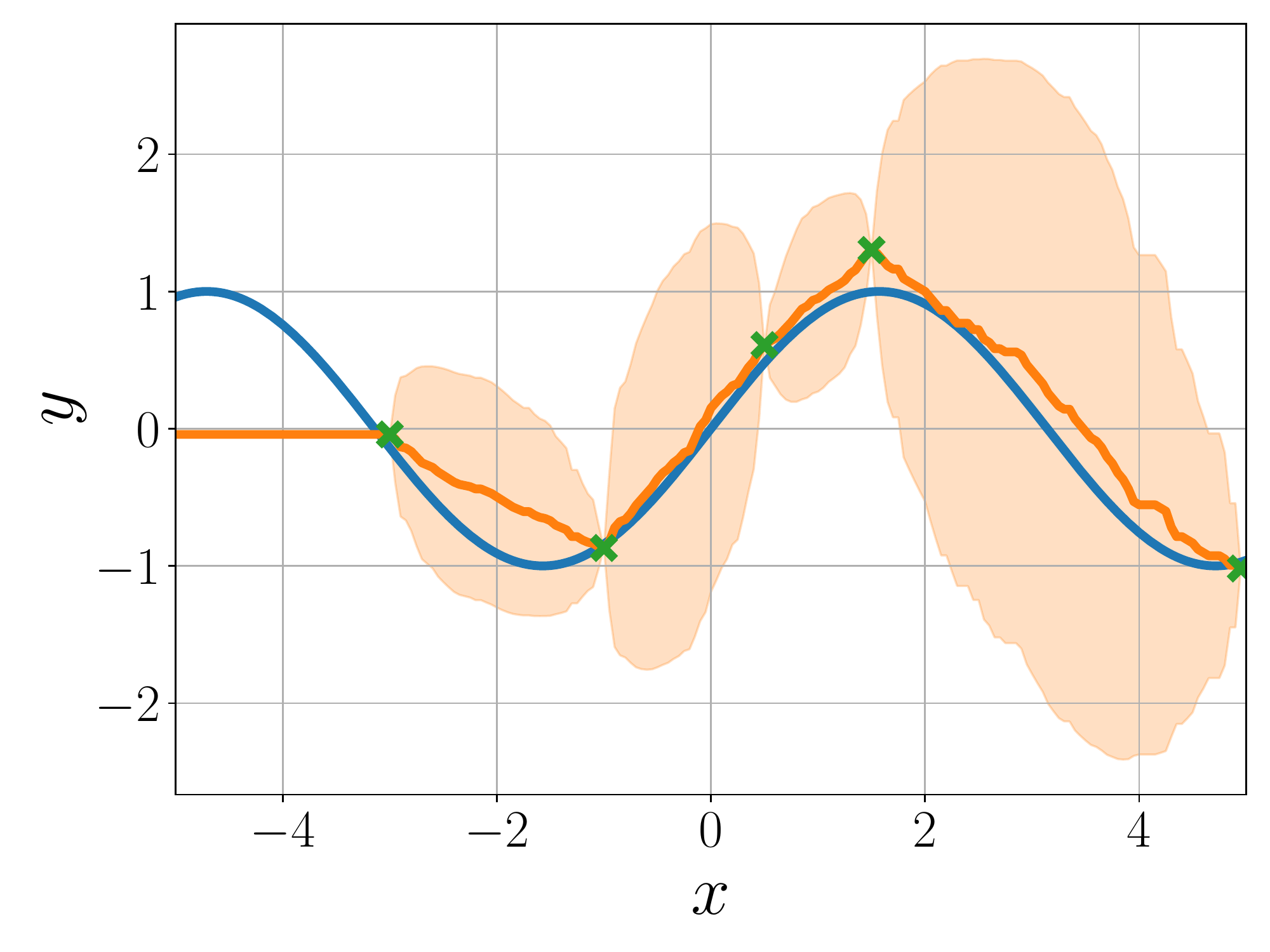}
		\label{fig:unc_1d_few_ext}
	}
	\subfigure[BART]{
		\includegraphics[width=0.12\textwidth]{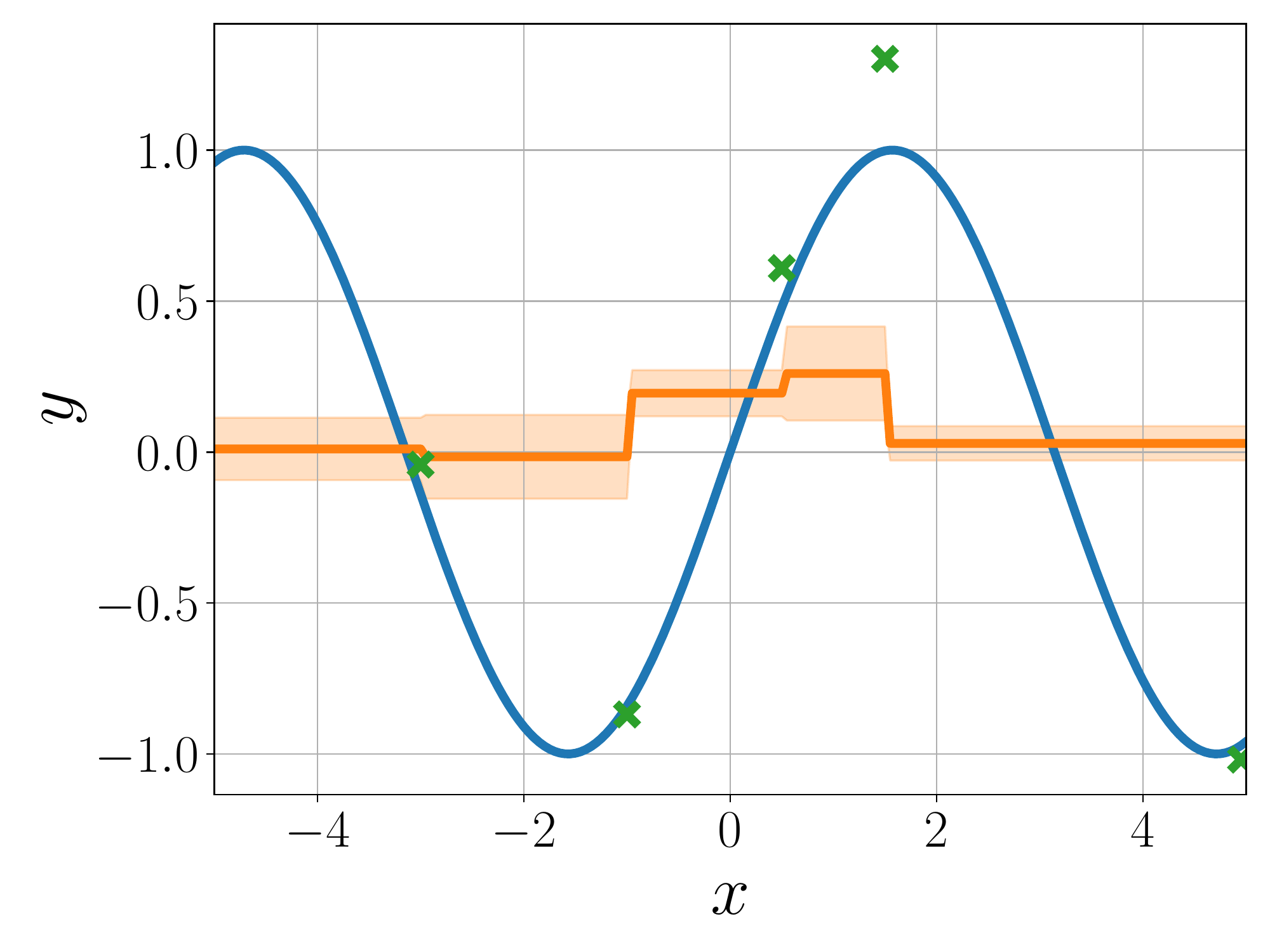}
		\label{fig:unc_1d_few_bart}
	}
	\subfigure[MF]{
		\includegraphics[width=0.12\textwidth]{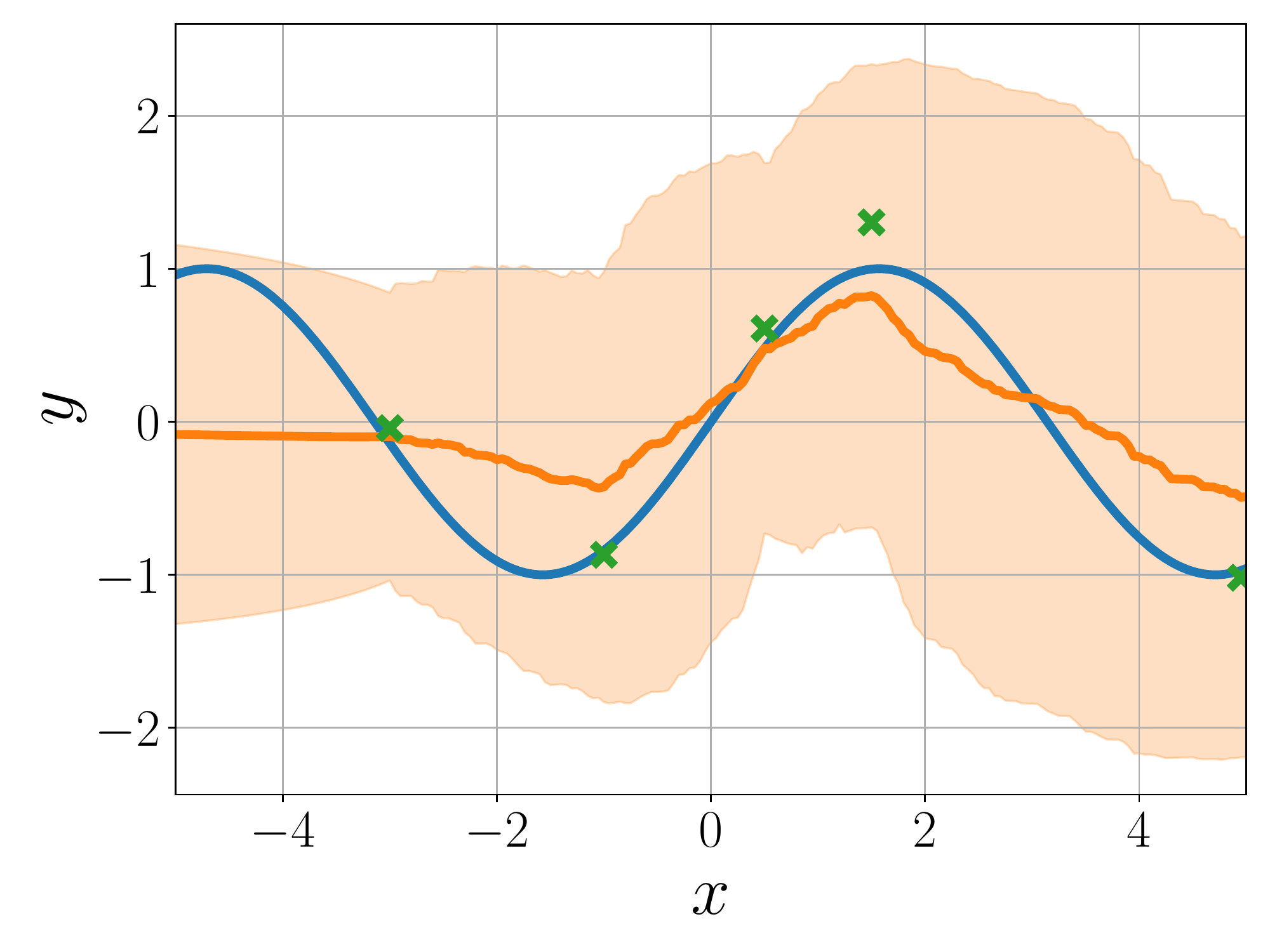}
		\label{fig:unc_1d_few_mf}
	}
	\subfigure[NGBoost]{
		\includegraphics[width=0.12\textwidth]{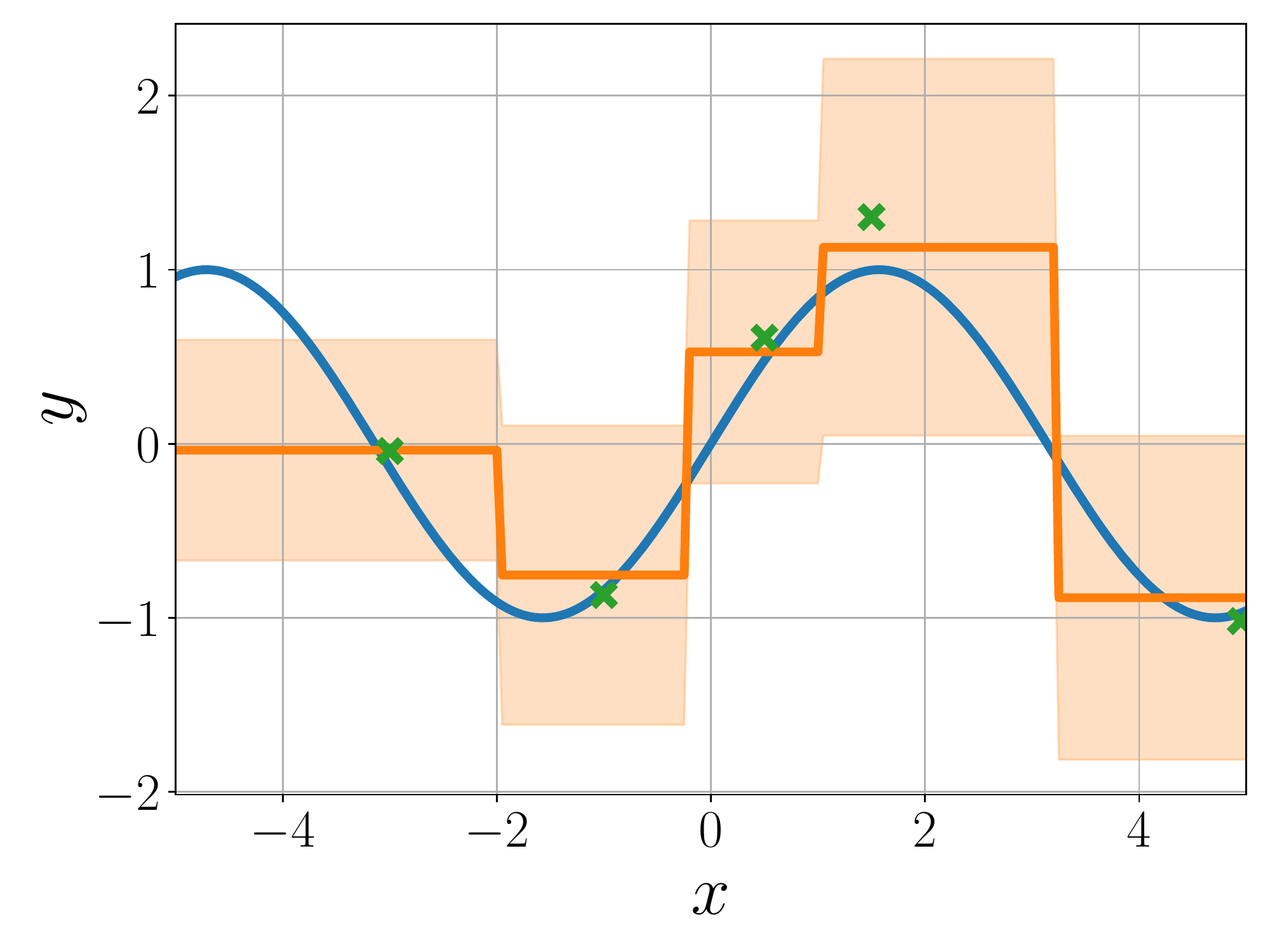}
		\label{fig:unc_1d_few_ngboost}
	}
	\subfigure[BwO forest]{
		\includegraphics[width=0.12\textwidth]{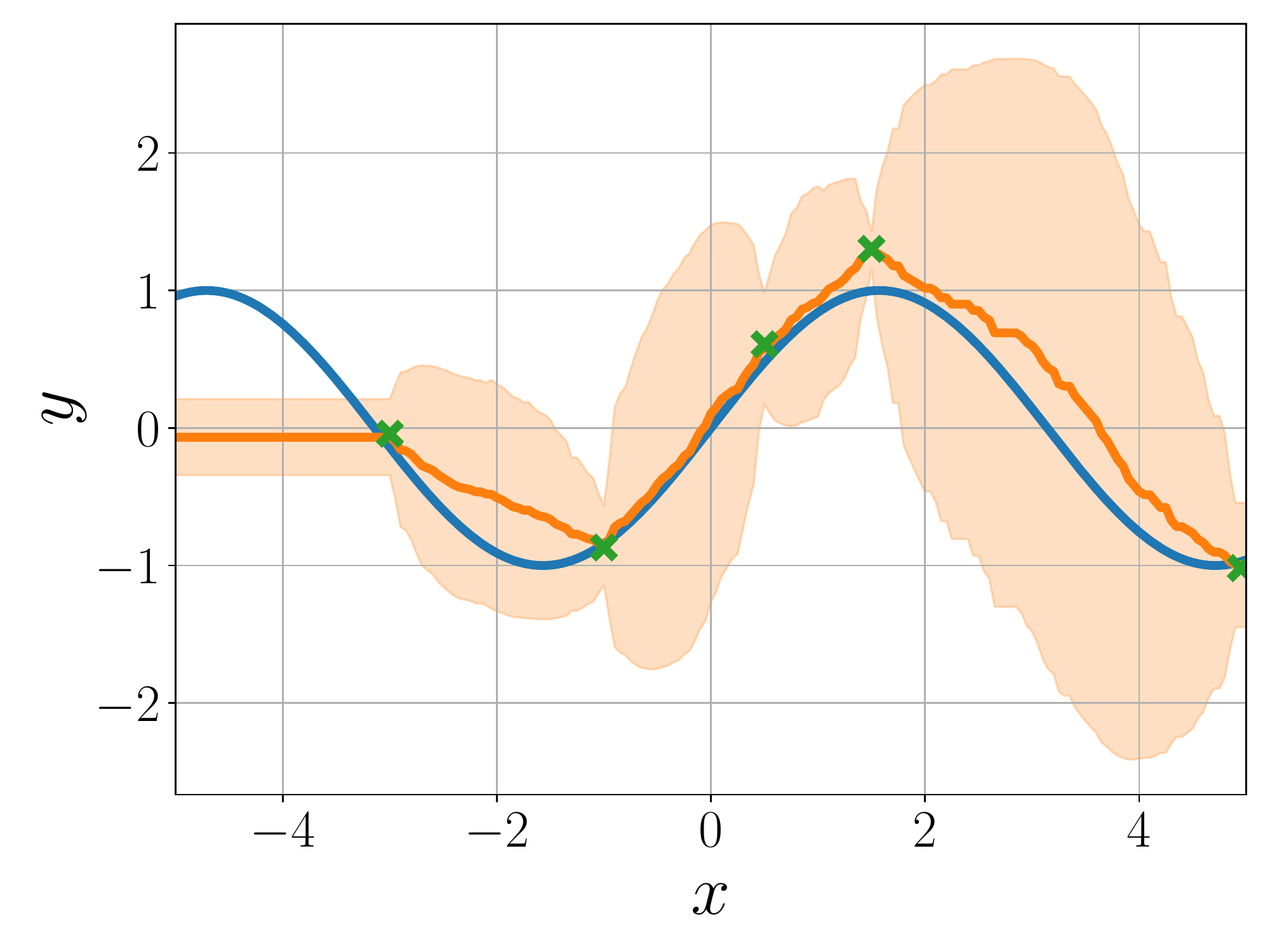}
		\label{fig:unc_1d_few_ours}
	}
	\vspace{-7pt}
	\caption{Results with GP regression and tree-based surrogate models such as random forest (RF), extremely randomized trees (ERTs), BART, Mondrian forest (MF), NGBoost, and BwO forest (ours). We randomly sample 5 points (green \textsf{x}), where a true function is the sine function (blue) and observation noises exist. Mean function (orange) and region with $\pm 1.96$ standard deviation is plotted. Qualitative analyses are presented in \tabref{tab:qual}.\label{fig:unc_1d_few}}
	\vspace{-7pt}
\end{figure*}

In this paper, we investigate SMO with a random forest-based model
and provide a new understanding of tree-based models.
Under such an understanding, we suggest our strategies with sophisticated ensemble models of trees, e.g., Bayesian additive regression trees (BART)~\citep{ChipmanHA2010aas},
Mondrian forest~\citep{LakshminarayananB2016aistats}, and NGBoost~\citep{DuanT2020icml},
and then propose a new tree-based surrogate model, named bagging with oversampling (BwO) forest.
To be clear, the goal of this work is not to outperform sequential GP-based optimization 
in every task. We aim to re-examine the random forest-based approach 
in terms of prediction uncertainty estimation, discuss about SMO strategies 
with diverse tree-based surrogate models, and propose a new method that inherits \emph{the nature of the ensemble of randomized trees} 
and also follows \emph{the underlying intuition about uncertainties}.

To briefly present our main analyses on prediction uncertainty estimation by tree-based surrogate models,
we demonstrate 1D examples using GP,
random forest, extremely randomized trees~\citep{GeurtsP2006ml},
BART,
Mondrian forest, NGBoost,
and BwO forest,
as shown in~\figref{fig:unc_1d_few}.
While the results by random forest, BART, Mondrian forest, and NGBoost
are distinct from the result by GP regression,
our BwO forest yields satisfactory uncertainty estimation,
which represents small variance on the region that the previous decisions have already been evaluated
and large variance on an unexplored region.
By the definition of uncertainties~\citep{GalY2016phd},
we can interpret that random forest, BART, (Mondrian forest), and NGBoost
rely on aleatoric uncertainty, which comes from the uncertainty of data such as a noise in data,
and on the contrary, GP, (Mondrian forest), and BwO forest capture epistemic uncertainty as well.\footnote{The reason why Mondrian forest is included in both groups is that the results (e.g., \figref{fig:exp_bo_michalewicz}) show that Mondrian forest also works well by capturing the uncertainty derived from the amount of the knowledge of data.}
Finally, such intuitive results lead us to obtain the global optimization results 
that BwO forest tends to be more beneficial than the other tree-based models
in various circumstances defined on continuous, high-dimensional binary, and mixed search spaces.
In addition to such results, our BwO forest consistently enjoys the advantage of computational efficiency,
similar to some of tree-based surrogate models such as random forest and extremely randomized trees.

Our contributions are summarized as follows:
\begin{enumerate}[(i)]
\item We investigate the characteristics of tree-based surrogate models (e.g., random forest, extremely randomized trees, BART, Mondrian forest, and NGBoost) in terms of prediction uncertainties;
\item We propose a new ensemble of randomized trees, named BwO forest, which can elaborate uncertainty estimation and yield the satisfactory results that follow the intuition about prediction uncertainties;
\item We employ various tree-based surrogate models including BwO forest as a component of SMO in solving diverse global optimization problems.
\end{enumerate}

\section{PREDICTION UNCERTAINTY ESTIMATION BY TREE-BASED SURROGATE MODELS\label{sec:uncertainty}}

In this section, we begin by introducing the notation of tree-based surrogate models.
Denote that a decision tree $\calT = (\bstau, \bsdelta, \bseta)$,
where $\bstau = \{\bstau_r, \bstau_d, \bstau_l\}$ is the nodes of tree including the root node $\bstau_r$, all the decision nodes $\bstau_d$, and all the leaf nodes $\bstau_l$,
$\bsdelta$ is a collection of all the split dimensions of the parent nodes,
and $\bseta$ is a collection of the split locations thereof.
A tree-based model $\hat{f}: \bbR^d \to \bbR$ is an ensemble of $B$ decision trees $\{\calT_b\}_{b = 1}^B$,
where $N$ points $\bX \in \bbR^{N \times d}$ and their evaluations $\by \in \bbR^N$
are given.
For example, if we define a surrogate output as the average of all the outputs of base decision trees, the surrogate is
defined as
\begin{equation}
	\hat{f}(\bx; \bX, \by) = \frac{1}{B}\sum_{b = 1}^B g(\bx; \calT_b, \bX, \by),
	\label{eqn:average_trees}
\end{equation}
where a function $g$ guides a route to a leaf node over $\bx$:
\begin{equation}
	g(\bx; \calT, \bX, \by) = \sum_{\tau \in \bstau_{l}} \frac{\sum_{i = 1}^N y_i 1_{\bx_i \in \tau}}{\sum_{i = 1}^N 1_{\bx_i \in \tau}}1_{\bx \in \tau},
\end{equation}
and $1_{\bx \in \tau}$ is 1 if $\bx \in \tau$ is true; otherwise, it is 0.
For brevity, we denote $\frac{\sum_{i = 1}^N y_i 1_{\bx_i \in \tau}}{\sum_{i = 1}^N 1_{\bx_i \in \tau}}$ by $\mu_\tau$,
which can be preemptively computed using $\bX$ and $\by$.
In addition to $\mu_\tau$, without loss of generality, the variance of node $\tau$, $\sigma_\tau^2$ can also be defined with $\mu_\tau$.
As will be discussed, the definition of tree-based surrogate model can differ as to how we define its formulation;
however we first describe the surrogate that defines as the form of \emph{sum-of-trees model} \eqref{eqn:average_trees}.

\subsection{Sum-of-Trees Models\label{subsec:sot}}

Generic tree-based surrogate models such as bootstrap aggregating (bagging)~\citep{BreimanL1996ml} with decision trees 
and random forest~\citep{BreimanL2001ml}, however, 
do not model a posterior predictive distribution over real-valued variables $p(y|\bx, \bX, \by)$.
To define a function prediction with its uncertainty for the sum-of-trees model,
under the assumption that a joint distribution over all random variables is a multivariate normal distribution,
the posterior predictive distribution is defined as
\begin{align}
	p(y|\bx, \bX, \by) = \calN \big(y | &\mu\big(\bx; \{\calT_b\}_{b = 1}^B, \bX, \by\big), \nonumber\\
	&\sigma^2\big(\bx; \{\calT_b\}_{b = 1}^B, \bX, \by\big) \big),
	\label{eqn:predictive_dist}
\end{align}
where
\begin{align}
	\mu\big(\bx; \{\calT_b\}_{b = 1}^B, \bX, \by\big) &= \frac{1}{B} \sum_{b = 1}^B \mu_b(\bx)\nonumber\\
	&= \frac{1}{B} \sum_{b = 1}^B \sum_{\tau \in \bstau_{b, l}} \mu_{\tau} 1_{\bx \in \tau},\label{eqn:forest_mean}
\end{align}
\begin{align}
	&\sigma^2\big(\bx; \{\calT_b\}_{b = 1}^B, \bX, \by\big) \nonumber\\
	&= \frac{1}{B}\sum_{b = 1}^B \big(\sigma_b^2(\bx) + \mu_b^2(\bx)\big) - \mu\big(\bx; \{\calT_b\}_{b = 1}^B, \bX, \by\big)^2\nonumber\\
	&= \frac{1}{B}\sum_{b = 1}^B \bigg(\Big(\sum_{\tau \in \bstau_{b, l}} \sigma_{\tau} 1_{\bx \in \tau}\Big)^2 + \Big(\sum_{\tau \in \bstau_{b, l}} \mu_{\tau} 1_{\bx \in \tau}\Big)^2\bigg) \nonumber\\
	&\quad- \bigg(\frac{1}{B} \sum_{b = 1}^B \mu_b(\bx)\bigg)^2,\label{eqn:forest_variance}
\end{align}
by the law of total variance, as described in~\citep{HutterF2014ai}.
Note that $\bstau_{b, l}$ is a set of leaf nodes for tree $b$.
\citet{HutterF2011lion} have applied the formulation \eqref{eqn:predictive_dist} in SMO,
and it is straightforwardly employed to estimate a prediction uncertainty using BART
and Mondrian forest.

The uncertainty of such an ensemble model is derived from the randomness of
individual trees, which is achieved by one or more of these techniques:
\begin{enumerate}[(i)]
	\item bagging: it samples a bootstrap sample from $\bX$ with replacement and then aggregates base estimators;
	\item random feature selection: this technique randomly selects $\bsdelta$ from a set of dimensions;
	\item random selection of split locations: it randomly selects $\bseta$ between lower and upper bounds of the selected dimension;
	\item random tree sampling: this strategy randomly samples a tree under the assumption on a prior distribution over trees.
\end{enumerate}
As shown in~\figref{fig:unc_1d_few}, these techniques are effective
in estimating an uncertainty.
However, compared to the result by GP regression, 
the results by random forest, BART, and Mondrian forest
(see \figref{fig:unc_1d_few_rf}, \figref{fig:unc_1d_few_bart}, and \figref{fig:unc_1d_few_mf}, respectively)
tend not to follow the underlying property of epistemic uncertainty, which
has a small uncertainty on the region that the previous decisions by SMO have already been evaluated
and a large uncertainty on the region that has not been explored yet.
This property of epistemic uncertainty is required to explore an unseen region effectively.
Before explaining why it occurs, we specify all the algorithms based on their respective original papers;
random forest employs (i) and (ii),
BART employs (i), (ii), and (iv),
and Mondrian forest employs (i) and (iii);
see the Leo Breiman's seminal work and the corresponding original references
for the details of these algorithms.

By \eqref{eqn:forest_variance}, the results that do not follow the underlying property of epistemic uncertainty imply that
two adjacent points have the same variance;
formally, given two adjacent points $\bx$, $\bx'$ where $\|\bx - \bx'\| < \varepsilon$ for $0 < \varepsilon \ll 1$,
the following equation
$|\sigma^2\big(\bx; \{\calT_b\}_{b = 1}^B, \bX, \by\big) - \sigma^2\big(\bx'; \{\calT_b\}_{b = 1}^B, \bX, \by\big)| = 0$ is satisfied almost everywhere.
This consequence is mainly induced due to the deterministic selection of split locations.
Although there exist a large enough number of distinct bootstrap 
samples -- we can choose $B$ bootstrap samples among $\binom{2N -1}{N - 1}$ bootstrap samples~\citep{HolmesSP2004web},
for example, if $N = 10$, there exist 92,378 bootstrap samples,
possible aggregations of base estimators are finite and
the aforementioned equation is satisfied almost everywhere.
However, in addition to this statement, we need to explain
the result by Mondrian forest, which uses the technique, random selection of split location~\citep{GeurtsP2006ml}
but tend not to follow the intuition about uncertainties; also see \figref{fig:unc_1d_many}.
An appropriate explanation is that this outcome is derived from a bootstrapping technique,
which makes a surrogate model underfit due to random sampling with replacement.
These understandings lead us to propose BwO forest, as will be presented in~\secref{sec:elaborating}.

Before proposing our method, we first review a recent study on a sophisticated tree-based surrogate model.

\subsection{Gradient Boosting Models\label{subsec:gb}}

Compared to a class of surrogate models described in~\secref{subsec:sot}, a more direct approach to estimating parameters
has recently been proposed~\citep{DuanT2020icml}.
This approach updates parameters $\bstheta$ (e.g., mean and variance)
using their gradients in terms of the objective
of parametric distribution (e.g., likelihood function or continuous ranked probability score).
For example, one of potential objectives, a log likelihood function can be used to find its maximizer:
\begin{equation}
	\calL(\bstheta; \bX, \by) = \sum_{i = 1}^N \log p(y_i|\bstheta(\bx_i)).
\end{equation}
In particular, in~\citep{DuanT2020icml}, natural gradients~\citep{AmariS1998nc_a} are used in updating $\bstheta$
in order to consider an appropriate distance between two parameter vectors,
which is capable of representing the gradient direction in Riemannian space,
and a gradient boosting machine~\citep{FriedmanJH2001aos} with respect to the parameters is built.
\citet{DuanT2020icml} show that the numerical results with the gradient boosting machine updated by natural gradients, dubbed NGBoost, outperform the results by generic gradients.

\begin{algorithm}[t]
	\caption{Training BwO Forest}
	\label{alg:method}
	\begin{algorithmic}[1]
		\REQUIRE Size of ensemble model $B$, training data $\bX \in \bbR^{N \times d}$, training function evaluations $\by \in \bbR^N$, size of bootstrap sample $M = \alpha N$ for the rate of oversampling $\alpha > 1$.
		\ENSURE Set of decision trees $\{ \calT_b \}_{b = 1}^B$
		\STATE Initialize a set of decision trees.
		\FOR {$b = 1, \ldots, B$}
			\STATE Sample a bootstrap sample $\bB_b \in \bbR^{M \times d}$ from $\bX$; the evaluations of $\bB_b$ are also stored using the indices that have already been sampled to construct $\bB_b$ and $\by$.
			\STATE Set a root node $\bstau_r$ that contains all the elements in $\bB_b$, and $\bstau_r$ is set as the current split node.
			\WHILE{a stopping criterion has not been met}
				\STATE Randomly choose a fixed number of split dimensions $\delta$ from all the feature dimensions $\{1, \ldots, d\}$.
				\STATE Determine the best split by uniformly sampling a split location $\eta$ between lower and upper bounds of the values along the selected dimensions $\delta$.
				\STATE Split the current split node into two decision nodes using $\delta$ and $\eta$.
				\STATE Update all the parameters of decision tree, $(\bstau_b, \bsdelta_b, \bseta_b)$.
				\STATE Choose the next split node from decision nodes $\bstau_d$ included in $\bstau_b$.
			\ENDWHILE
			\STATE Update $\bstau_b = \{ \bstau_{r}, \bstau_{d}, \bstau_{l} \}$ by determining leaf nodes $\bstau_l$.
			\STATE Update a set of decision trees by adding $\calT_b = (\bstau_b, \bsdelta_b, \bseta_b)$.
		\ENDFOR
		\STATE \textbf{return} A set of decision trees $\{\calT_b\}_{b = 1}^B$
	\end{algorithmic}
\end{algorithm}

This gradient boosting method over parameters is undoubtedly a reasonable approach to estimating parameters 
as a probabilistic regression model. However, such a multi-parameter boosting algorithm is not robust
in a high-dimensional space, and relies on aleatoric uncertainties~\citep{MmalininA2020arxiv}.

\section{ELABORATING UNCERTAINTY ESTIMATION BY TREE-BASED SURROGATE MODELS\label{sec:elaborating}}

\begin{figure*}[t]
	\centering
	\subfigure{
		\includegraphics[width=0.31\textwidth]{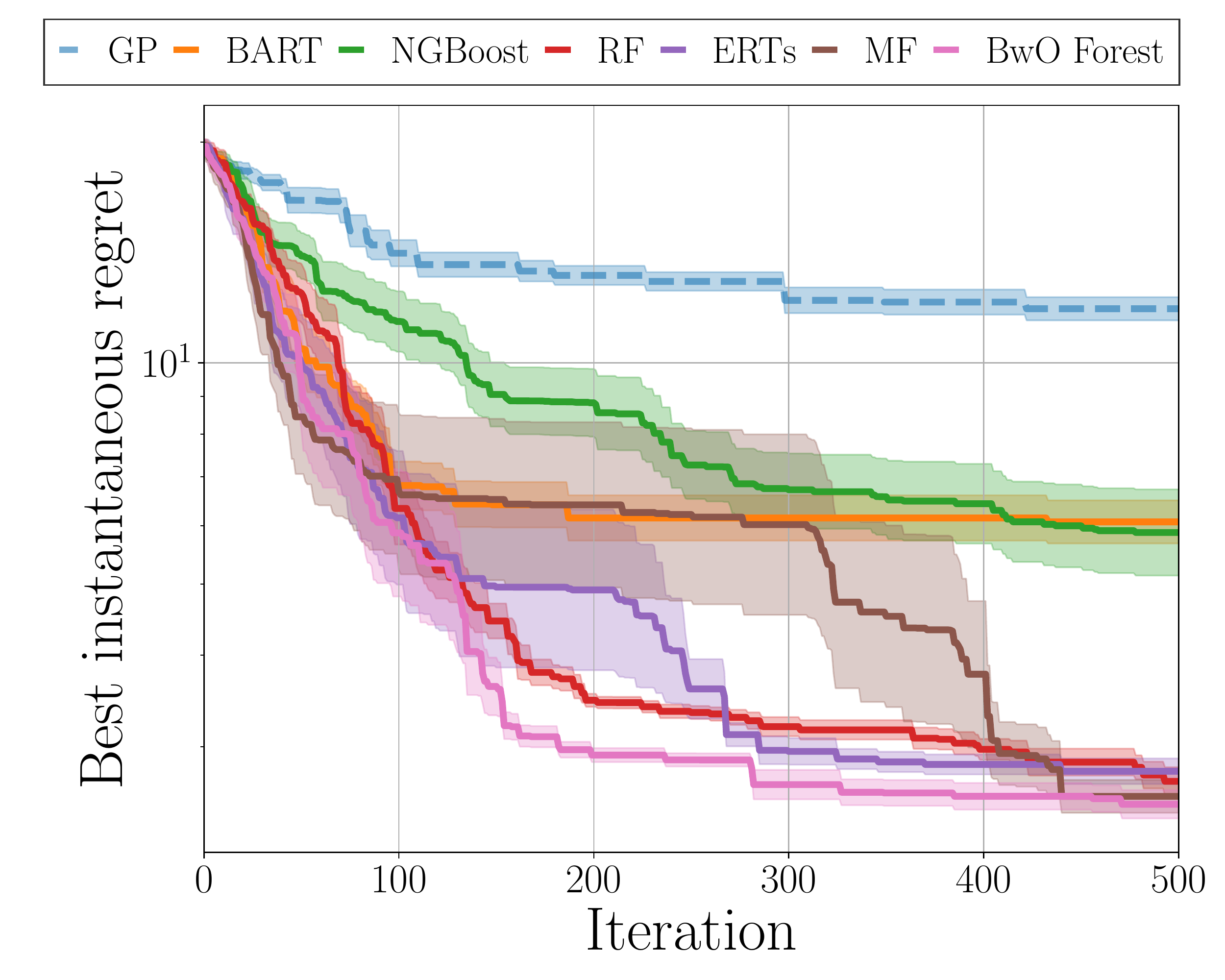}
	}
	\subfigure{
		\includegraphics[width=0.31\textwidth]{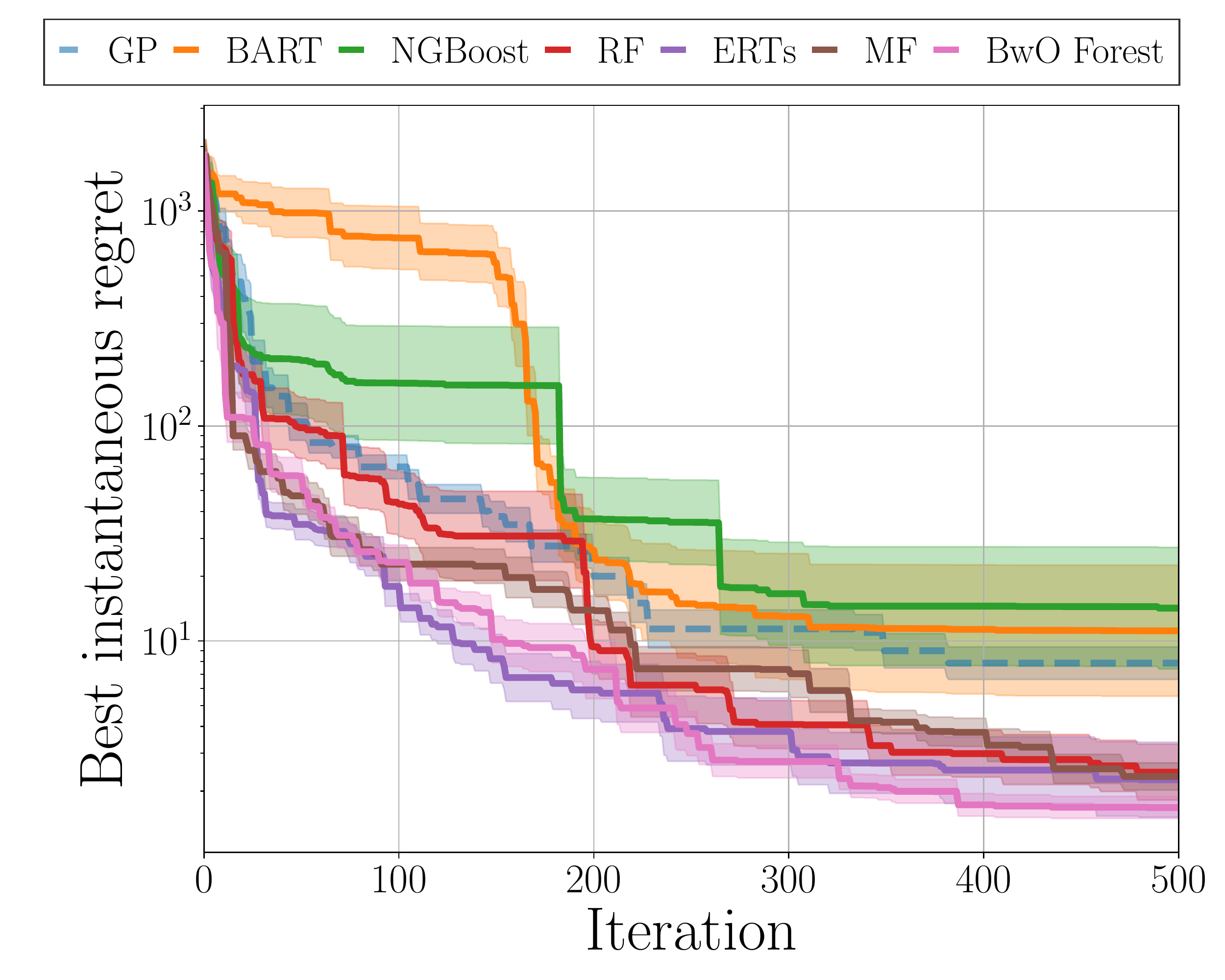}
	}
	\subfigure{
		\includegraphics[width=0.31\textwidth]{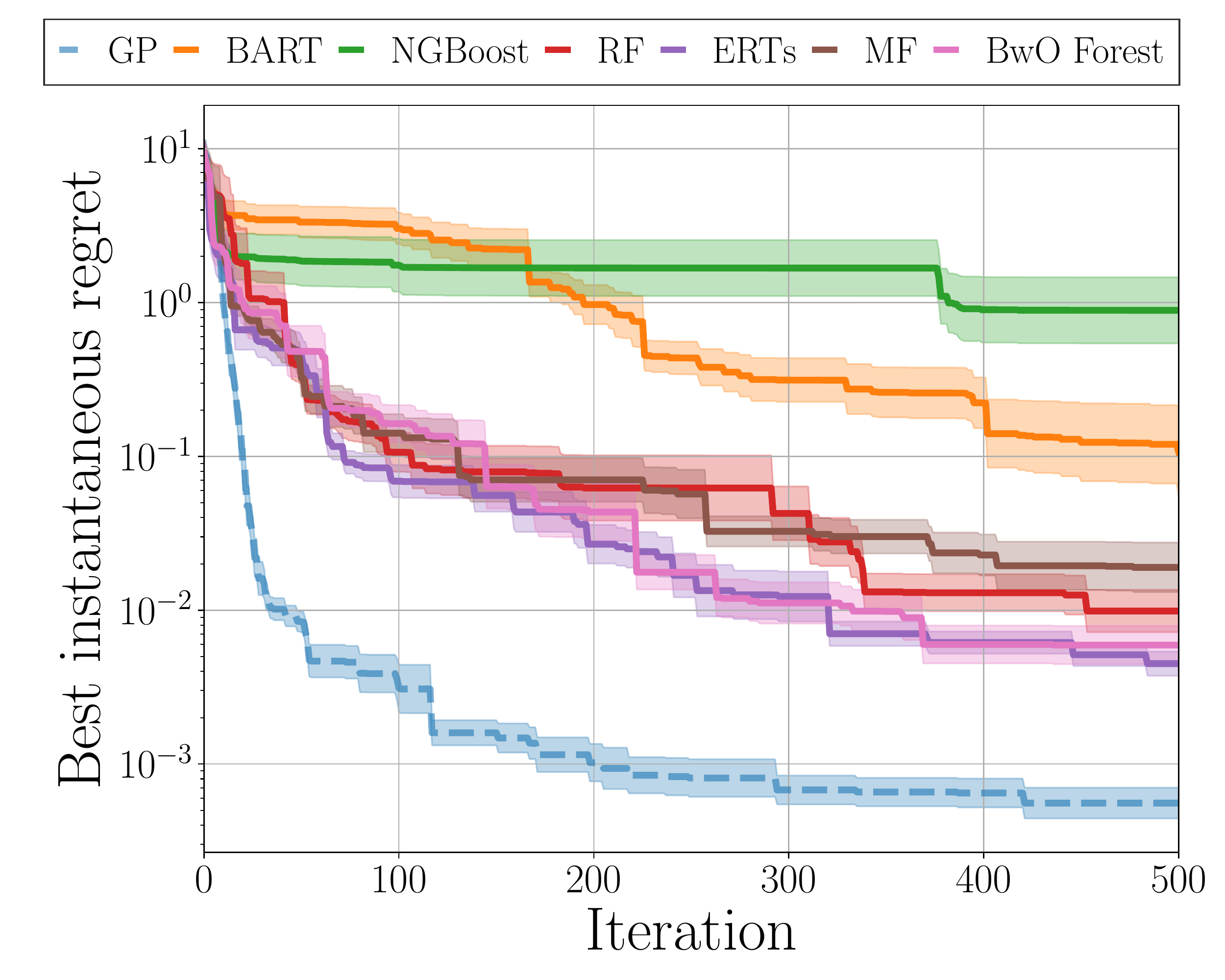}
	}
	\setcounter{subfigure}{0}
	\subfigure[Ackley (4 dim.)]{
		\includegraphics[width=0.31\textwidth]{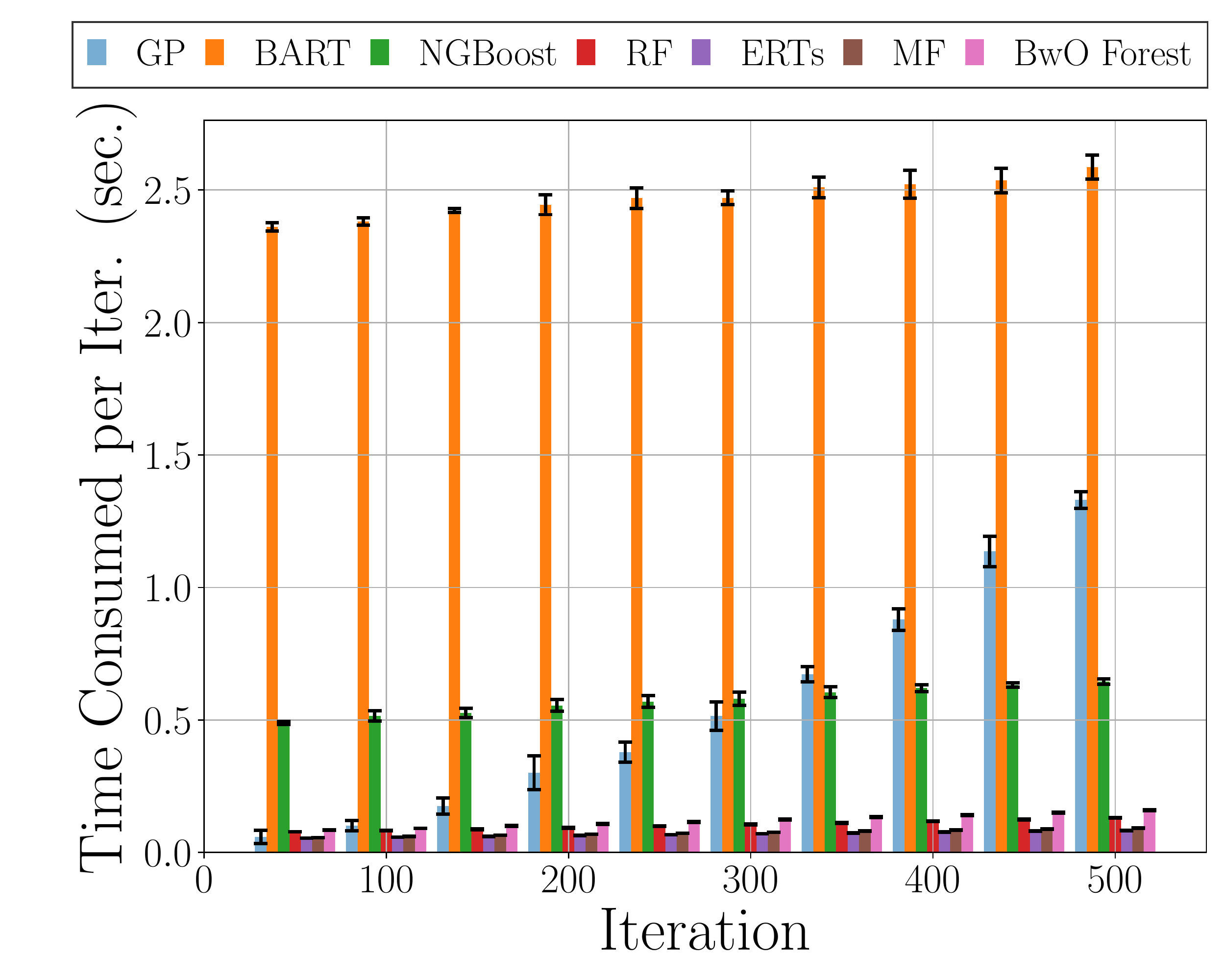}
		\label{fig:exp_bo_ackley}
	}
	\subfigure[Bohachevsky (2 dim.)]{
		\includegraphics[width=0.31\textwidth]{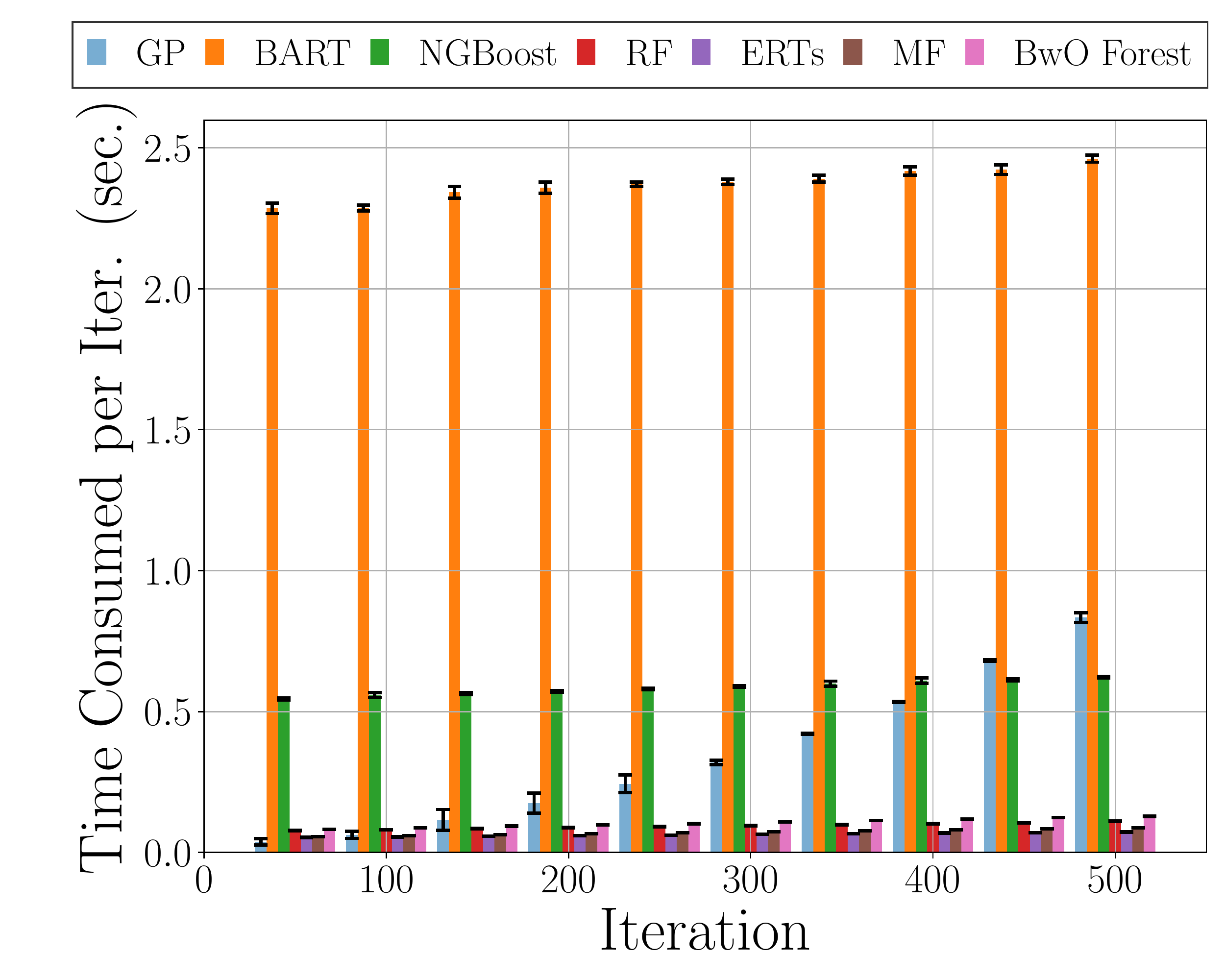}
		\label{fig:exp_bo_bohachevsky}
	}
	\subfigure[Branin (2 dim.)]{
		\includegraphics[width=0.31\textwidth]{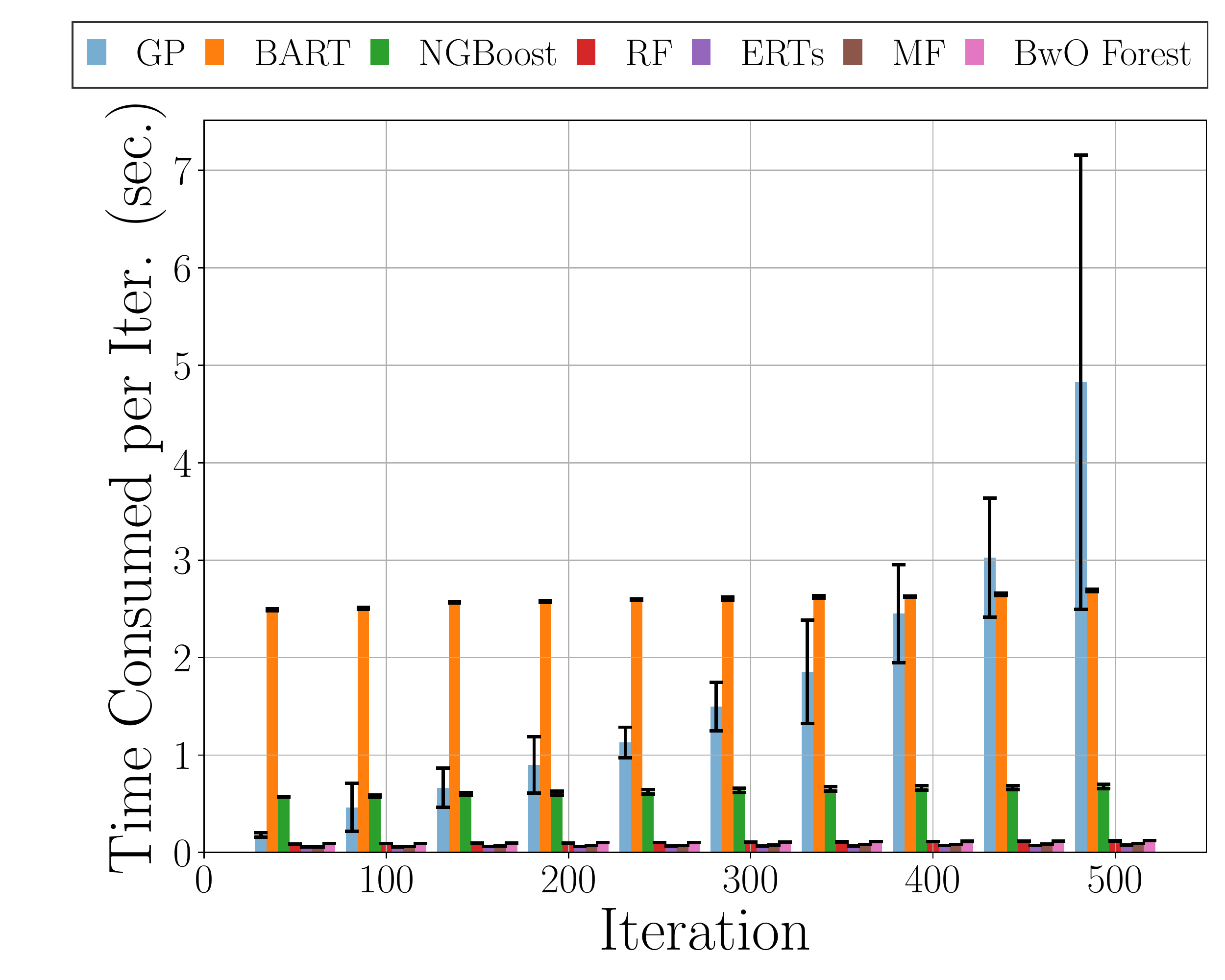}
		\label{fig:exp_bo_branin}
	}
	\vspace{-7pt}
	\caption{Results on various benchmark functions defined on continuous search spaces. The best instantaneous regret (left panel of each figure) and time consumed per iteration (right panel of each figure) versus iterations are plotted. All the runs are repeated 10 times. For the results on regrets, $y$-axis is set as a $\log$-10 scale and relative log error bars are presented. For brevity, the results on runtime are plotted every 50 iterations.\label{fig:exp_bo_benchmarks_1}}
	\vspace{-7pt}
\end{figure*}

\begin{figure*}[t]
	\centering
	\subfigure{
		\includegraphics[width=0.31\textwidth]{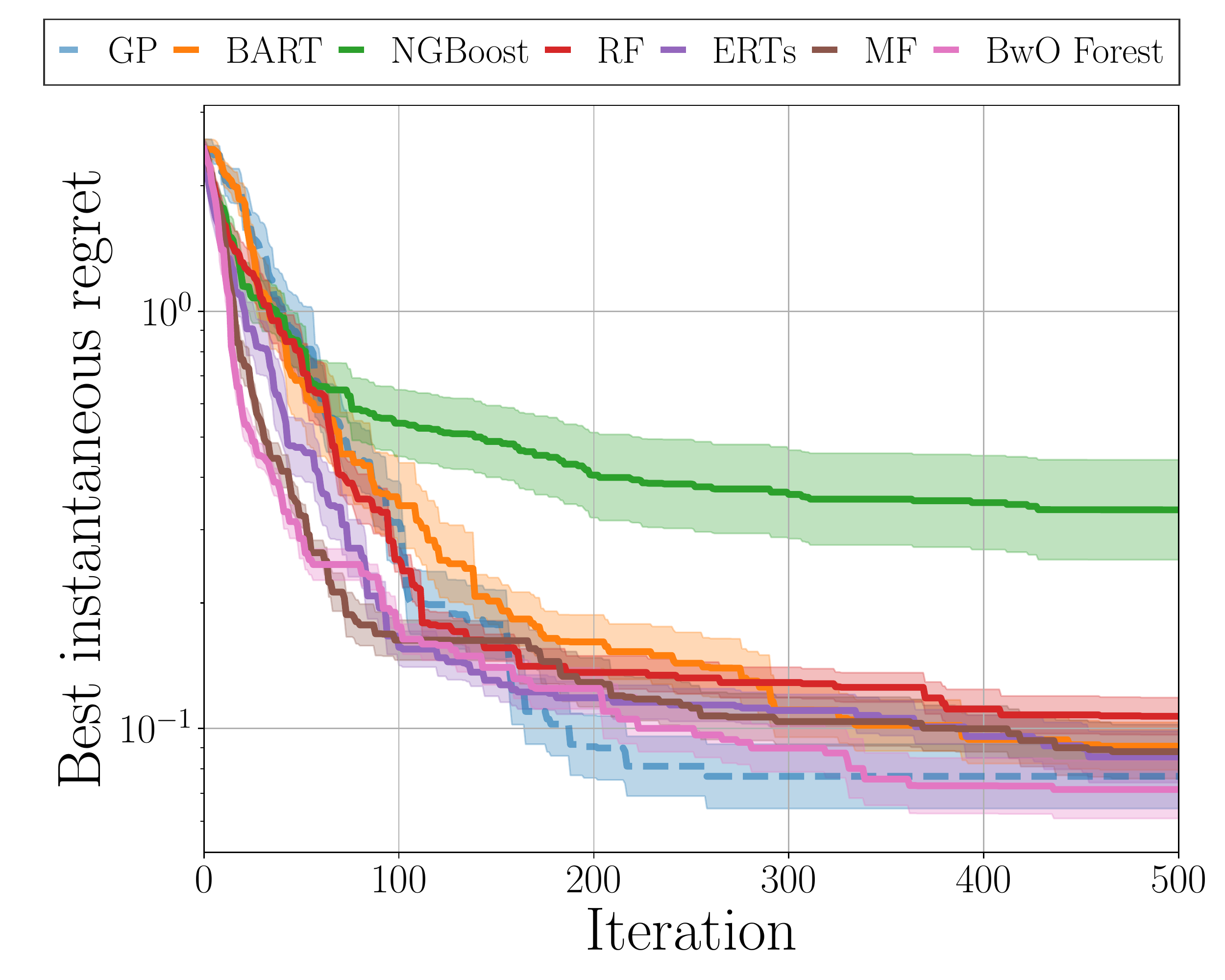}
	}
	\subfigure{
		\includegraphics[width=0.31\textwidth]{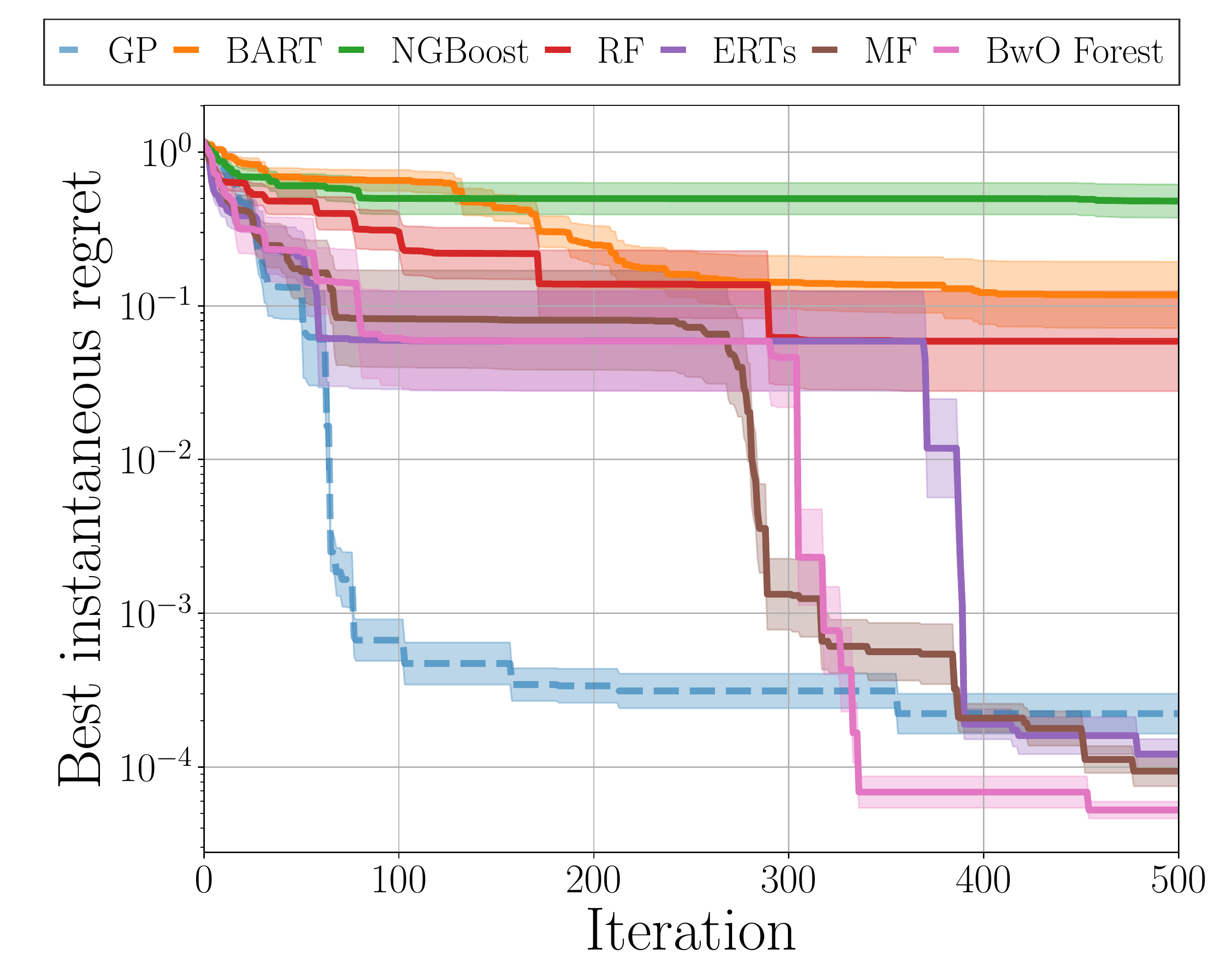}
	}
	\subfigure{
		\includegraphics[width=0.31\textwidth]{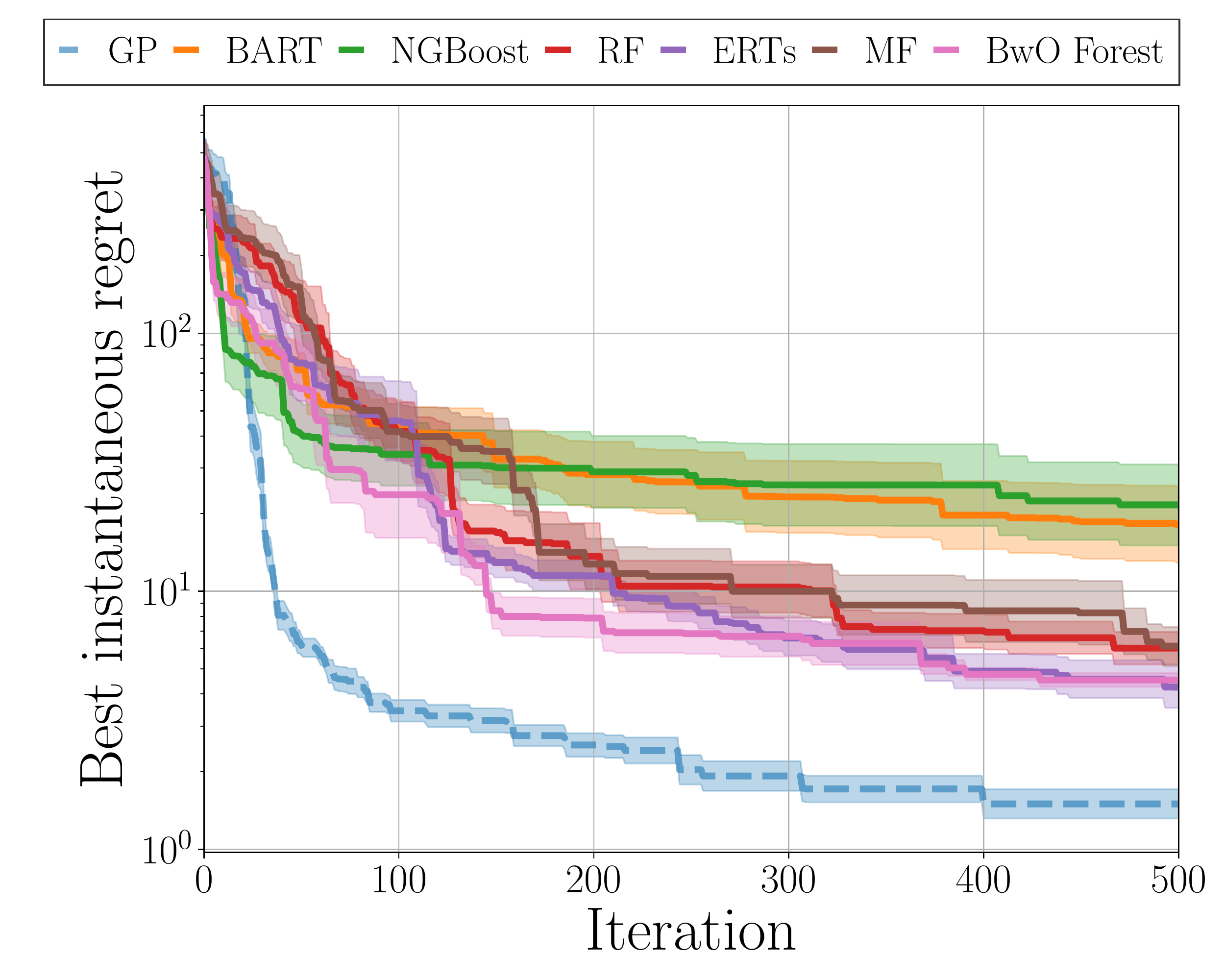}
	}
	\setcounter{subfigure}{0}
	\subfigure[Hartmann6D (6 dim.)]{
		\includegraphics[width=0.31\textwidth]{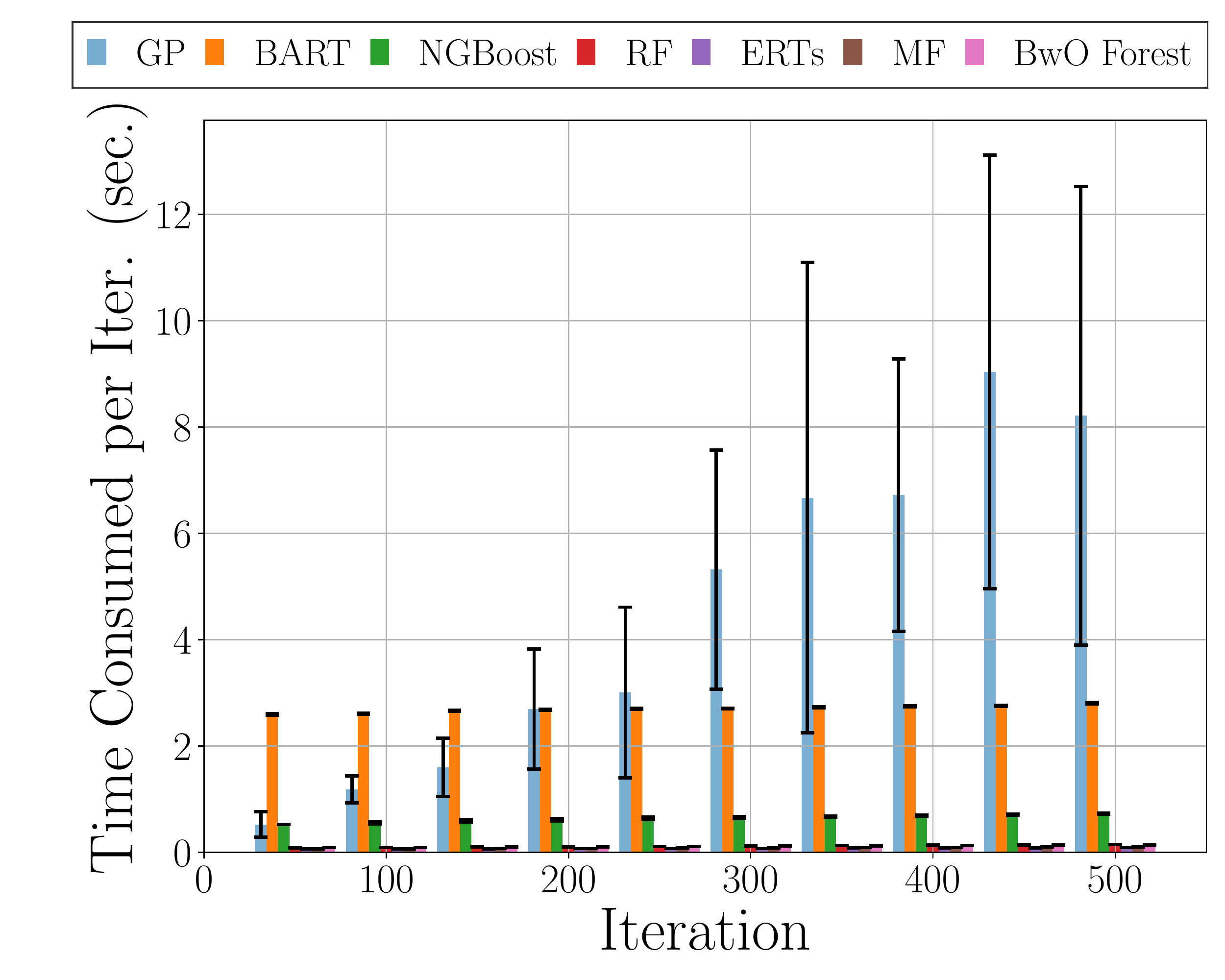}
		\label{fig:exp_bo_hartmann6d}
	}
	\subfigure[Michalewicz (2 dim.)]{
		\includegraphics[width=0.31\textwidth]{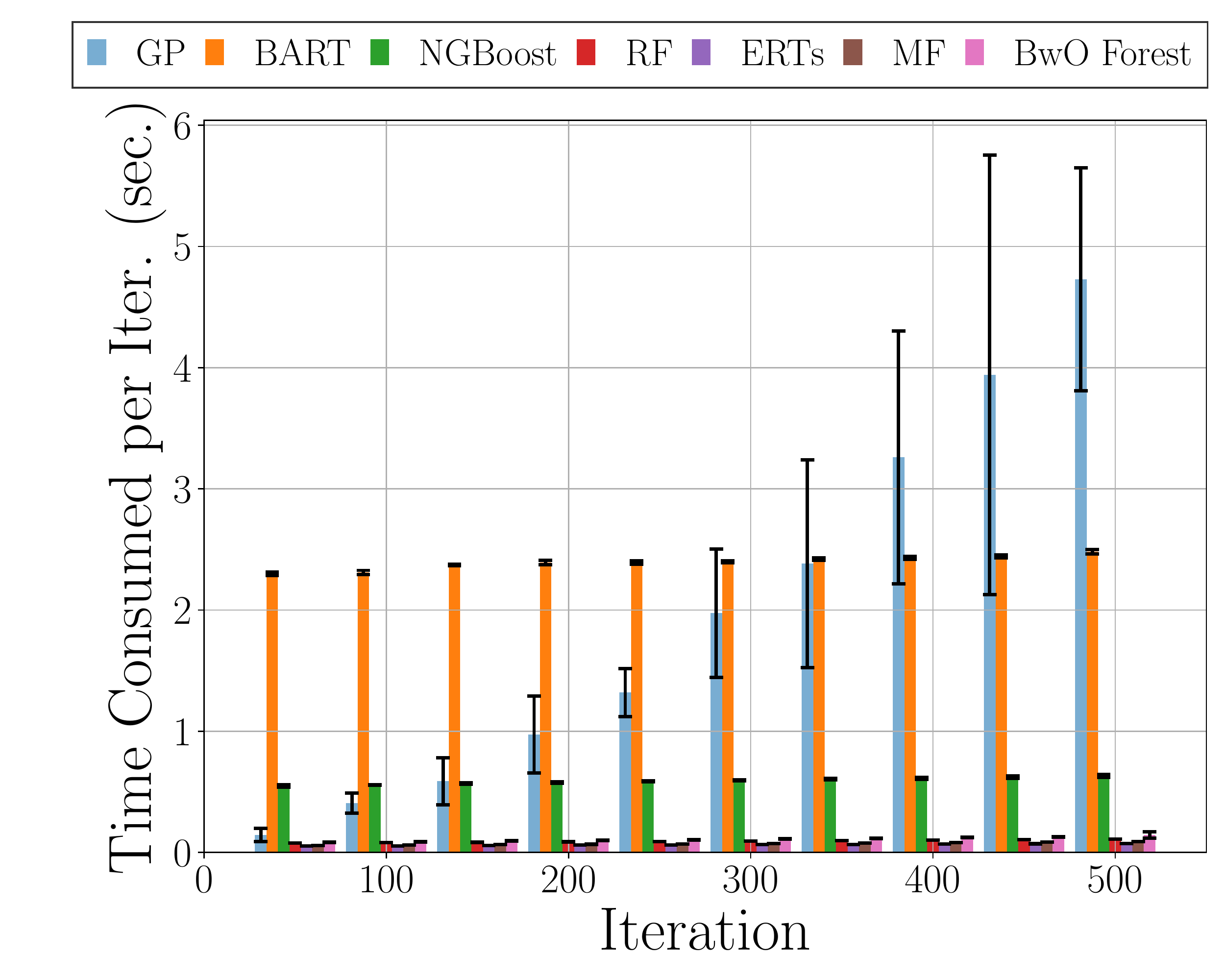}
		\label{fig:exp_bo_michalewicz}
	}
	\subfigure[Rosenbrock (4 dim.)]{
		\includegraphics[width=0.31\textwidth]{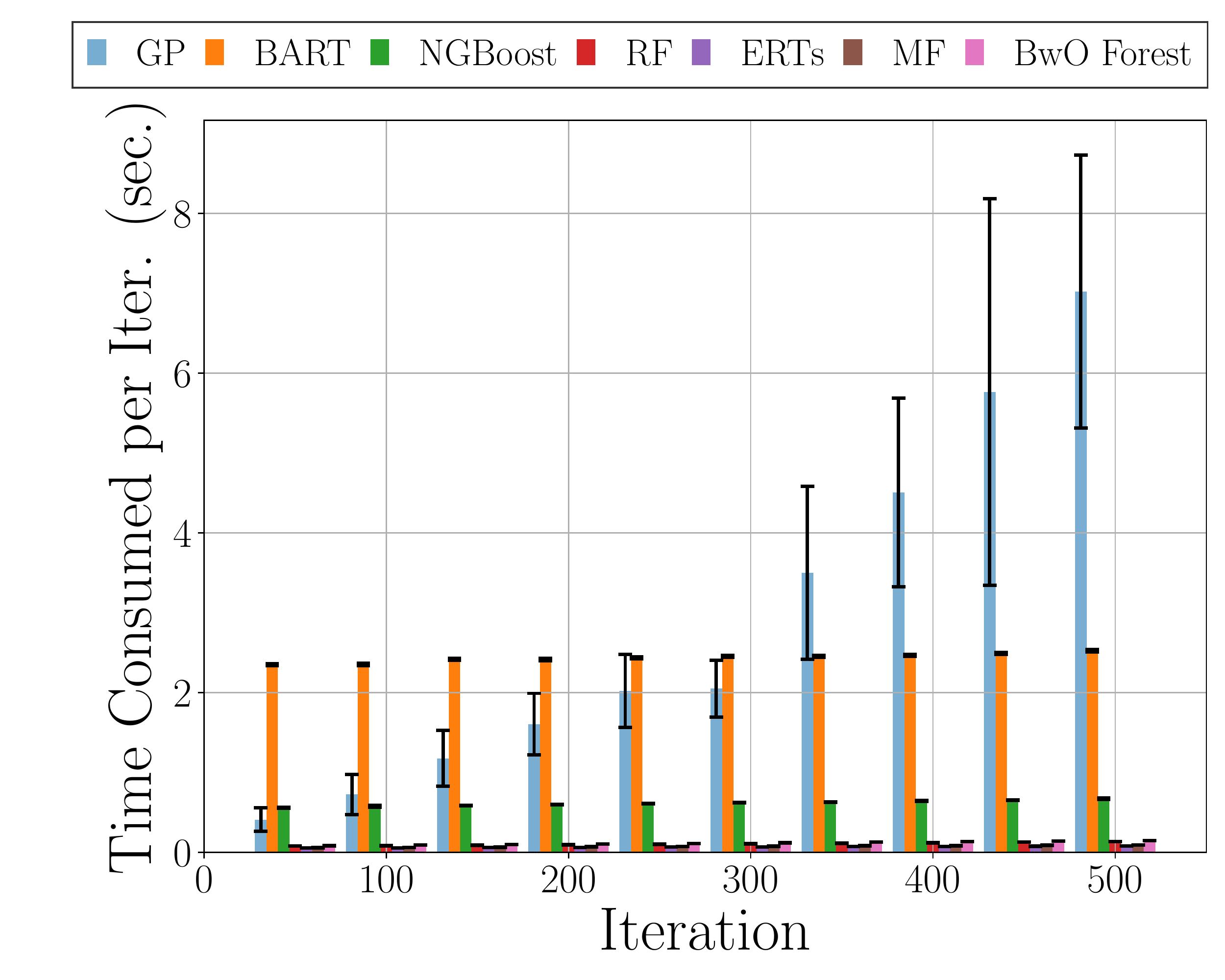}
		\label{fig:exp_bo_rosenbrock}
	}
	\vspace{-7pt}
	\caption{Results on various benchmark functions defined on continuous search spaces. All the experimental settings follow the settings described in~\figref{fig:exp_bo_benchmarks_1}.\label{fig:exp_bo_benchmarks_2}}
	\vspace{-7pt}
\end{figure*}

Before elaborating uncertainty estimation by tree-based surrogate models, 
we introduce a context of how our proposed model is motivated.
According to \secref{subsec:sot}, if a split location is deterministic and bootstrapping
is applied, an uncertainty is not estimated properly.
Similar to these observations, \citet{TangC2018neurips} have discussed 
that either no or severe subsampling leads to inconsistent forest construction,
while they have not mentioned about the uncertainty of tree-based surrogate model.
Here, we verify this issue with the property related to the number of unique original elements in a bootstrap sample.
As pointed out in the work~\citep{MendelsonAF2016arxiv}, 
the expectation and variance of an indicator for the existence of $\bx_i$ in a bootstrap sample $\bB$ are expressed as
\begin{align}
	\bbE[1_{\bx_i \in \bB}] &= p\big(1_{\bx_i \in \bB} = 1\big) \nonumber\\
	&= 1 - p\big(1_{\bx_i \in \bB} = 0\big) \nonumber\\
	&= 1 - \left(1 - \frac{1}{N}\right)^M,\label{eqn:exp_x_b}\\
	\textrm{Var}[1_{\bx_i \in \bB}] &= p(1_{\bx_i \in \bB} = 0) p(1_{\bx_i \in \bB} = 1) \nonumber\\
	&= \left(1 - \frac{1}{N}\right)^M - \left(1 - \frac{1}{N}\right)^{2M},\label{eqn:var_x_b}
\end{align}
where $N$ is the size of $\bX$ and $M$ is the size of a bootstrap sample.
By \eqref{eqn:exp_x_b} and \eqref{eqn:var_x_b}, the distribution of unique original elements in a bootstrap sample $\bB$ can be described:
\begin{align}
	\bbE[|\textrm{unique}(\bB)|] &= \bbE\left[\sum_{i = 1}^N 1_{\bx_i \in \bB}\right] = N - \frac{(N - 1)^M}{N^{M - 1}},\label{eqn:exp_uniq_b}\\
	\textrm{Var}[|\textrm{unique}(\bB)|] &= \textrm{Var}\left[\sum_{i = 1}^N 1_{\bx_i \in \bB}\right] \nonumber\\
	&= (N - 1) \frac{(N - 2)^M}{N^{M - 1}} + \frac{(N - 1)^M}{N^{M-1}} \nonumber\\
	&\quad- \frac{(N - 1)^{2M}}{N^{2M - 2}},\label{eqn:var_uniq_b}
\end{align}
where $\textrm{unique}(\bB)$ filters duplicates and leaves unique original elements.
For example, if $N = M = 5$, \eqref{eqn:exp_uniq_b} and \eqref{eqn:var_uniq_b} return 3.362 and 0.509, respectively,
and if $N = 5$ and $M = 20$, they are 4.942 and 0.055.
Consequently, the distribution specified by \eqref{eqn:exp_uniq_b} and \eqref{eqn:var_uniq_b}
implies that combining two techniques, bagging and oversampling can help a tree construction process
to fit well in $\bX$.

From now, we propose our tree-based surrogate model, named BwO forest, which 
elaborates prediction uncertainty estimation by applying the technique, bagging with oversampling.
As described in~\algref{alg:method}, our BwO forest is trained by following 
a general pipeline of forest construction,
given the size of ensemble model $B$, a training dataset $(\bX, \by)$, and the size of bootstrap samples $M = \alpha N$ for the rate of oversampling $\alpha > 1$.
Note that a stopping criterion is satisfied when some pre-defined conditions such as maximum depth or the minimum number of elements in a node are encountered.
To accommodate a page limit, we highlight the main components of our model.
BwO forest utilizes random selection of split dimensions (Line 6) and random selection of split locations (Line 7),
as well as bagging with oversampling (Line 3).
Interestingly, as will be discussed in~\secref{sec:discussion},
all the components are important for appropriately estimating a function value and its uncertainty.
As a prediction procedure, BwO forest estimates a function value and its uncertainty
by computing \eqref{eqn:predictive_dist}, \eqref{eqn:forest_mean}, and \eqref{eqn:forest_variance},
where a set of trained decision trees $\{\calT_b\}_{b = 1}^B$ is given.

Finally, BwO forest is utilized as a surrogate model in the process of SMO;
see \secref{sec:suppl_smo} for the details.
Compared to SMO with GP regression,
it does not need a step for optimizing a kernel hyperparameter of GP,
which is one of the most time-consuming steps in the corresponding procedure.
Therefore, our method is more efficient than the GP-based approach, and besides
it produces more satisfactory uncertainty estimation than the models with other tree-based surrogate models.
More detailed empirical analyses can be found in the subsequent sections.

\section{EXPERIMENTAL RESULTS\label{sec:experiments}}

In this section, we show the experimental results on continuous, high-dimensional binary,
and mixed search spaces using SMO strategies with diverse tree-based surrogate models
and GP regression. Experimental setup is described below.

\begin{figure*}[t]
	\centering
	\subfigure{
		\includegraphics[width=0.31\textwidth]{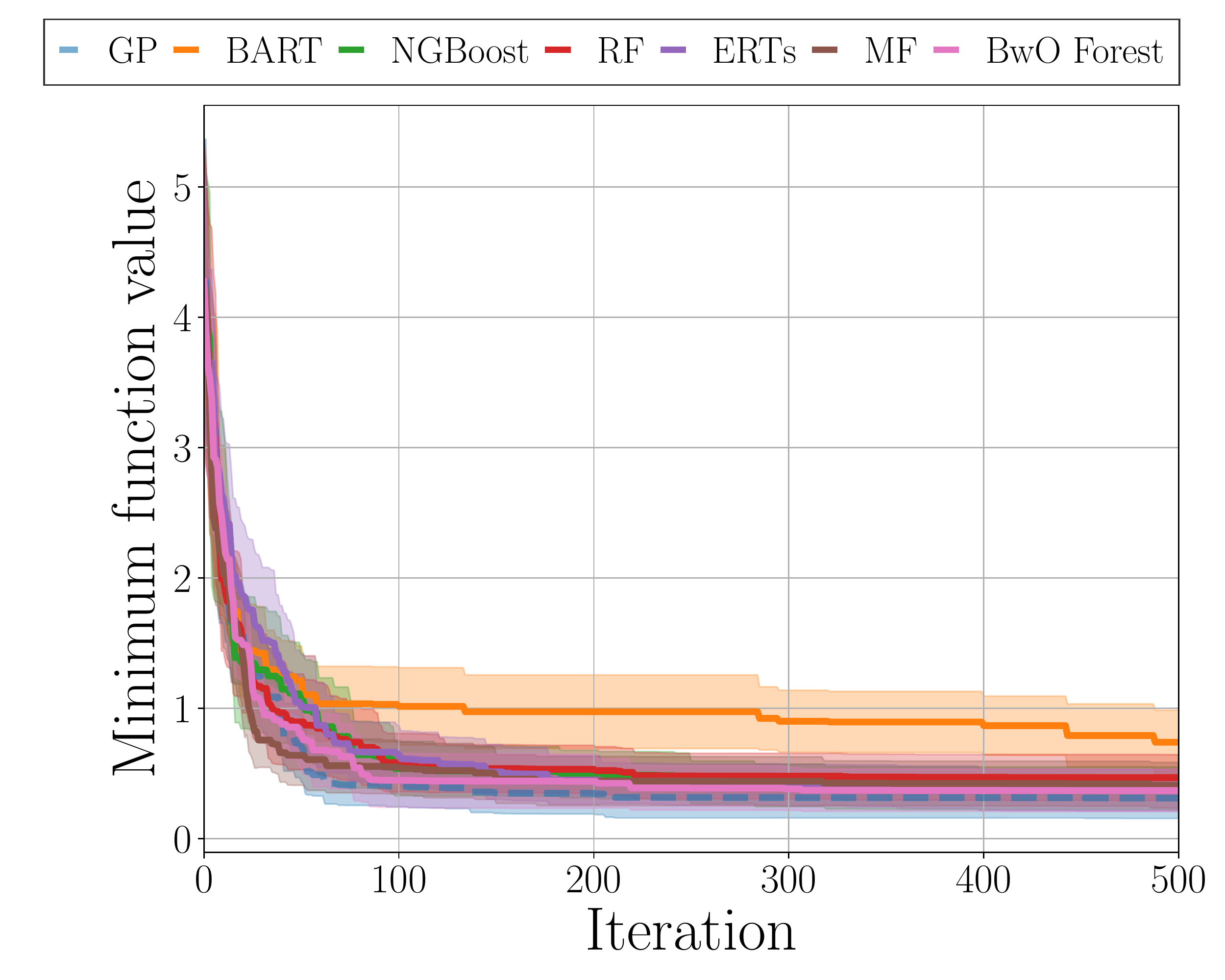}
	}
	\subfigure{
		\includegraphics[width=0.31\textwidth]{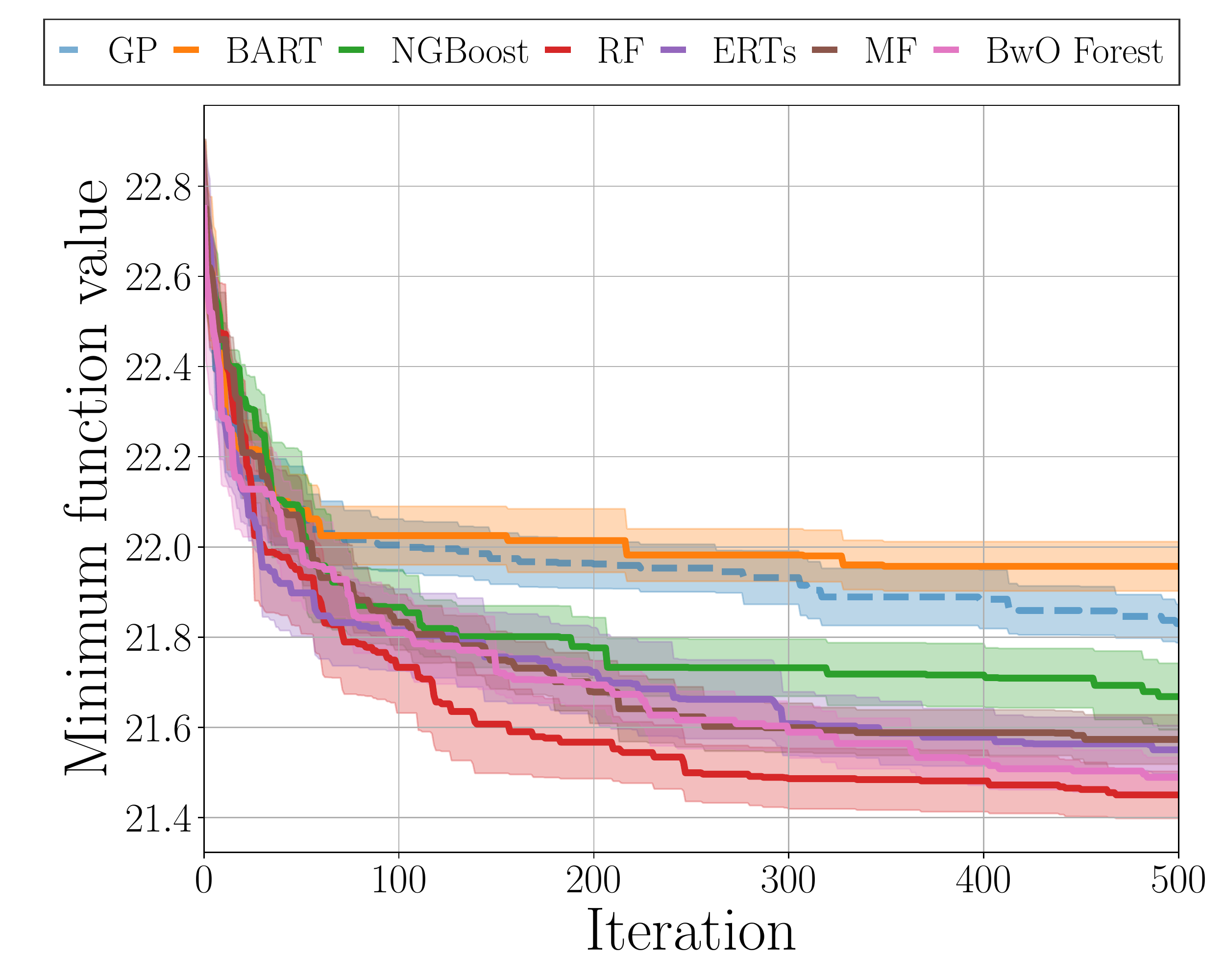}
	}
	\subfigure{
		\includegraphics[width=0.31\textwidth]{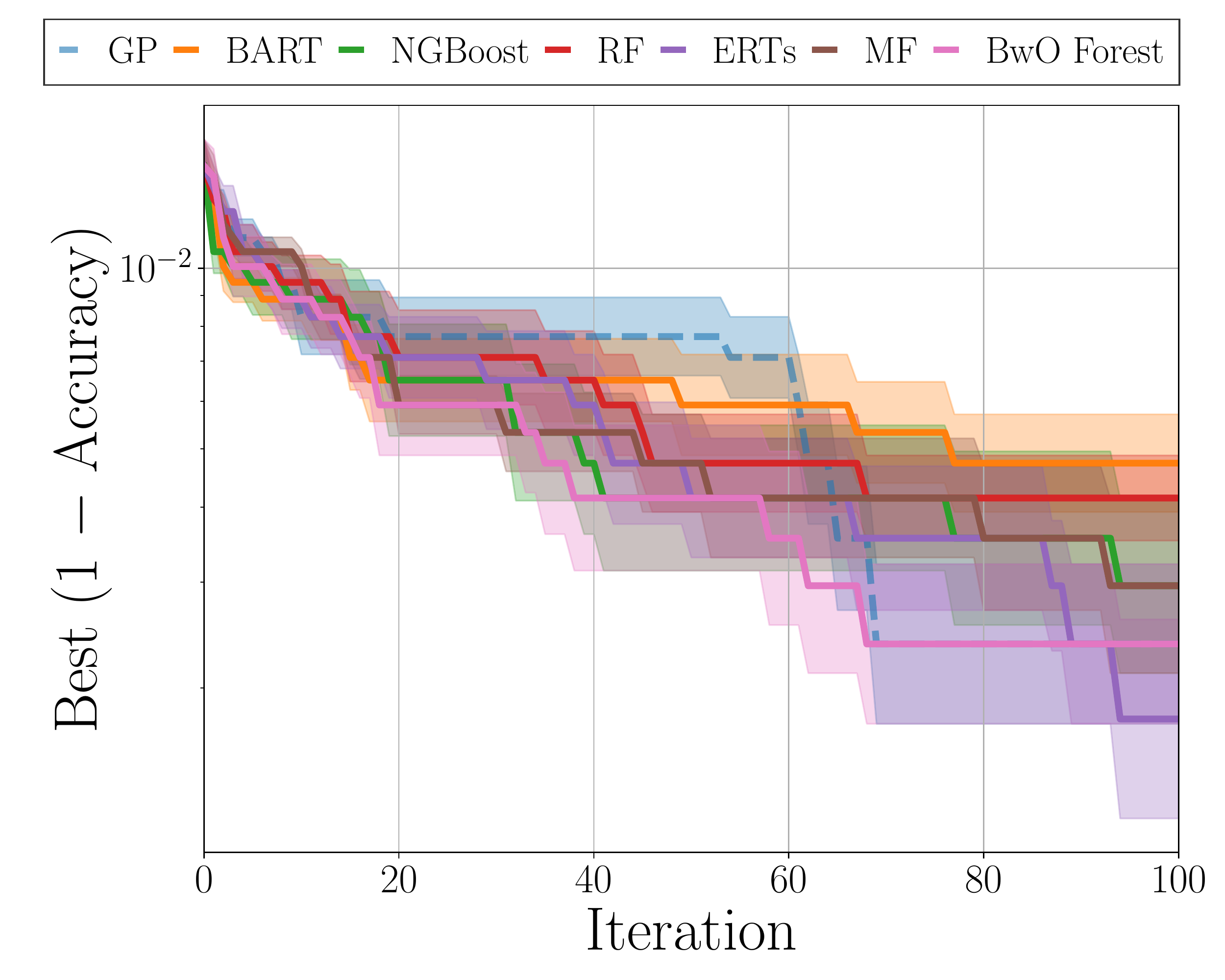}
	}
	\setcounter{subfigure}{0}
	\subfigure[Ising (24 dim., $\lambda = 0.01$)]{
		\includegraphics[width=0.31\textwidth]{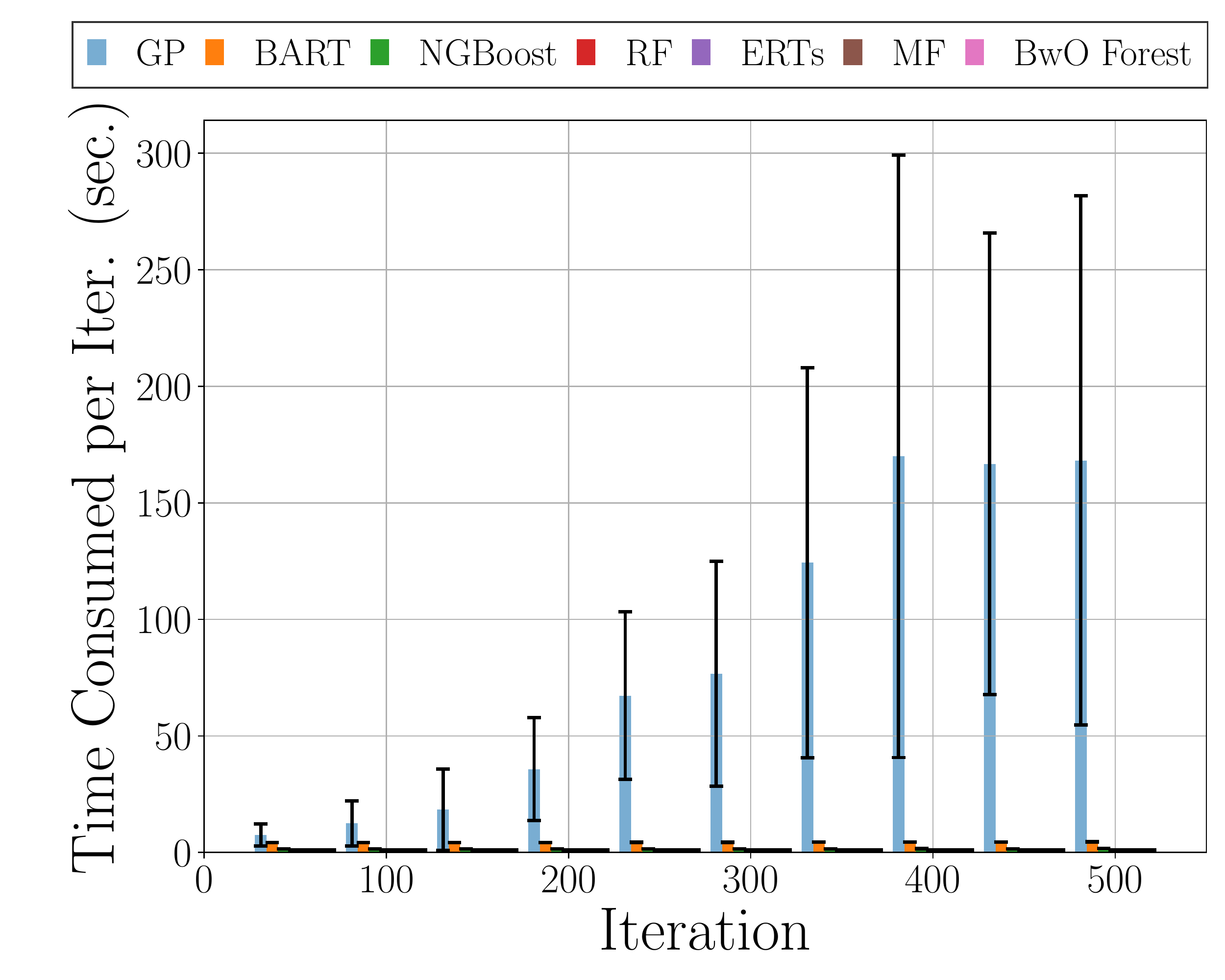}
		\label{fig:exp_binary_1}
	}
	\subfigure[Contamination (25 dim., $\lambda = 0.01$)]{
		\includegraphics[width=0.31\textwidth]{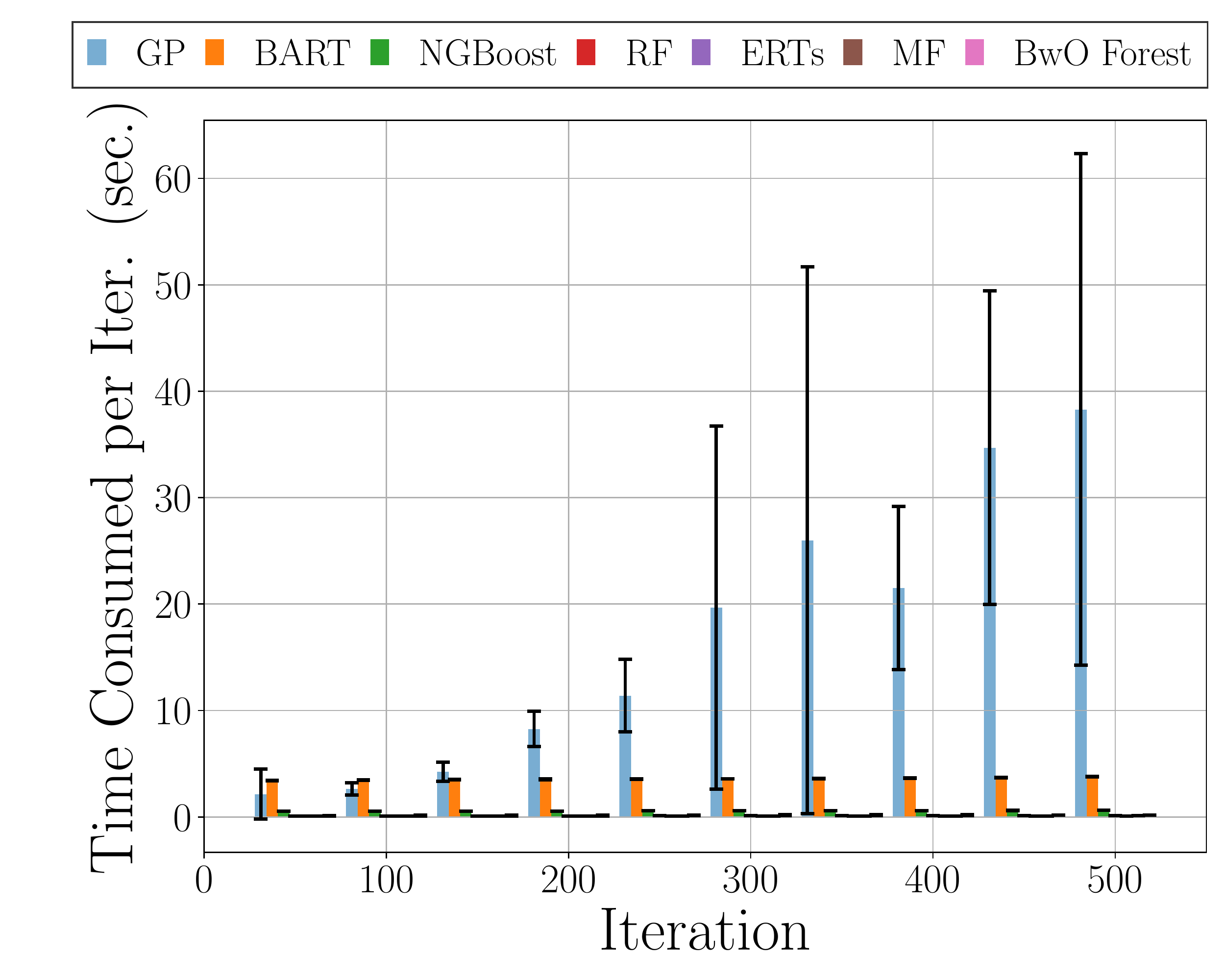}
		\label{fig:exp_binary_2}
	}
	\subfigure[Authorship]{
		\includegraphics[width=0.31\textwidth]{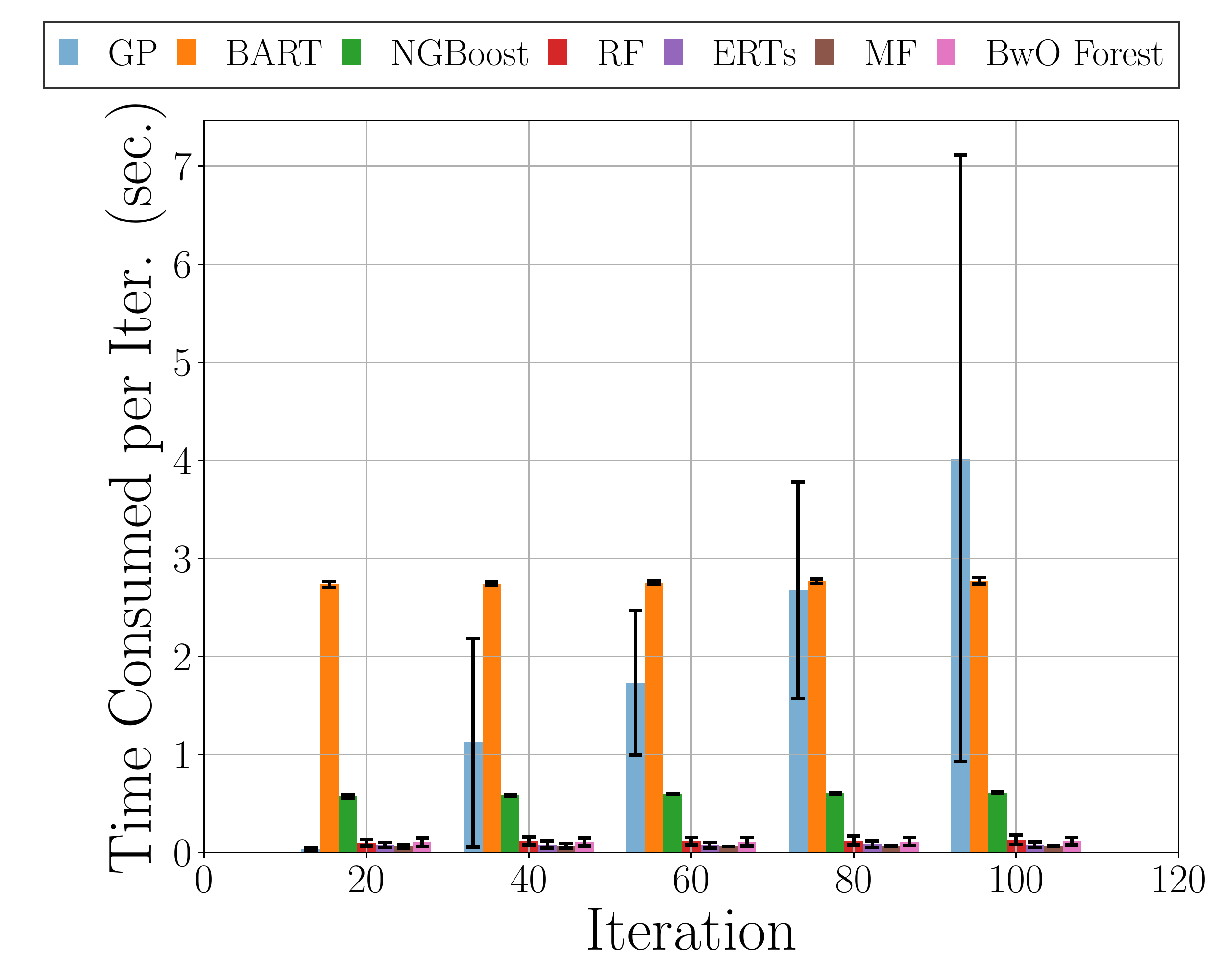}
		\label{fig:automl_1}
	}
	\vspace{-7pt}
	\caption{Results on two functions defined on high-dimensional binary search spaces and automated machine learning for the Authorship dataset. All the runs are repeated 10 times.\label{fig:exp_binary}}
	\vspace{-7pt}
\end{figure*}

\begin{figure*}[t]
	\centering
	\subfigure{
		\includegraphics[width=0.31\textwidth]{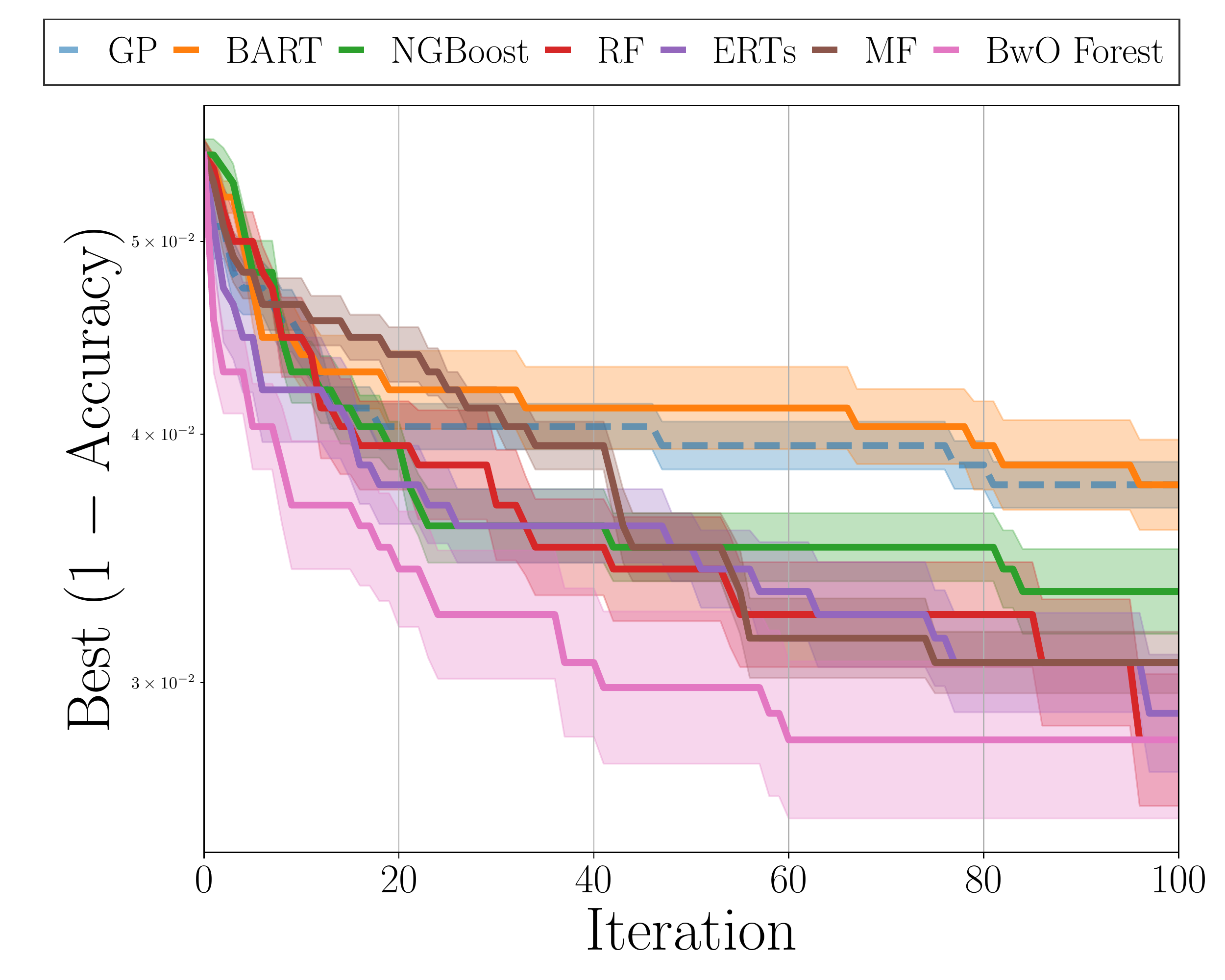}
	}
	\subfigure{
		\includegraphics[width=0.31\textwidth]{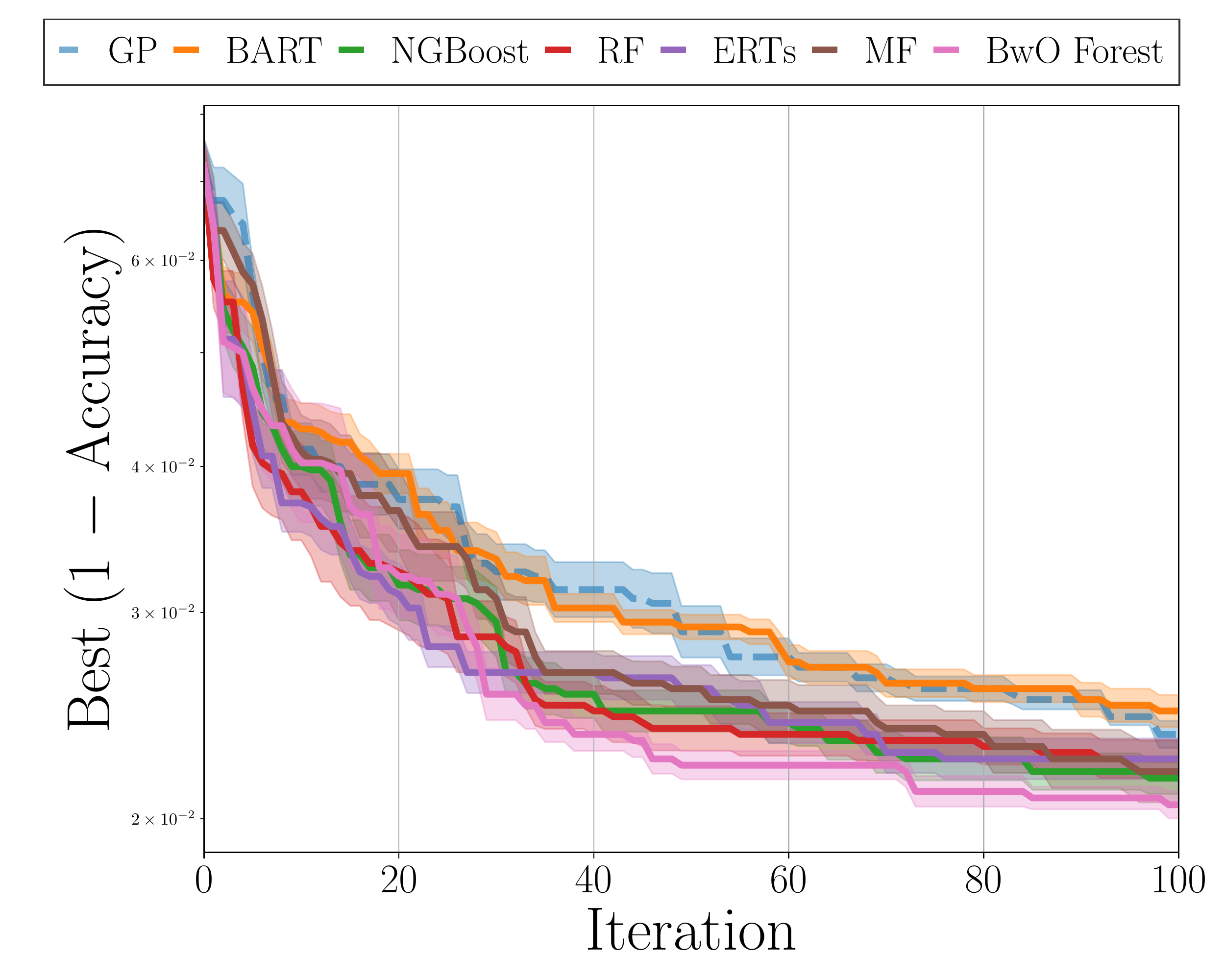}
	}
	\subfigure{
		\includegraphics[width=0.31\textwidth]{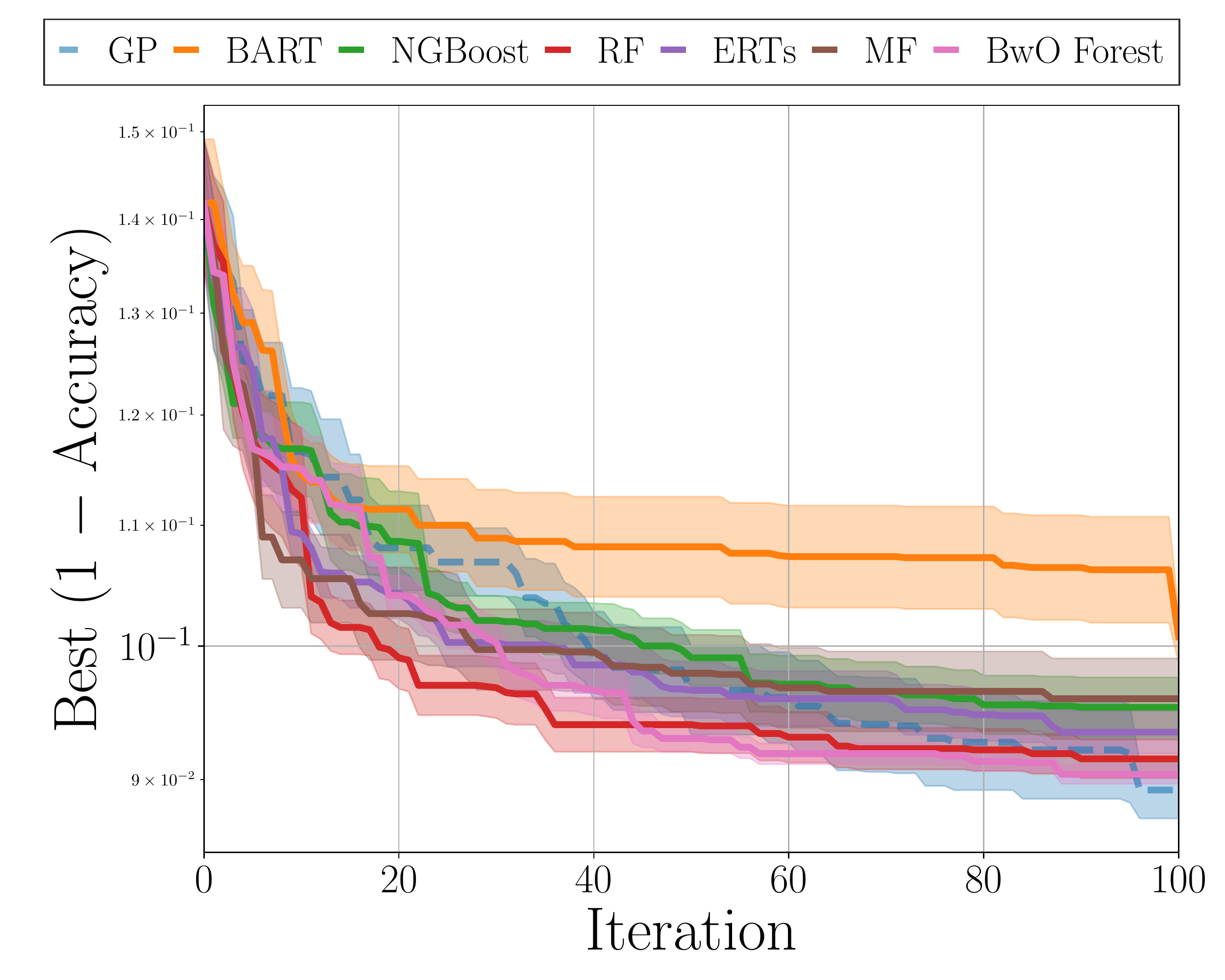}
	}
	\setcounter{subfigure}{0}
	\subfigure[Breast Cancer]{
		\includegraphics[width=0.31\textwidth]{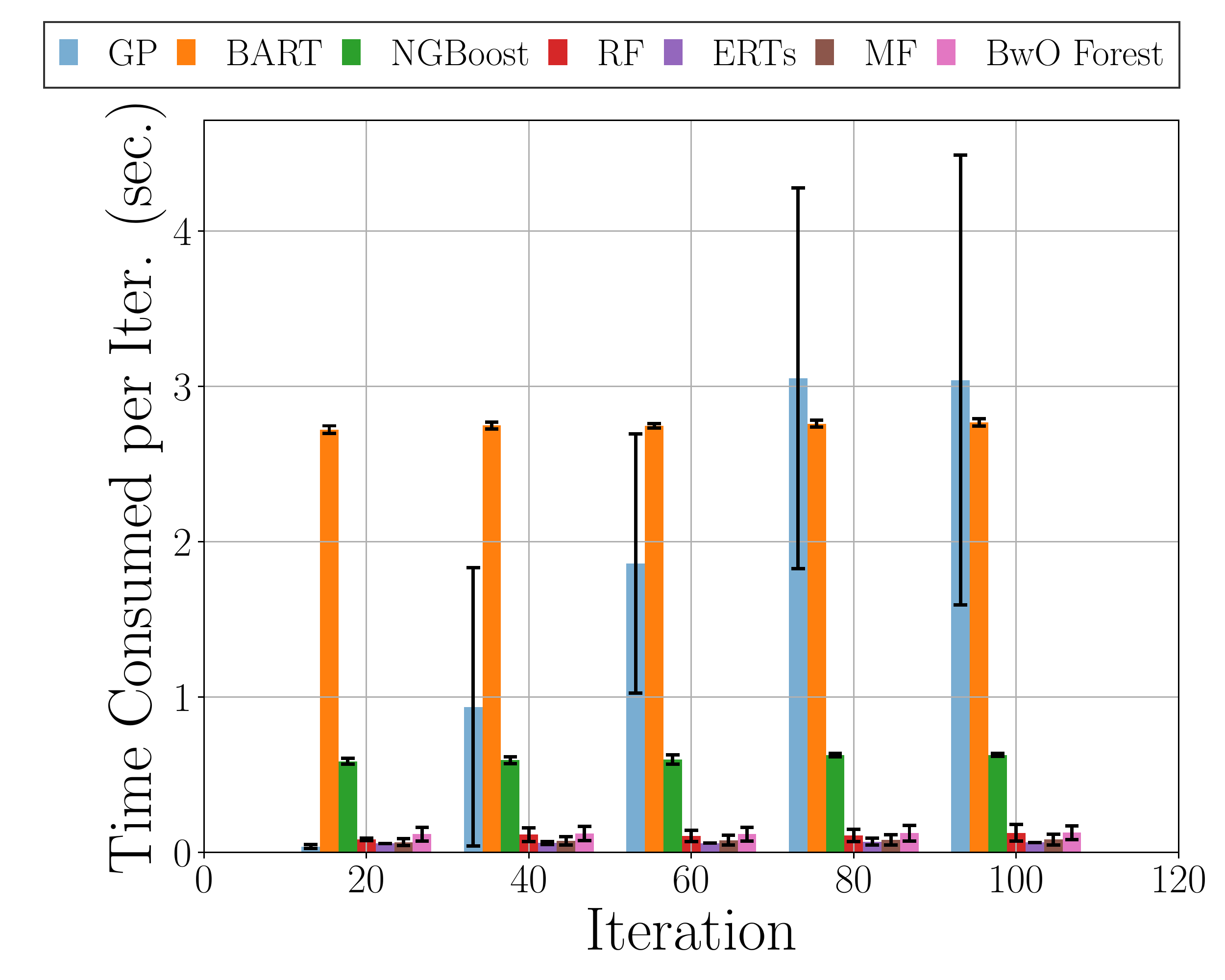}
		\label{fig:automl_2}
	}
	\subfigure[Digits]{
		\includegraphics[width=0.31\textwidth]{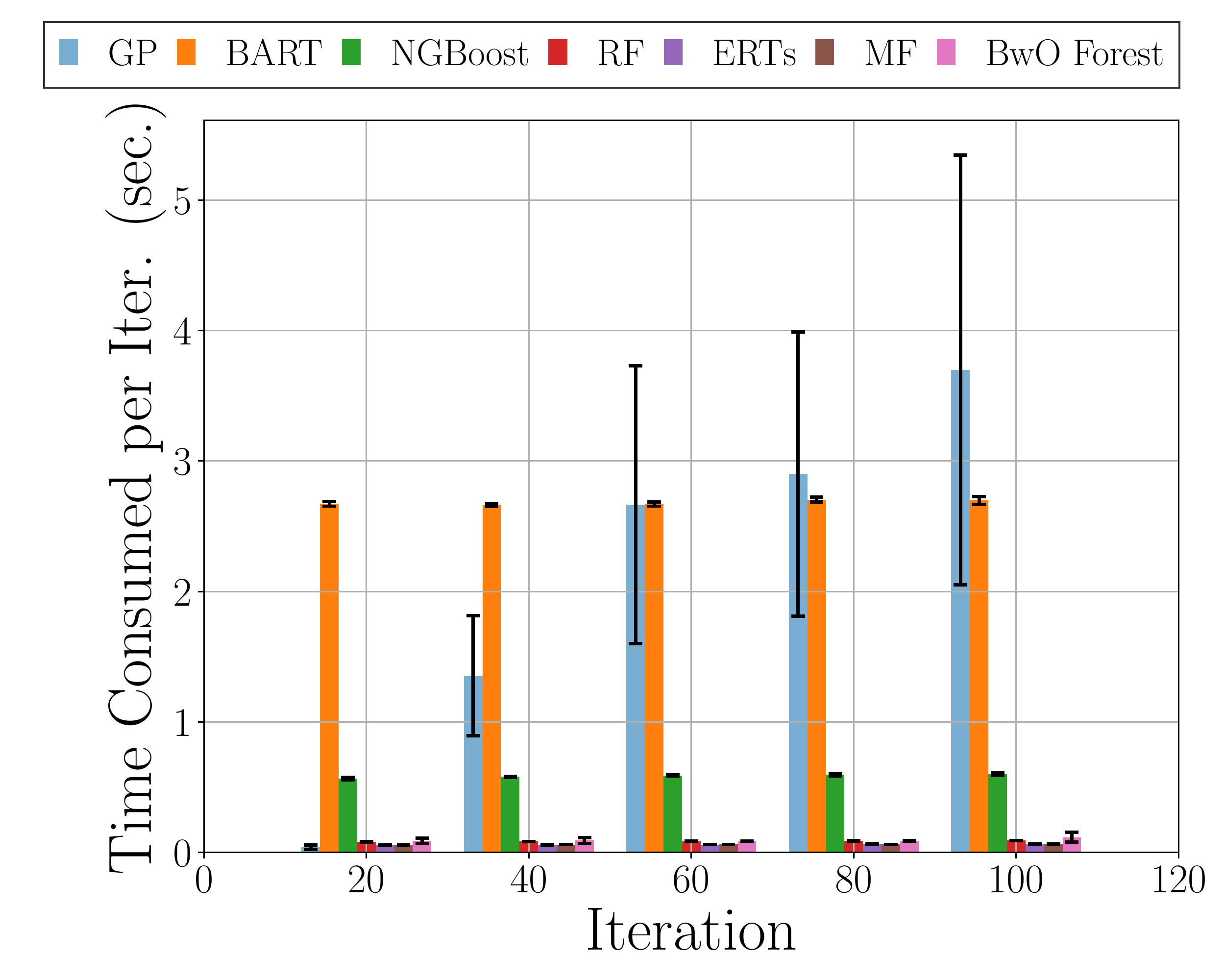}
		\label{fig:automl_3}
	}
	\subfigure[Phoneme]{
		\includegraphics[width=0.31\textwidth]{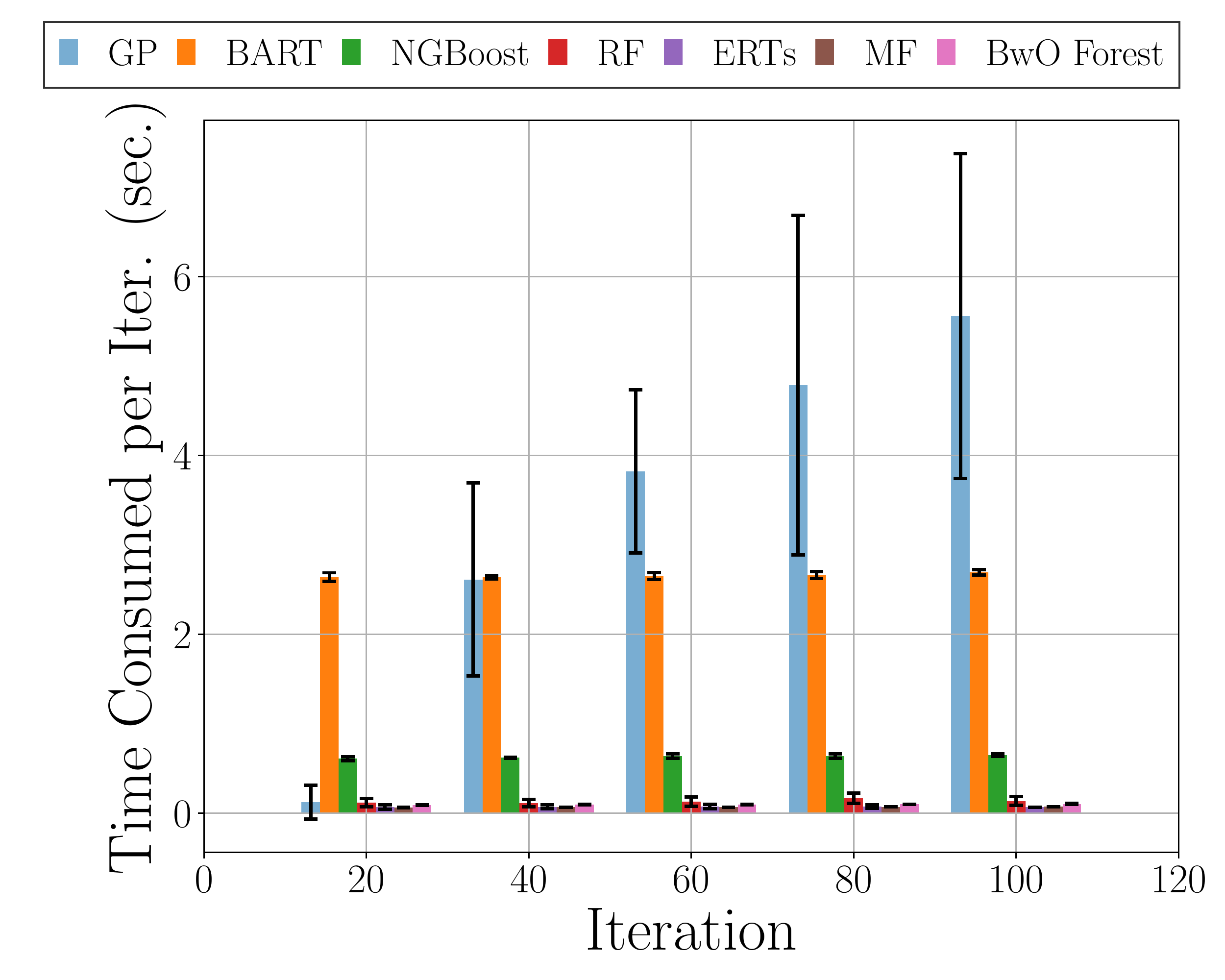}
		\label{fig:automl_4}
	}
	\vspace{-7pt}
	\caption{Results on automated machine learning for three datasets, Breast Cancer, Digits, and Phoneme. Note that the result for the Authorship dataset is presented in~\figref{fig:automl_1}. All the runs are repeated 10 times.\label{fig:automl}}
	\vspace{-7pt}
\end{figure*}

\paragraph{Experimental setup.}
We use the implementation of decision trees and ensemble methods,
included in scikit-learn~\citep{PedregosaF2011jmlr},
and employ them in the implementation of our BwO forest.
To fairly compare the results, we set the size of ensemble model $B$ as 100
and the rate of oversampling $\alpha$ as 4.
All the tree-based surrogate models employ random feature selection as the square root of the feature dimensionality.
In addition, for simplicity of BwO implementation, we duplicate a training dataset $\beta$ times,
where $\beta$ is larger than $\alpha$, and then pick a bootstrap sample that contains $\alpha / \beta$ of the duplicated 
dataset, e.g., if $\beta = 16$, a quarter of the duplicated dataset is sampled by bootstrapping for every bootstrap sample.

GP regression with Mat\'ern 5/2 kernel is used, and its hyperparameters are optimized
by marginal likelihood maximization. Because the Cholesky decomposition is applied to compute a marginal likelihood
and a posterior distribution, it is slightly faster than the vanilla GP regression model.
To focus on the tree-based surrogate models, we do not apply more sophisticated techniques 
to speed up GP regression.
Nevertheless, we include the results with the aforementioned GP regression,
in order to briefly compare it to the results with tree-based surrogate models.

For the SMO setting, we use the expected improvement~\citep{MockusJ1978tgo}
as an acquisition function. To optimize an acquisition function for tree-based surrogate models,
a fixed number of points are sampled to compute acquisition function values;
we sample 50,000 points using the Sobol' sequence.
Additionally, for GP regression, L-BFGS-B with multiple initializations is used.
Note that every set of initial points is fixed across surrogate models, so that they are started from same regret values,
and 5 initial points are given for every run.

All the computations are conducted on a system with CPU, and each experiment set 
is tested on the same machine in order to measure wall-clock time precisely.
In addition, all the missing details are described in the supplementary material.

\subsection{Continuous Search Spaces\label{susbec:continuous}}
We test popular benchmark functions such as Ackley (4 dim.), Bohachevsky (2 dim.),
Branin (2 dim.), Hartmann6D (6 dim.), Michalewicz (2 dim.), and Rosenbrock (4 dim.) functions.
As shown in~\figref{fig:exp_bo_benchmarks_1} and \figref{fig:exp_bo_benchmarks_2}, 
our method with BwO forest works well,
compared to SMO strategies with the other tree-based models.

\subsection{High-Dimensional Binary Search Spaces}

We conduct the methods studied in this work on two high-dimensional binary problems
such as Ising (24 dim.) and contamination (25 dim.) problems,
as in~the first two columns of \figref{fig:exp_binary}.
For the case of contamination problem, the SMO method with random forest regression
shows better results than other methods; see \figref{fig:exp_binary_2}.
The details of these problems are described in the supplementary material.

\subsection{Mixed Search Spaces}

\begin{table}[t]
	\centering
	\small
	\caption{Details of automated machine learning. AB, GB, DT, ET, and RF indicate AdaBoost, GradientBoosting, Decision Tree, ExtraTrees, Random Forest classifiers, respectively, and Cat. and Int. stand for categorical and integer variables.\label{tab:hyps}}
	\vspace{7pt}
	\begin{tabular}{ccc}
		\toprule
		\textbf{Hyperparam.} & \textbf{Range} & \textbf{Type} \\
		\midrule
		Algorithm type & \{AB, GB, DT, ET, RF\} & Cat. \\
		Ensemble size & [10, 200] & Int. \\
		Max. depth & [2, 8] & Int. \\
		Max. features & [0.2, 1.0] & Float \\
		$\log$ learning rate & [-3, 1] & Float \\
		\bottomrule
	\end{tabular}
\end{table}

We carry out the experiments of automated machine learning, which are defined on a mixed search space; 
see \tabref{tab:hyps} for the detailed description of search space.
We choose one of machine learning algorithms through a categorical variable 
and simultaneously tune their numerical or ordinal hyperparameters using SMO.
To optimize a categorical variable, we employ a one-hot encoding, which can be defined as a simplex.
Four datasets such as Authorship, Breast Cancer, Digits, and Phoneme are used to train and test 
an automated machine learning model.
The tendency of the results by our method is better than the other methods,
as presented in~\figref{fig:automl_1} and \figref{fig:automl}.

\section{DISCUSSION AND LIMITATIONS\label{sec:discussion}}

Here, we provide the discussion on tree-based models as well as BwO forest, 
which is about oversampling, tree construction techniques, epistemic uncertainty, and cheap required computation.
Furthermore, we introduce a future direction of this research to resolve the limitations of tree-based surrogate models,
which are related to the choice of tree-based surrogate models and extrapolation.

\paragraph{Oversampling.}
While this technique is widely used in solving an imbalanced data
problem~\citep{YapBW2014daeng,PerezM2015ieeennls,YanY2019aaai},
it is not popular for bagging.
Since it samples a bootstrap sample from an original set of data points $\bX$,
it increases the number of unique original elements as described in~\secref{sec:elaborating},
but it is a solid method to construct a bootstrap sample.
To sum up, bagging with oversampling prevents a surrogate model underfitting to $\bX$.
Moreover, it is more effective in the case that $N$ is relatively small, which is common in SMO,
than the case $N$ is large, as in~\figref{fig:unc_1d_few}, \figref{fig:unc_1d_many}, and \tabref{tab:qual}.

\begin{figure}[t]
	\centering
	\subfigure[B + O]{
		\includegraphics[width=0.22\textwidth]{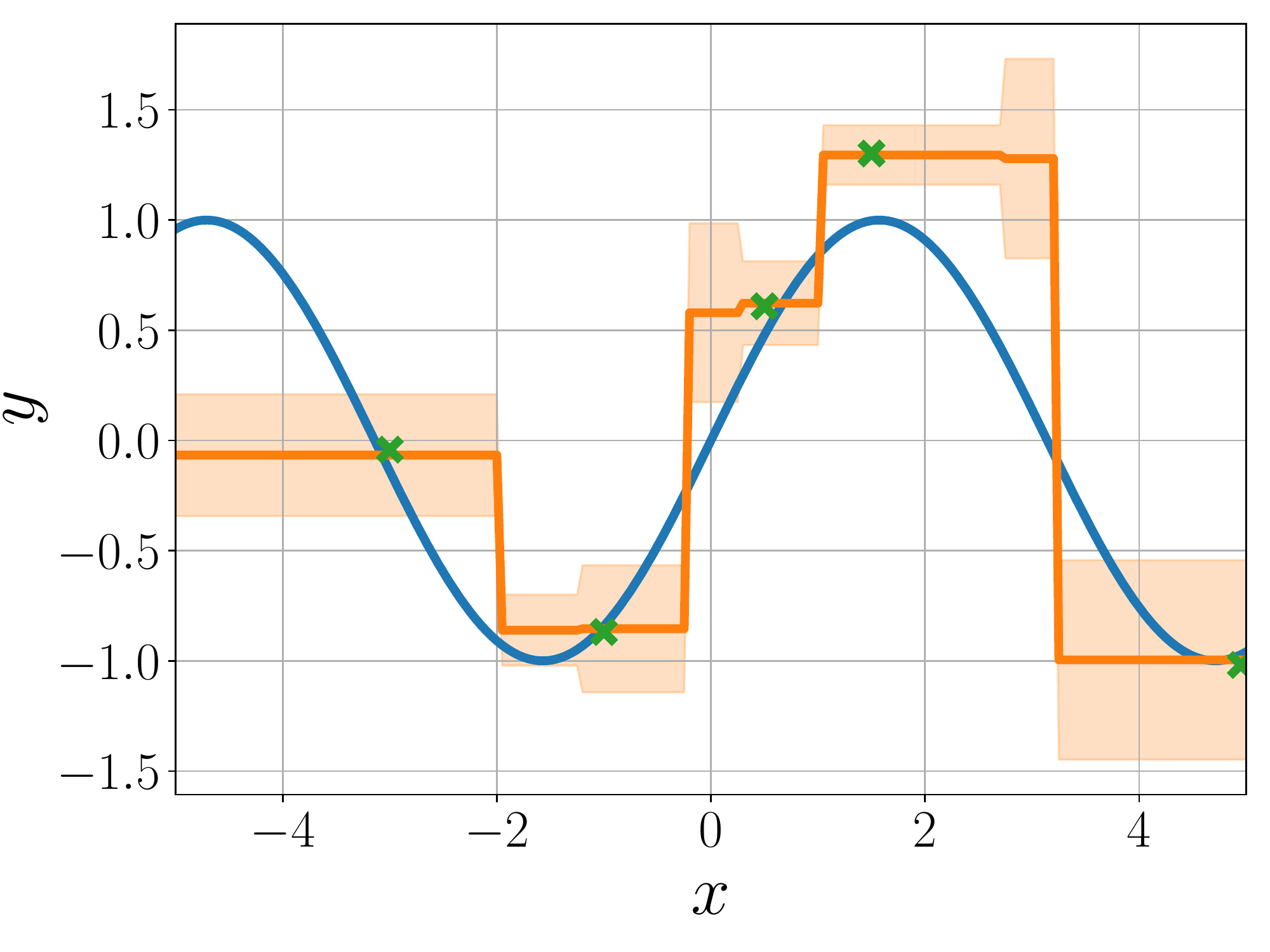}
		\label{fig:ablation_bwo_only}
	}
	\subfigure[R]{
		\includegraphics[width=0.22\textwidth]{unc_1d_few_ext.pdf}
		\label{fig:ablation_ext}
	}
	\subfigure[R + B]{
		\includegraphics[width=0.22\textwidth]{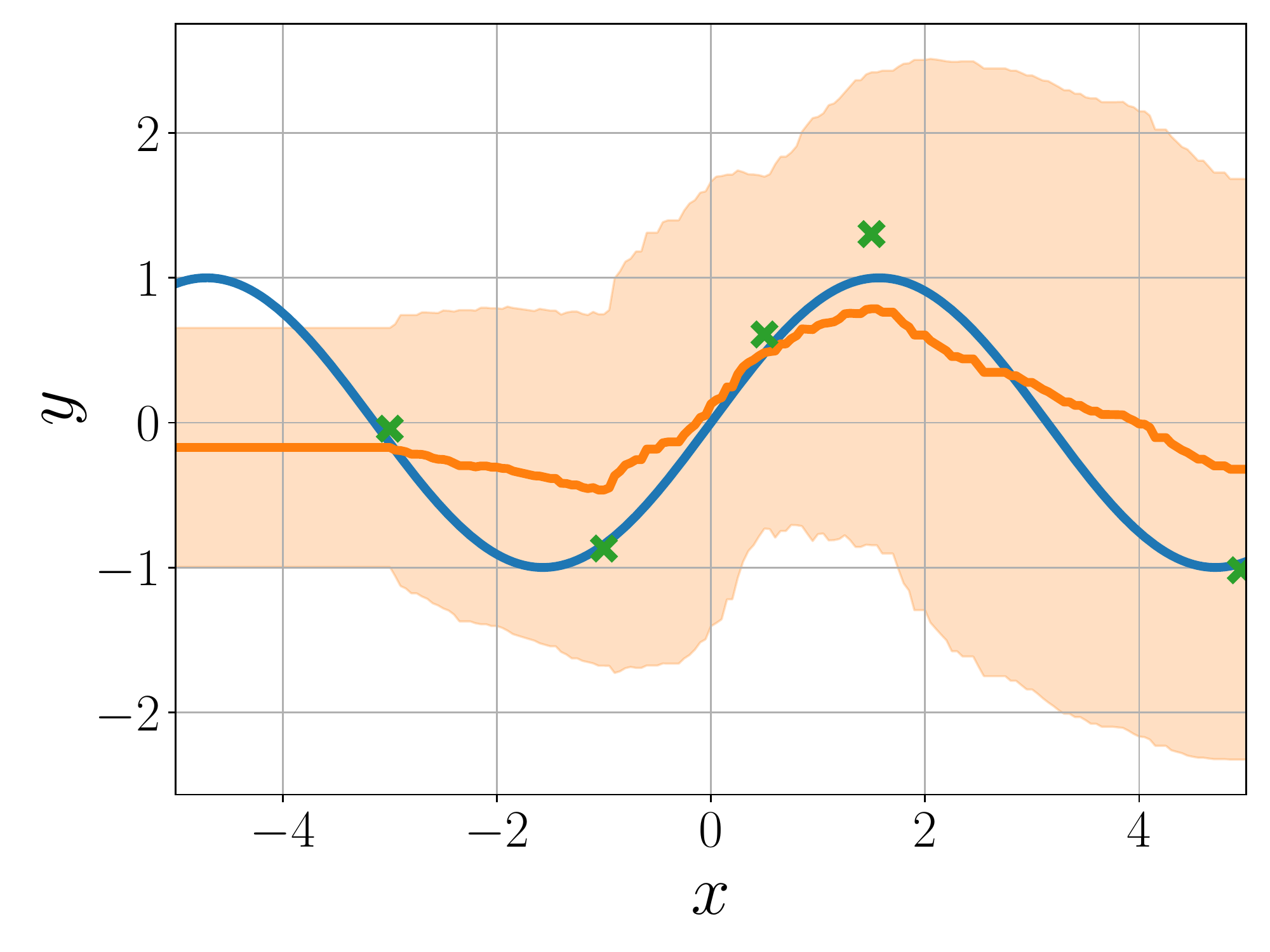}
		\label{fig:ablation_ext_bs}
	}
	\subfigure[R + B + O (Ours)]{
		\includegraphics[width=0.22\textwidth]{unc_1d_few_ours.pdf}
		\label{fig:ablation_ours}
	}
	\vspace{-7pt}
	\caption{Results on tree construction techniques.
	B, O, and R indicate bagging, oversampling, and random split location, respectively.	
	The case of only B is shown in~\figref{fig:unc_1d_few_rf} as the result by random forest.\label{fig:unc_1d_ablation}}
	\vspace{-7pt}
\end{figure}

\paragraph{Tree construction techniques.}
We show the effects of tree construction techniques
such as bagging (denoted as B), oversampling (denoted as O), and random sampling of split location (denoted as R),
as presented in~\figref{fig:unc_1d_ablation}.
The prediction uncertainty estimation by B + O is abnormal compared to other results, because it overfits to the duplicated training data and split locations are not properly determined.
The R + B result is similar to the result by Mondrian forest, but it tends to underfit to training data.
The result in \figref{fig:ablation_ext} looks similar to \figref{fig:ablation_ours}, 
however the uncertainty on the region out of the range of training data goes to zero.
We provide more diverse results by individual trees in \figref{fig:unc_1d_few_trees},
\figref{fig:unc_1d_many_trees}, and \figref{fig:unc_1d_cubic_trees}.

\paragraph{Qualitative analysis on regression.}

\begin{table}[t]
	\centering
	\small
	\caption{Kullback–Leibler divergence from GP to respective results by tree-based surrogate models.\label{tab:qual}}
	\vspace{7pt}
	\begin{tabular}{cccc}
		\toprule
		\textbf{Method} & \textbf{\figref{fig:unc_1d_few}} & \textbf{\figref{fig:unc_1d_many}} & \textbf{\figref{fig:unc_1d_cubic}} \\
		\midrule
		RF & 1.38 & \textbf{0.00} & \textbf{0.00} \\
		ERTs & \underline{0.06} & 0.07 & 0.09 \\
		BART & 0.93 & \underline{0.01} & \underline{0.01} \\
		MF & 0.93 & \underline{0.01} & \textbf{0.00} \\
		NGBoost & 0.21 & 0.05 & \textbf{0.00} \\
		BwO forest & \textbf{0.04} & \underline{0.01} & \underline{0.01} \\
		\bottomrule
	\end{tabular}
\end{table}

To compare regression results qualitatively, 
we measure Kullback–Leibler divergence 
from GP to a result by each tree-based surrogate model, 
which shows how similar a result is to GP.
As shown in~\tabref{tab:qual}, 
our BwO forest tends to follow the GP appropriately, 
compared to random forest, extremely randomized trees, BART, Mondrian forest, and NGBoost.

\paragraph{Epistemic uncertainty by tree-based surrogate models.}
Epistemic uncertainty is reducible if we collect more data or identify a model well~\citep{GalY2016phd}.
This type of uncertainty is induced from the randomness of model or bootstrapping.
Thus, our BwO forest can be considered that the randomness of model
is maximized with the tree construction techniques such as
random feature selection, random selection of split locations as well as bagging with oversampling.
Especially, the bootstrap sample constructed by our technique, 
bagging with oversampling, contains more unique original elements to enhance the degree
of fitting to training data than a standard bagging method.

\paragraph{Cheap required computation.}
In general, tree-based surrogate models are more time-consuming in predicting an output
than other types of estimators, because it has to compute the outputs of all the base
estimators. However, fortunately the setup of SMO does
not assume a large number of data points, which implies
that the size of ensemble model can be maintained as a small size.
On the other hand, as shown in~\secref{sec:experiments},
tree-based models are consistently faster than GP regression
in many cases, and they are able to be directly applied to speed up the overall procedure of SMO
with comparable performance.

\paragraph{Choice of tree-based surrogate models.}
Inevitably, a specific SMO method successfully optimizes some specific functions,
and simultaneously fails to optimize other functions, depending on the characteristics of target functions.
As shown in~\secref{sec:experiments}, our method is robust in many cases,
but either random forest or Mondrian forest can be a good option in some cases, e.g., \figref{fig:exp_bo_michalewicz} and \figref{fig:exp_binary_2}.

\paragraph{Extrapolation.}
Lastly, tree-based surrogate models are vulnerable in predicting a function value and its corresponding uncertainty
out of the range of training data. At least, even if a function estimate is not correct,
an uncertainty estimate should become larger than the results shown in the paper.
In order to expand the usage of tree-based models in SMO,
improving the ability to extrapolate is left to a future work.
According to our preliminary experiments, injecting a noise in the duplicates is likely to improve the extrapolation ability.

\section{RELATED WORK\label{sec:related}}

From now,
we briefly review tree-based estimators and SMO 
with tree-based surrogate models.

\paragraph{Tree-based estimators.}
An estimator that aggregates a set of decision trees is attractive to many machine learning practitioners
in both classification
and regression tasks, since it shows reliable performance despite its relatively efficient training
and test procedures~\citep{BiauG2012jmlr,NatekinA2013fin,LouppeG2014arxiv}. In particular,
compared to a deep neural network, such a randomized tree-based approach has a practical strength
in real-world problems,
which is shown by the popularity of gradient boosting machines such as
XGBoost~\citep{ChenT2016kdd}
and LightGBM~\citep{KeG2017neurips}.
As mentioned in the previous sections,
\citet{BreimanL1996ml,BreimanL2001ml,FriedmanJH2001aos,GeurtsP2006ml,ChipmanHA2010aas,LakshminarayananB2014neurips,LakshminarayananB2016aistats,DuanT2020icml}
have proposed the foundations and theories of the approaches used in this work;
see the work by~\citet{DietterichTG2000mcs,ZhouZH2012book,LouppeG2014arxiv} for the details.
Moreover, while we omit it in this work, \citet{MentchL2016jmlr} suggest
a method to estimate uncertainties using subbagging for random forest.

\paragraph{Sequential model-based optimization with tree-based surrogate models.}
\citet{HutterF2011lion} propose SMO with random forest
by applying itself in optimizing an algorithm configuration.
This method is widely adopted in many applications~\citep{FeurerM2015neurips,CandelieriA2018jgo,YingC2019icml}
including Auto-sklearn~\citep{FeurerM2020arxiv}.
Importantly, it is beneficial in certain circumstances presented and tested in this paper.

\section{CONCLUSION\label{sec:conclusion}}

We re-examined sequential random forest-based optimization and suggested our
methods defined with diverse randomized tree-based surrogate models including 
extremely randomized trees, BART, Mondrian forest, and NGBoost.
Then, we proposed a new tree-based surrogate model, named BwO forest, which uses an ensemble construction
technique, bagging with oversampling. The empirical analyses on such methods help us 
to understand the tree-based models thoroughly 
and provide a future research direction of SMO with tree-based surrogate models.

\bibliography{kjt}

\clearpage
\appendix

\thispagestyle{empty}

\onecolumn \makesupplementtitle

In this material, we describe the examples and contents that
are missing in the main article.

\section{1D EXAMPLES}

Two 1D examples: (i) $y = \sin(x) + \epsilon$, (ii) $y = x^3 + \epsilon$,
where $\epsilon$ is an observation noise,
are demonstrated.

\begin{figure}[ht]
	\centering
	\subfigure[Gaussian process]{
		\includegraphics[width=0.30\textwidth]{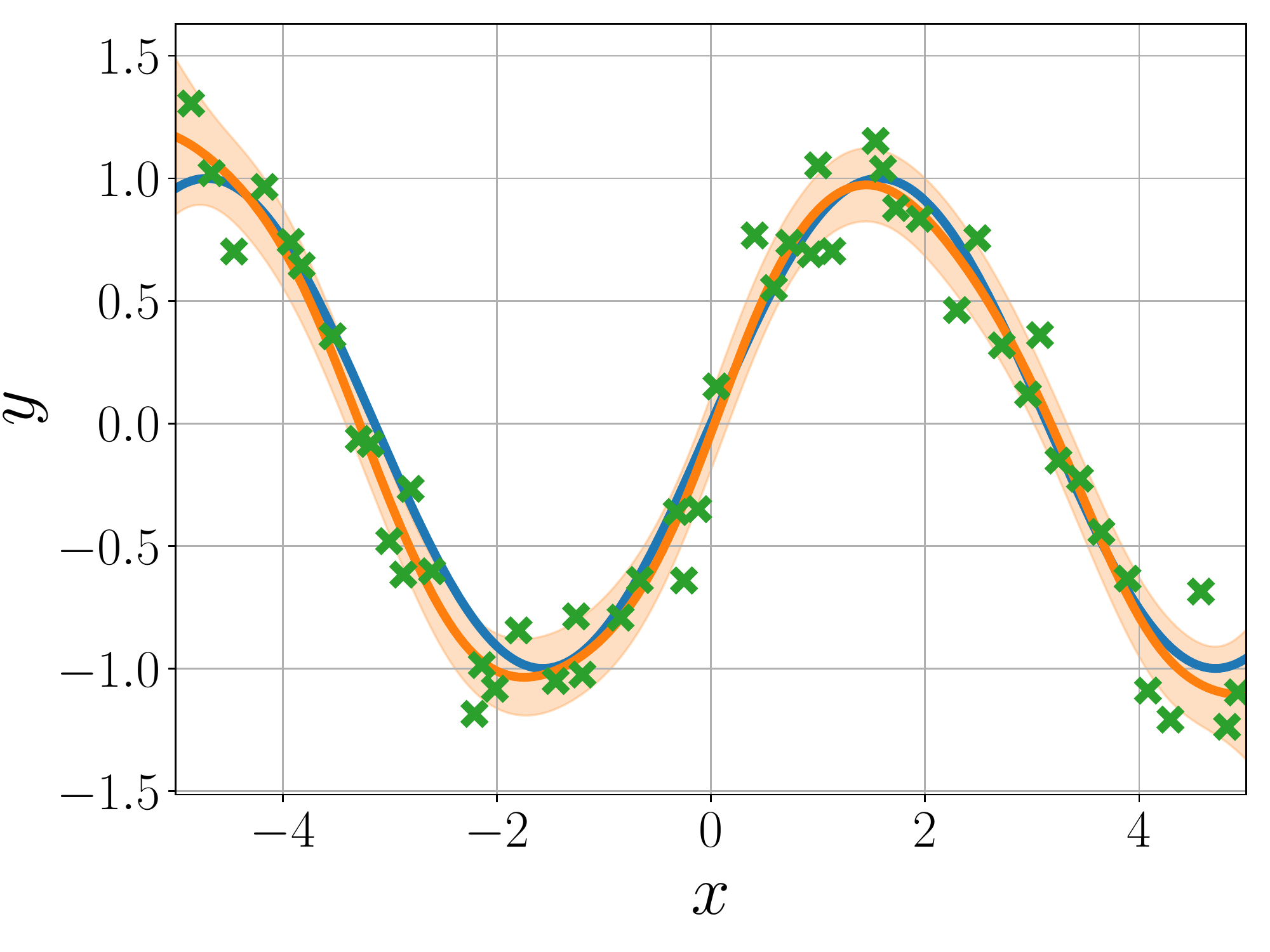}
		\label{fig:unc_1d_many_gp}
	}
	\subfigure[Random forest]{
		\includegraphics[width=0.30\textwidth]{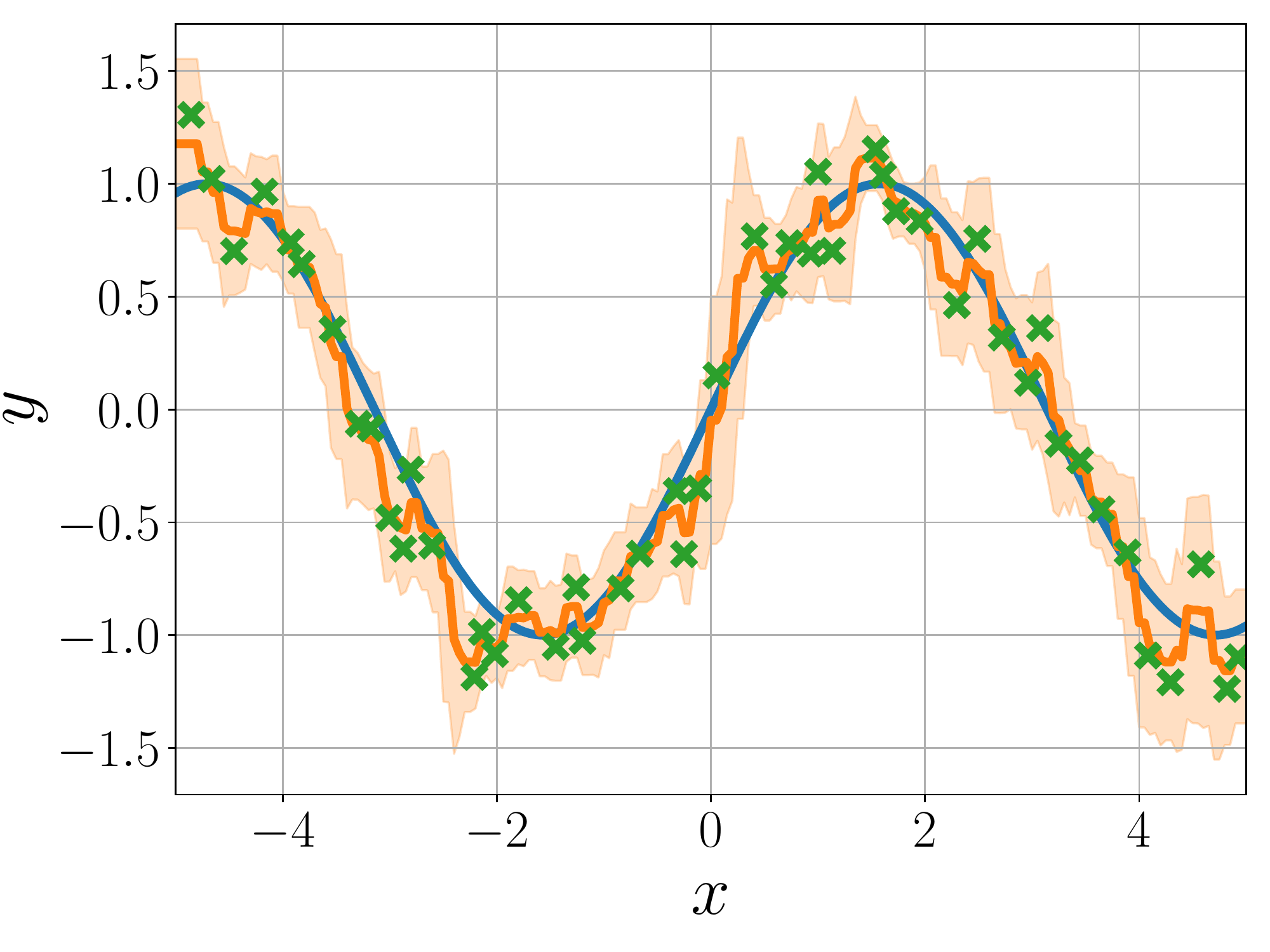}
		\label{fig:unc_1d_many_rf}
	}
	\subfigure[Extremely randomized trees]{
		\includegraphics[width=0.30\textwidth]{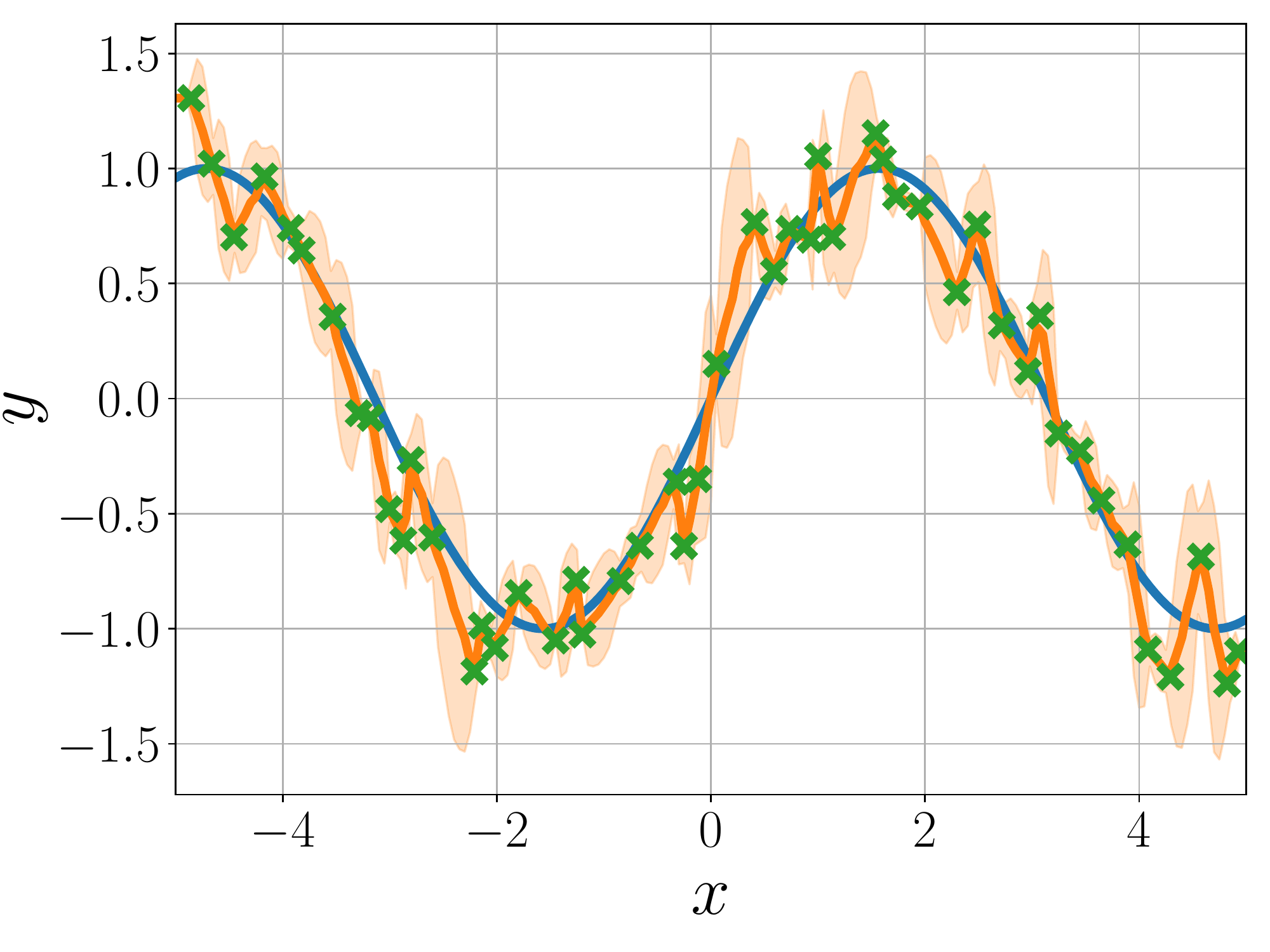}
		\label{fig:unc_1d_many_r}
	}
	\subfigure[BART]{
		\includegraphics[width=0.30\textwidth]{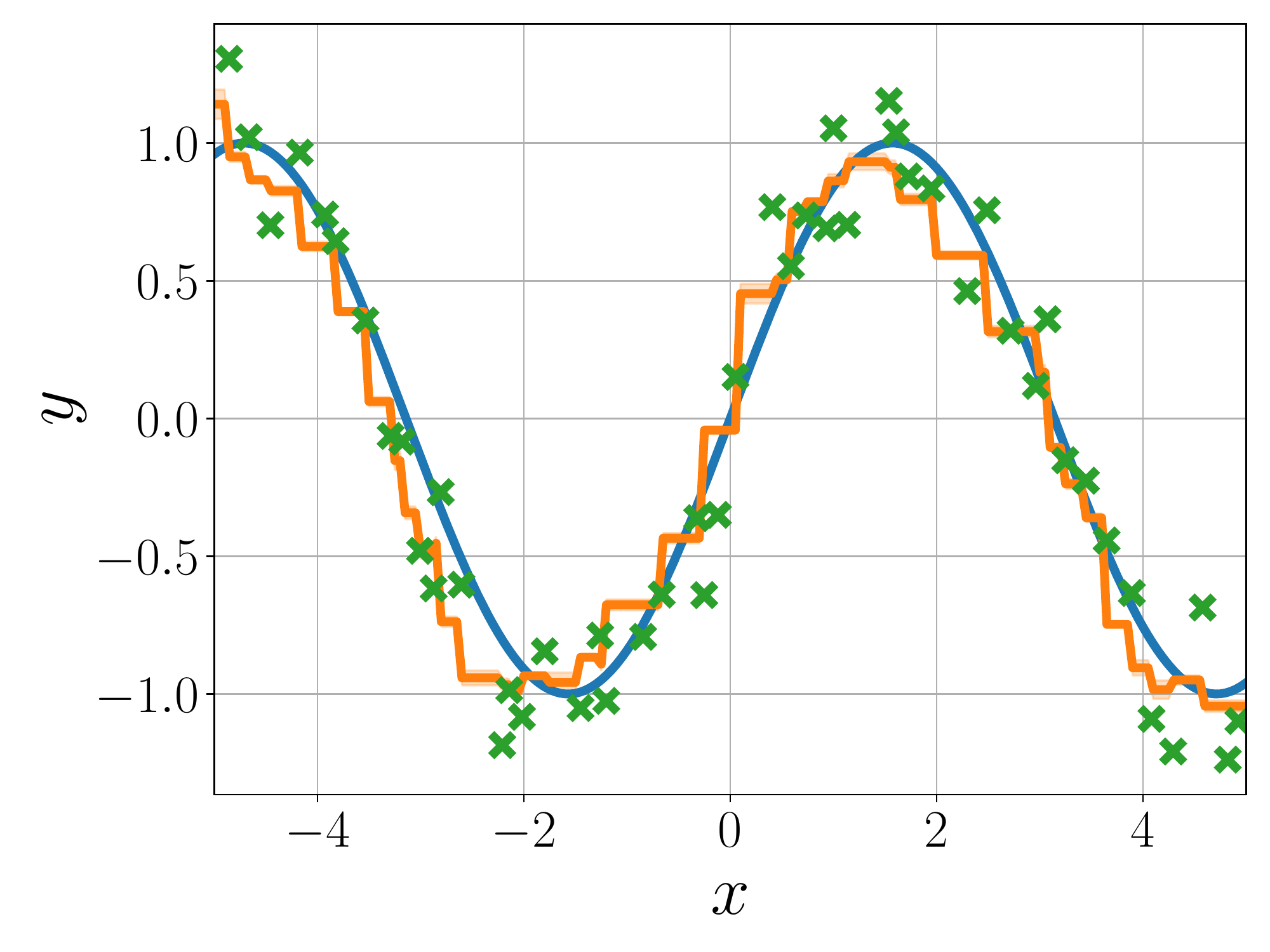}
		\label{fig:unc_1d_many_bart}
	}
	\subfigure[Mondrian forest]{
		\includegraphics[width=0.30\textwidth]{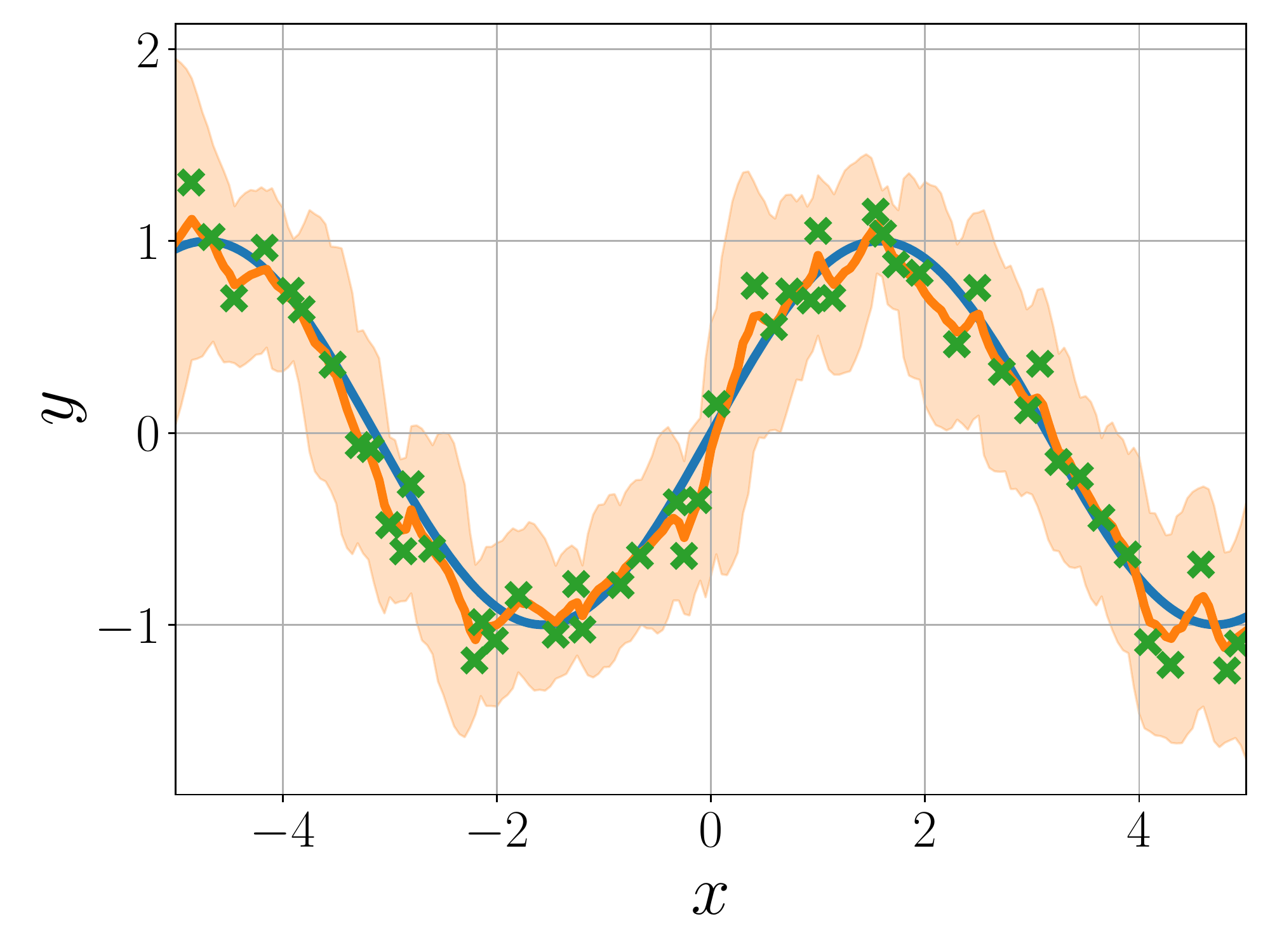}
		\label{fig:unc_1d_many_mf}
	}
	\subfigure[NGBoost]{
		\includegraphics[width=0.30\textwidth]{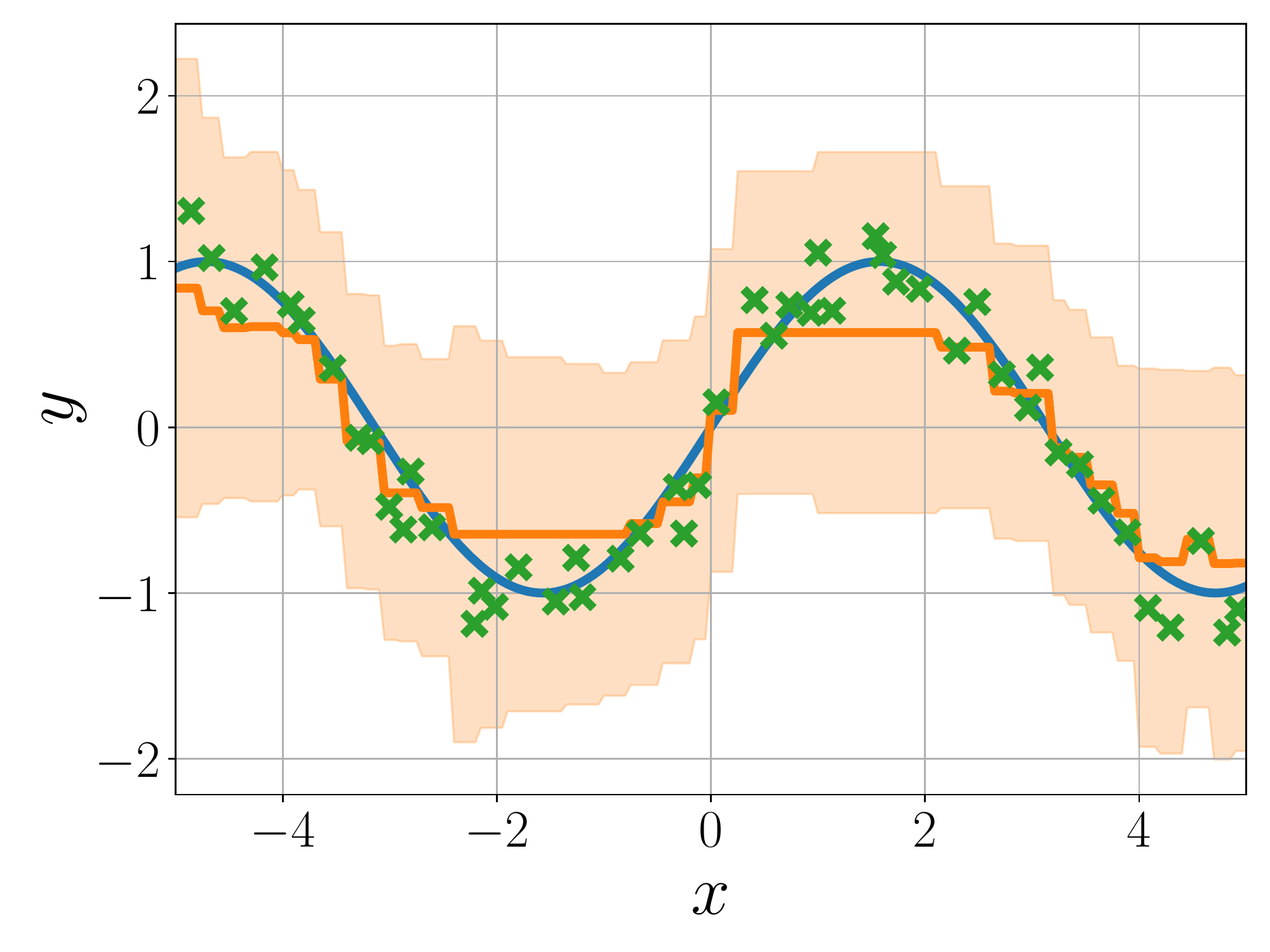}
		\label{fig:unc_1d_many_ngboost}
	}
	\subfigure[B + O]{
		\includegraphics[width=0.30\textwidth]{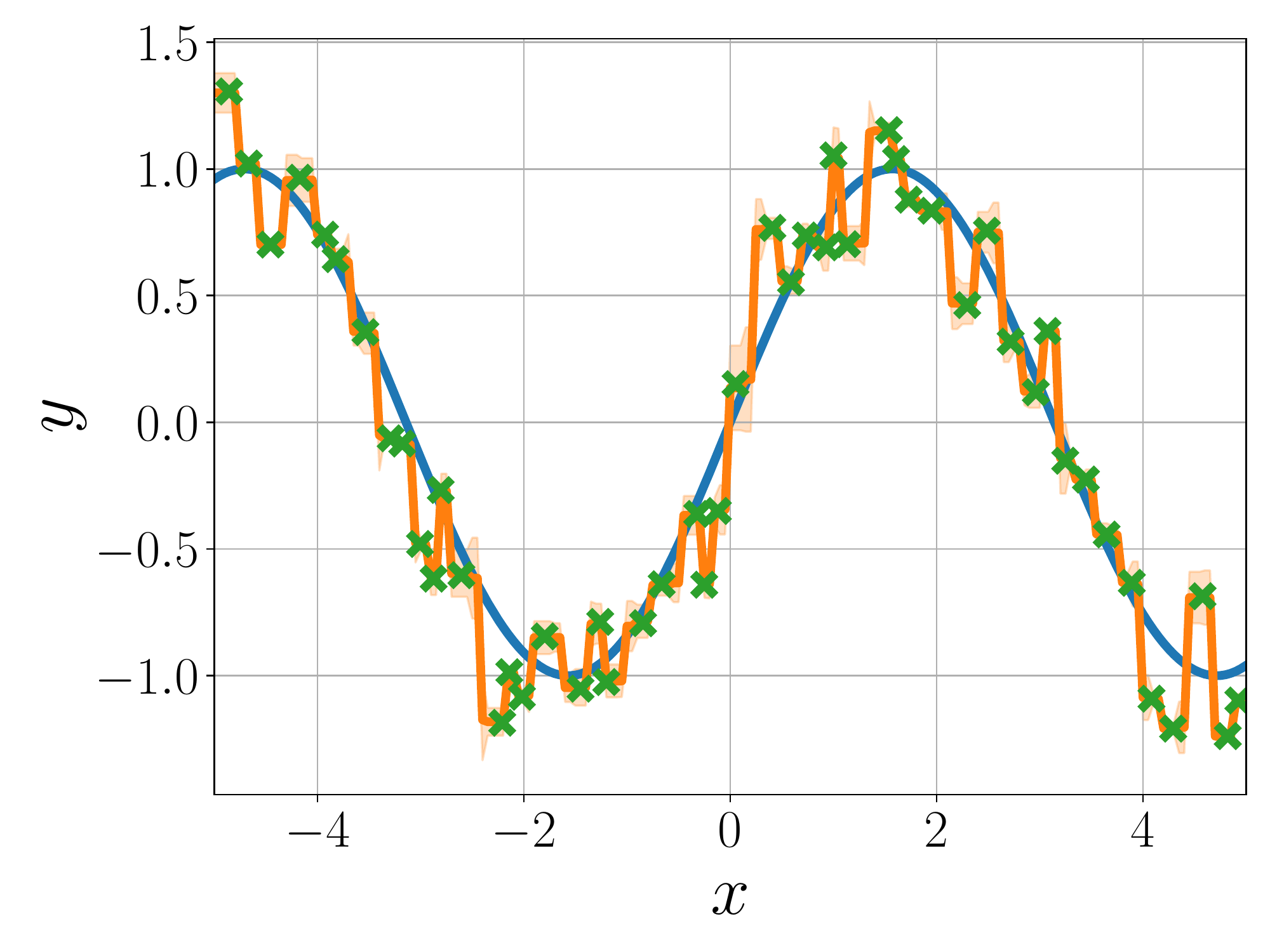}
		\label{fig:unc_1d_many_bo}
	}
	\subfigure[R + B]{
		\includegraphics[width=0.30\textwidth]{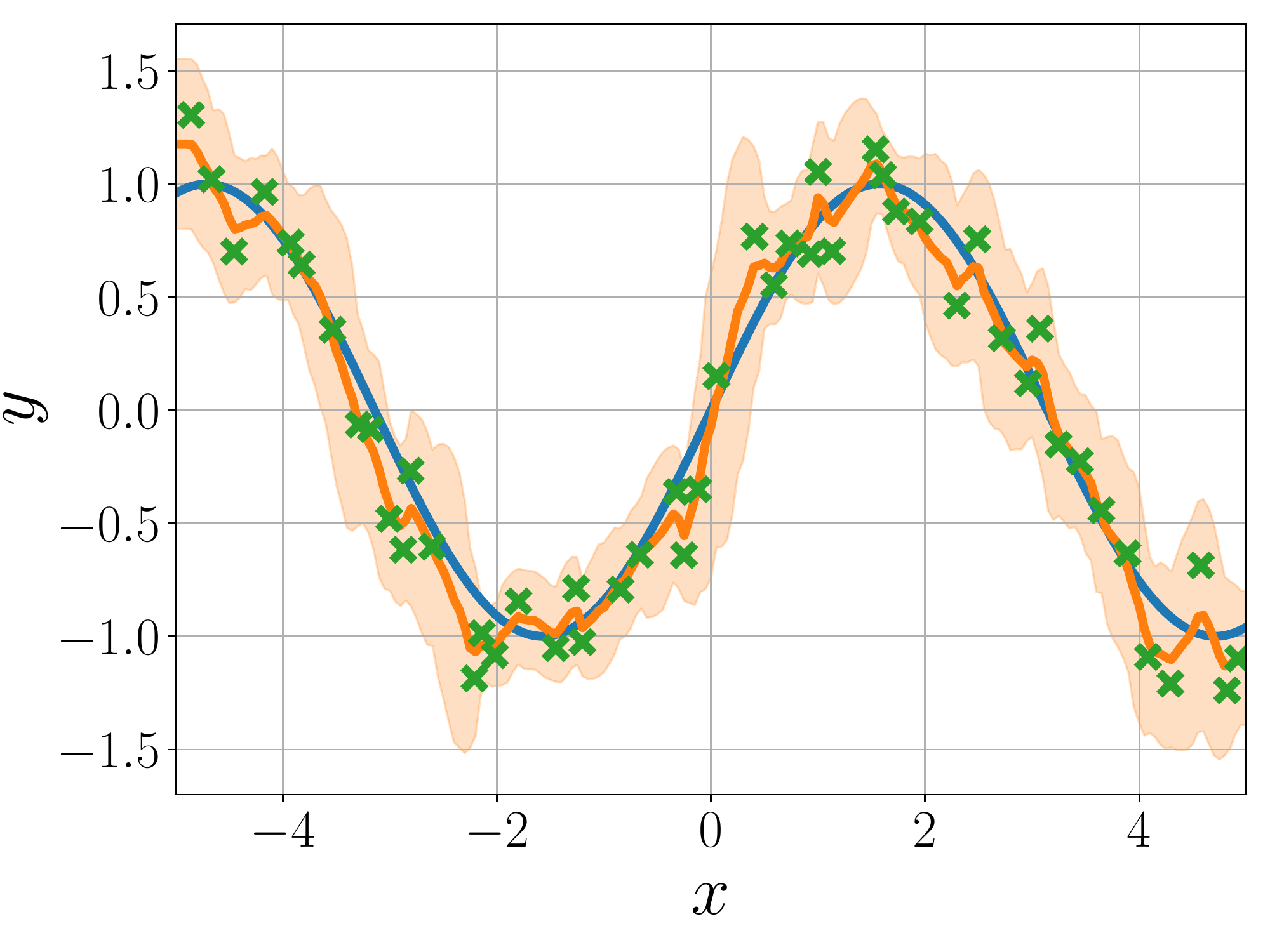}
		\label{fig:unc_1d_many_rb}
	}
	\subfigure[BwO forest (Ours)]{
		\includegraphics[width=0.30\textwidth]{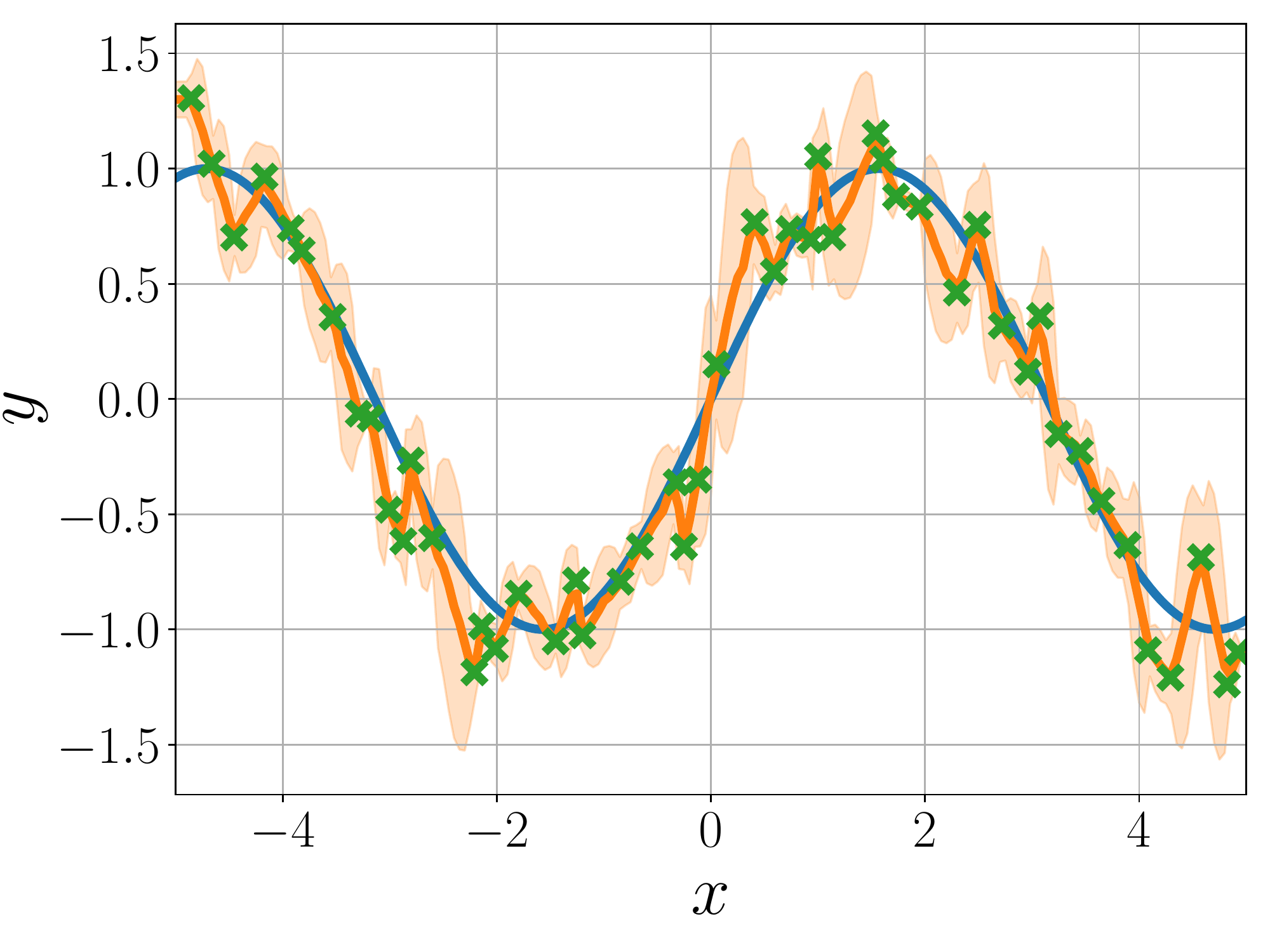}
		\label{fig:unc_1d_many_ours}
	}
	\caption{Examples on the sine with 50 points.\label{fig:unc_1d_many}}
\end{figure}

\begin{figure}[t]
	\centering
	\subfigure[Gaussian process]{
		\includegraphics[width=0.30\textwidth]{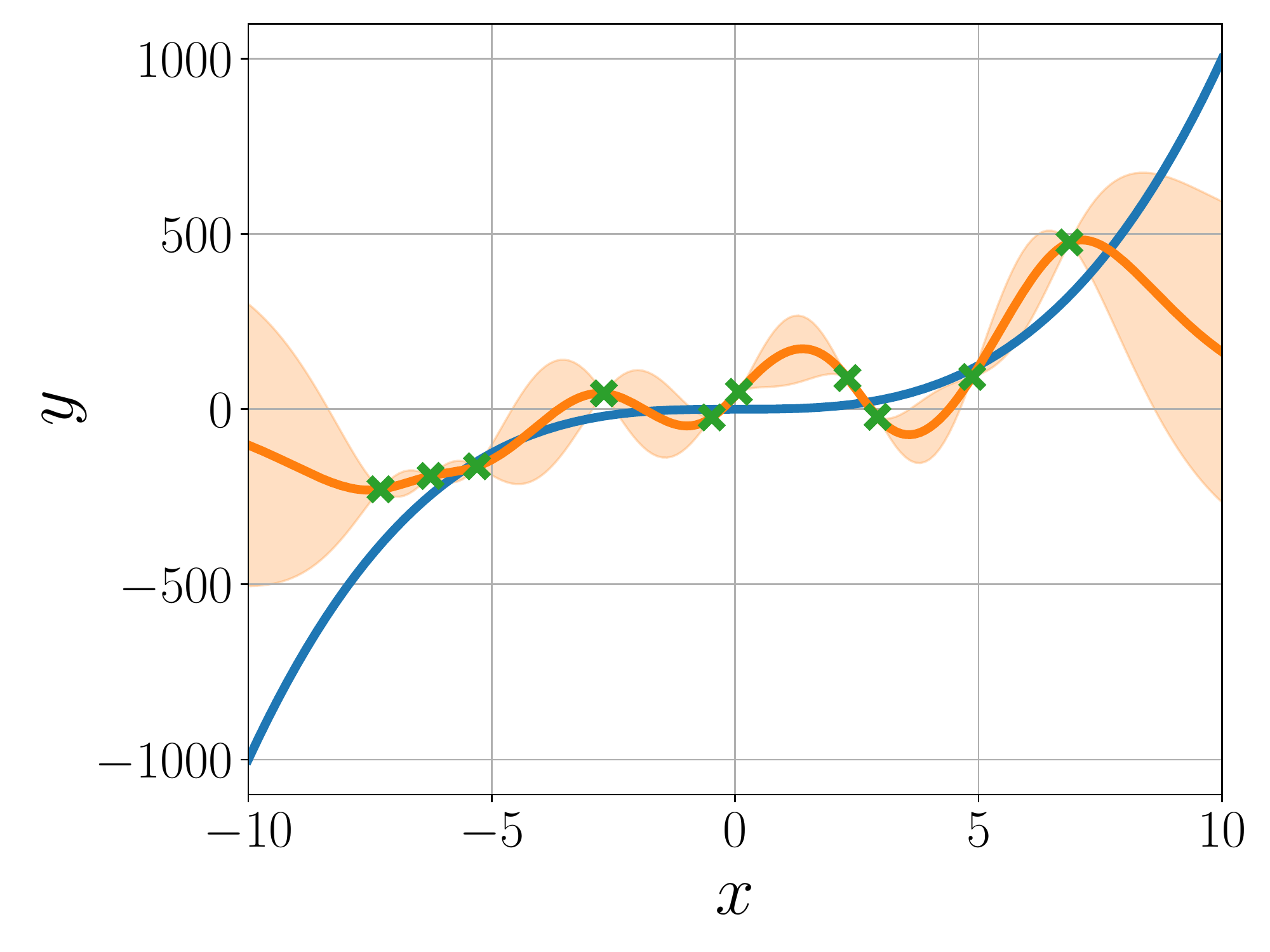}
		\label{fig:unc_1d_cubic_gp}
	}
	\subfigure[Random forest]{
		\includegraphics[width=0.30\textwidth]{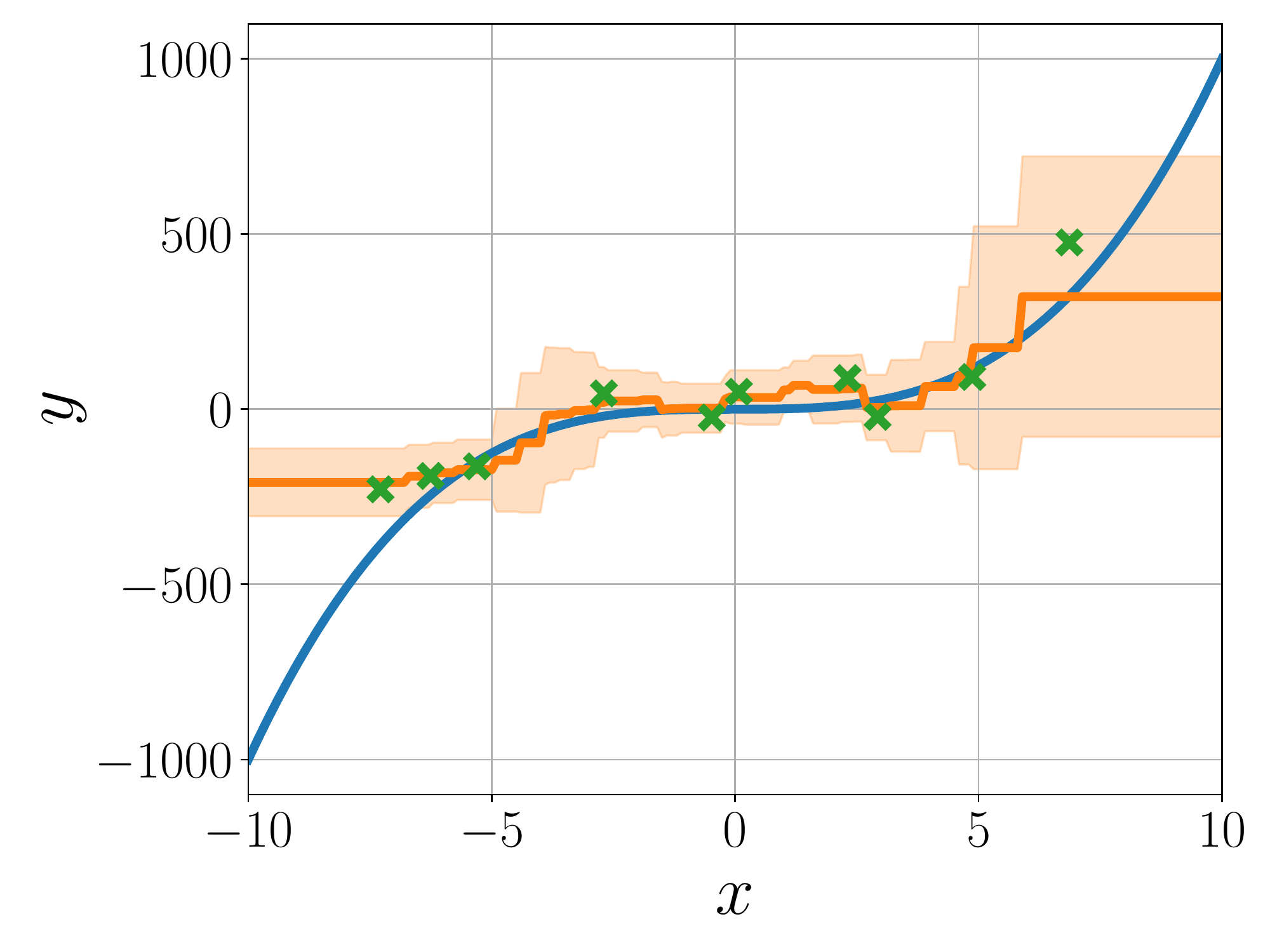}
		\label{fig:unc_1d_cubic_rf}
	}
	\subfigure[Extremely randomized trees]{
		\includegraphics[width=0.30\textwidth]{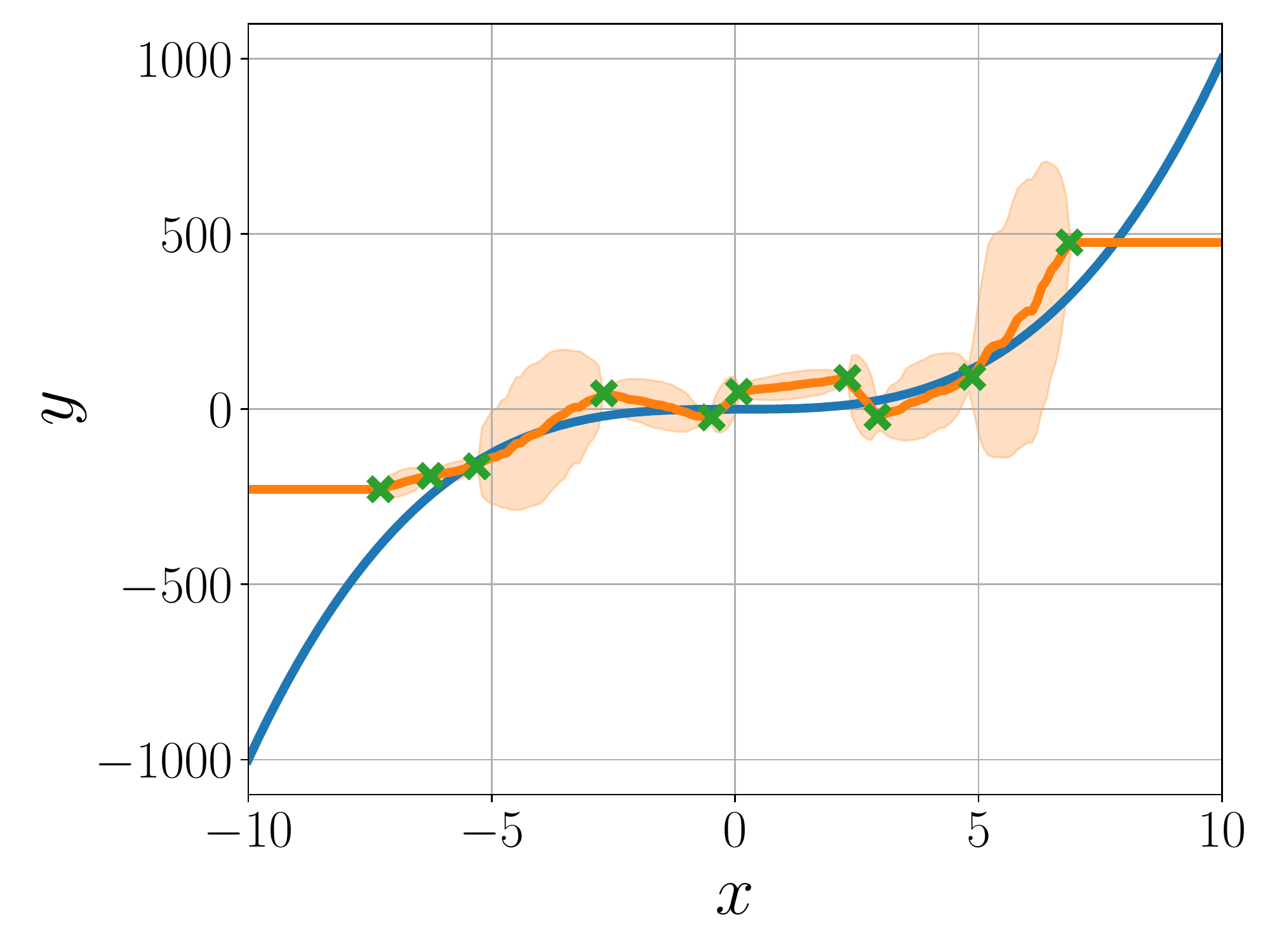}
		\label{fig:unc_1d_cubic_r}
	}
	\subfigure[BART]{
		\includegraphics[width=0.30\textwidth]{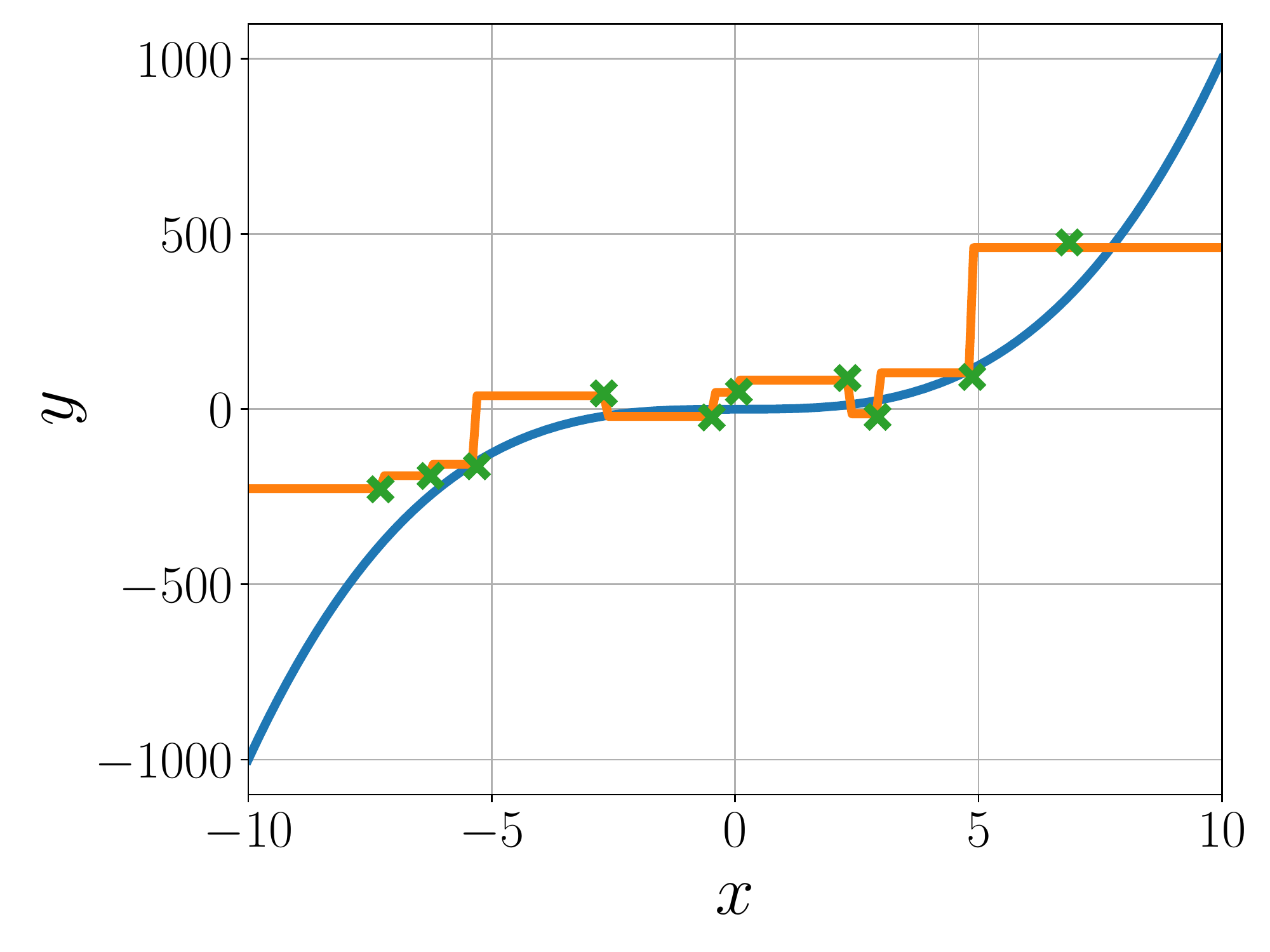}
		\label{fig:unc_1d_cubic_bart}
	}
	\subfigure[Mondrian forest]{
		\includegraphics[width=0.30\textwidth]{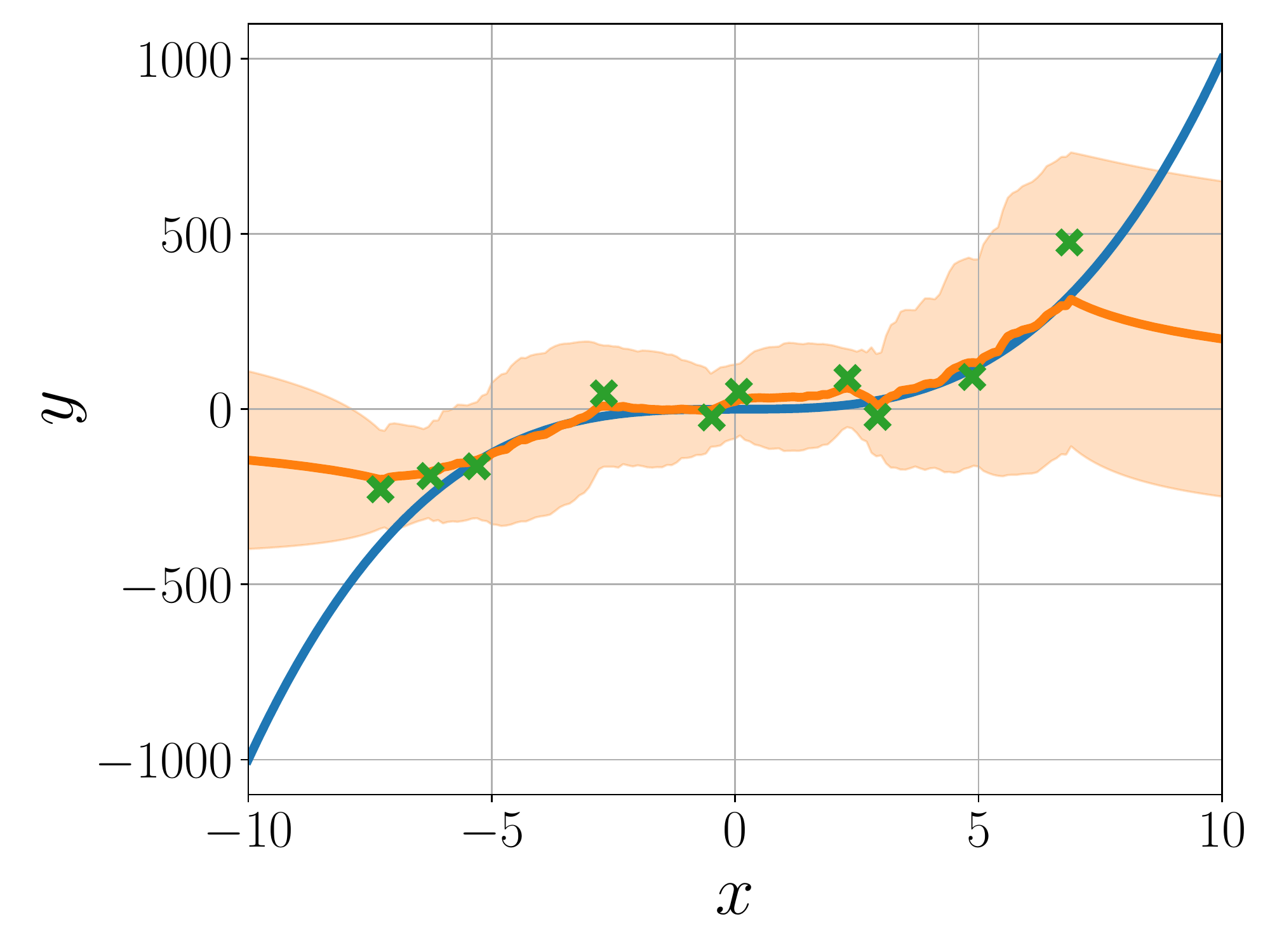}
		\label{fig:unc_1d_cubic_mf}
	}
	\subfigure[NGBoost]{
		\includegraphics[width=0.30\textwidth]{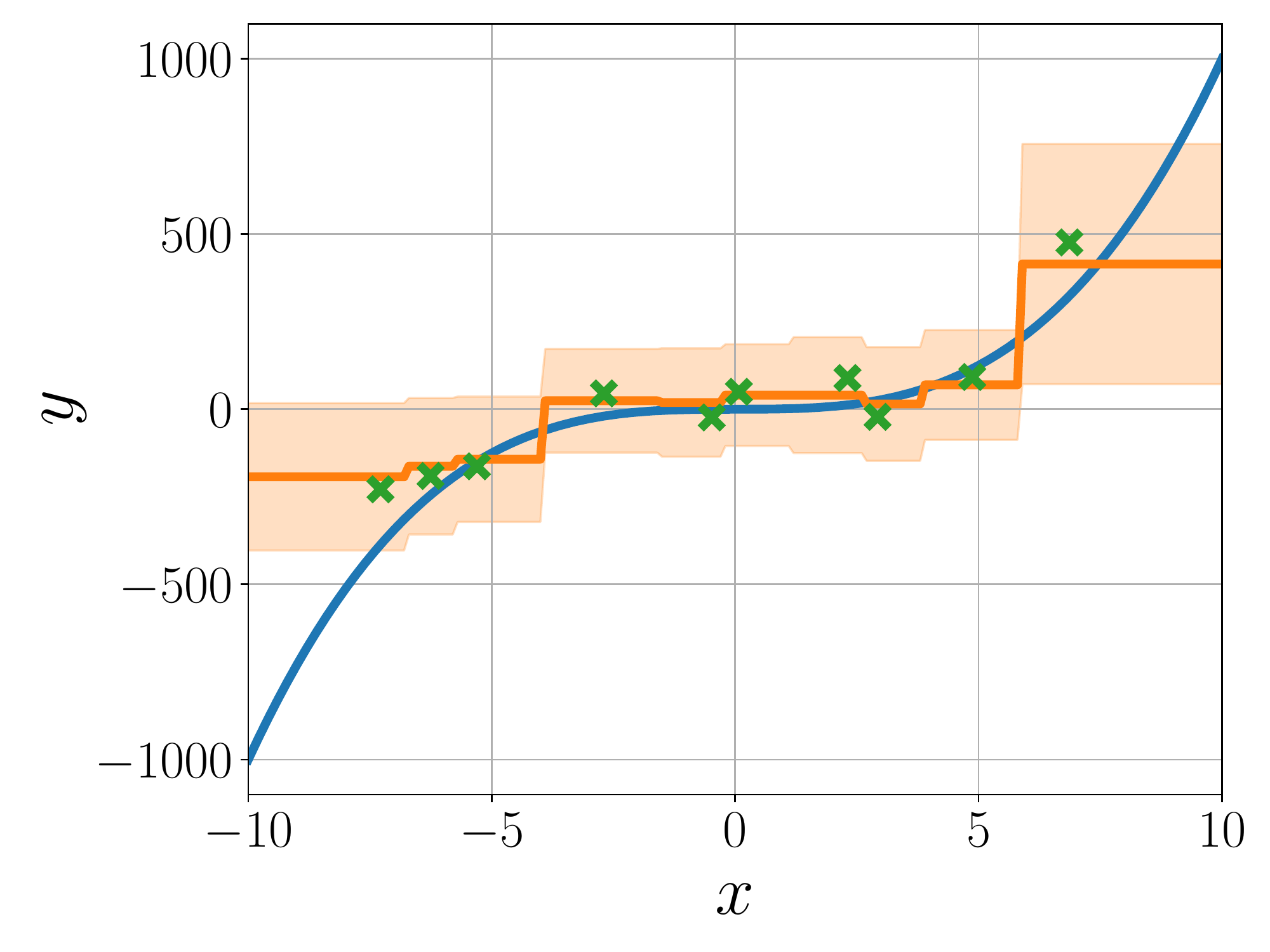}
		\label{fig:unc_1d_cubic_ngboost}
	}
	\subfigure[B + O]{
		\includegraphics[width=0.30\textwidth]{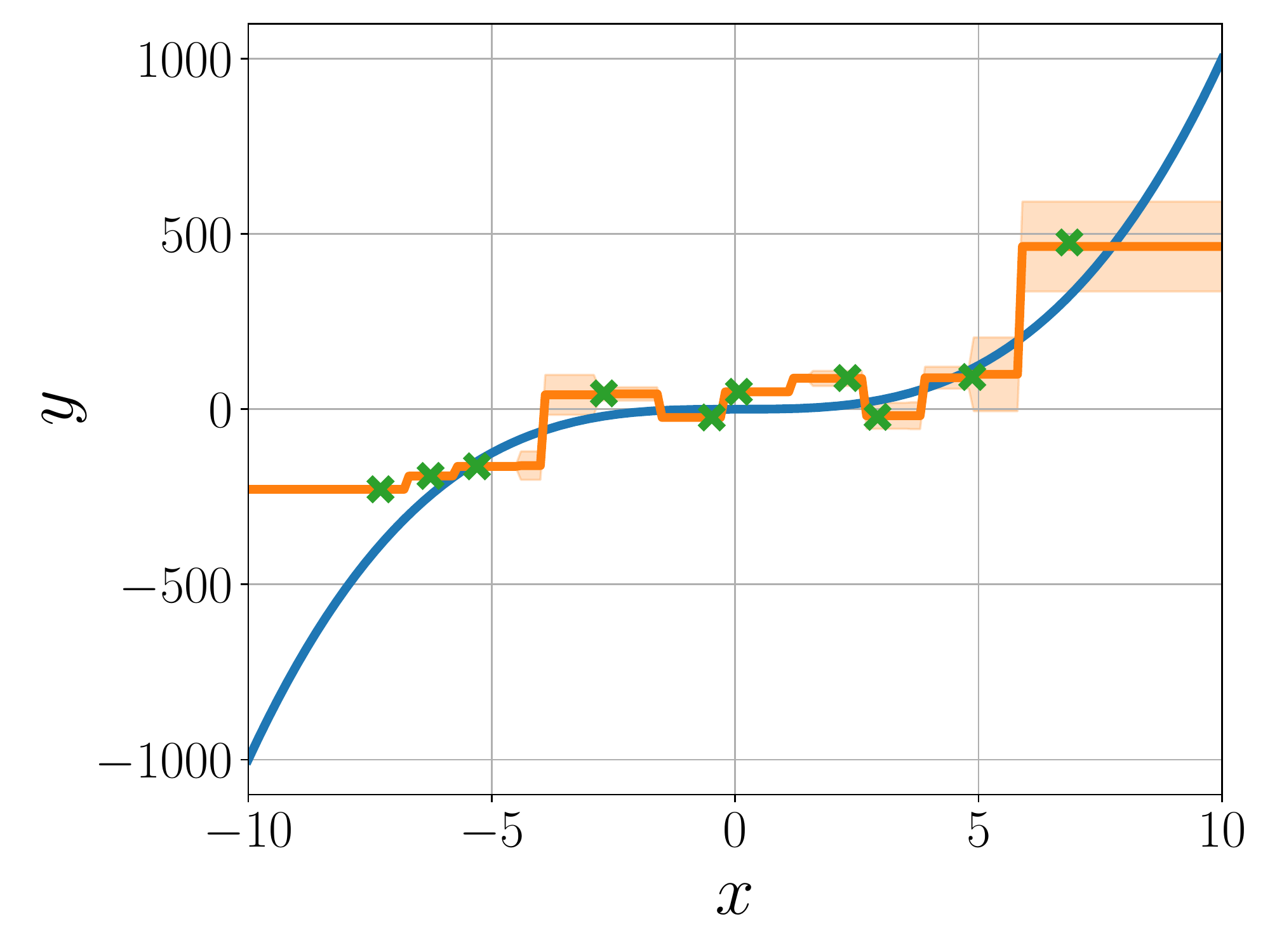}
		\label{fig:unc_1d_cubic_bo}
	}
	\subfigure[R + B]{
		\includegraphics[width=0.30\textwidth]{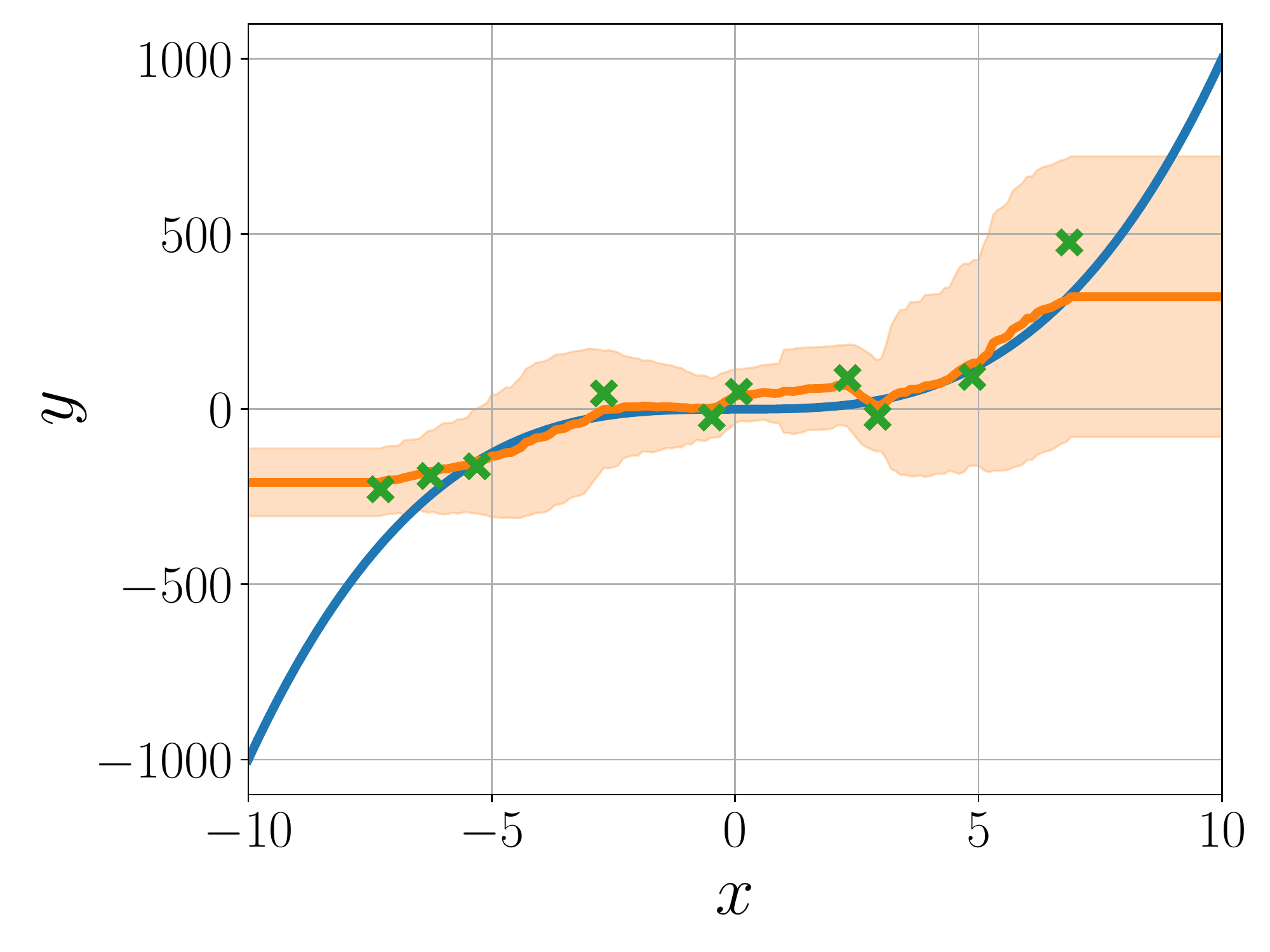}
		\label{fig:unc_1d_cubic_rb}
	}
	\subfigure[BwO forest (Ours)]{
		\includegraphics[width=0.30\textwidth]{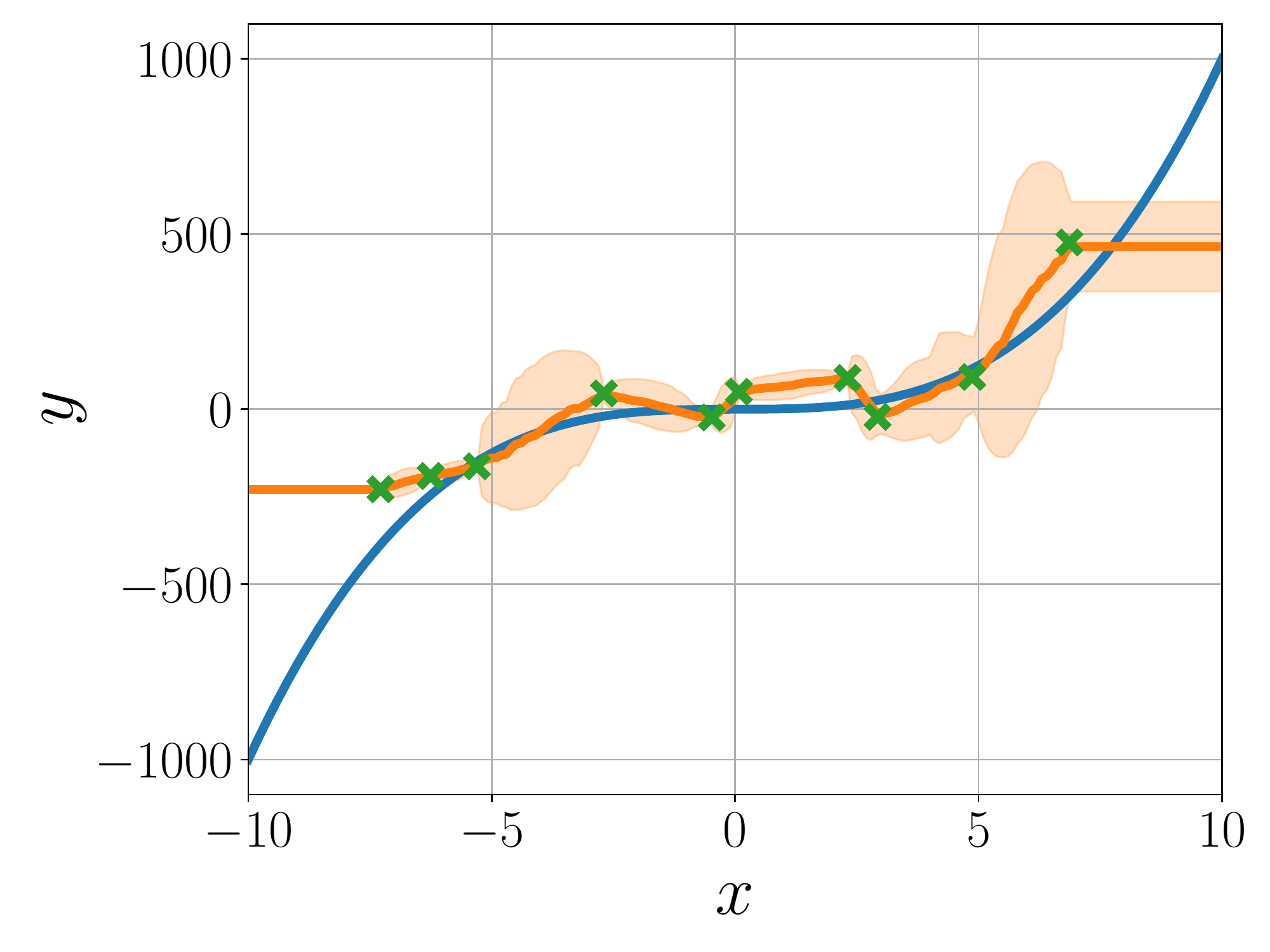}
		\label{fig:unc_1d_cubic_ours}
	}
	\caption{Examples on a cubic equation with 10 points.\label{fig:unc_1d_cubic}}
\end{figure}

The examples by diverse tree-based surrogate models as well as Gaussian process surrogate
are shown in~\figref{fig:unc_1d_many} and \figref{fig:unc_1d_cubic}.
Such examples present the characteristics of surrogate models,
as described in the main article.

\section{SEQUENTIAL MODEL-BASED OPTIMIZATION WITH OUR PROPOSED SURROGATE MODEL\label{sec:suppl_smo}}

In this section, we present the sequential model-based optimization procedure with BwO forests.
It follows generic steps of sequential model-based optimization~\citep{BrochuE2010arxiv}.

\begin{algorithm}[t]
	\caption{Sequential Model-based Optimization with BwO Forests}
	\label{alg:bo_bwo}
	\begin{algorithmic}[1]
		\REQUIRE Initial points and their evaluations $(\bX_0, \by_0)$, the number of iterations $T$, and the size of ensemble model $B$.
		\ENSURE The best query point $\bx_{\textrm{best}}$.
		\FOR {$t = 1, \ldots, T$}
			\STATE Fit BwO forest $\{\calT_b\}_{b = 1}^B$ using $\bX_{t - 1}$ and $\by_{t - 1}$.
			\STATE Acquire a query point by optimizing an acquisition function, $\bx_t = \argmax a(\bx; \{\calT_b\}_{b = 1}^B)$.
			\STATE Evaluate $y_t = f(\bx_t) + \epsilon$ where $\epsilon$ is an observation noise.
			\STATE Update $\bX_t \leftarrow [\bX_{t - 1}; \bx_t]$ and $\by_t \leftarrow [\by_{t - 1}; y_t]$.
		\ENDFOR
		\STATE \textbf{return} The best query point $\bx_{\textrm{best}}$ among $\bX_T$
	\end{algorithmic}
\end{algorithm}

As shown in~\algref{alg:bo_bwo}, we are given initial points $\bX_0$, their corresponding evaluations $\by_0$,
the number of iterations $T$, and the size of ensemble model $B$.
Sequentially, we acquire a query point and evaluate it every iteration
by fitting BwO forest and optimizing an acquisition function.
Finally, the best query point $\bx_{\textrm{best}}$ among $\bX_T$ is determined,
by considering the evaluations $\by_T$.

\section{REGRESSION RESULTS BY INDIVIDUAL TREES}

\begin{figure}[p]
	\centering
	\subfigure[B (originally proposed as random forest)]{
		\includegraphics[width=0.235\textwidth]{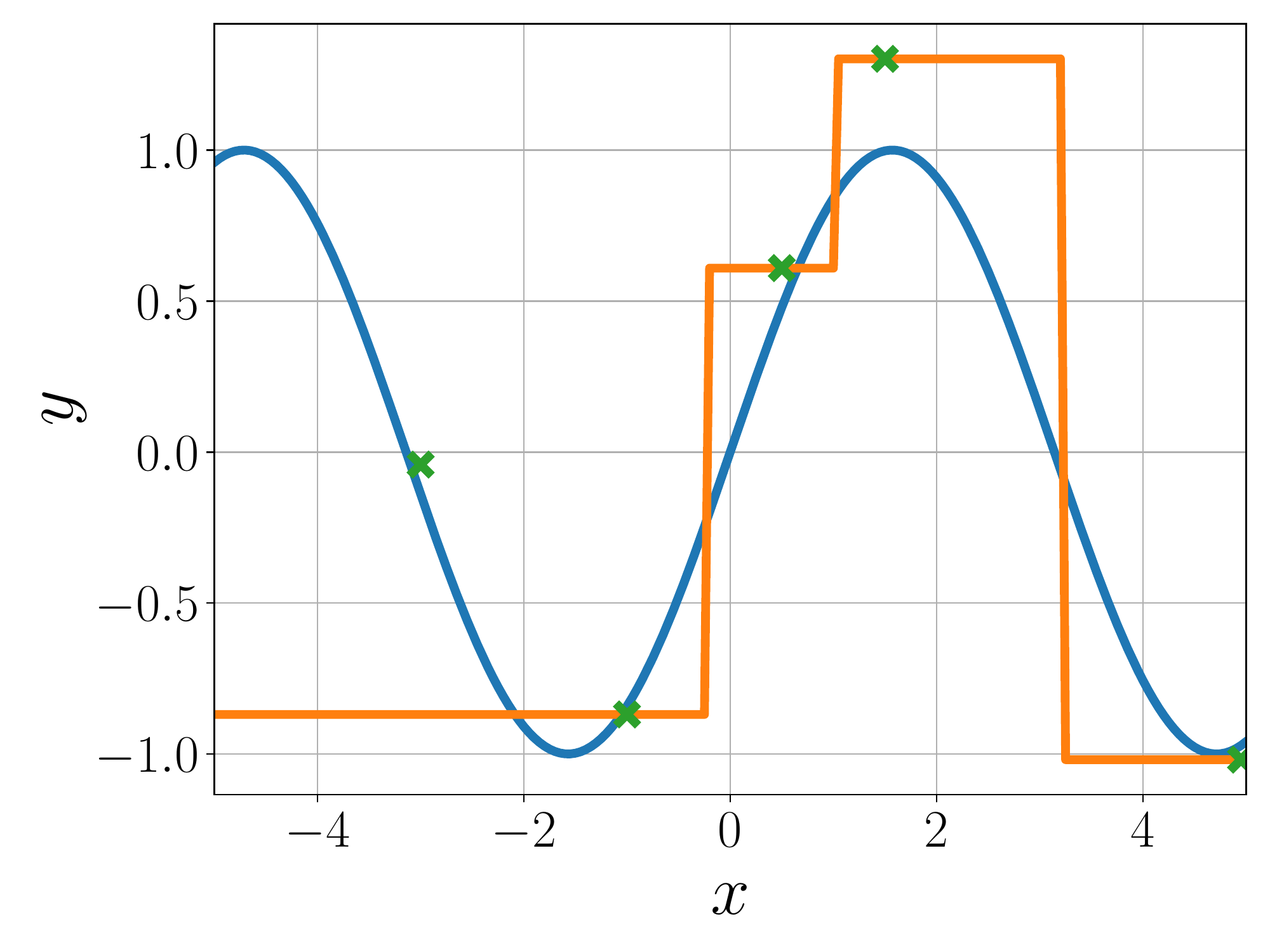}
		\includegraphics[width=0.235\textwidth]{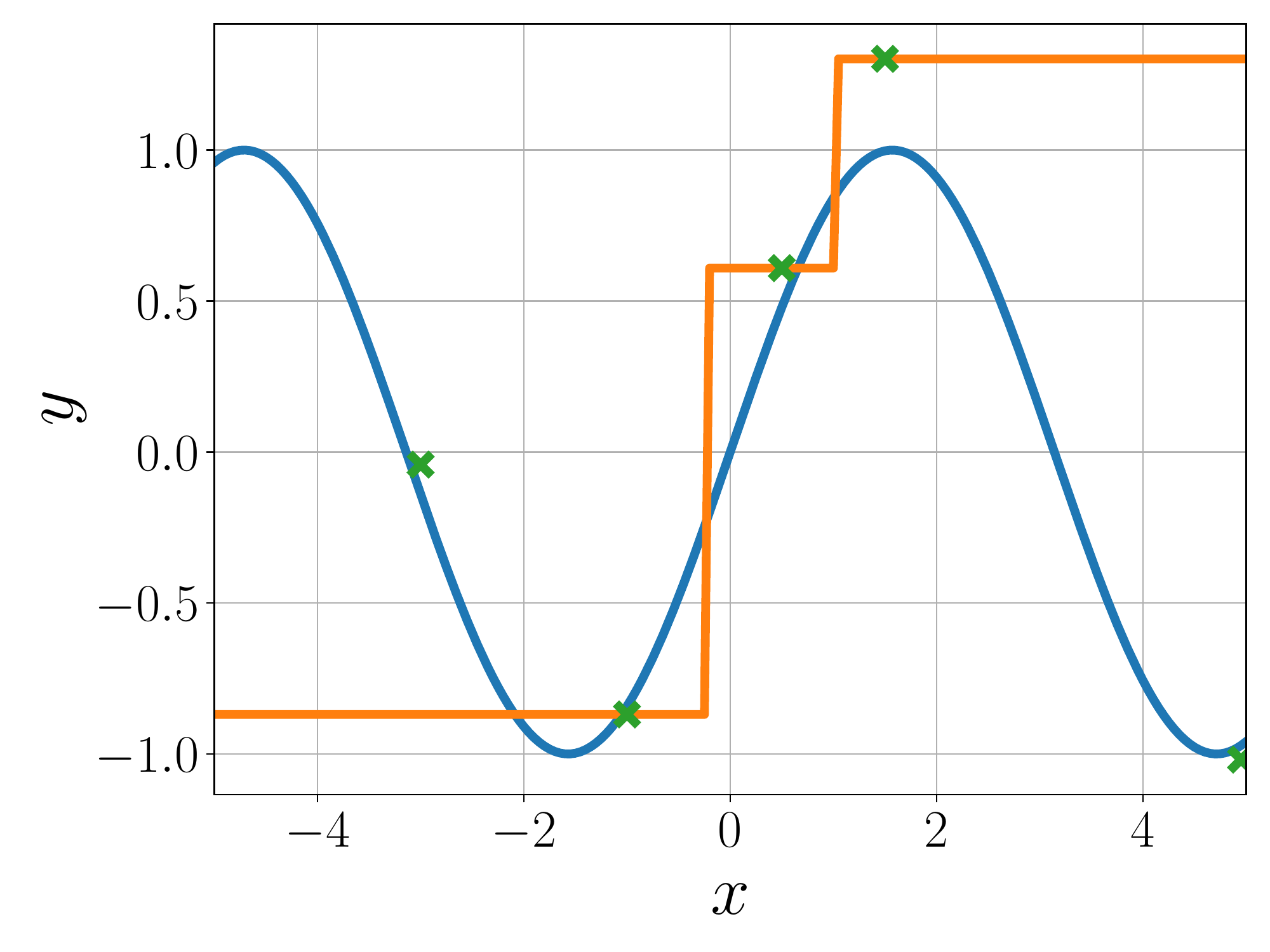}
		\includegraphics[width=0.235\textwidth]{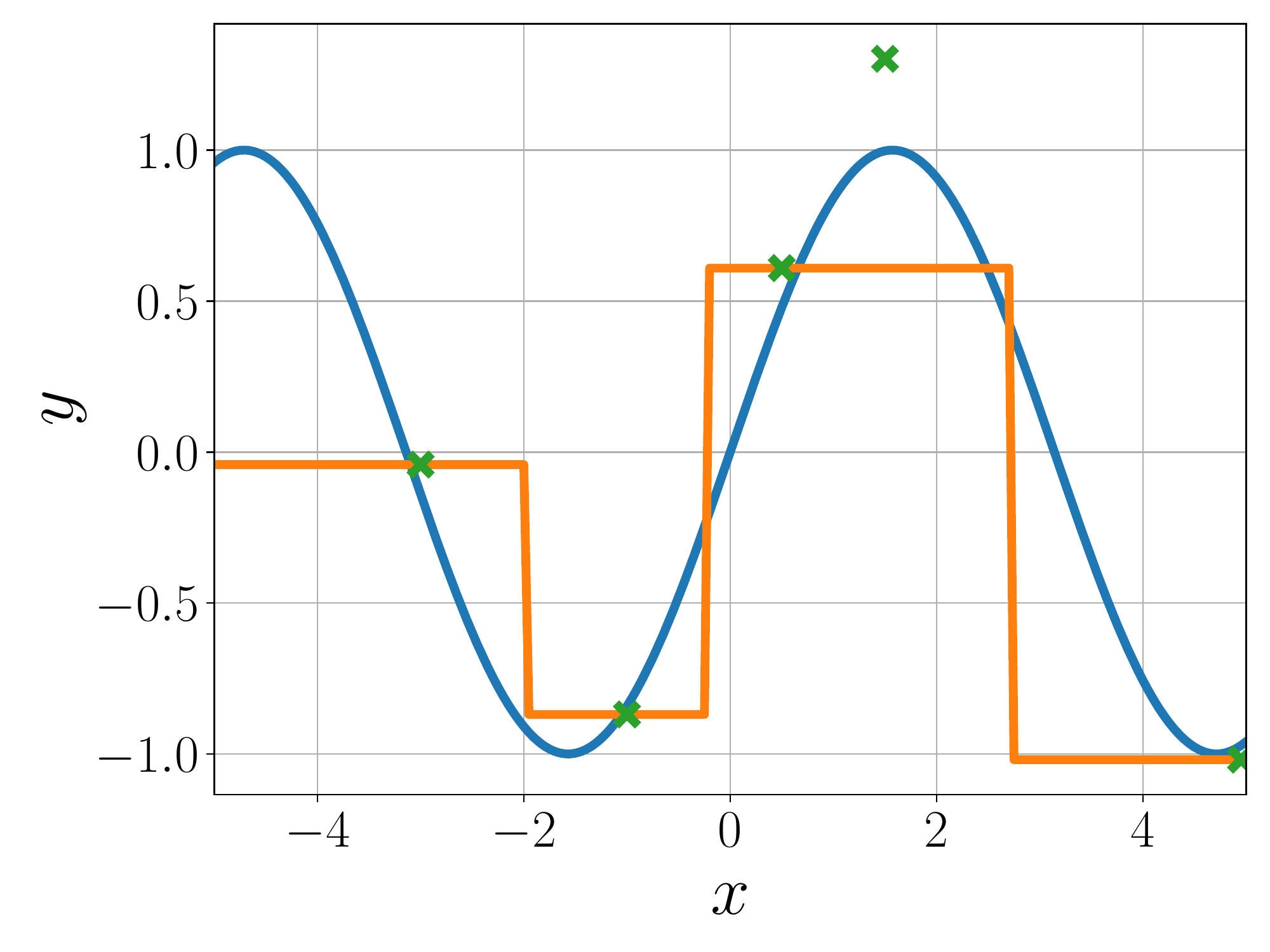}
		\includegraphics[width=0.235\textwidth]{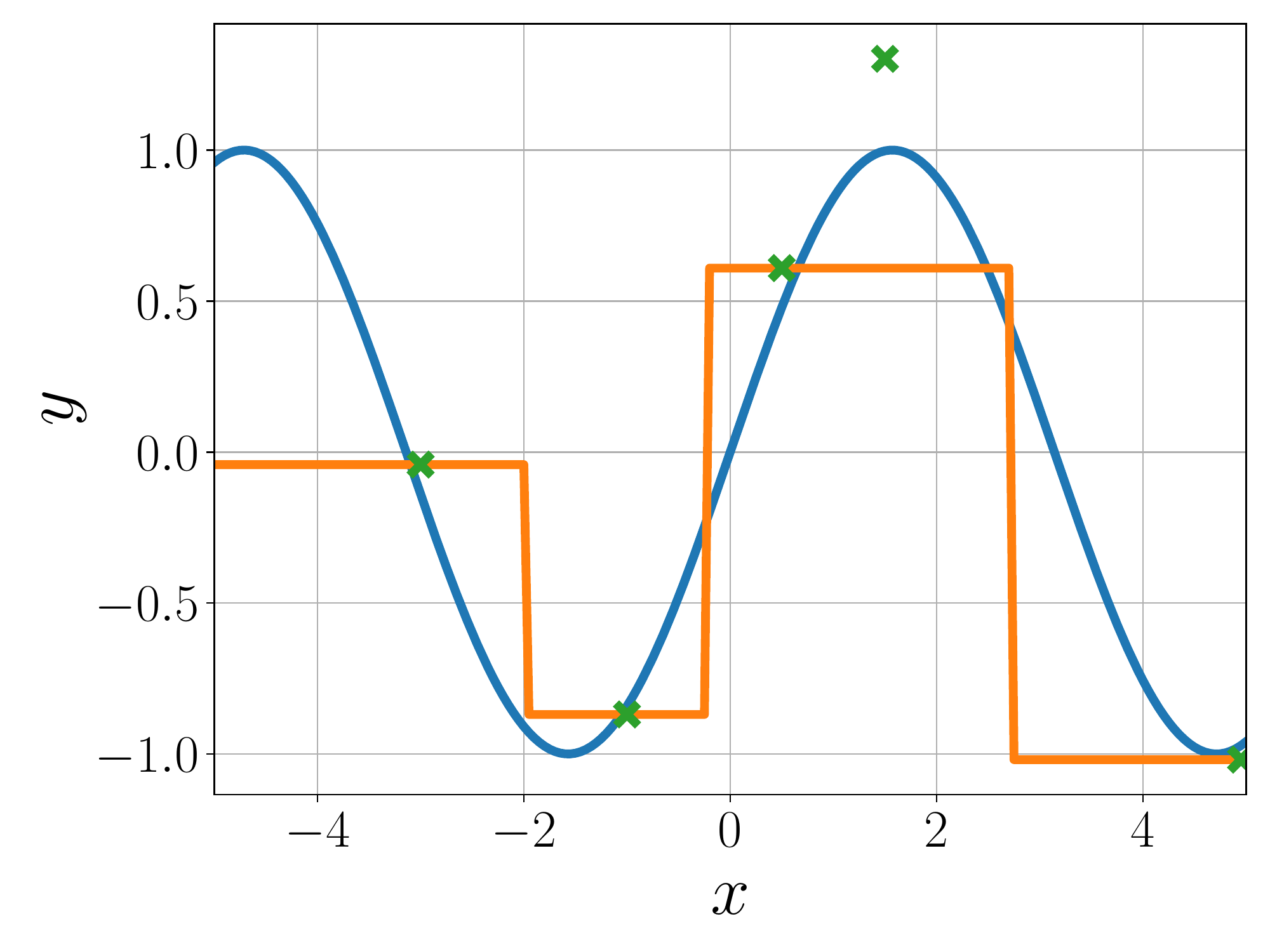}
		\label{fig:unc_1d_few_trees_b}
	}
	\subfigure[R (originally proposed as extremely randomized trees)]{
		\includegraphics[width=0.235\textwidth]{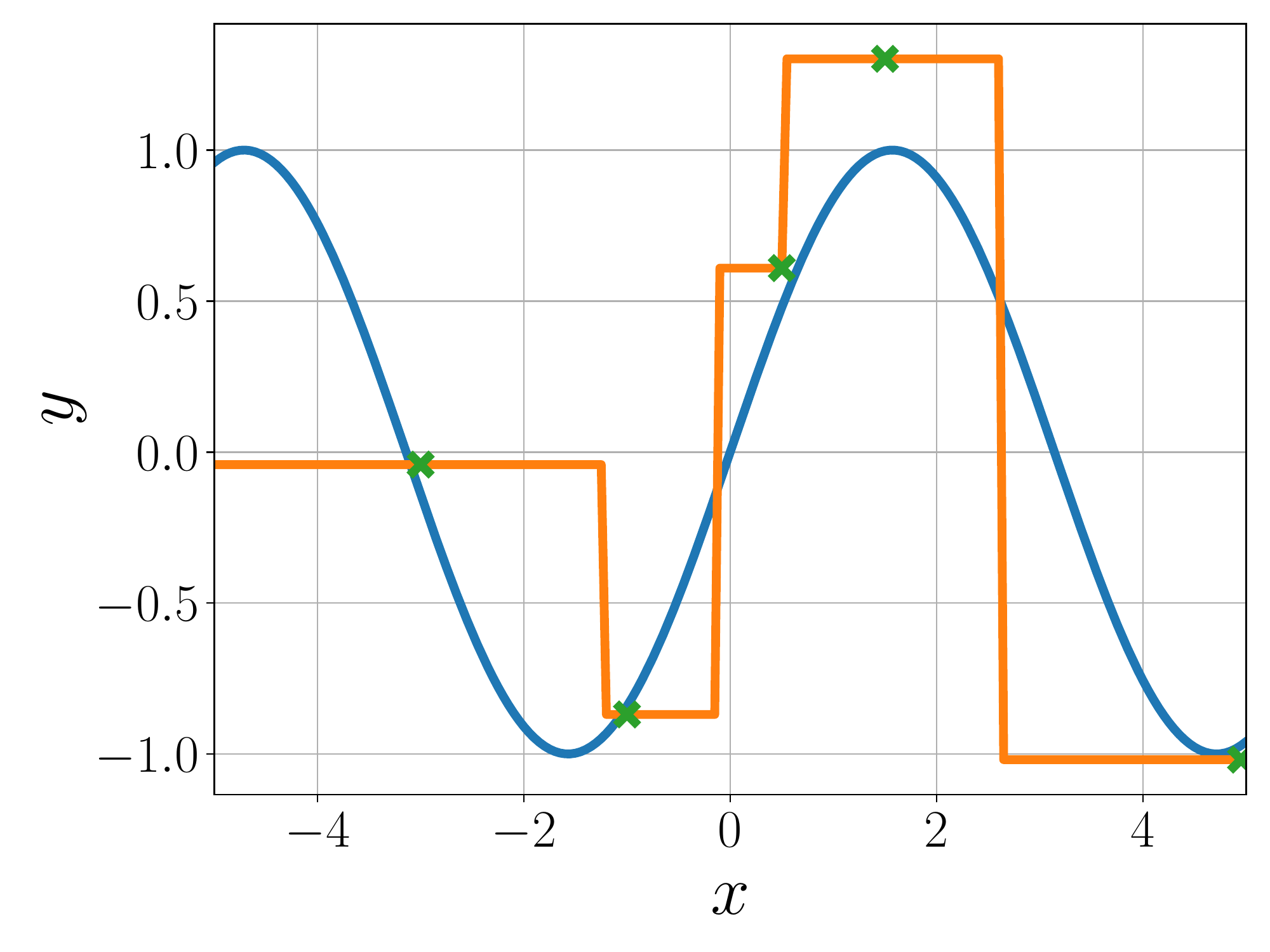}
		\includegraphics[width=0.235\textwidth]{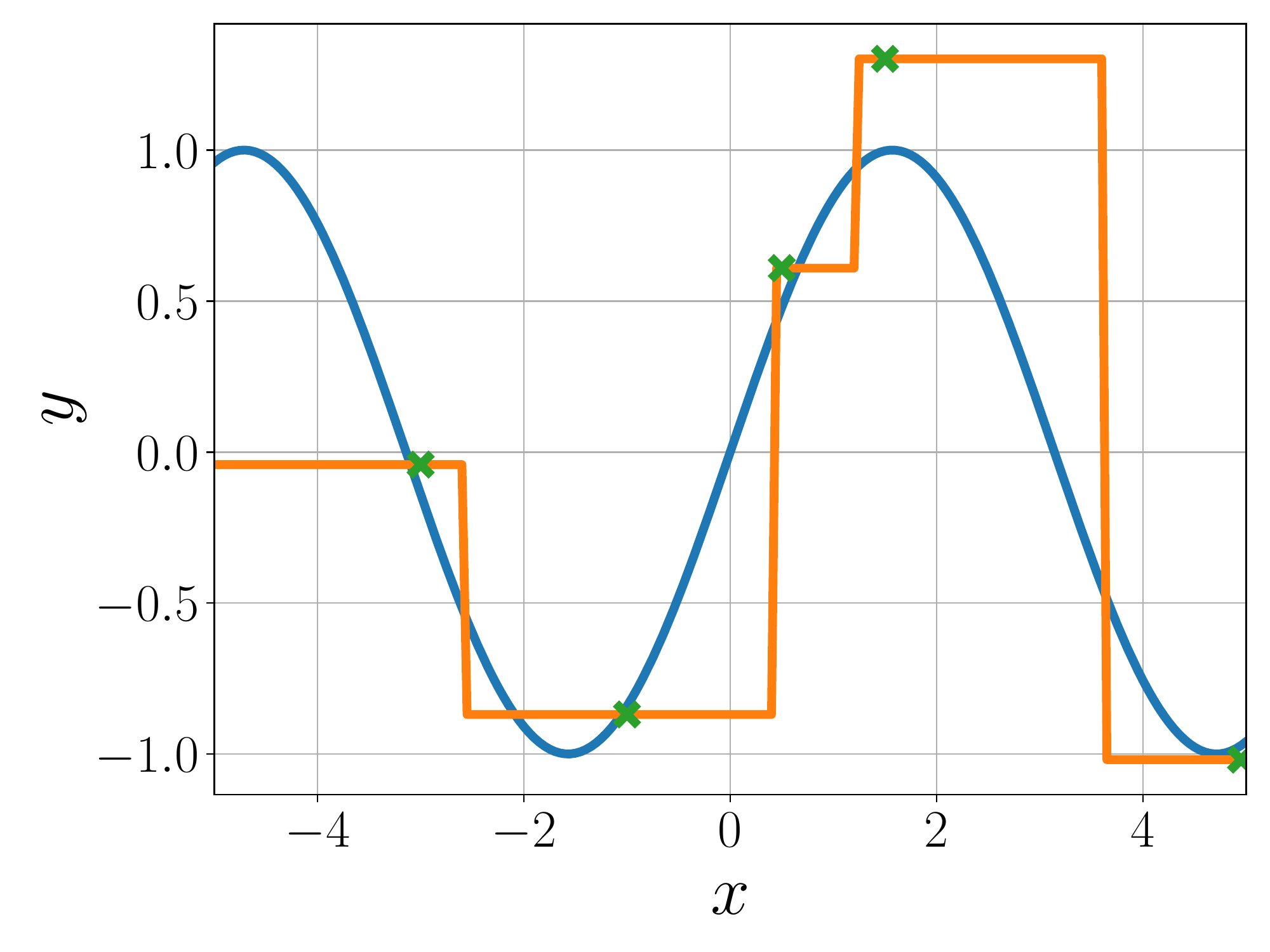}
		\includegraphics[width=0.235\textwidth]{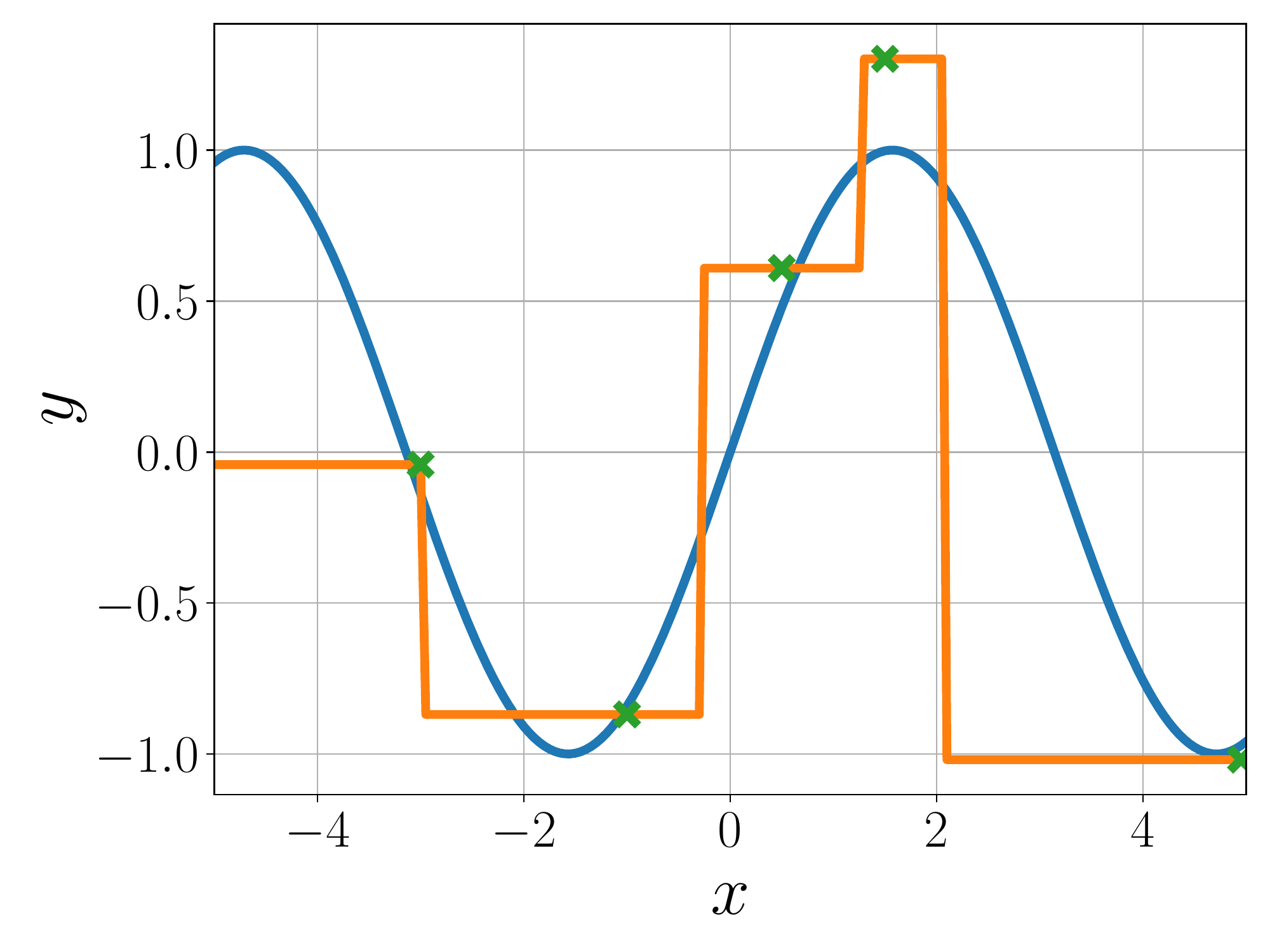}
		\includegraphics[width=0.235\textwidth]{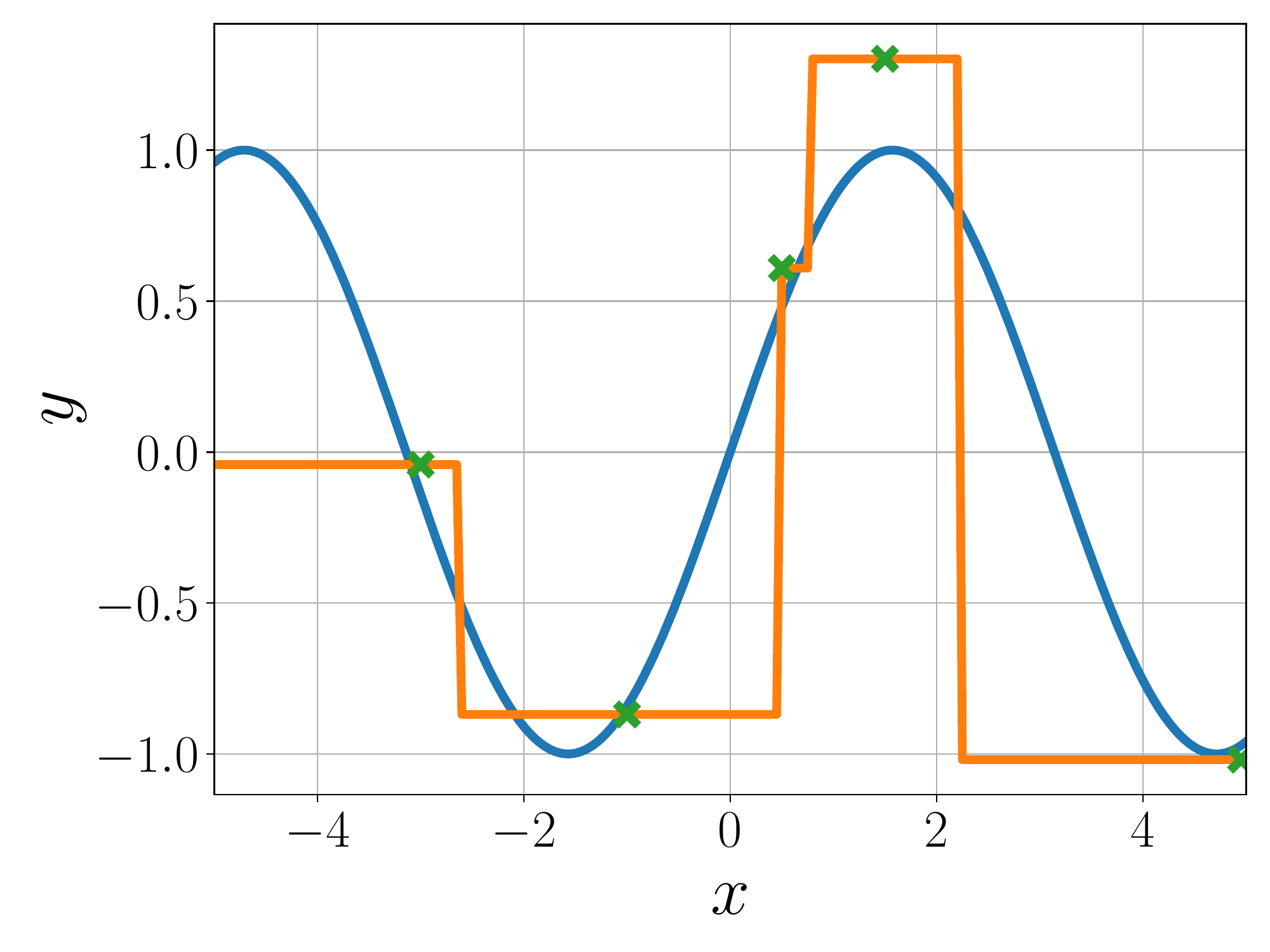}
		\label{fig:unc_1d_few_trees_r}
	}
	\subfigure[B + O]{
		\includegraphics[width=0.235\textwidth]{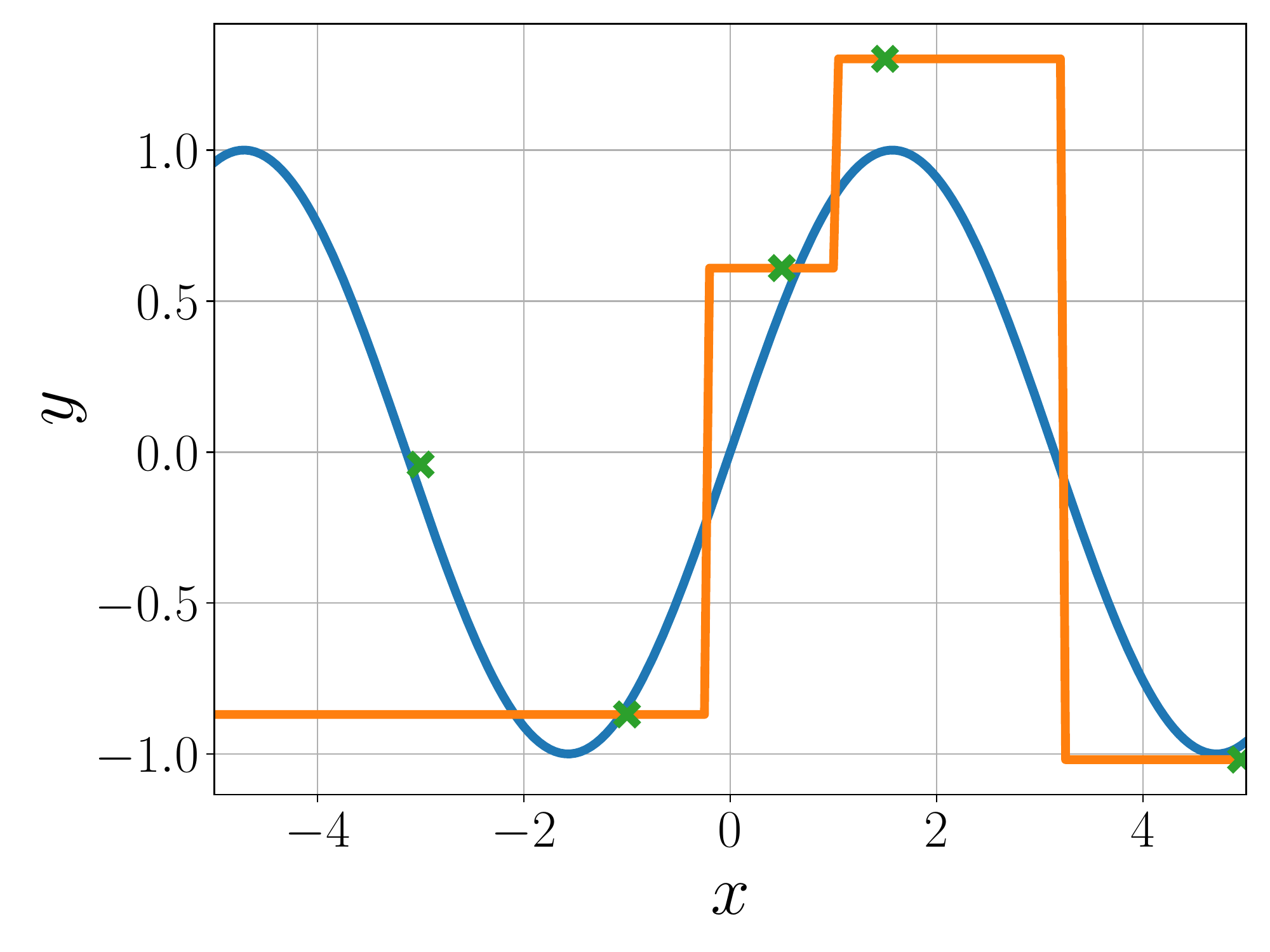}
		\includegraphics[width=0.235\textwidth]{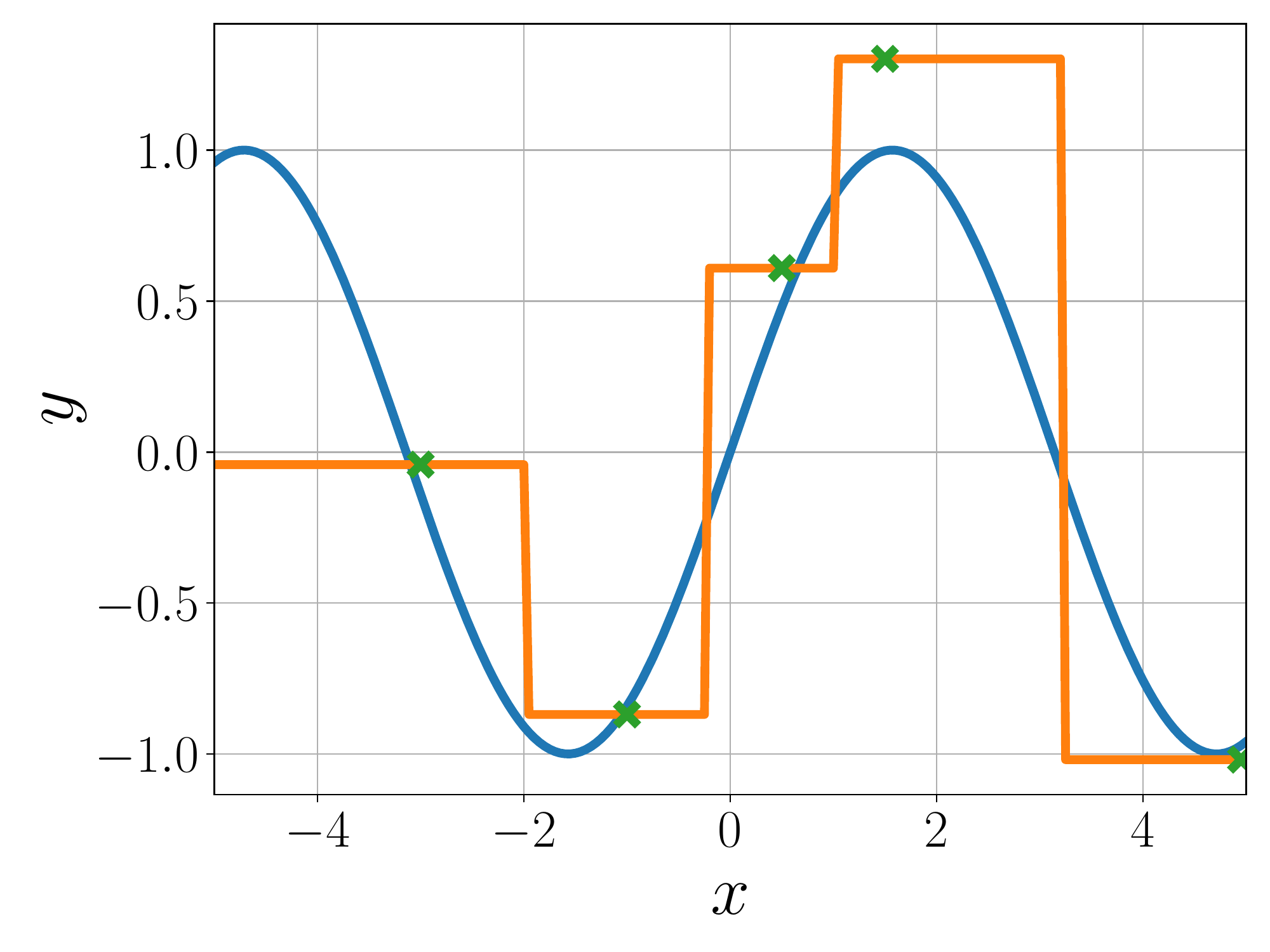}
		\includegraphics[width=0.235\textwidth]{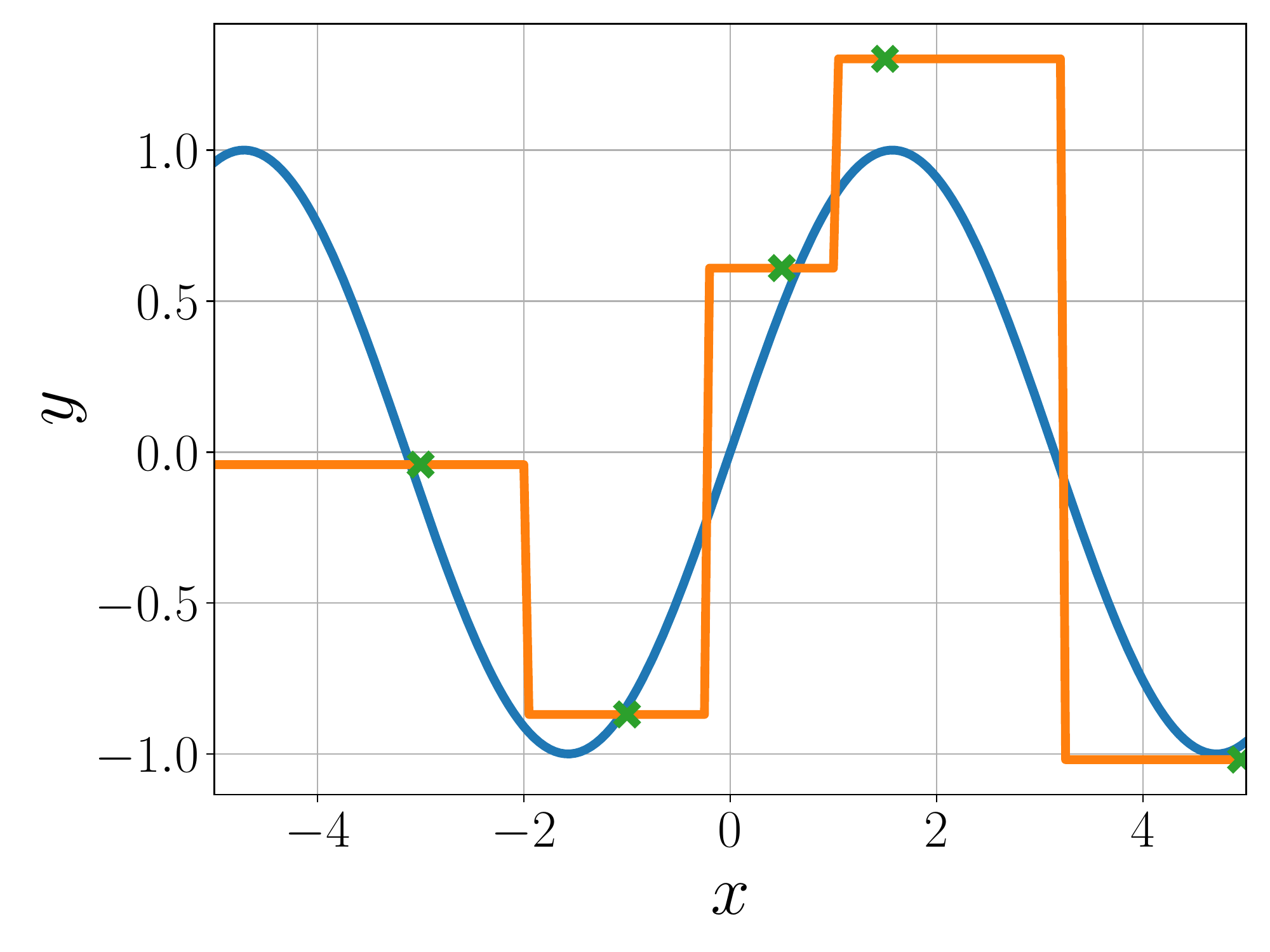}
		\includegraphics[width=0.235\textwidth]{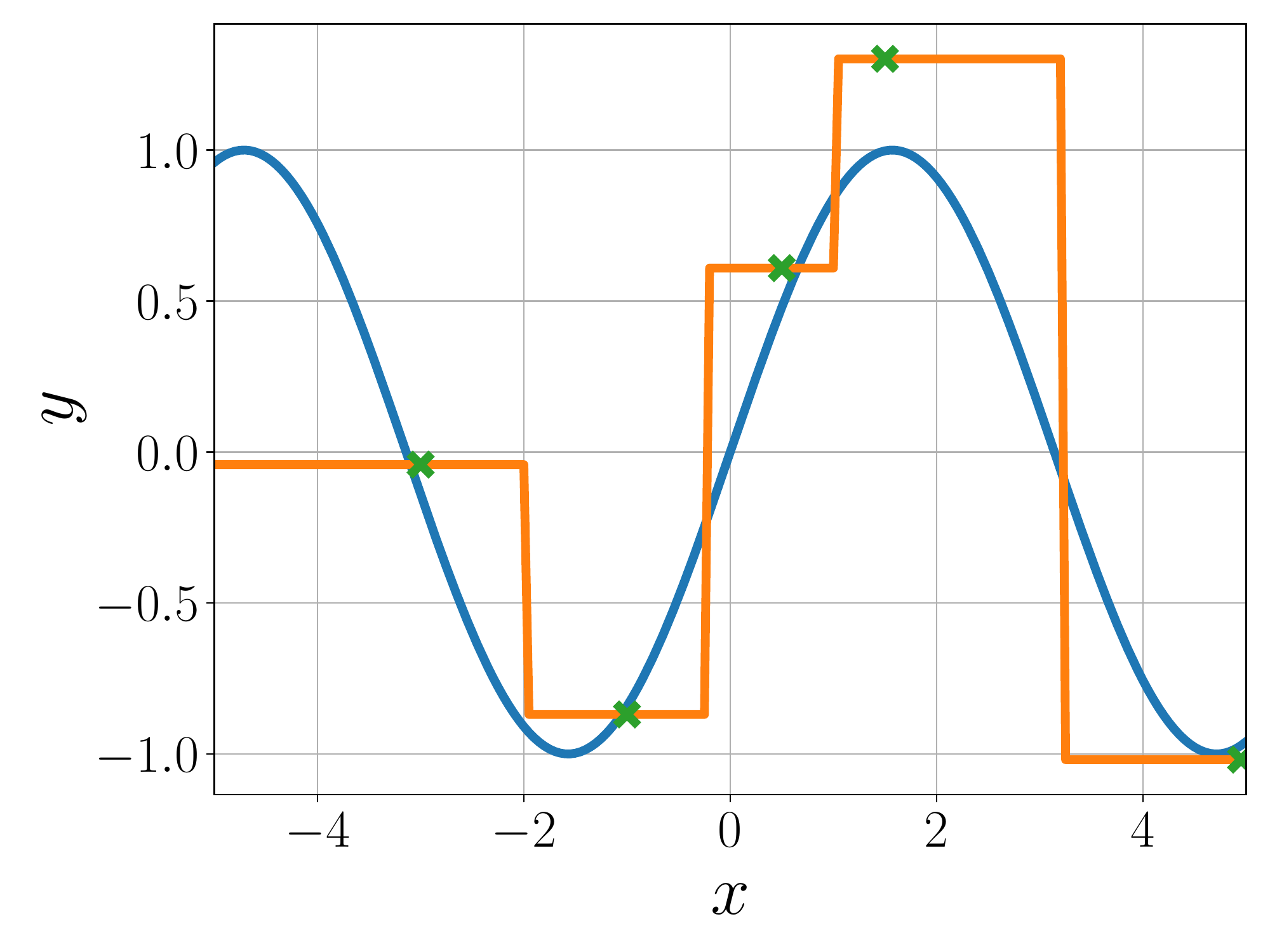}
		\label{fig:unc_1d_few_trees_bo}
	}
	\subfigure[R + B]{
		\includegraphics[width=0.235\textwidth]{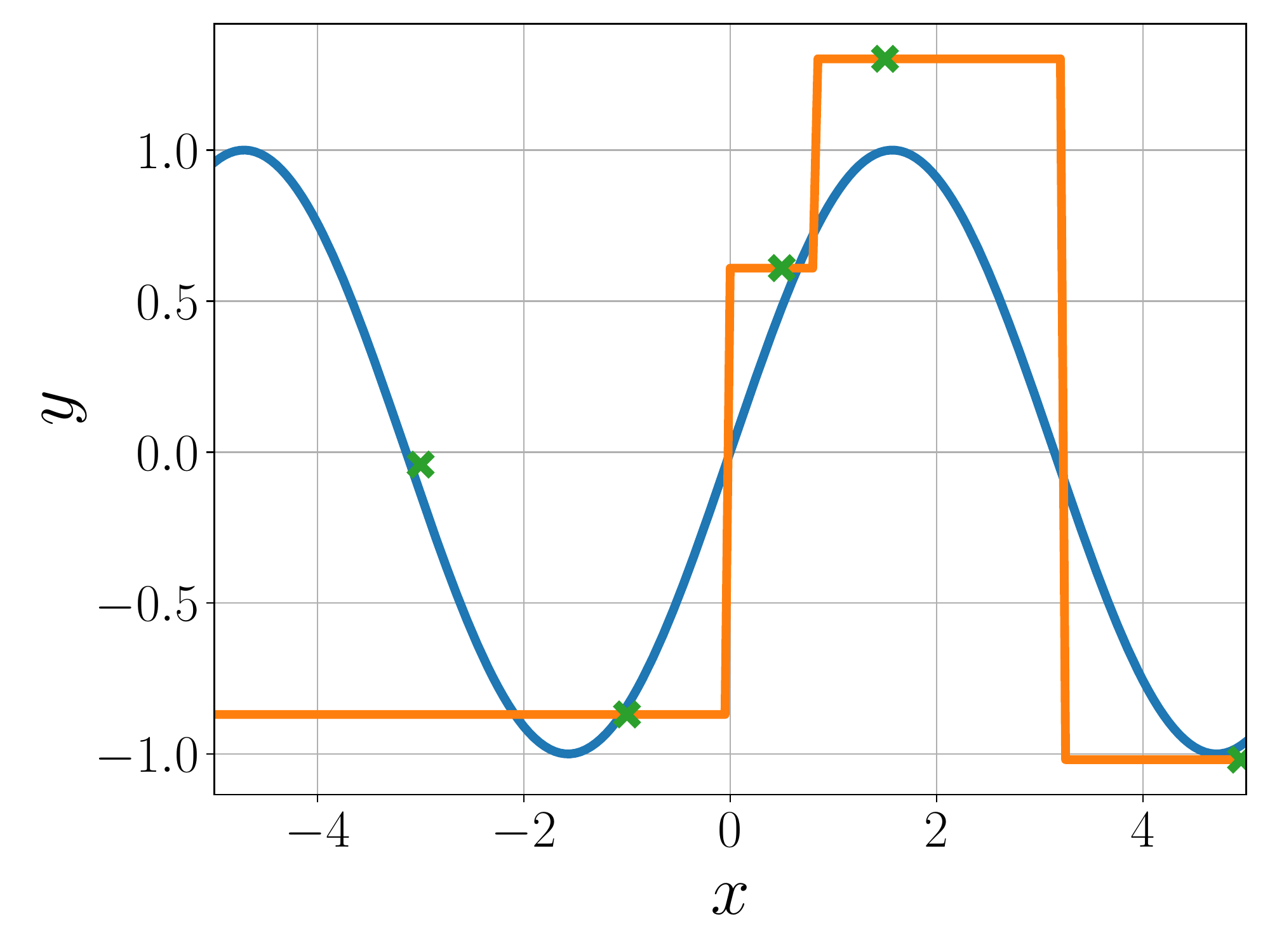}
		\includegraphics[width=0.235\textwidth]{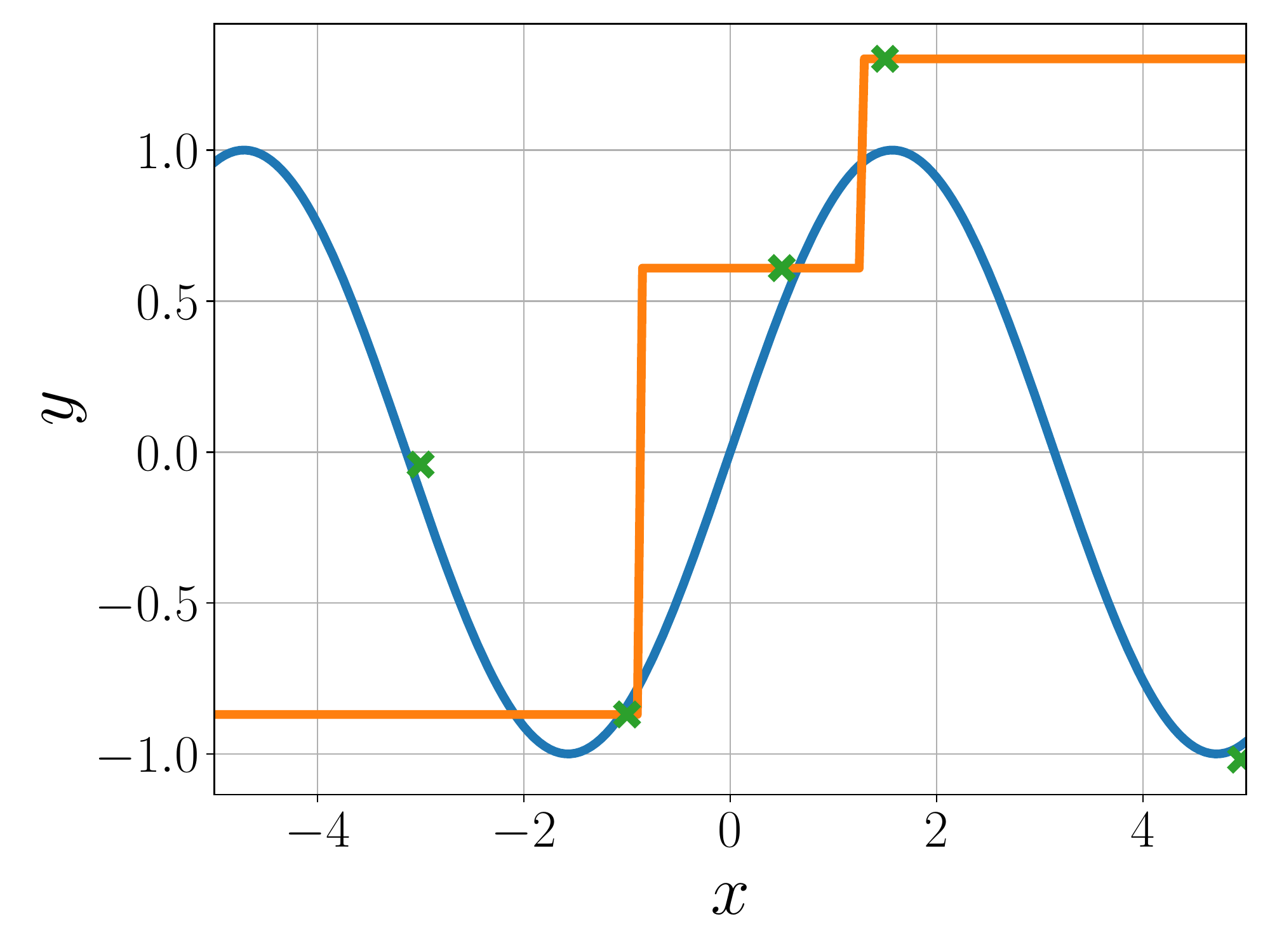}
		\includegraphics[width=0.235\textwidth]{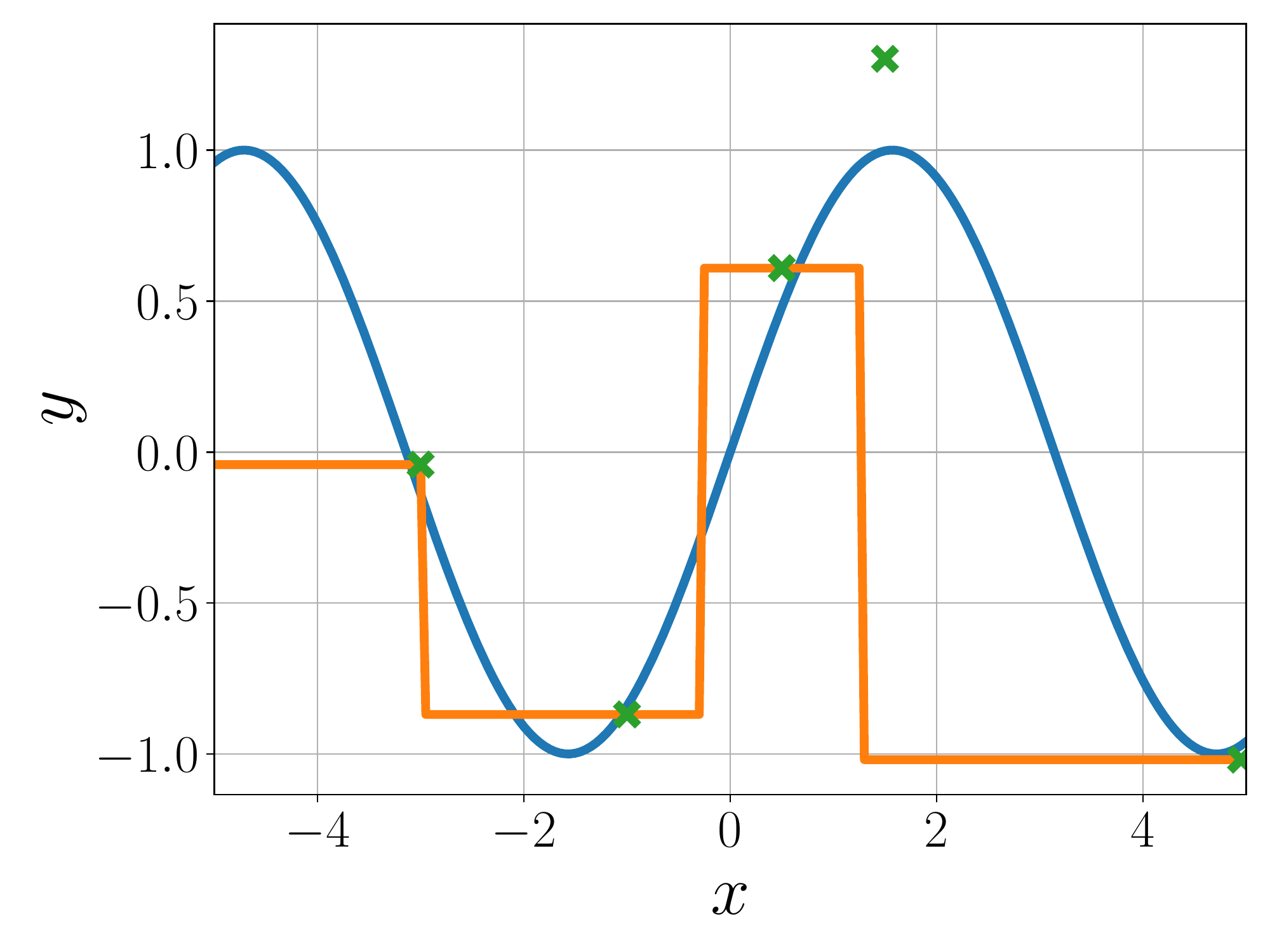}
		\includegraphics[width=0.235\textwidth]{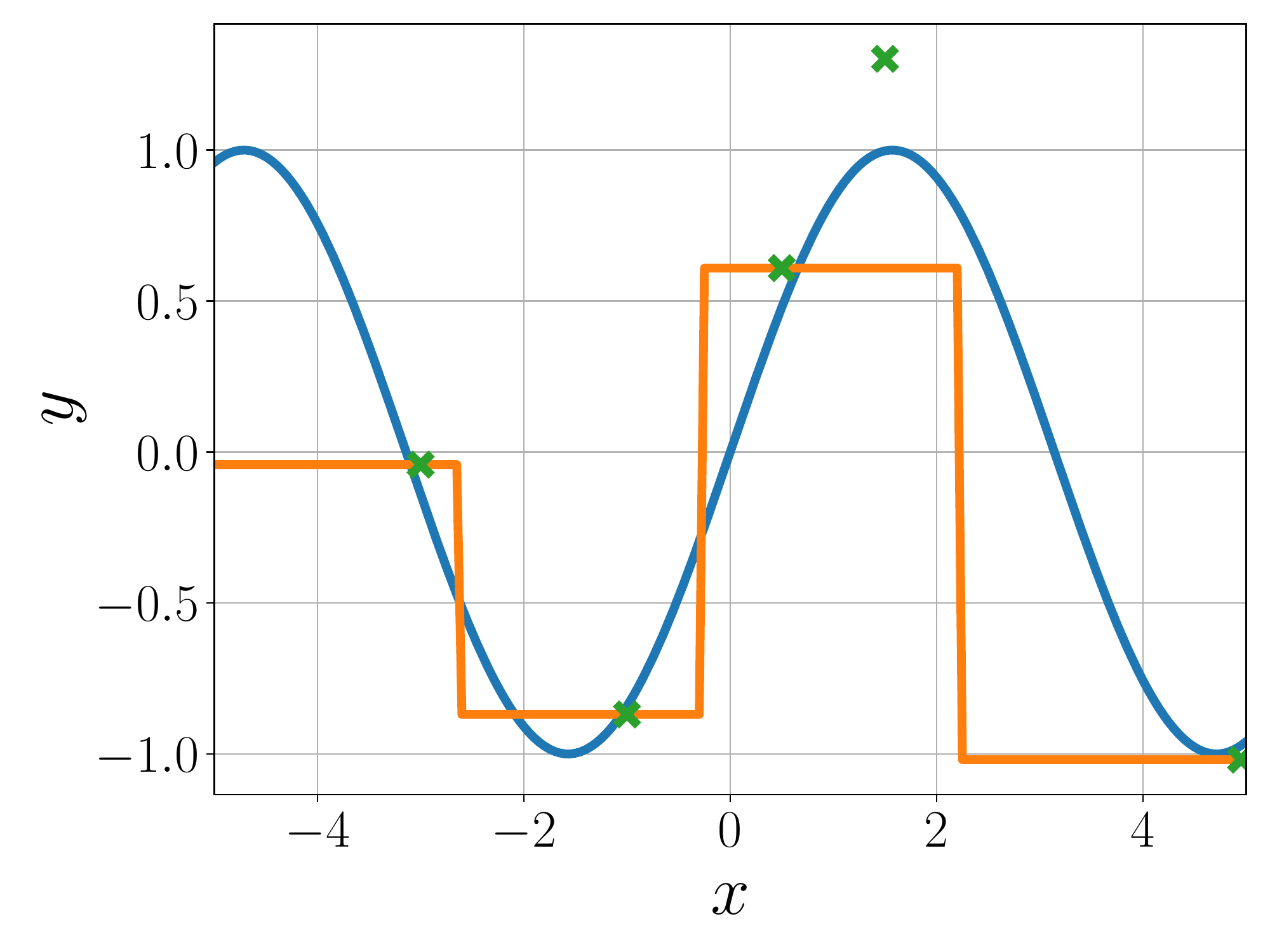}
		\label{fig:unc_1d_few_trees_rb}
	}
	\subfigure[R + B + O (i.e., BwO forest)]{
		\includegraphics[width=0.235\textwidth]{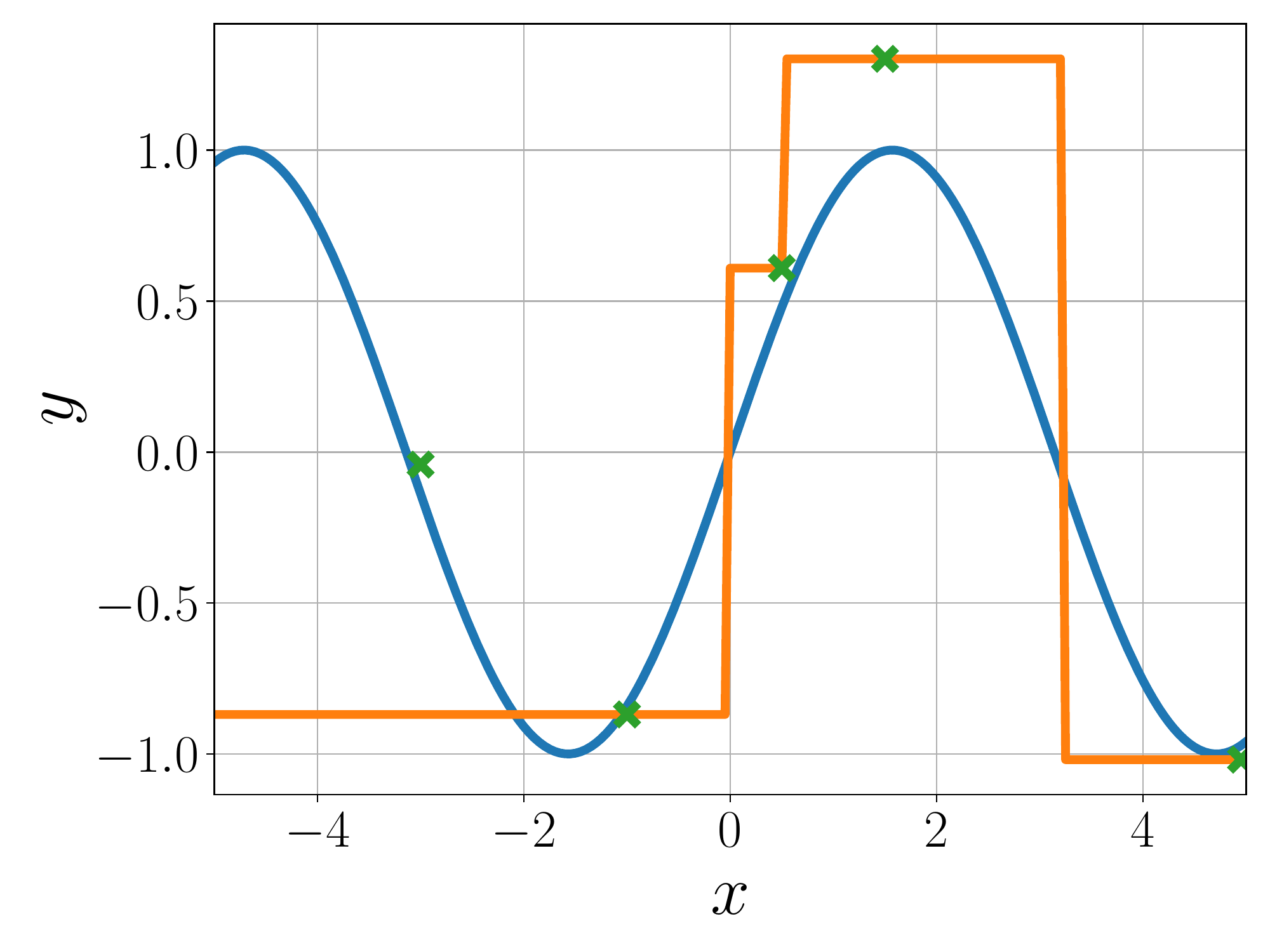}
		\includegraphics[width=0.235\textwidth]{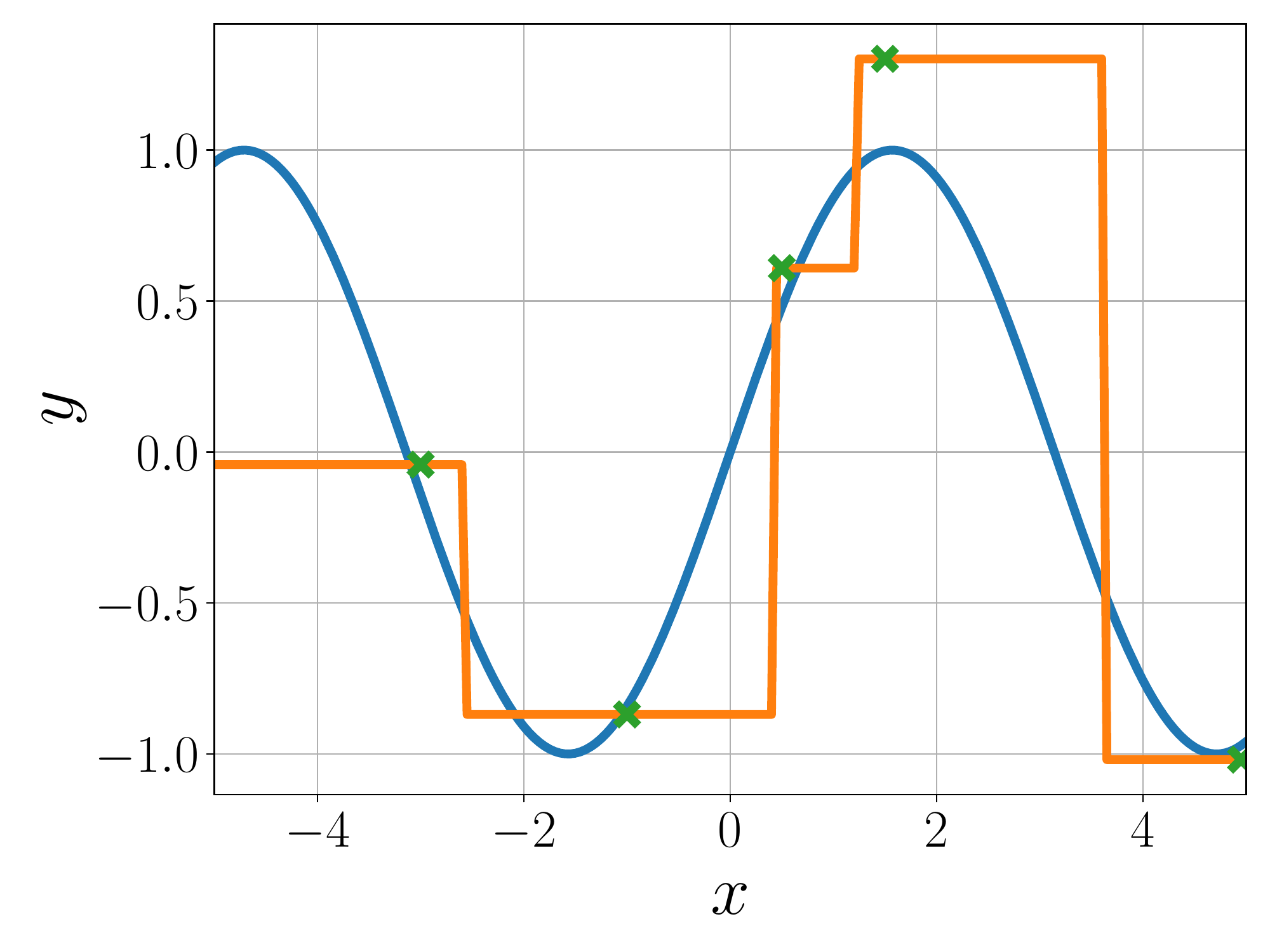}
		\includegraphics[width=0.235\textwidth]{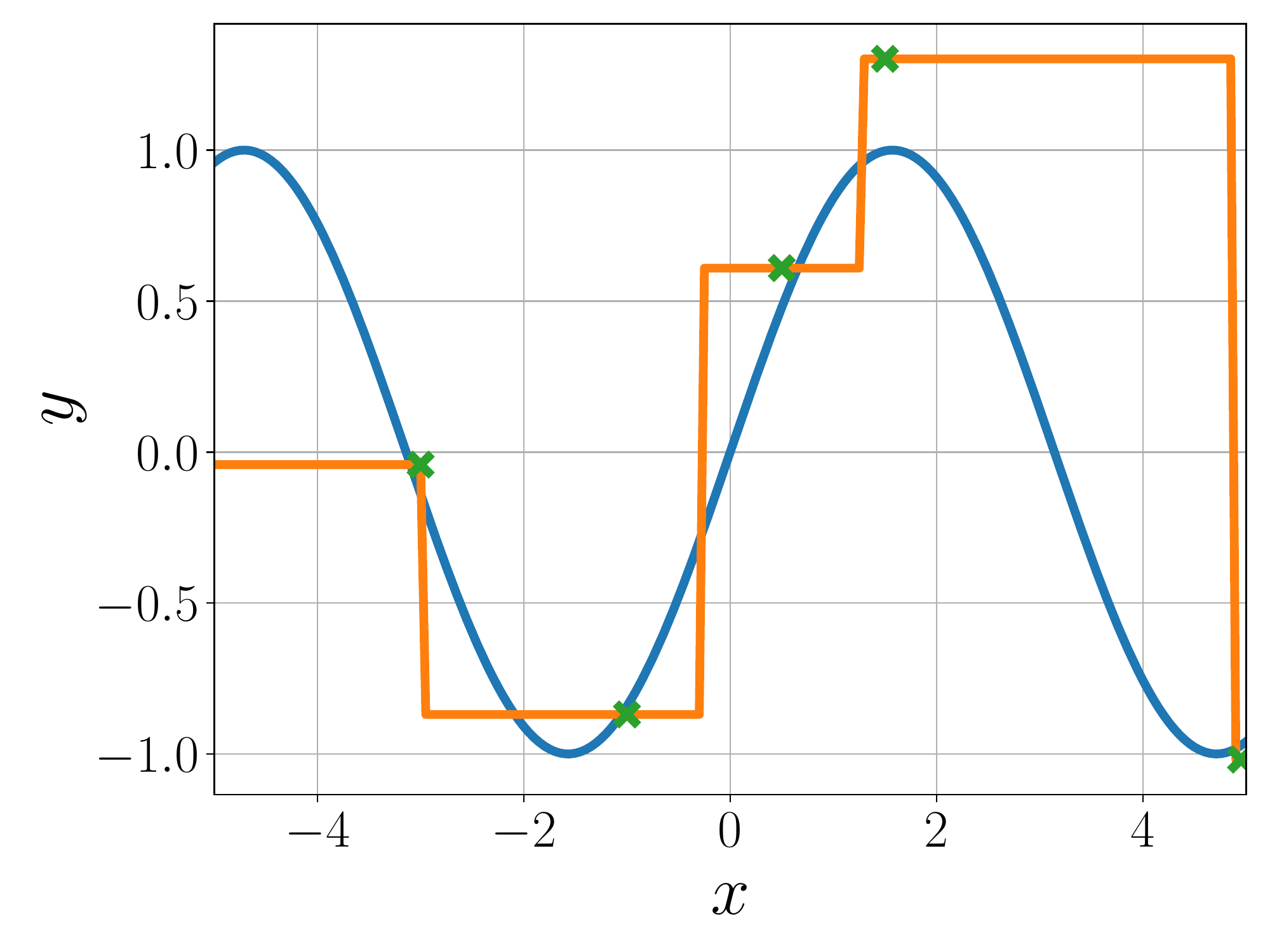}
		\includegraphics[width=0.235\textwidth]{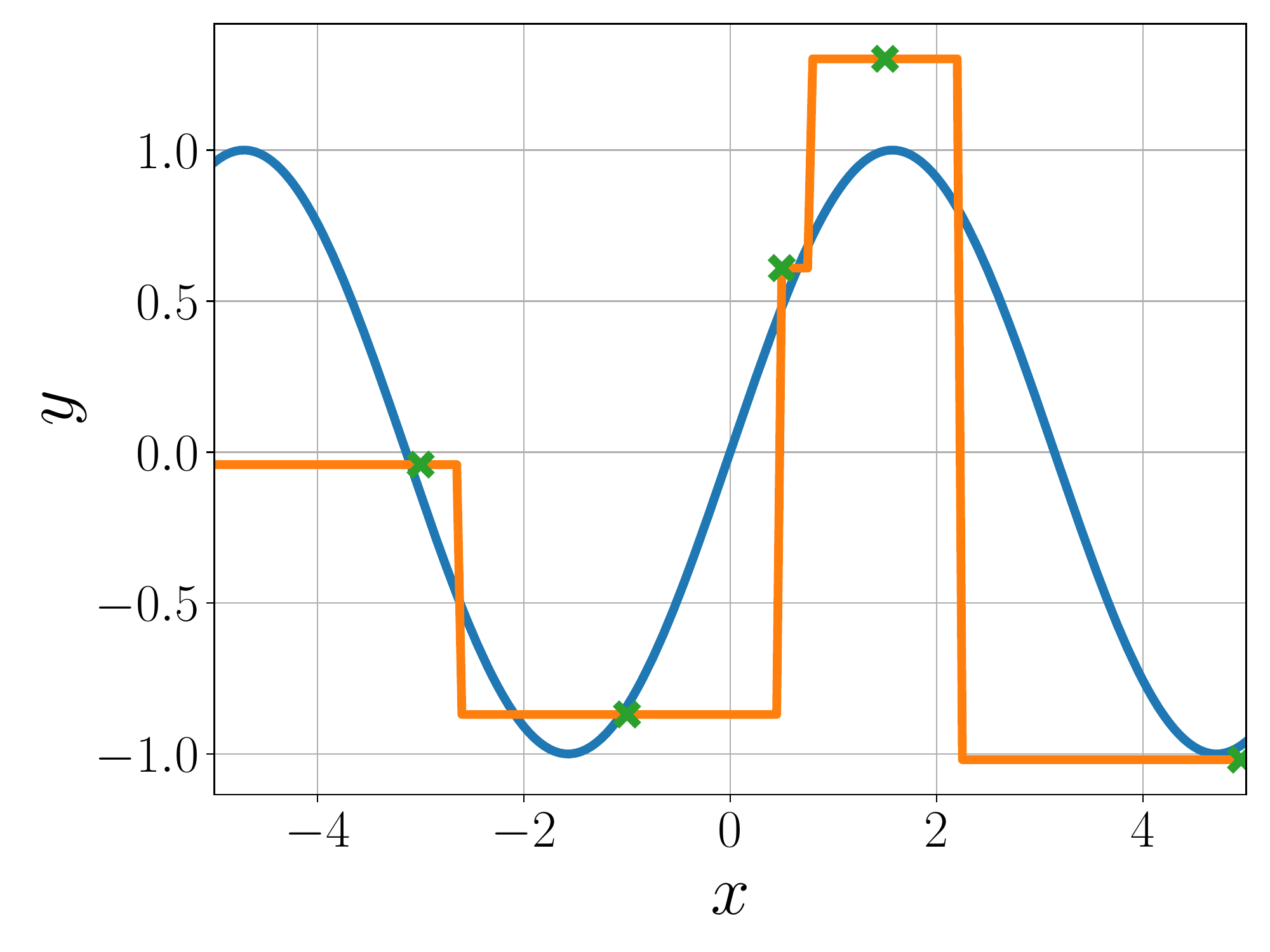}
		\label{fig:unc_1d_few_trees_rbo}
	}
	\caption{Results by individual trees for the case shown in \figref{fig:unc_1d_few}. For brevity, each result is randomly sampled.\label{fig:unc_1d_few_trees}}
\end{figure}

\begin{figure}[p]
	\centering
	\subfigure[B (originally proposed as random forest)]{
		\includegraphics[width=0.235\textwidth]{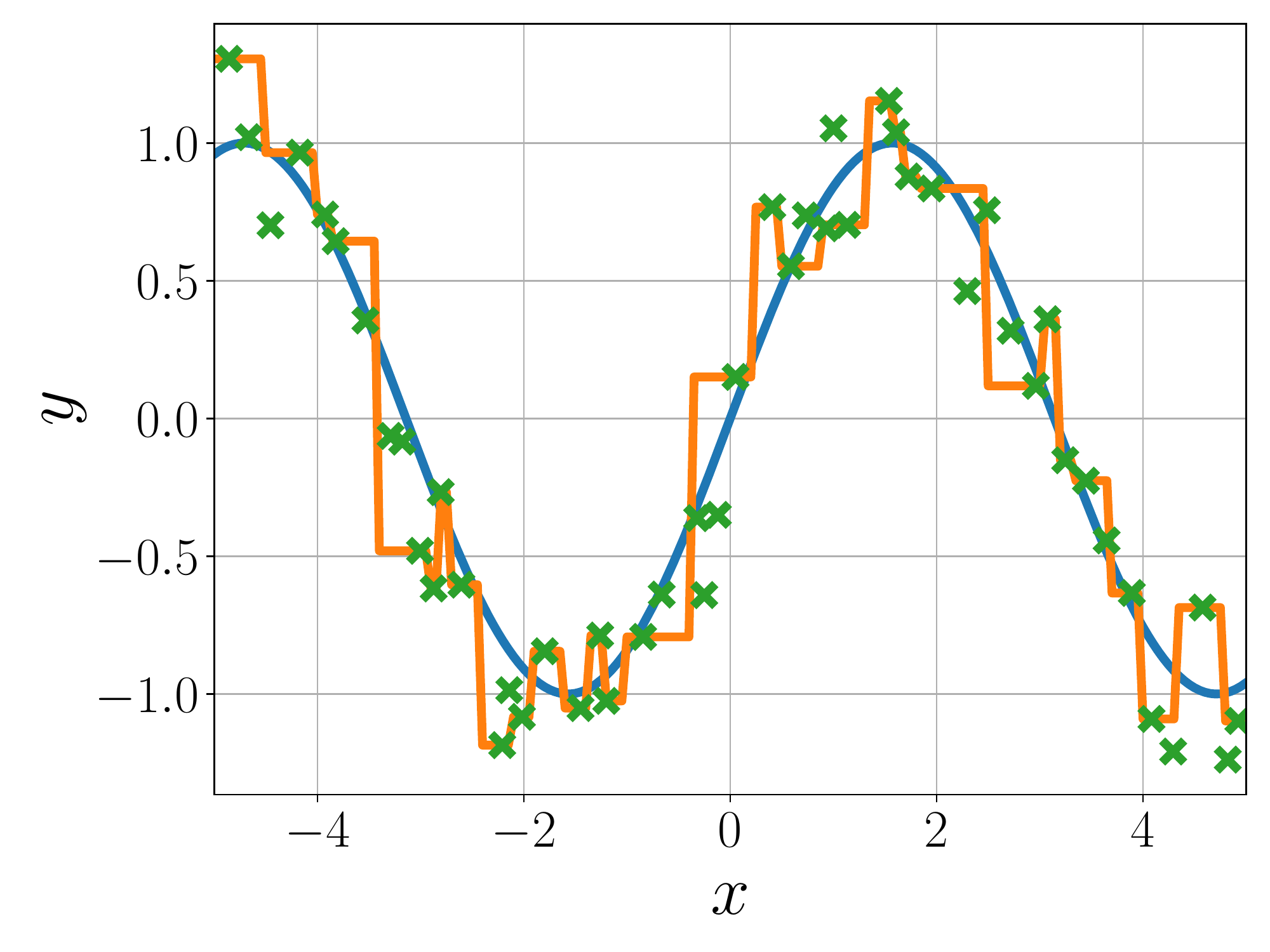}
		\includegraphics[width=0.235\textwidth]{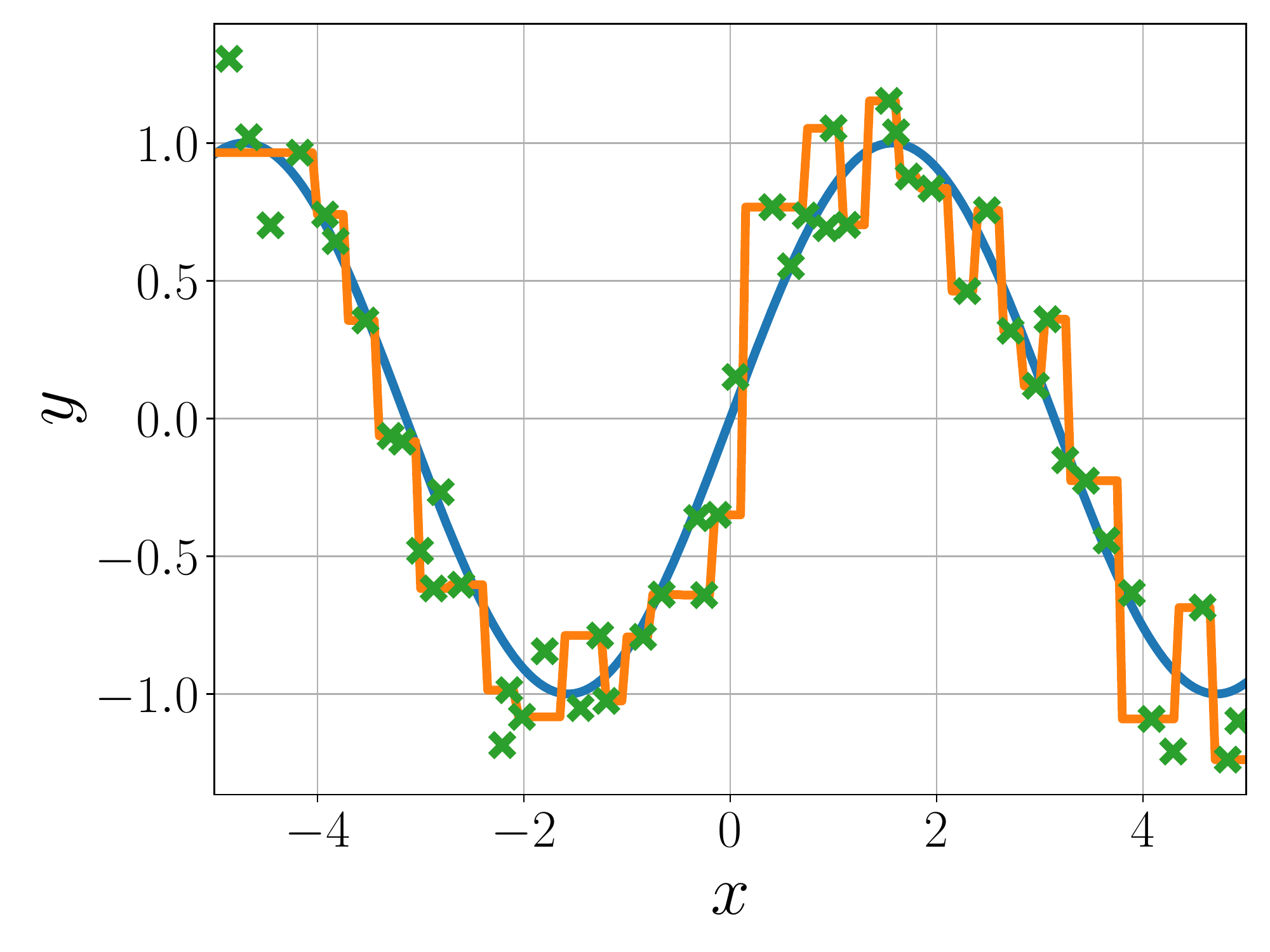}
		\includegraphics[width=0.235\textwidth]{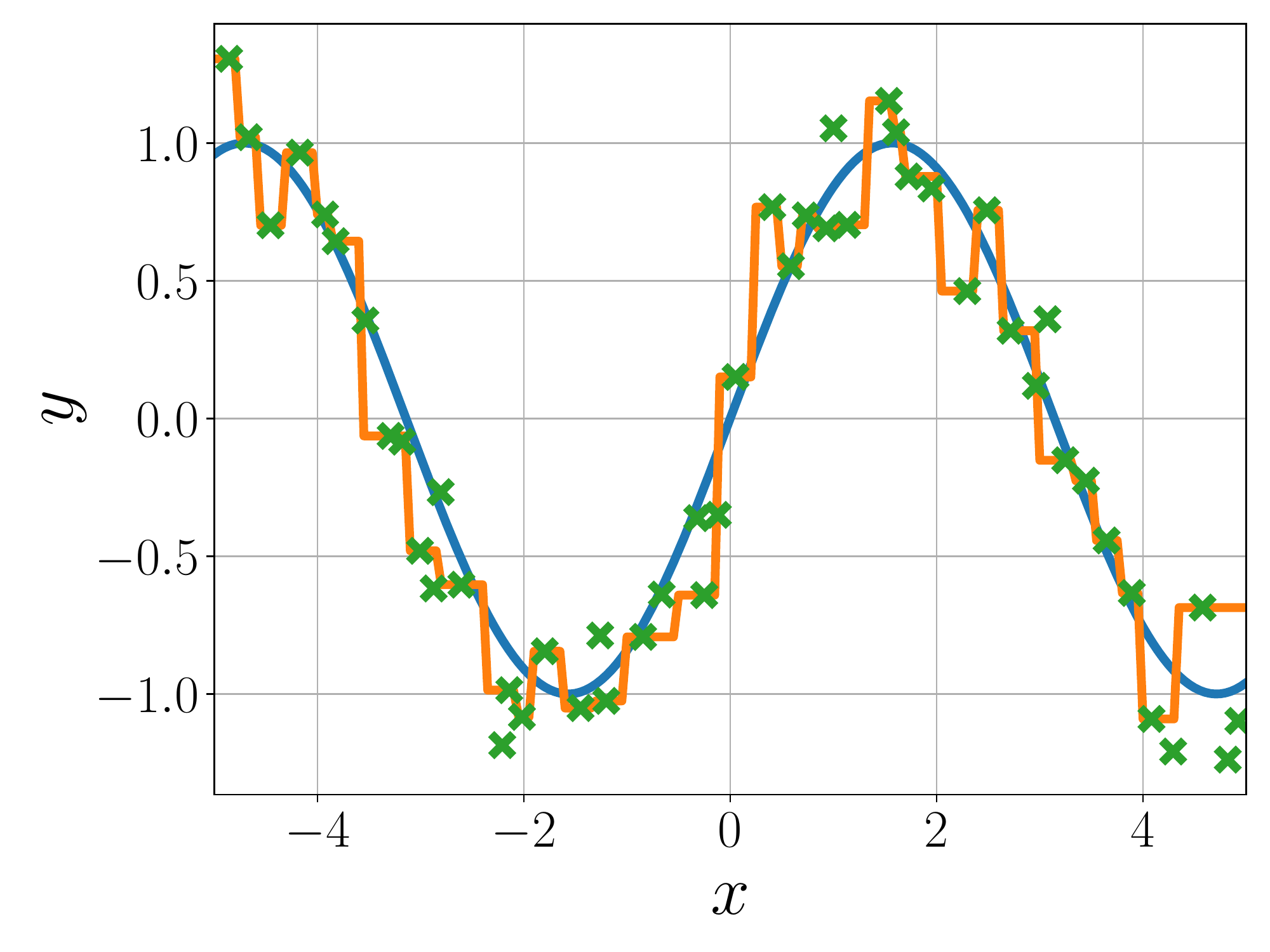}
		\includegraphics[width=0.235\textwidth]{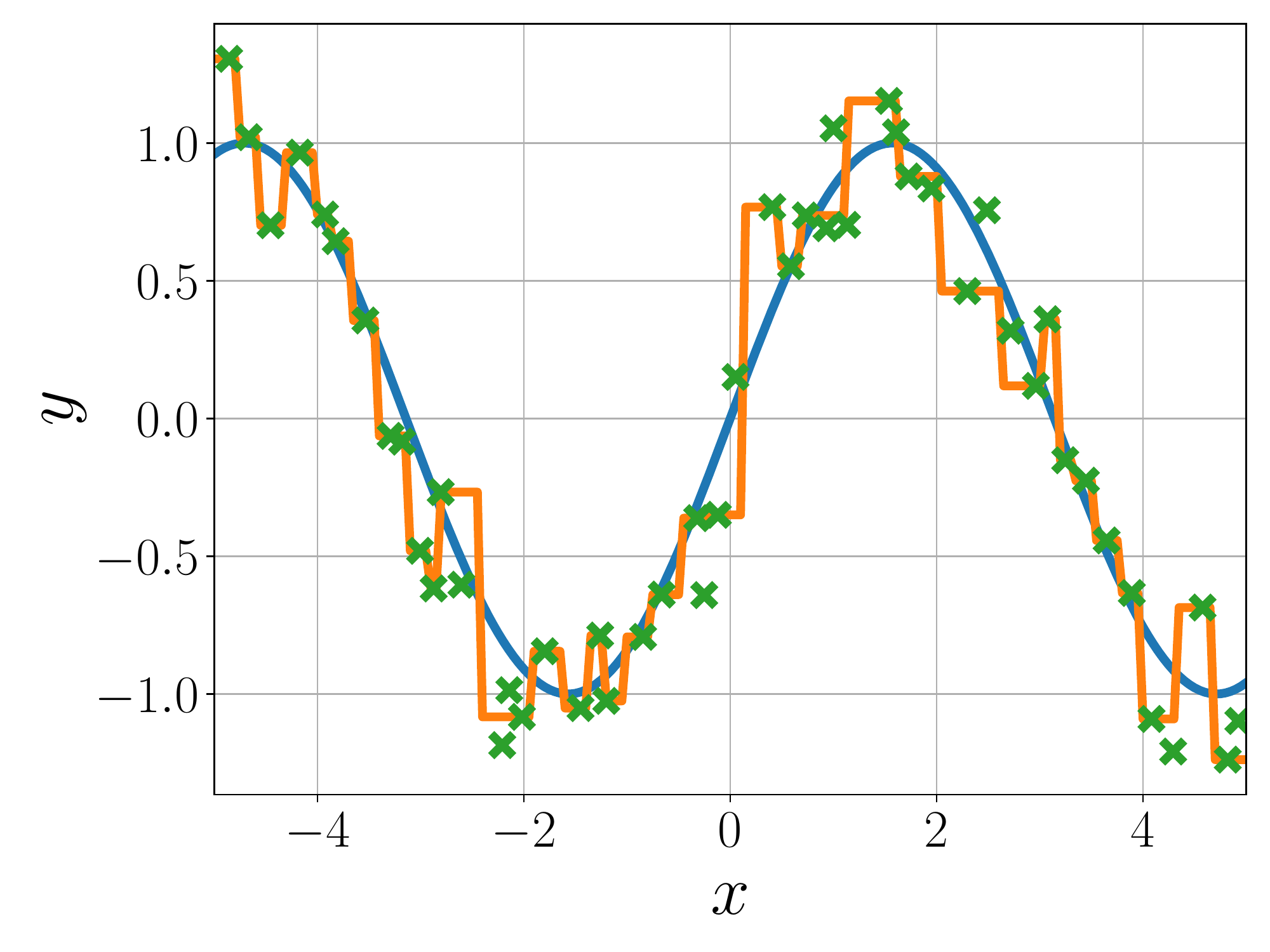}
		\label{fig:unc_1d_many_trees_b}
	}
	\subfigure[R (originally proposed as extremely randomized trees)]{
		\includegraphics[width=0.235\textwidth]{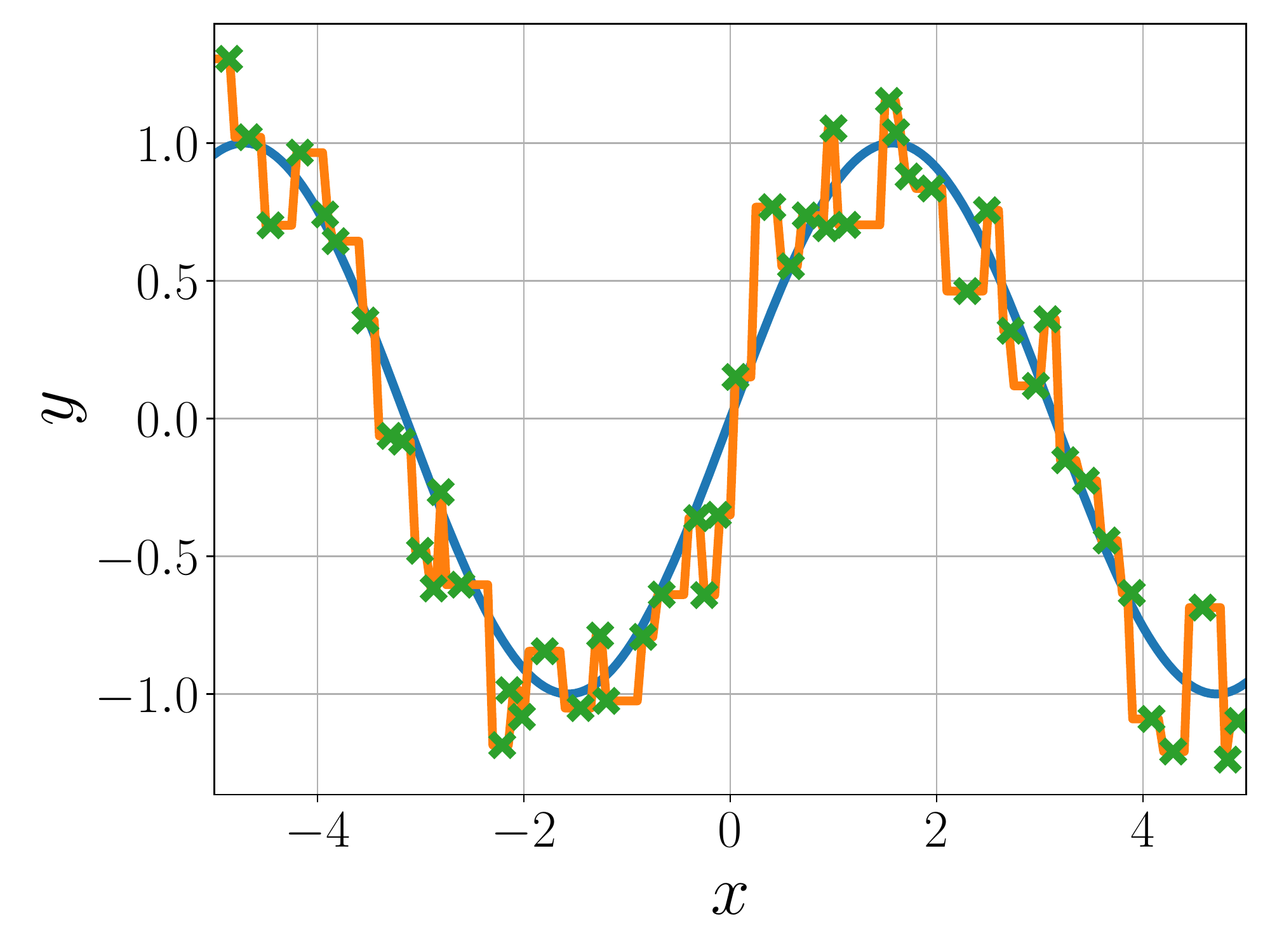}
		\includegraphics[width=0.235\textwidth]{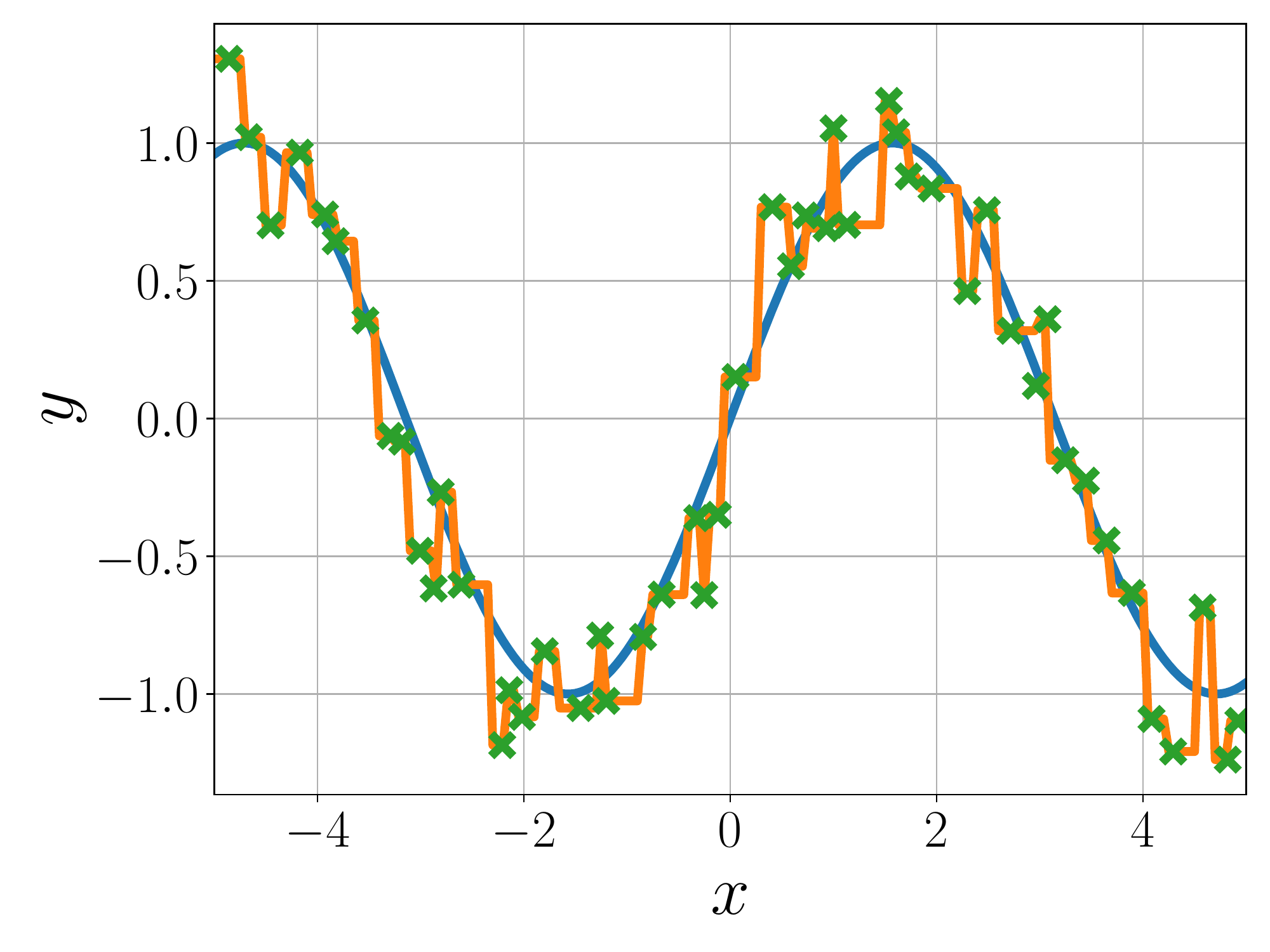}
		\includegraphics[width=0.235\textwidth]{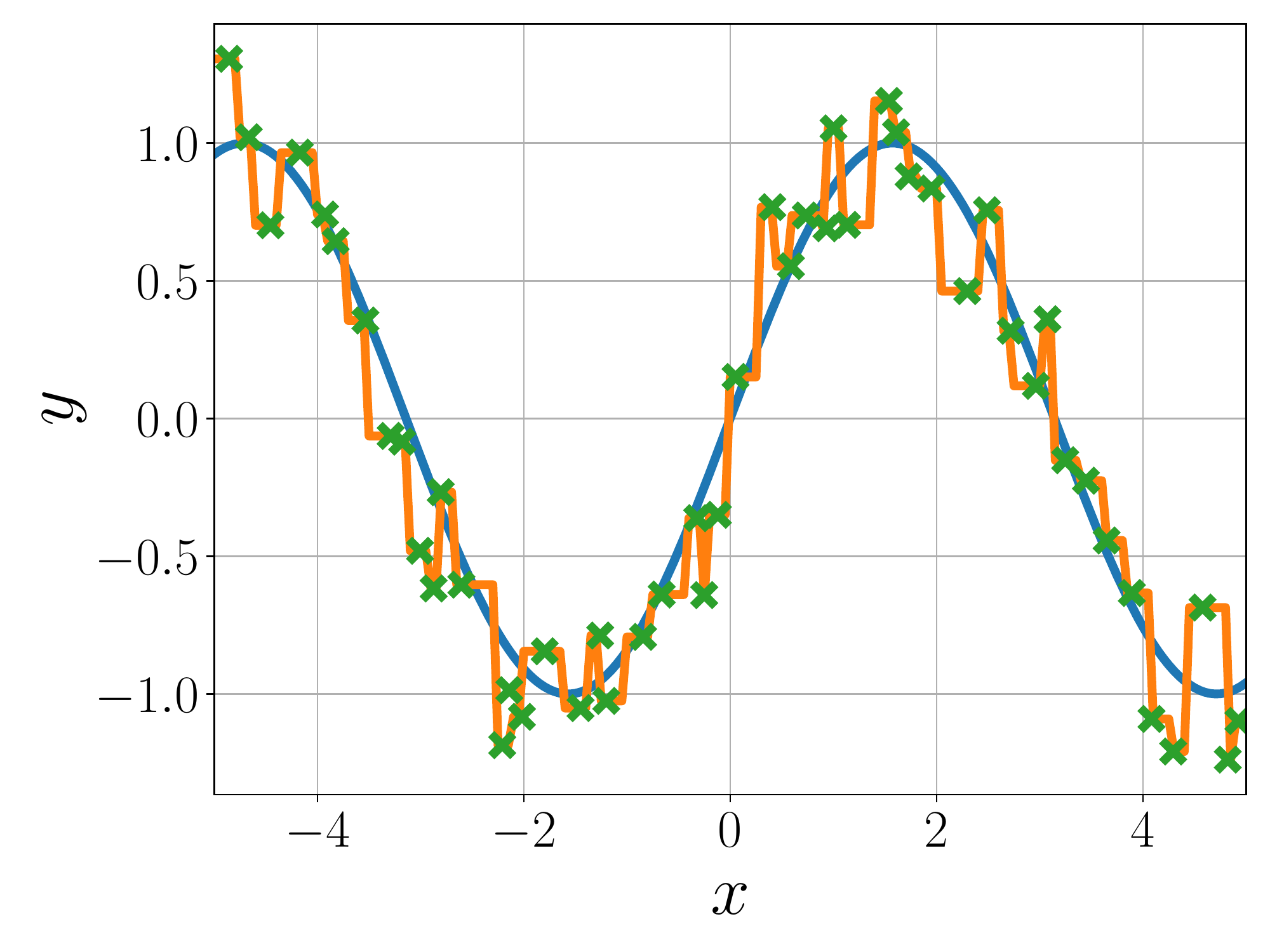}
		\includegraphics[width=0.235\textwidth]{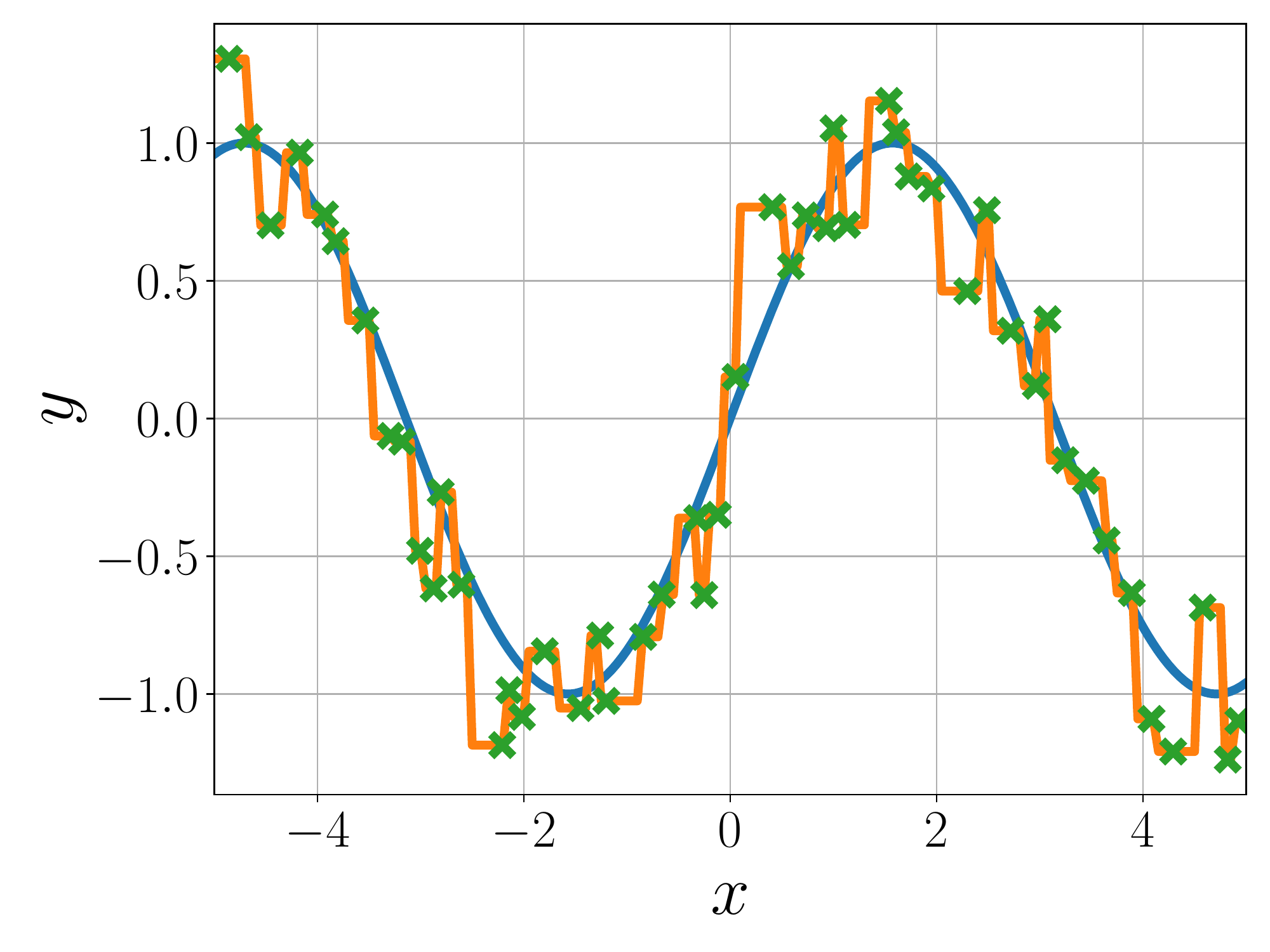}
		\label{fig:unc_1d_many_trees_r}
	}
	\subfigure[B + O]{
		\includegraphics[width=0.235\textwidth]{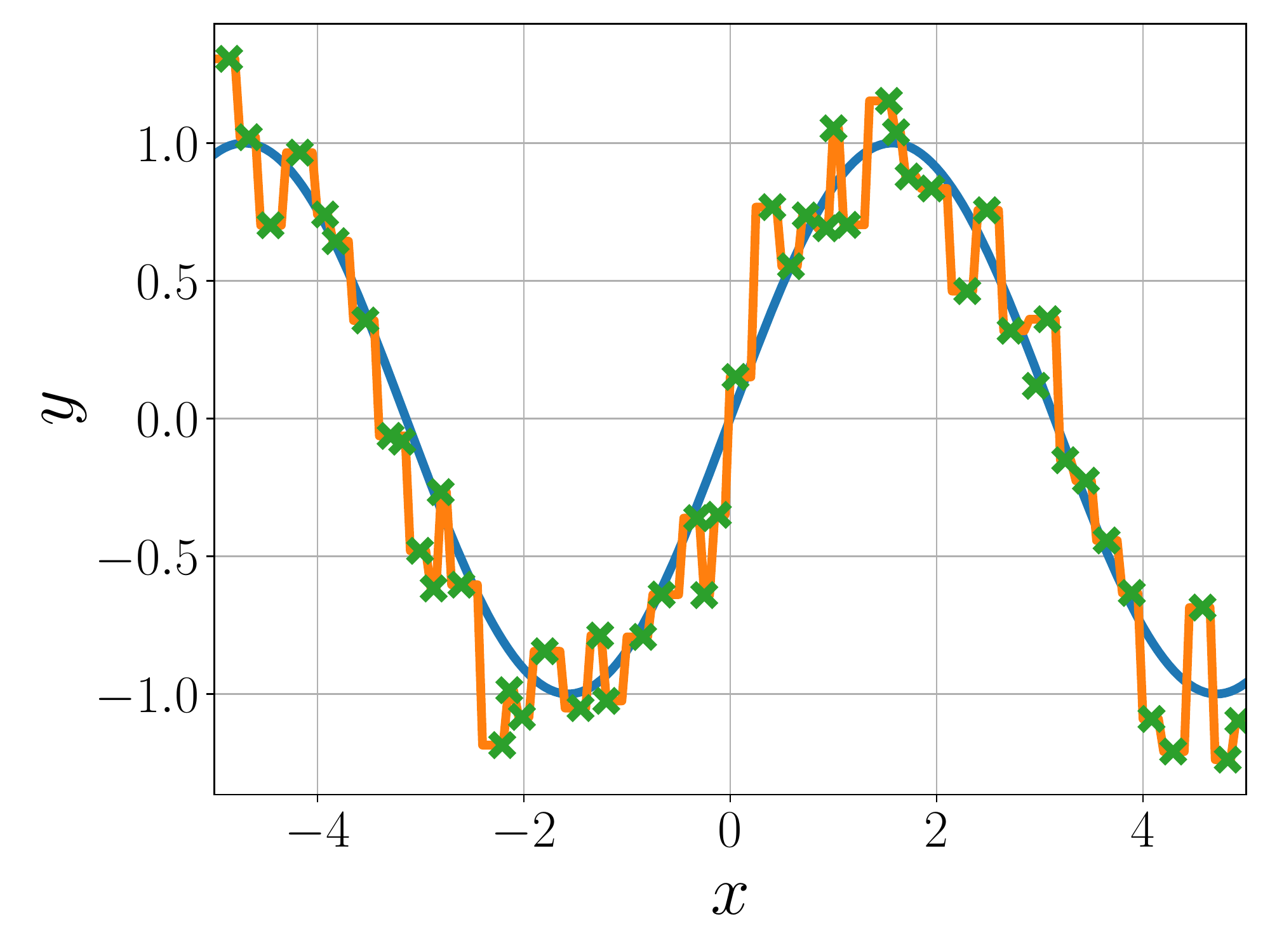}
		\includegraphics[width=0.235\textwidth]{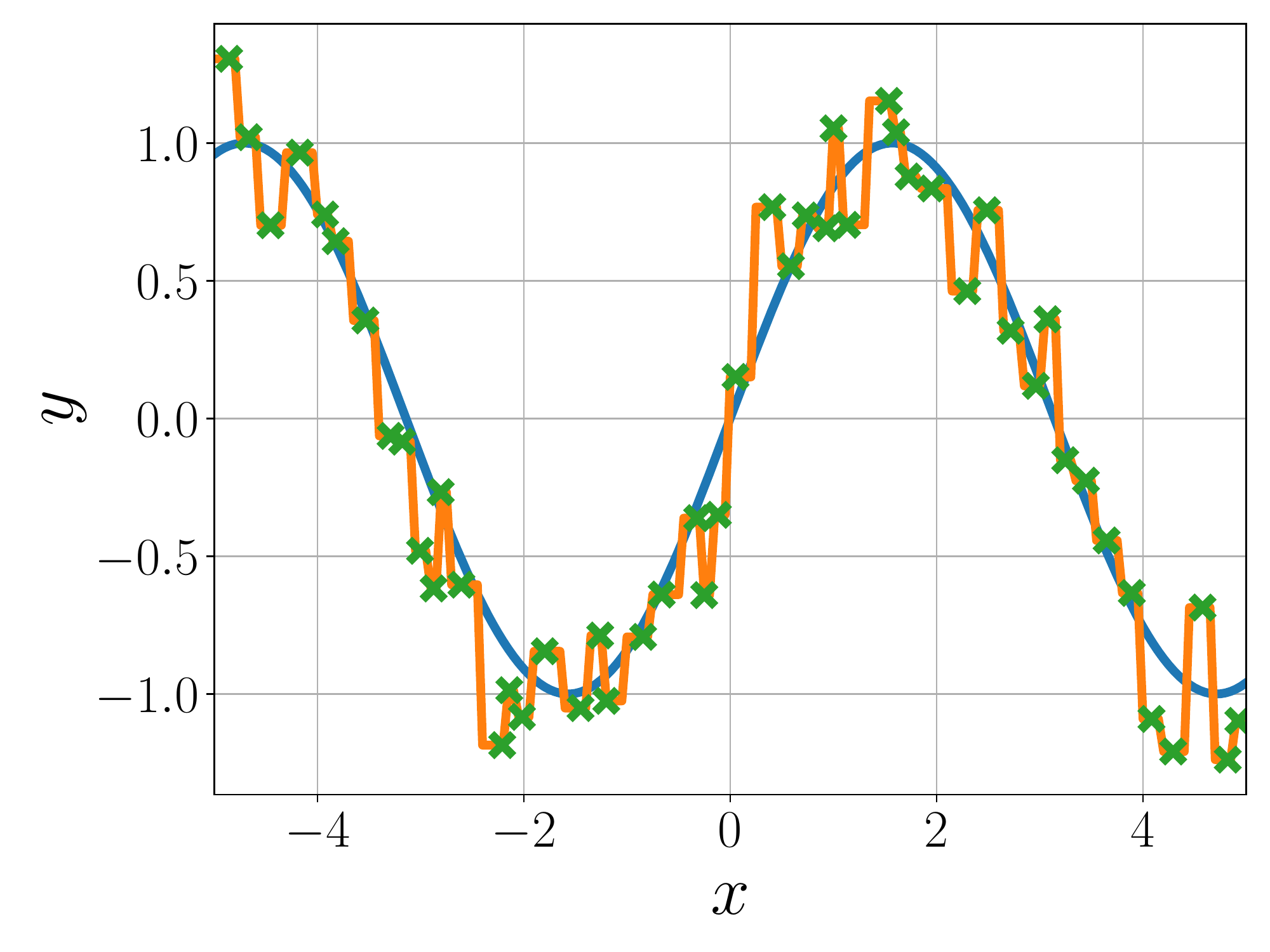}
		\includegraphics[width=0.235\textwidth]{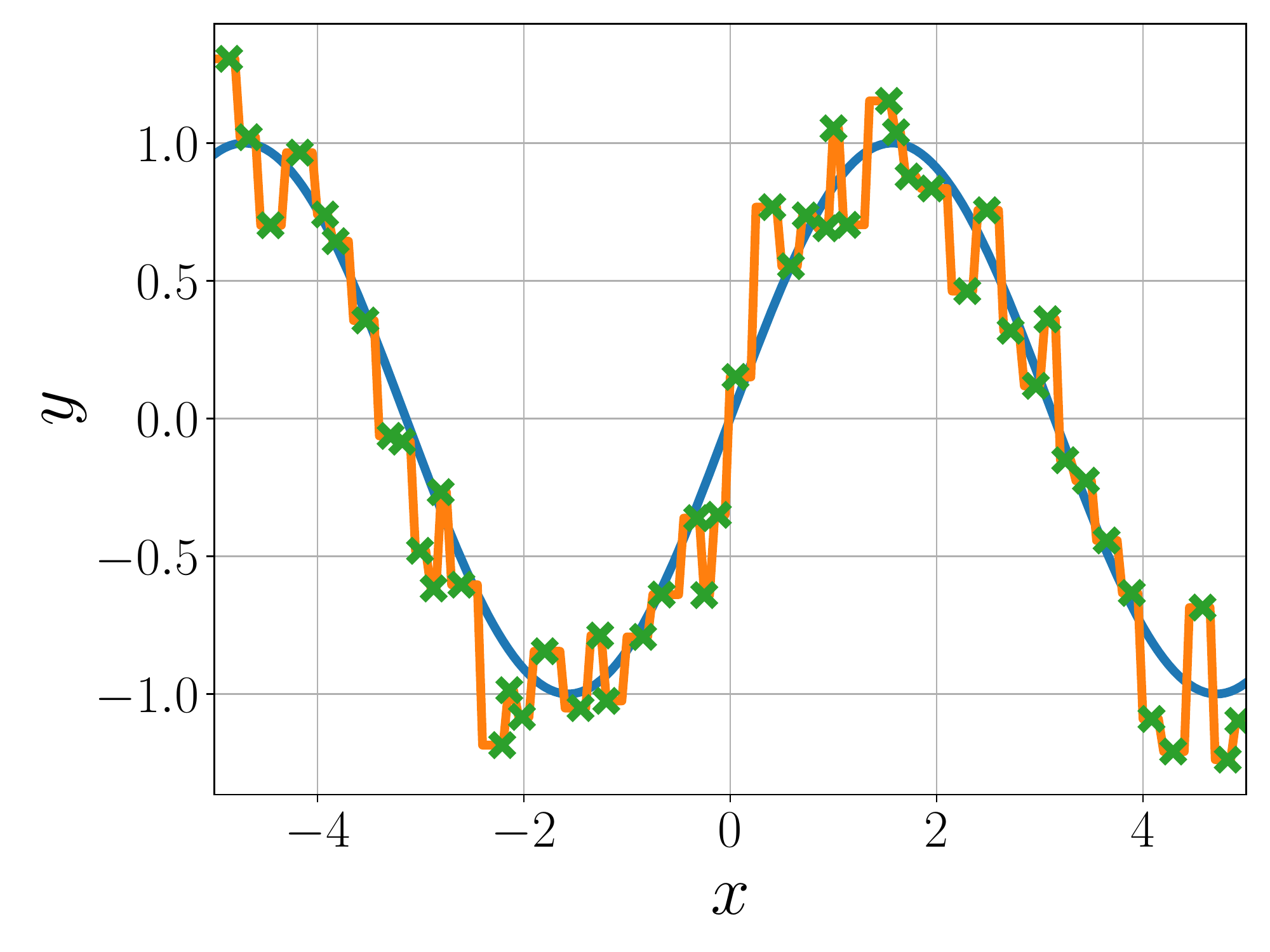}
		\includegraphics[width=0.235\textwidth]{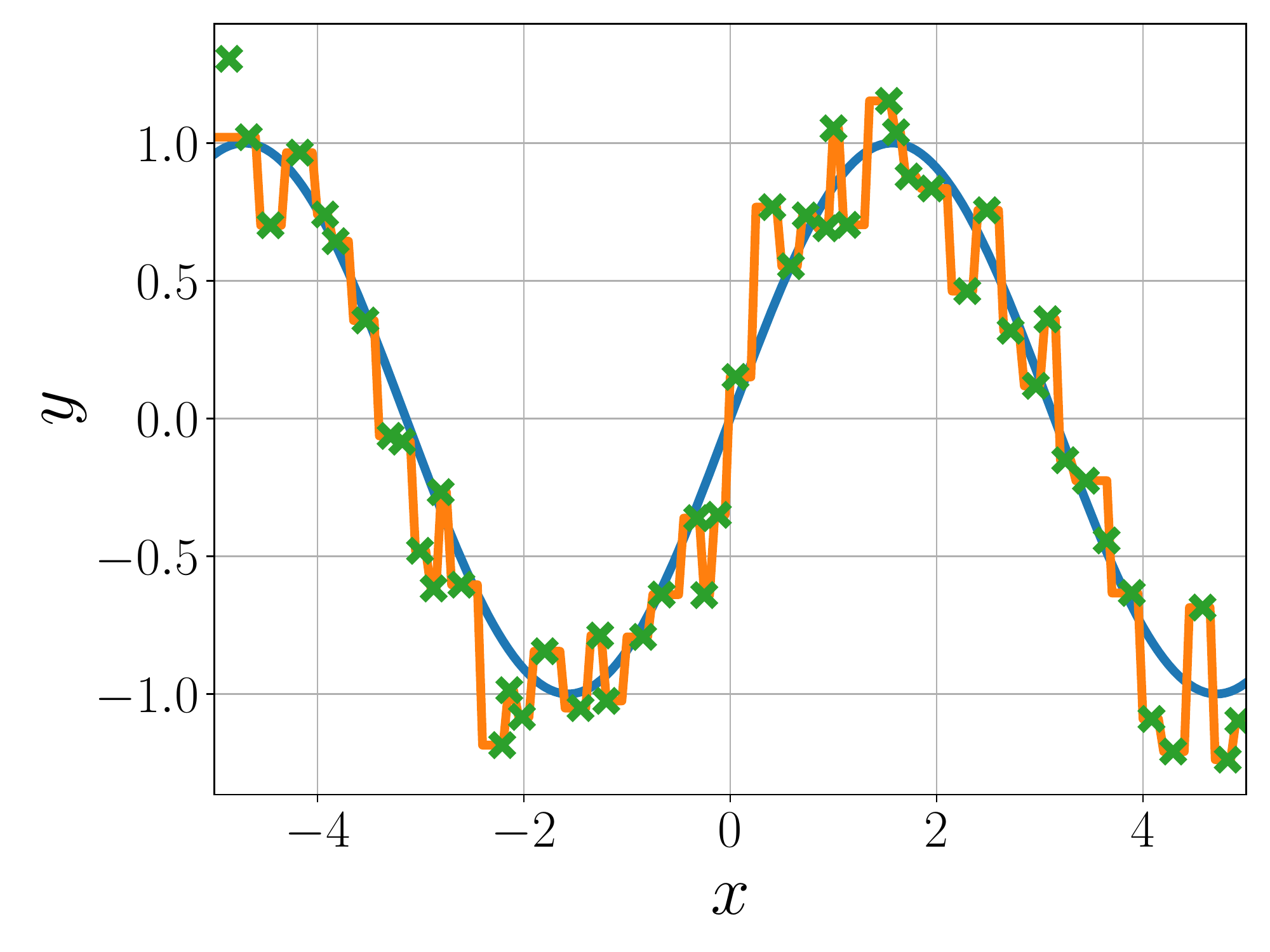}
		\label{fig:unc_1d_many_trees_bo}
	}
	\subfigure[R + B]{
		\includegraphics[width=0.235\textwidth]{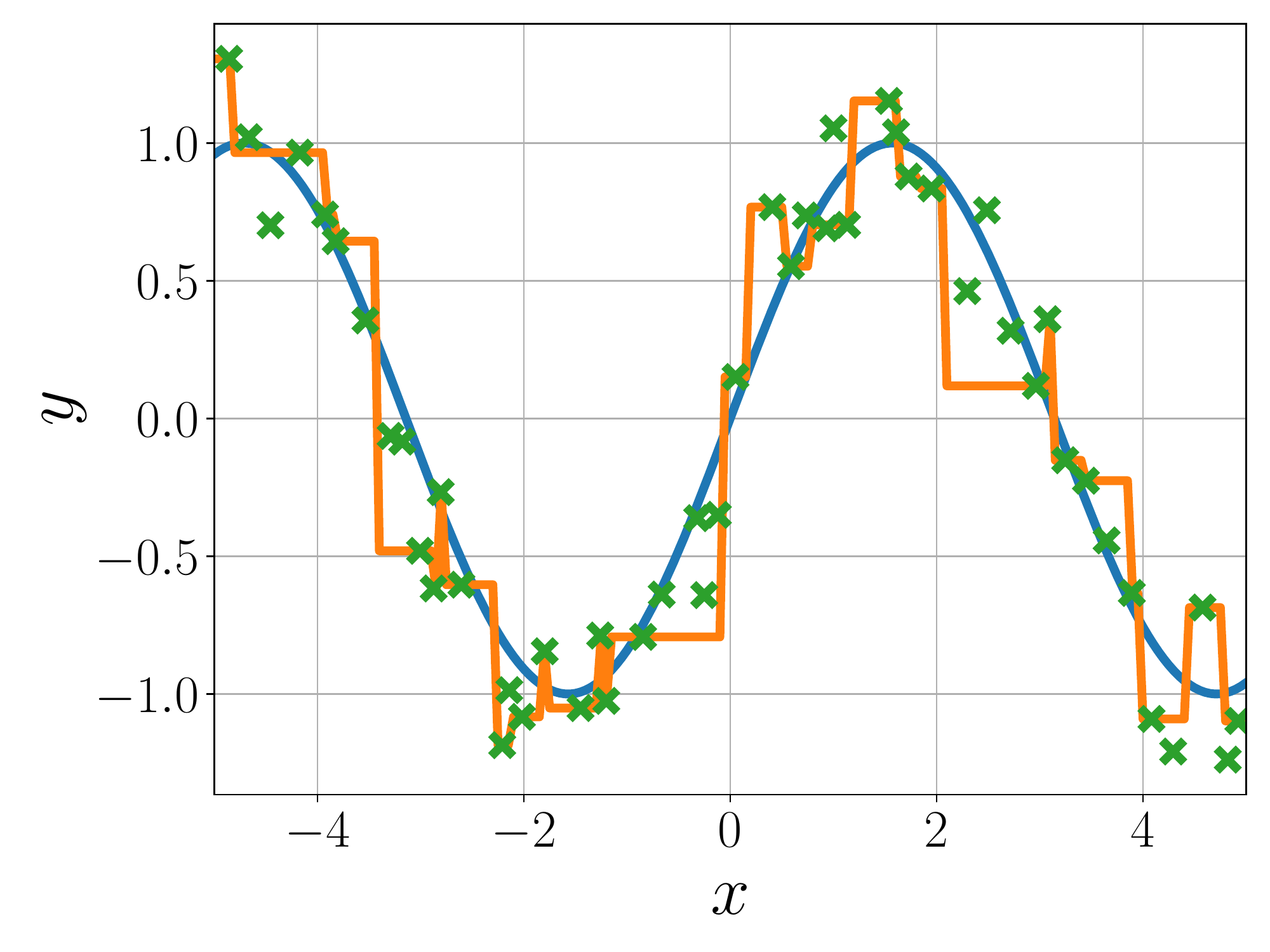}
		\includegraphics[width=0.235\textwidth]{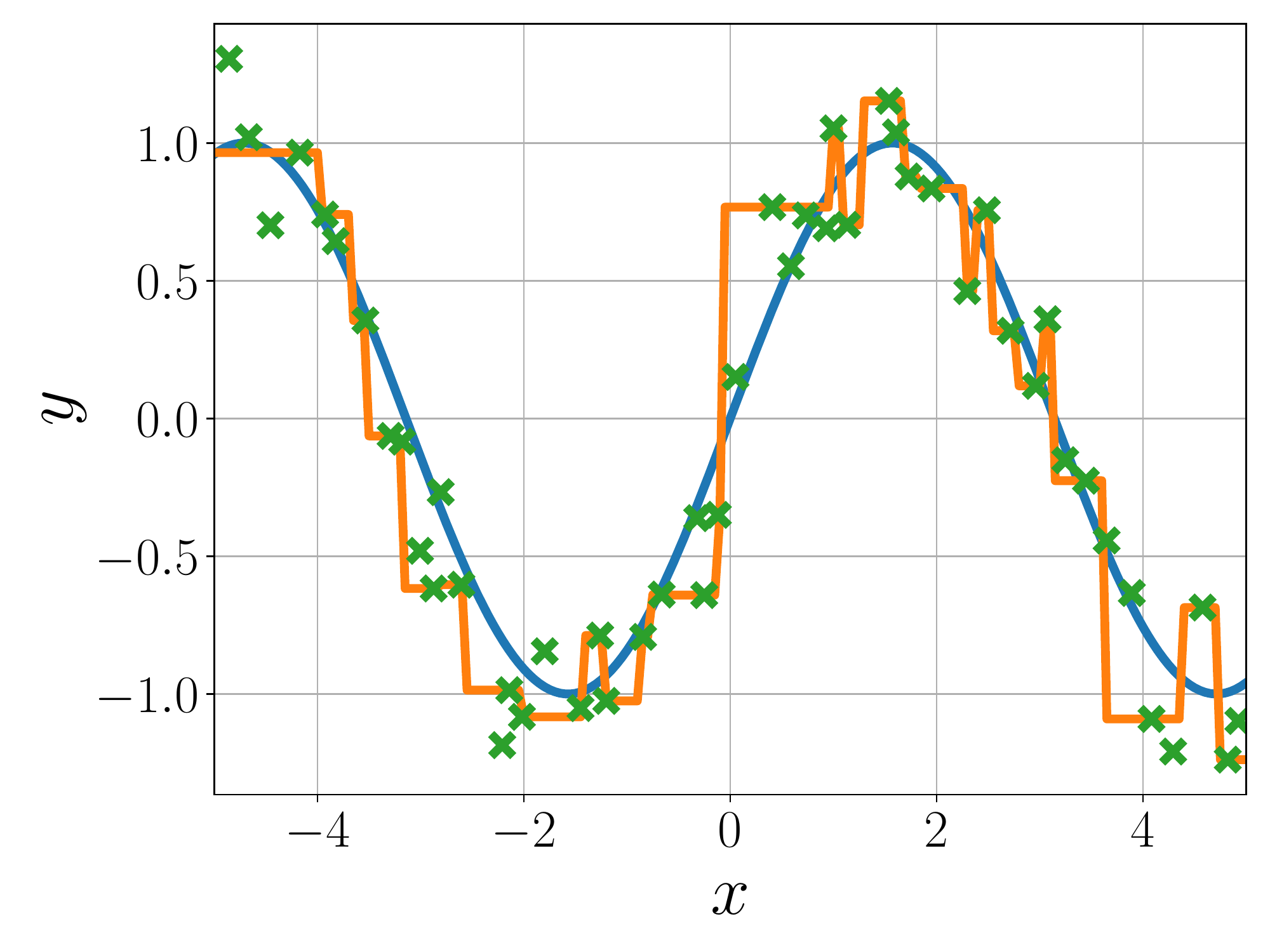}
		\includegraphics[width=0.235\textwidth]{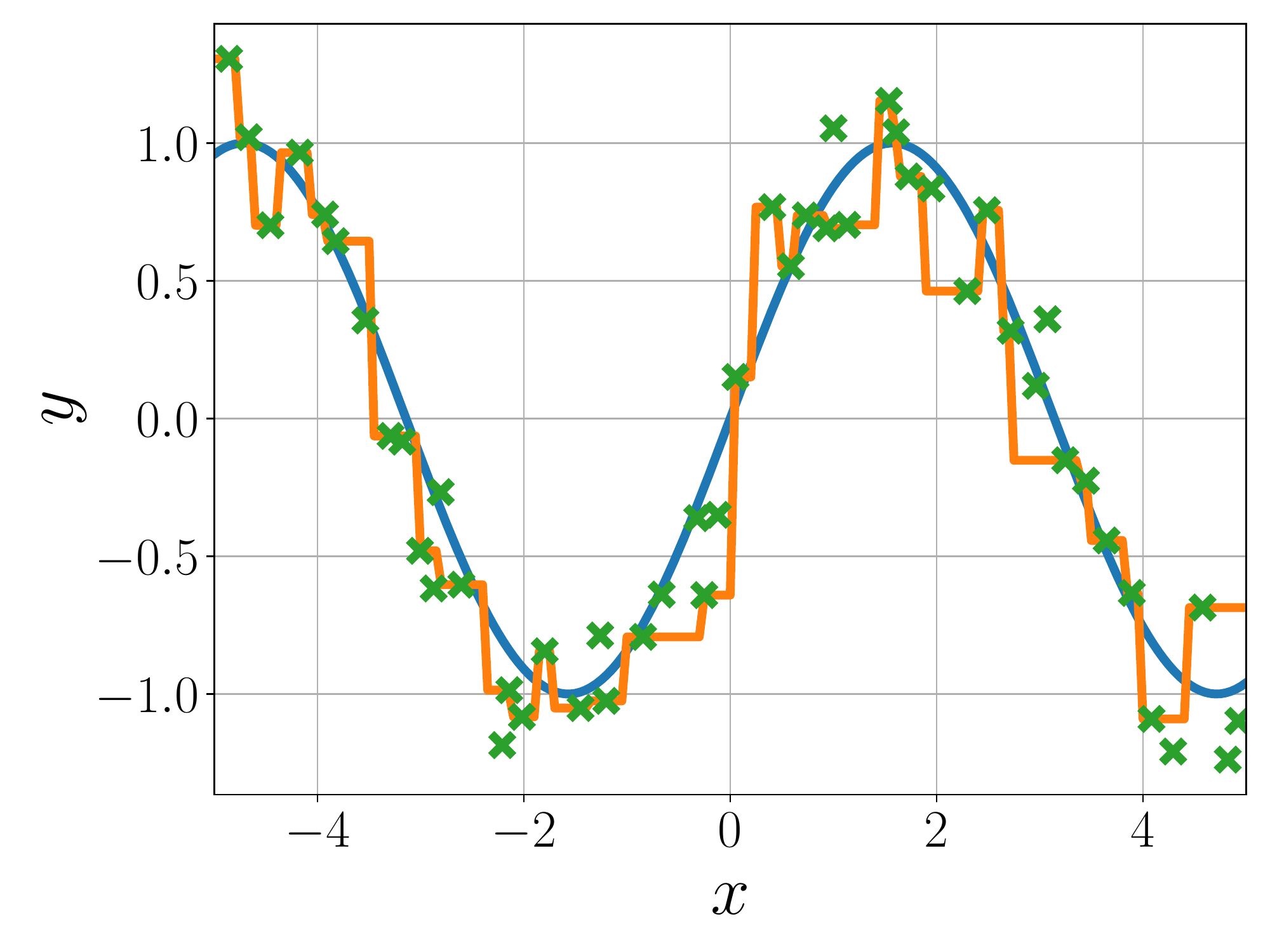}
		\includegraphics[width=0.235\textwidth]{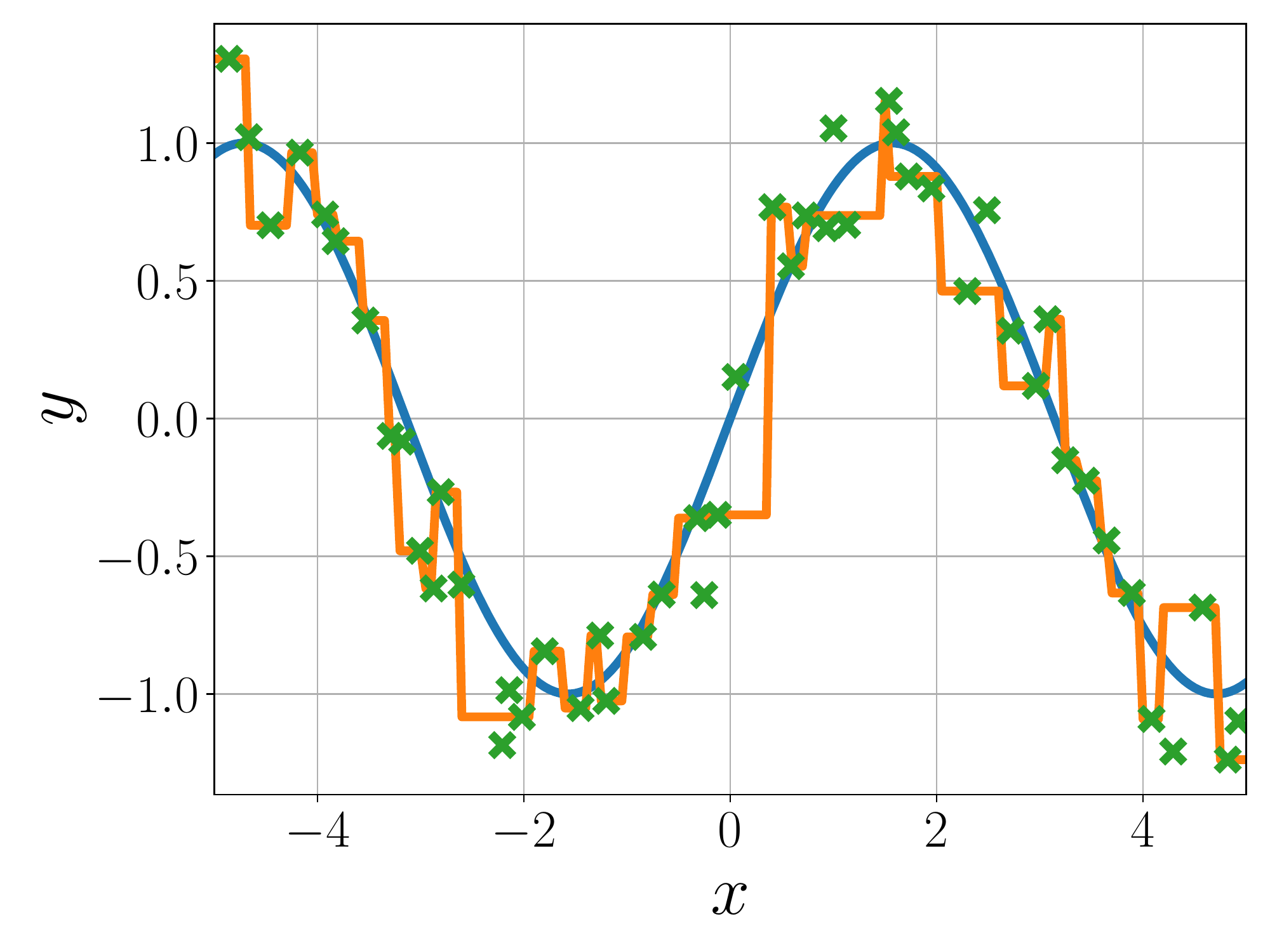}
		\label{fig:unc_1d_many_trees_rb}
	}
	\subfigure[R + B + O (i.e., BwO forest)]{
		\includegraphics[width=0.235\textwidth]{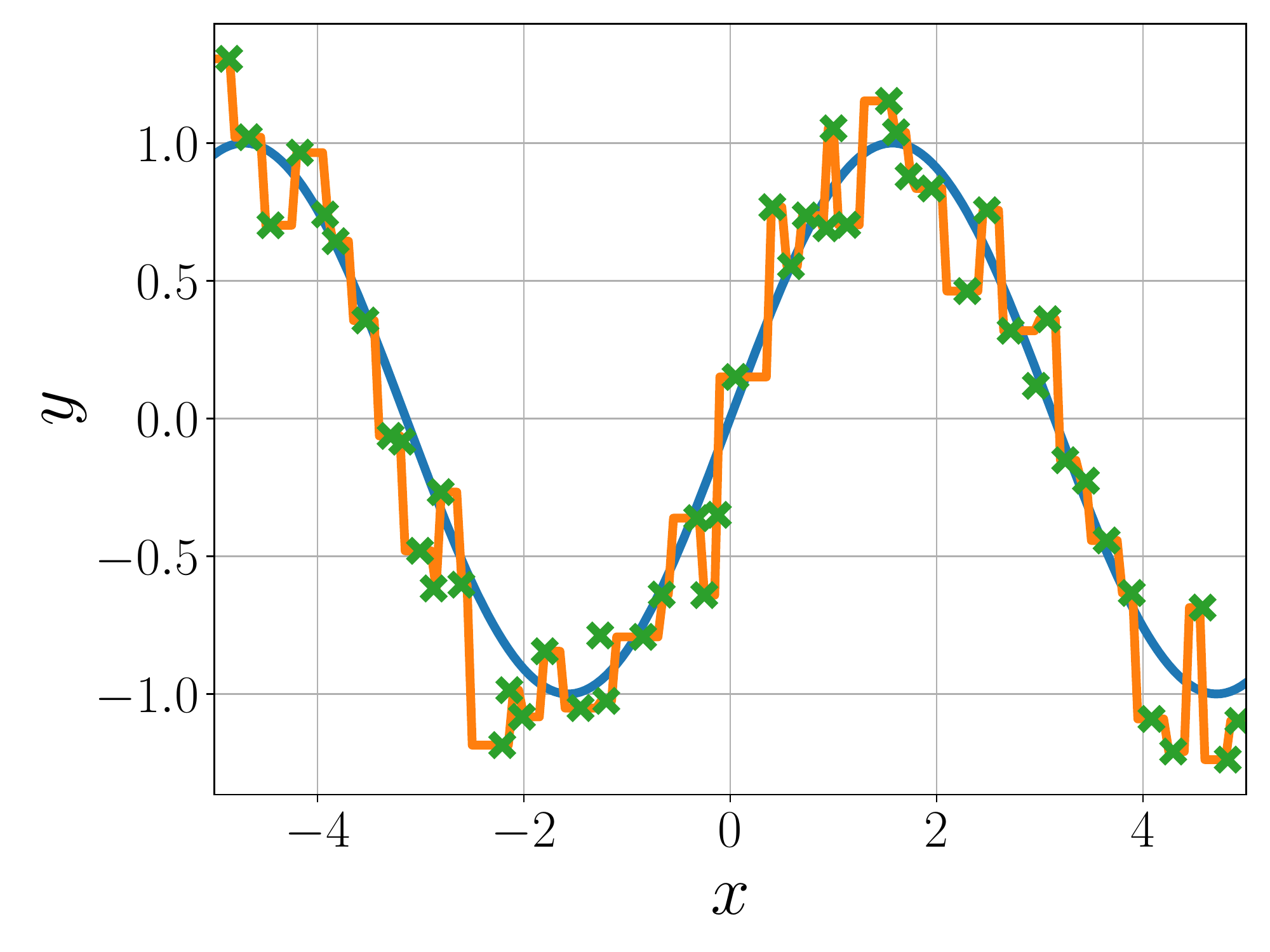}
		\includegraphics[width=0.235\textwidth]{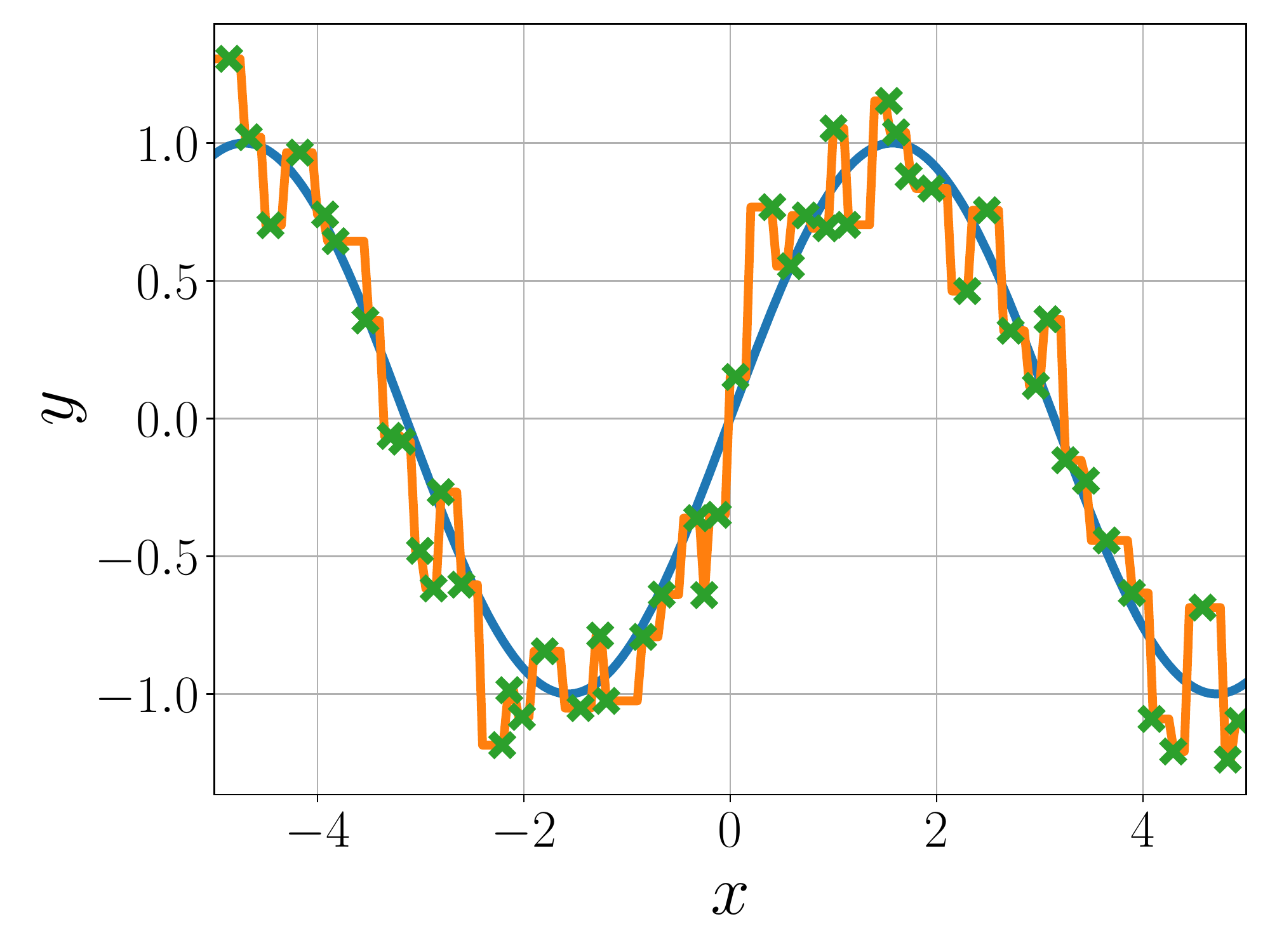}
		\includegraphics[width=0.235\textwidth]{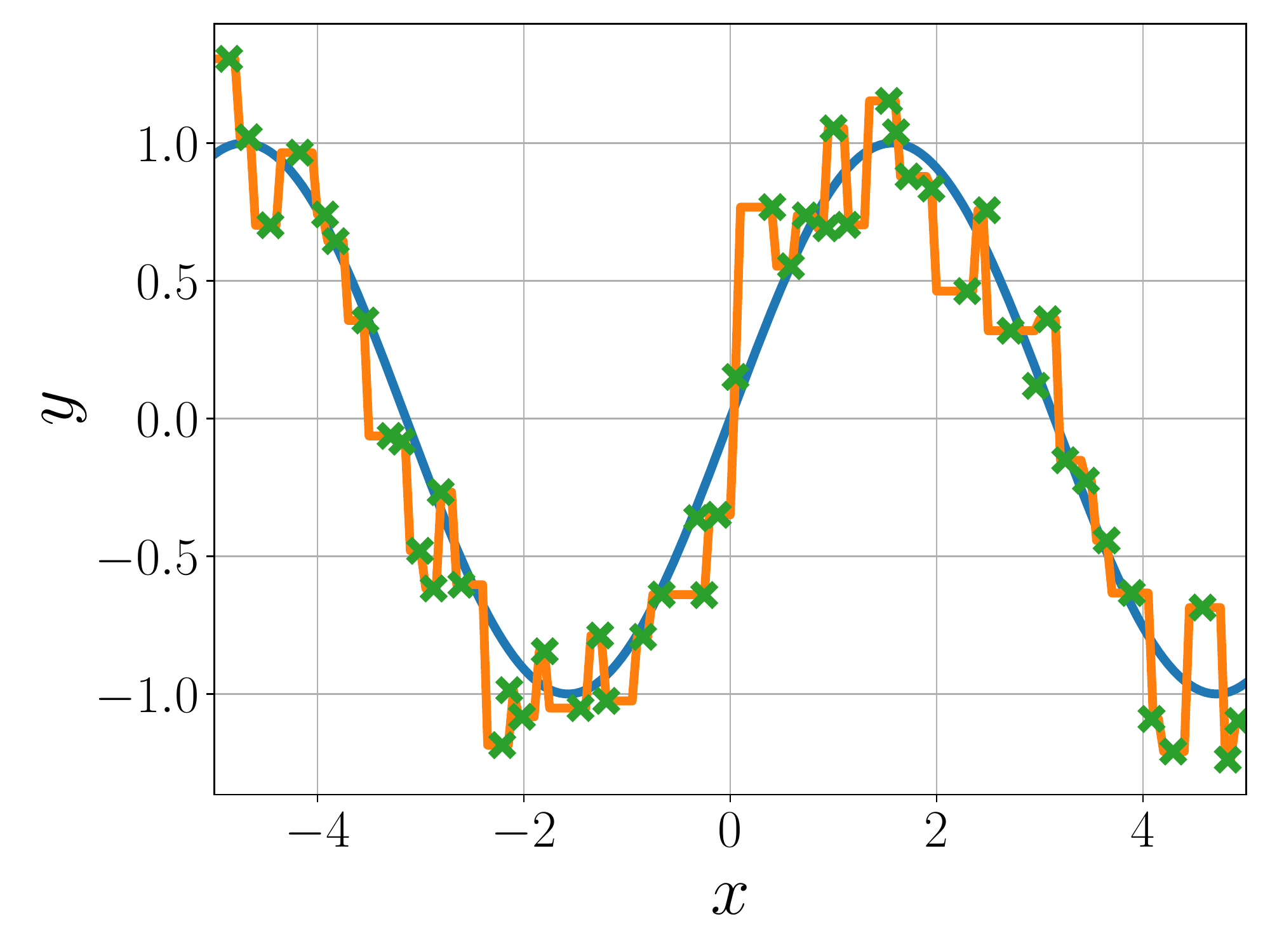}
		\includegraphics[width=0.235\textwidth]{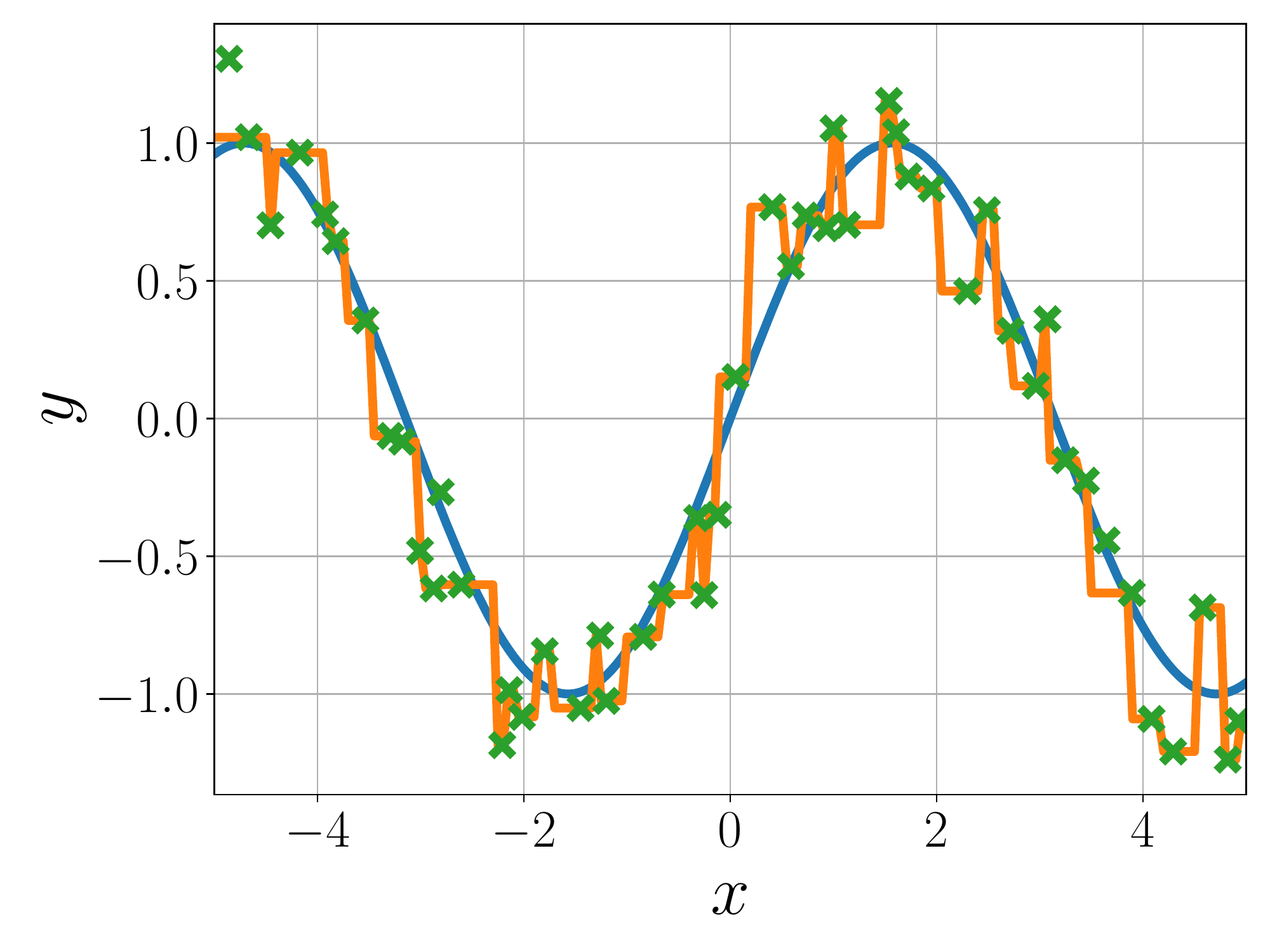}
		\label{fig:unc_1d_many_trees_rbo}
	}
	\caption{Results by individual trees for the case shown in \figref{fig:unc_1d_many}.\label{fig:unc_1d_many_trees}}
\end{figure}

\begin{figure}[p]
	\centering
	\subfigure[B (originally proposed as random forest)]{
		\includegraphics[width=0.235\textwidth]{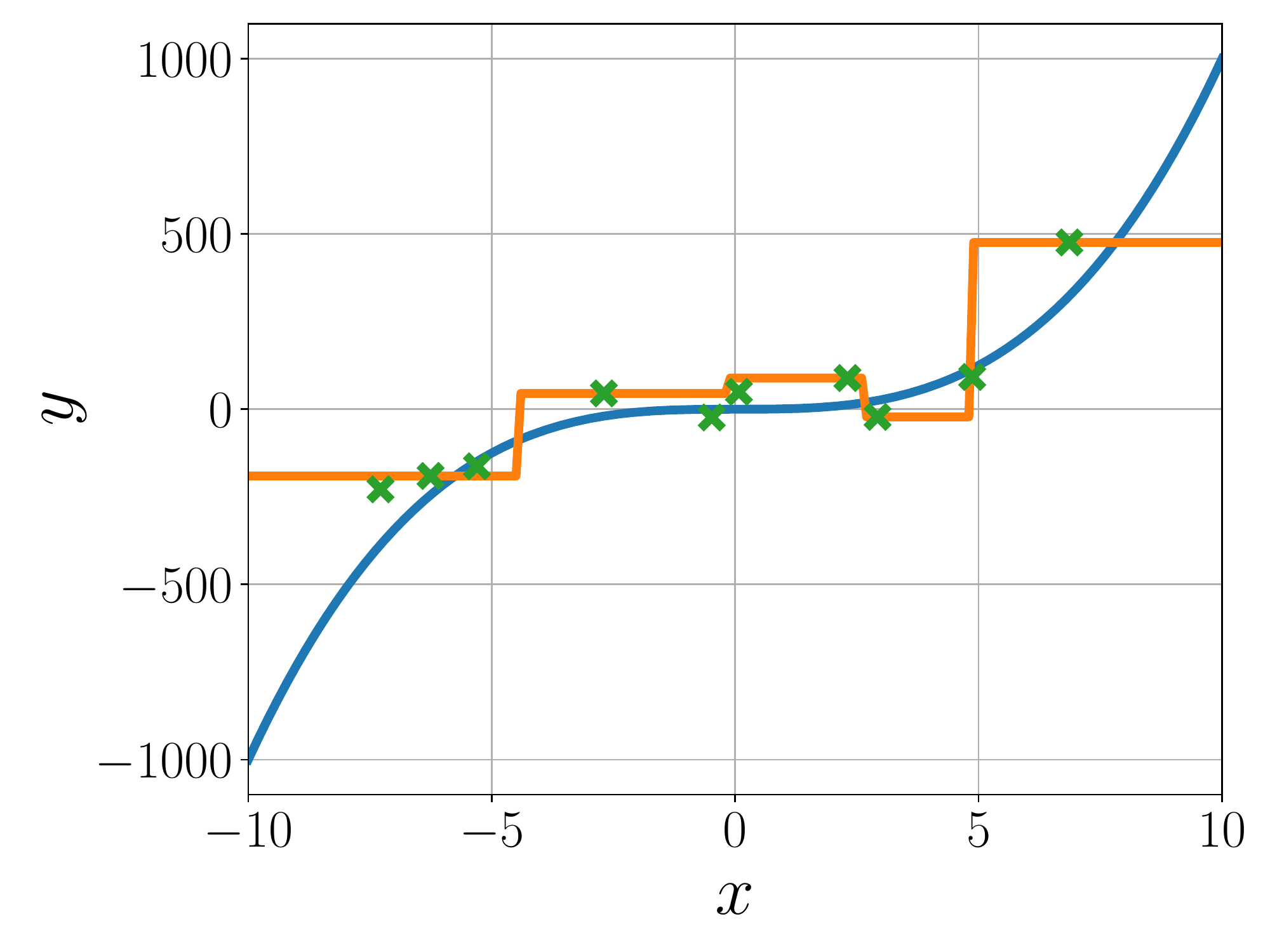}
		\includegraphics[width=0.235\textwidth]{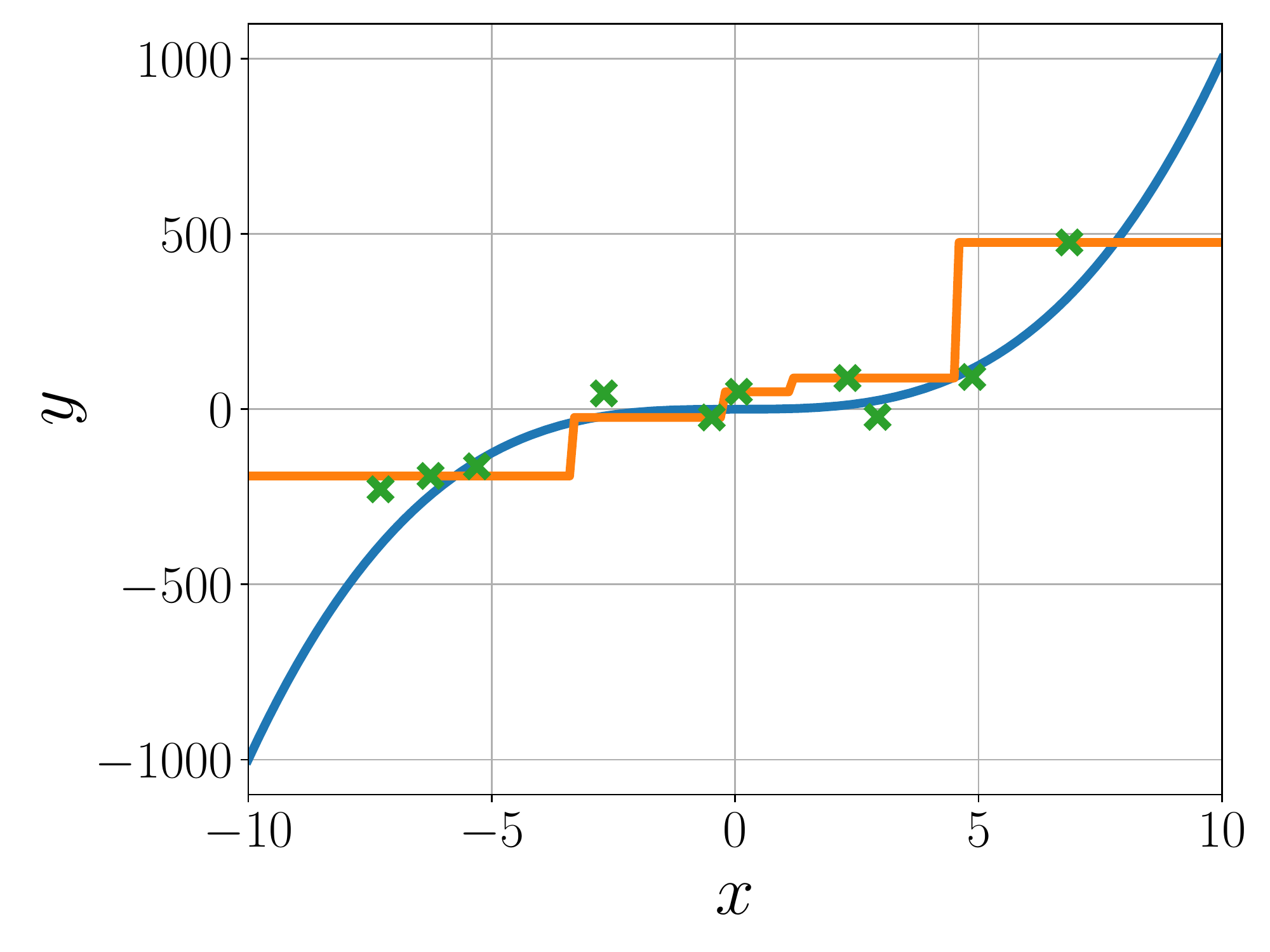}
		\includegraphics[width=0.235\textwidth]{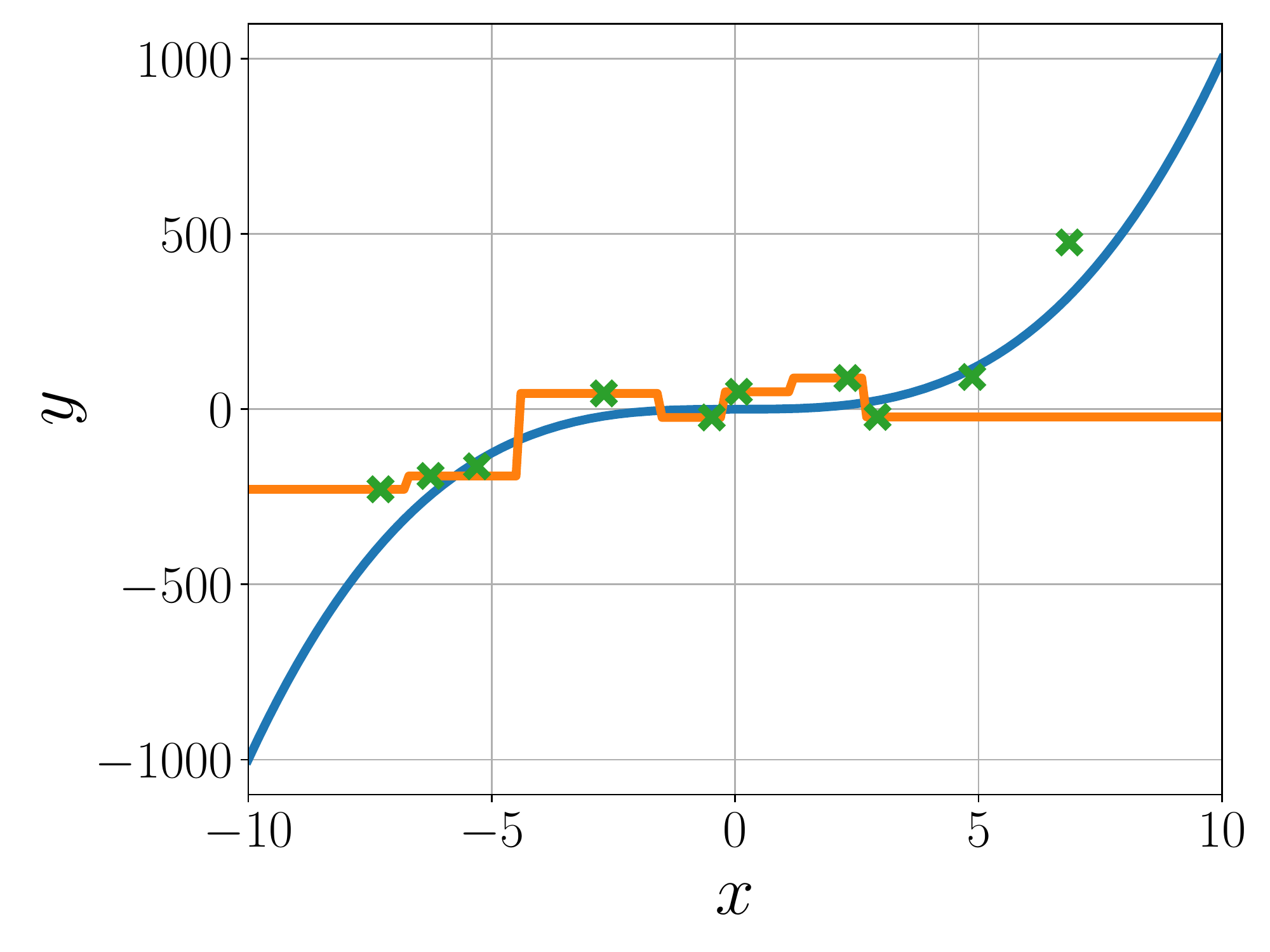}
		\includegraphics[width=0.235\textwidth]{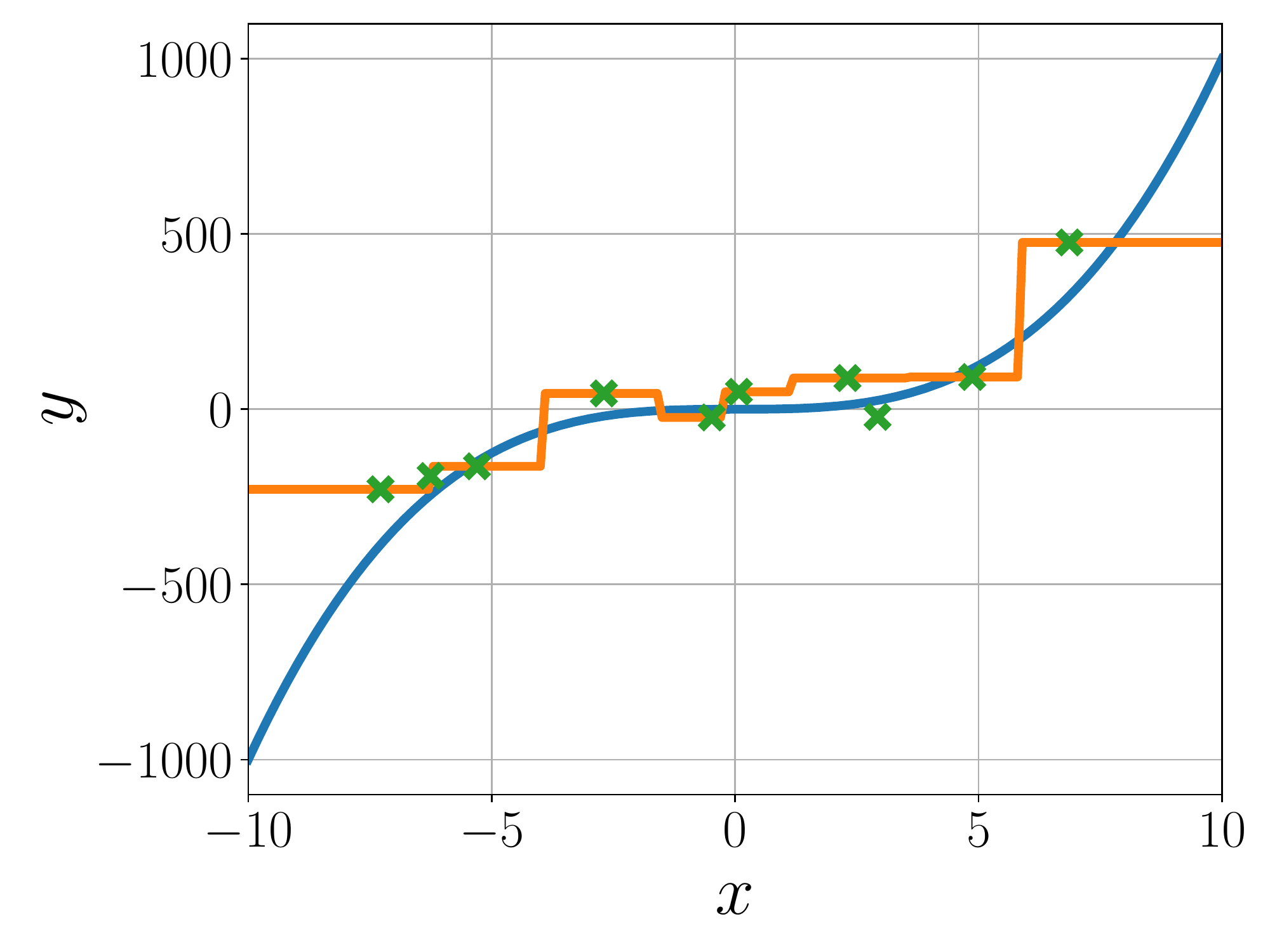}
		\label{fig:unc_1d_cubic_trees_b}
	}
	\subfigure[R (originally proposed as extremely randomized trees)]{
		\includegraphics[width=0.235\textwidth]{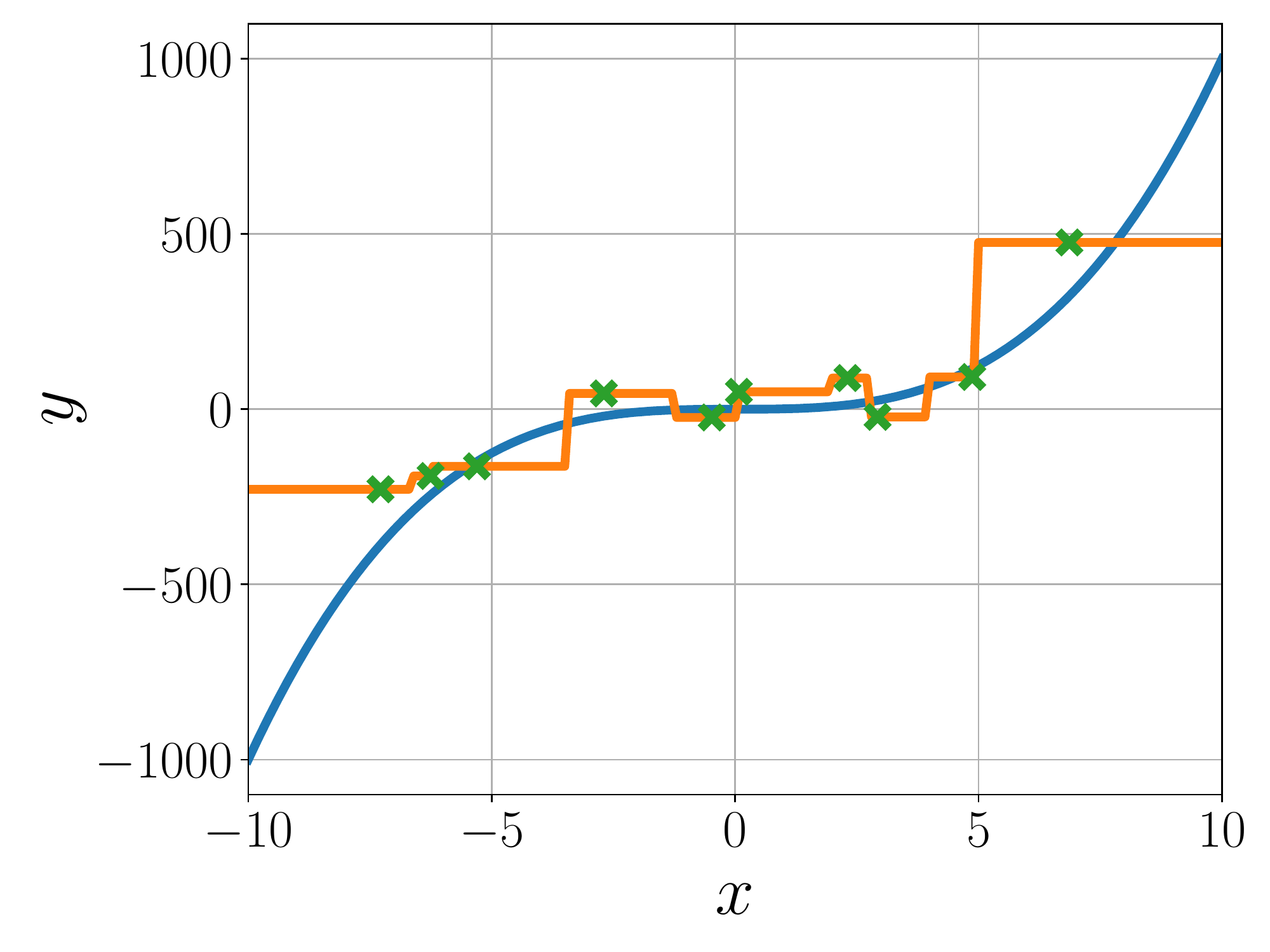}
		\includegraphics[width=0.235\textwidth]{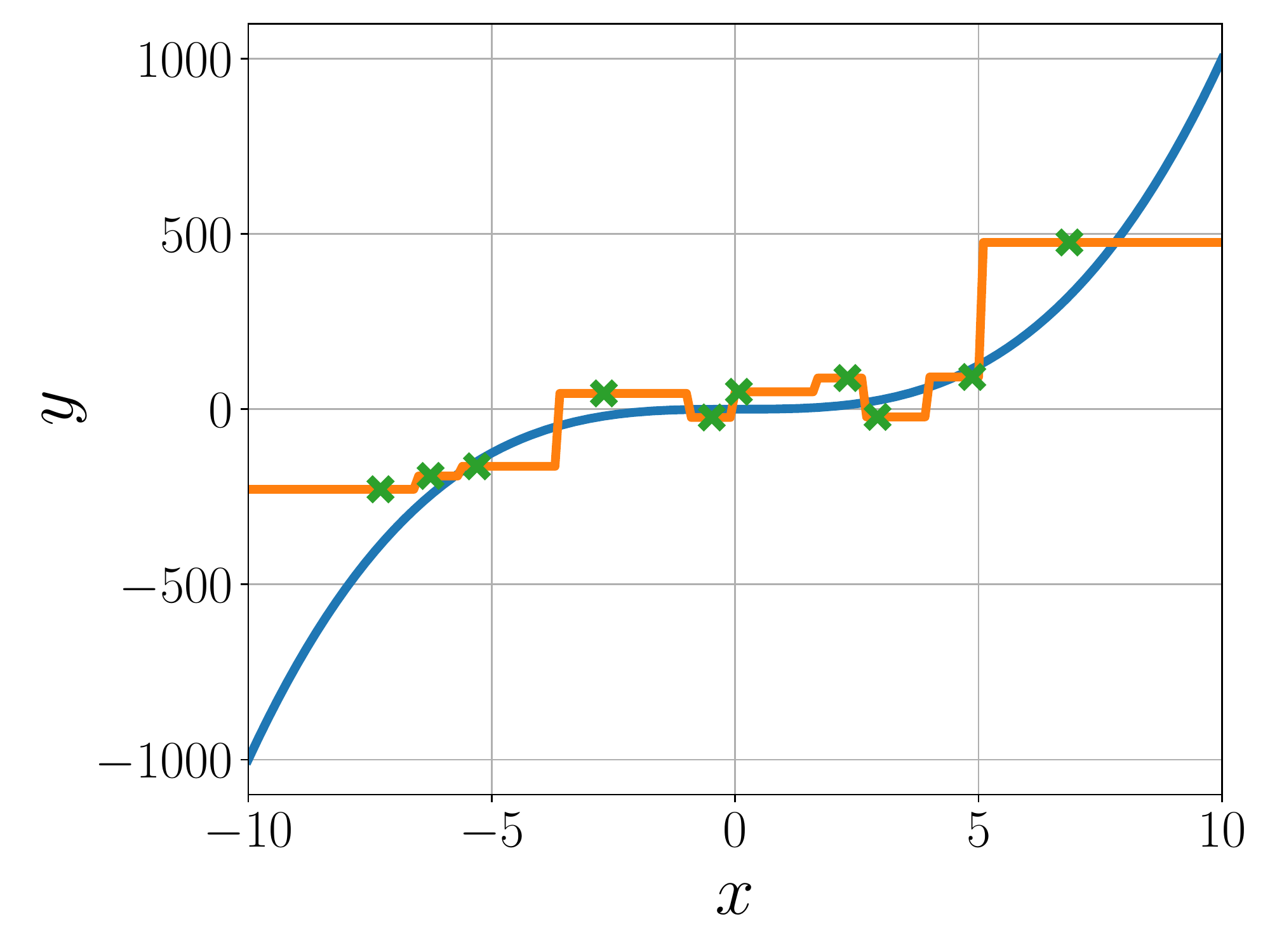}
		\includegraphics[width=0.235\textwidth]{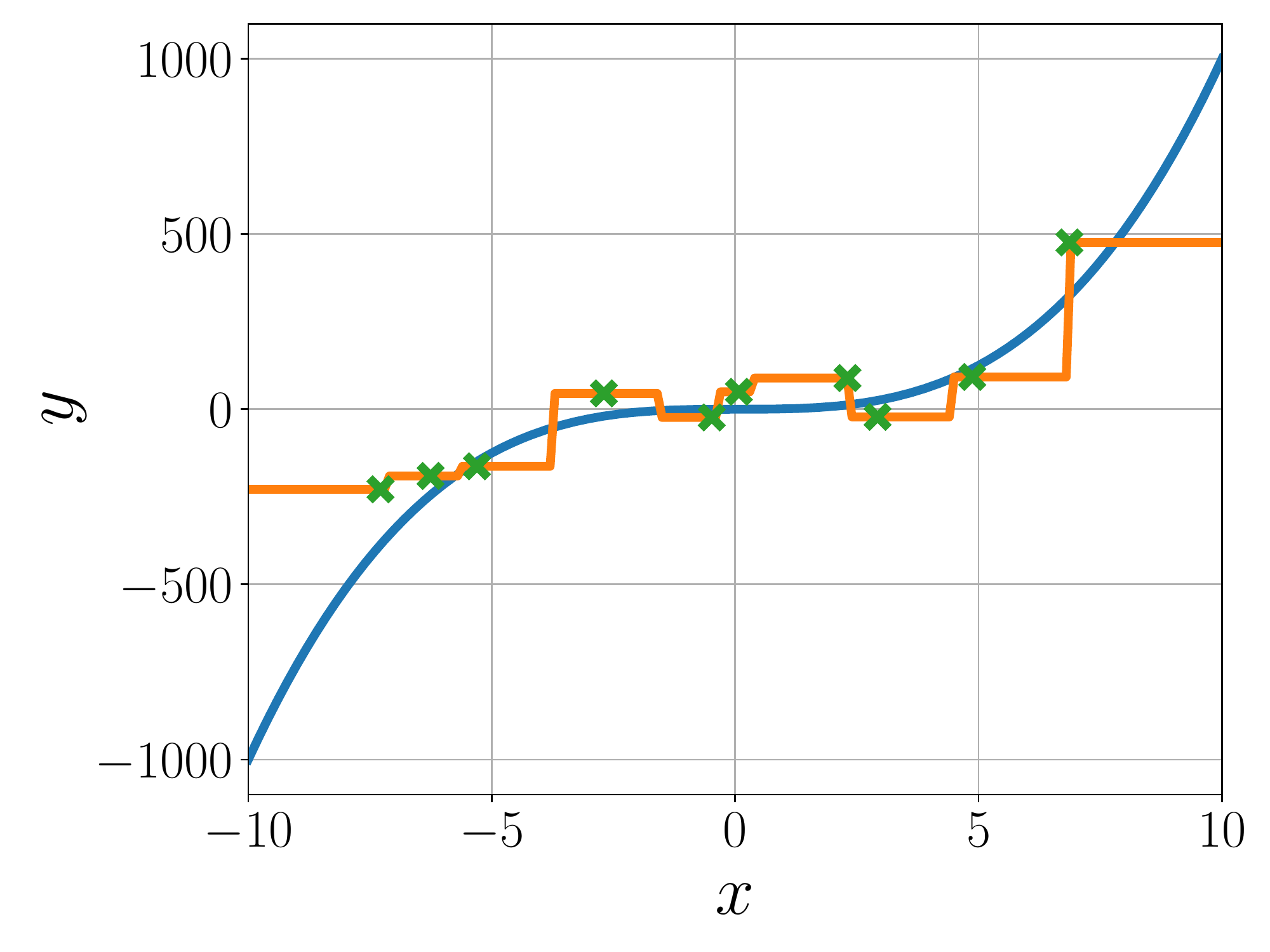}
		\includegraphics[width=0.235\textwidth]{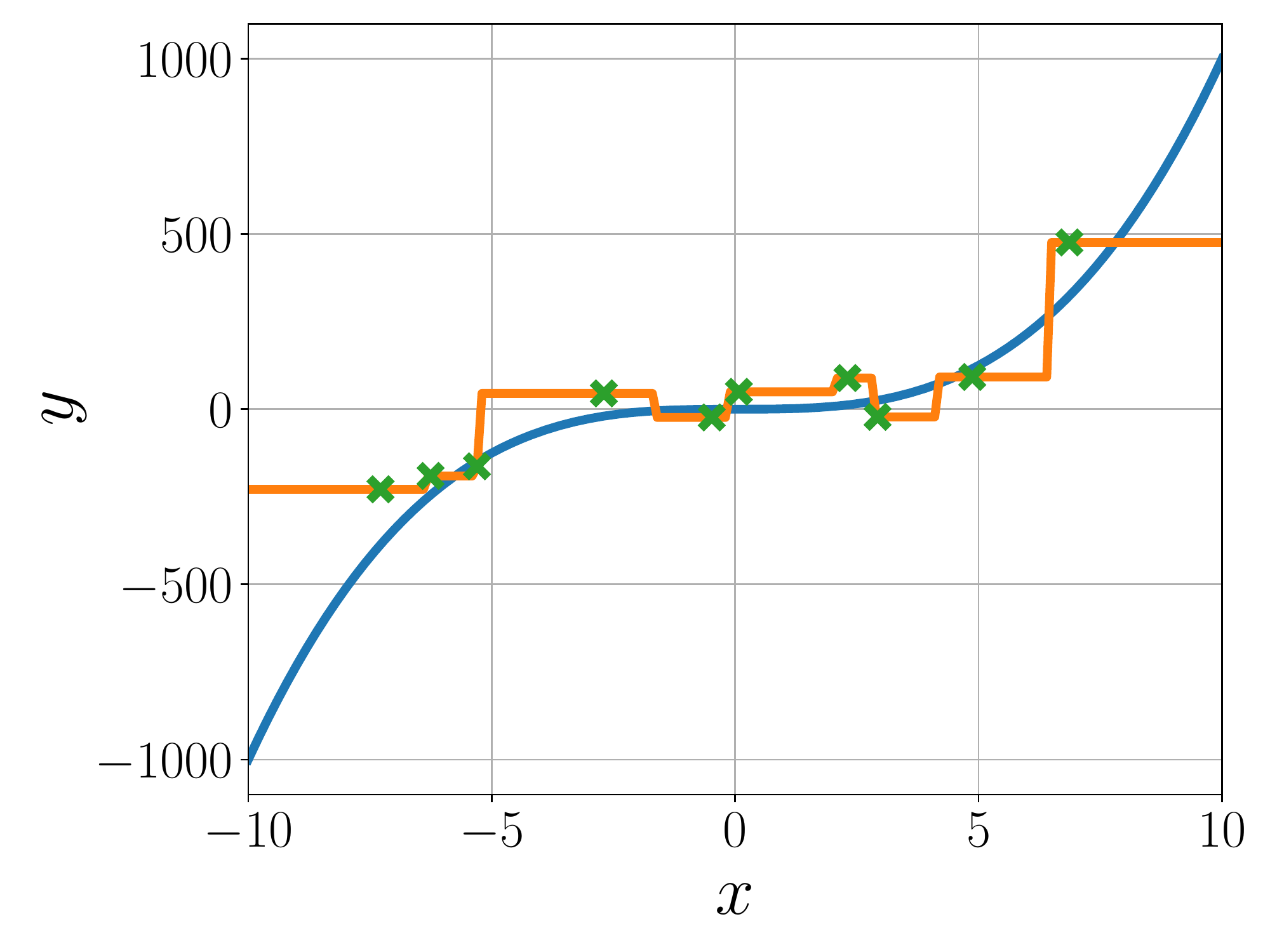}
		\label{fig:unc_1d_cubic_trees_r}
	}
	\subfigure[B + O]{
		\includegraphics[width=0.235\textwidth]{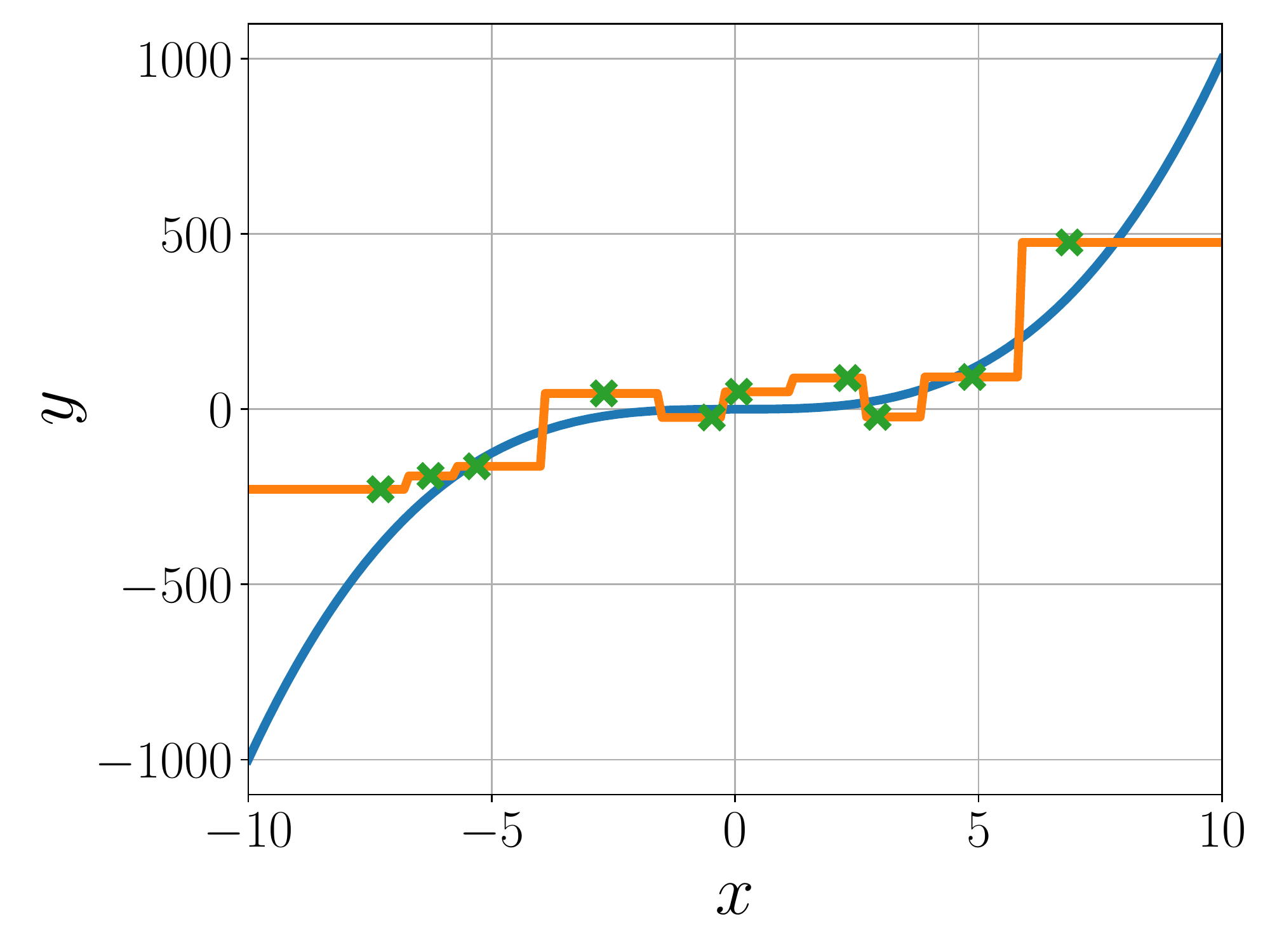}
		\includegraphics[width=0.235\textwidth]{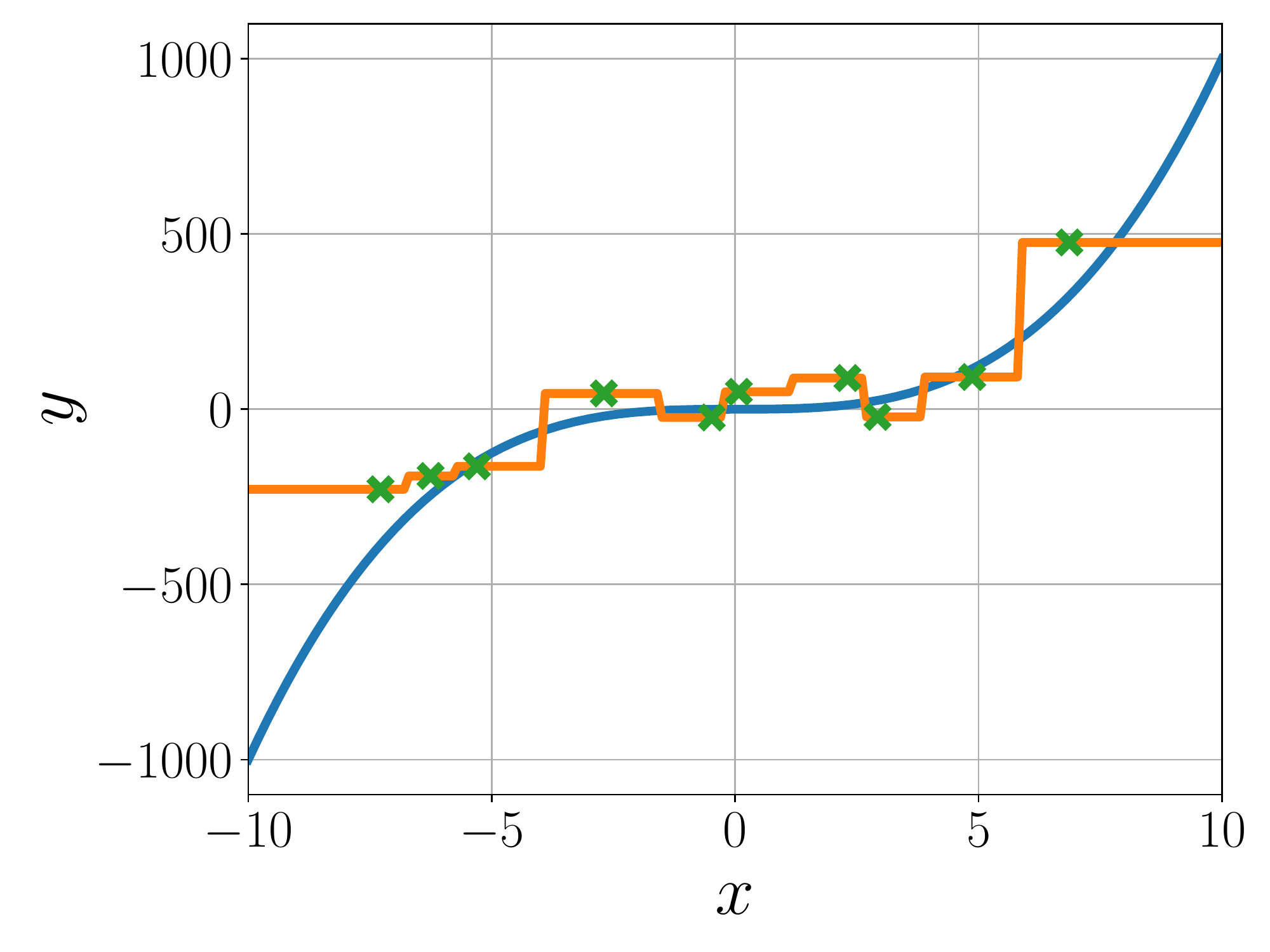}
		\includegraphics[width=0.235\textwidth]{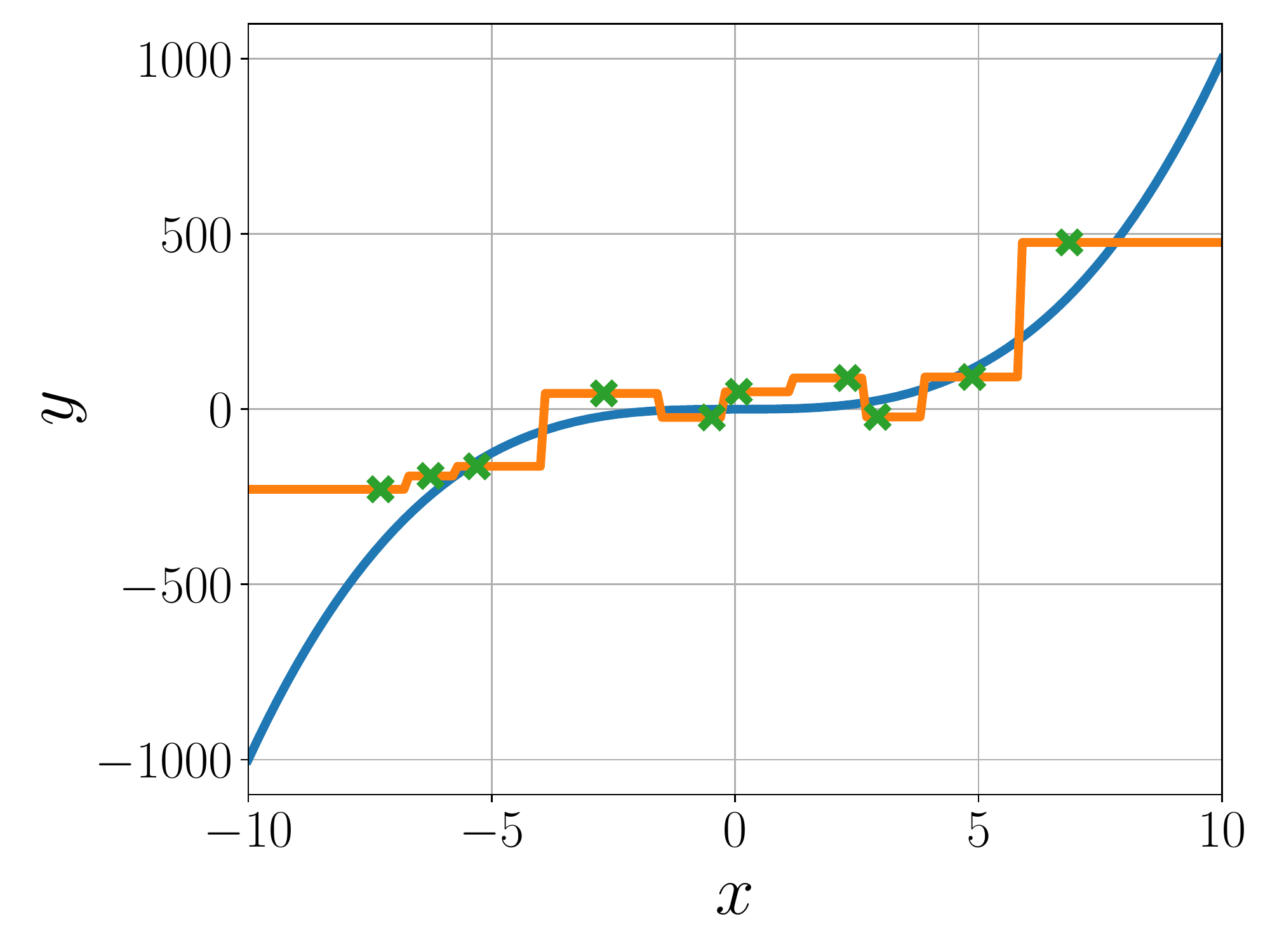}
		\includegraphics[width=0.235\textwidth]{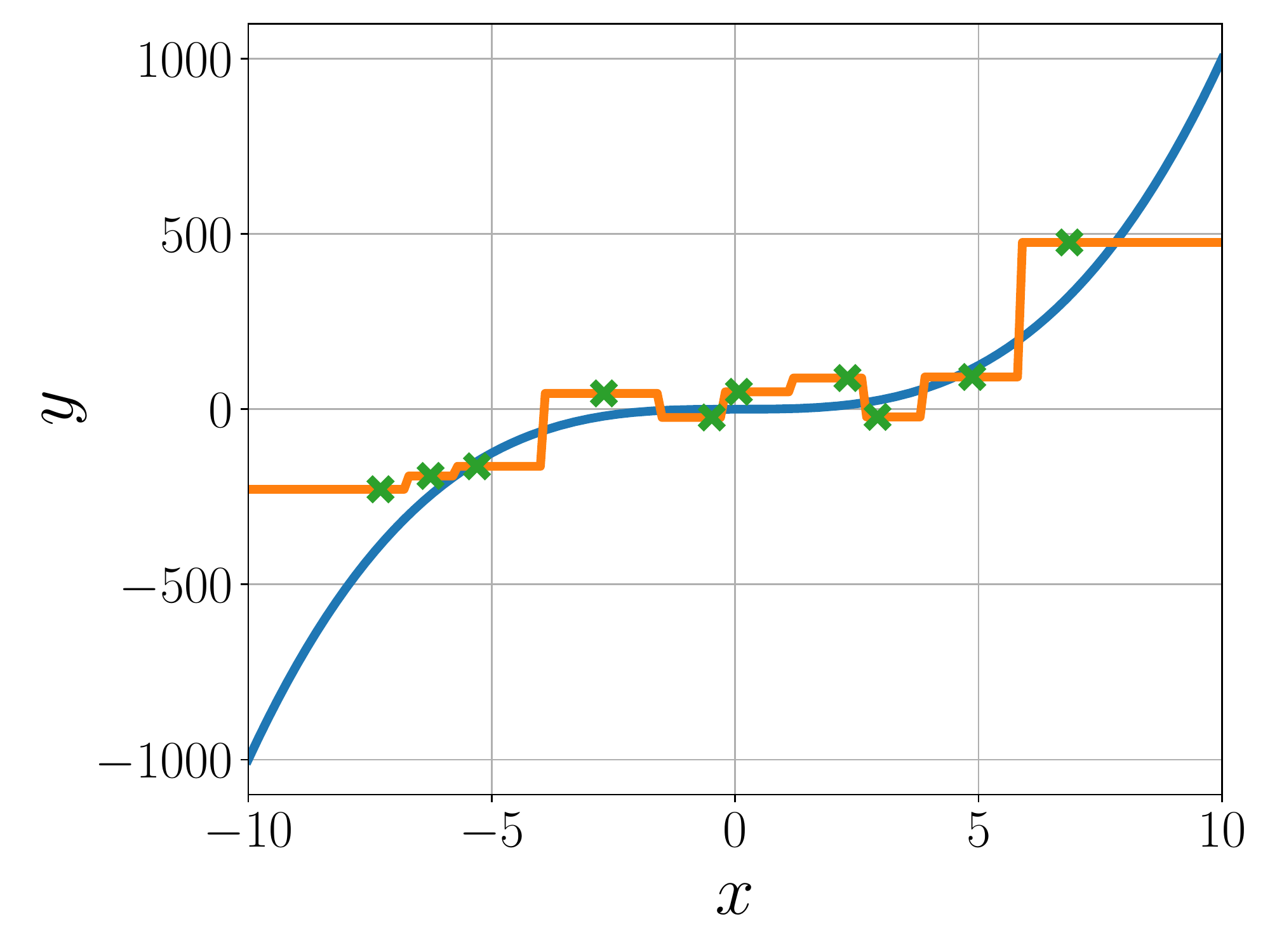}
		\label{fig:unc_1d_cubic_trees_bo}
	}
	\subfigure[R + B]{
		\includegraphics[width=0.235\textwidth]{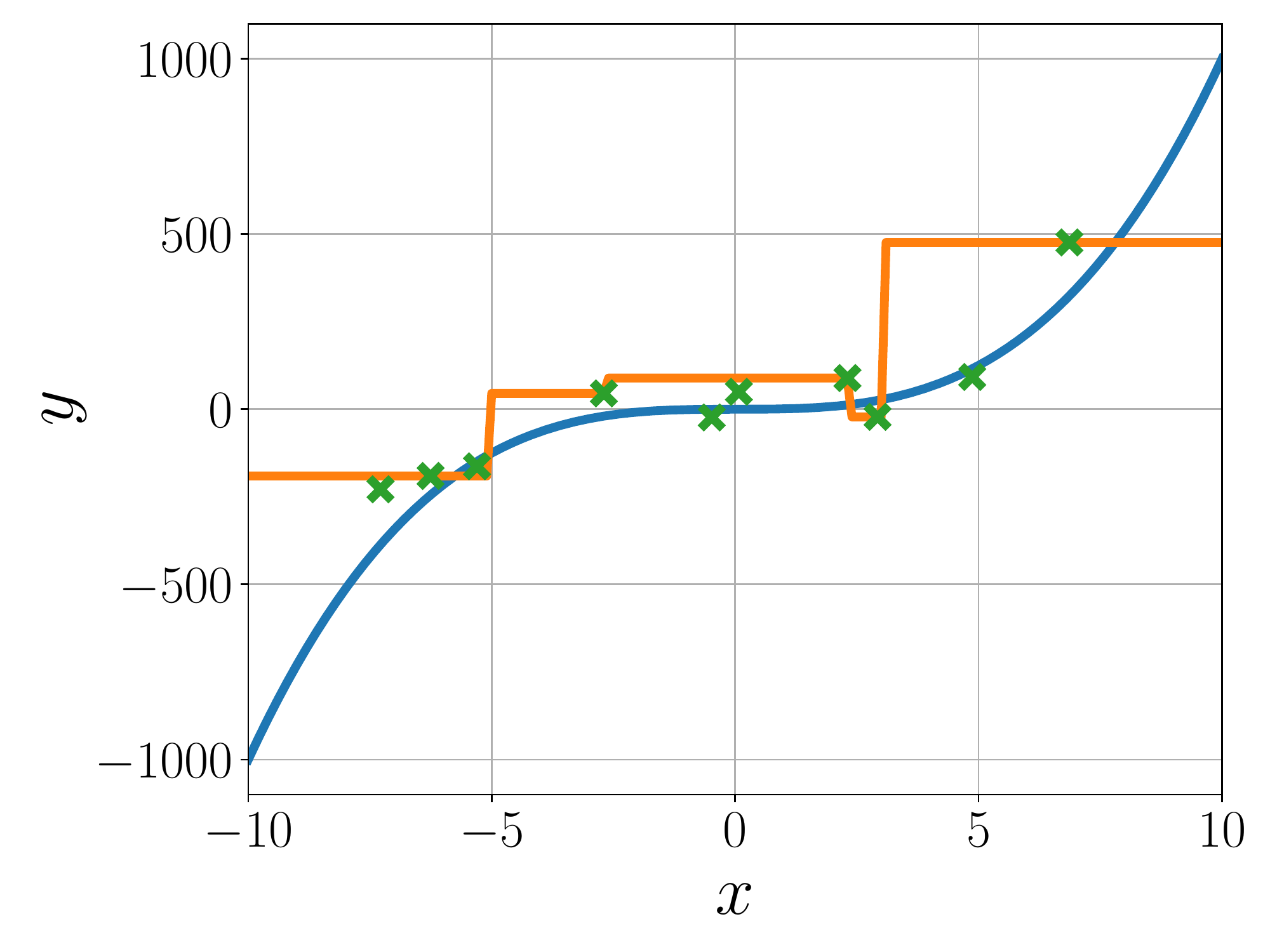}
		\includegraphics[width=0.235\textwidth]{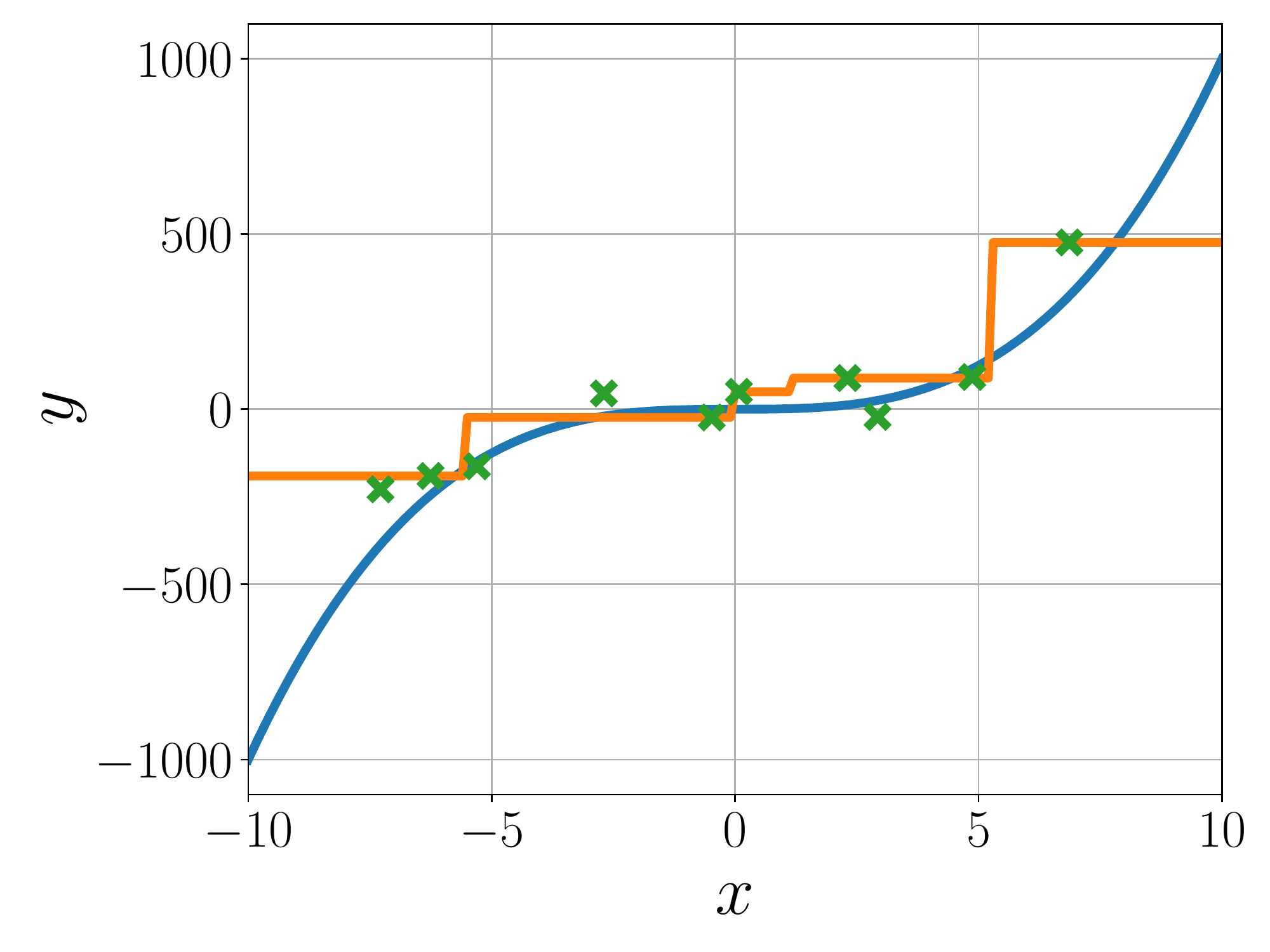}
		\includegraphics[width=0.235\textwidth]{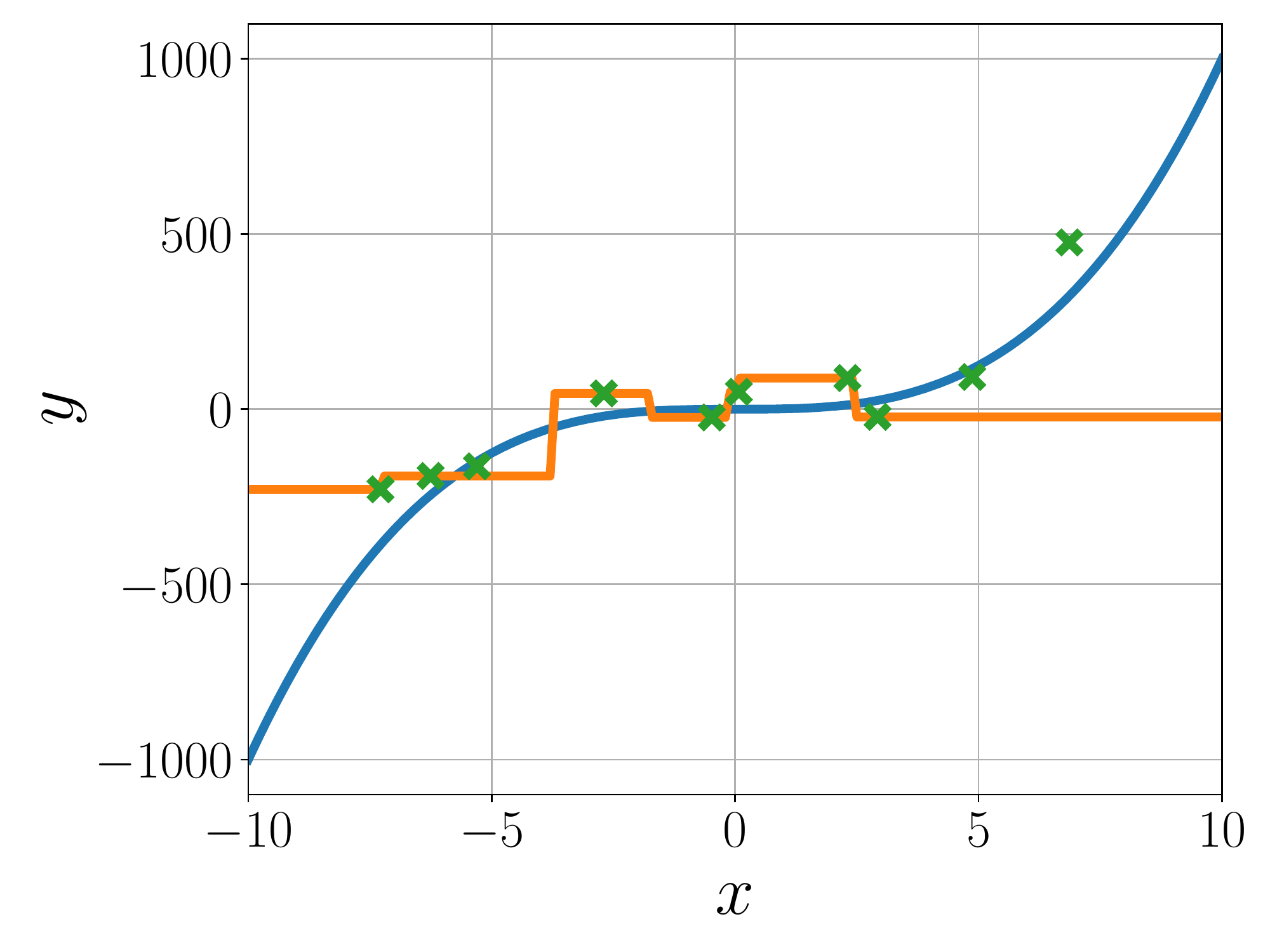}
		\includegraphics[width=0.235\textwidth]{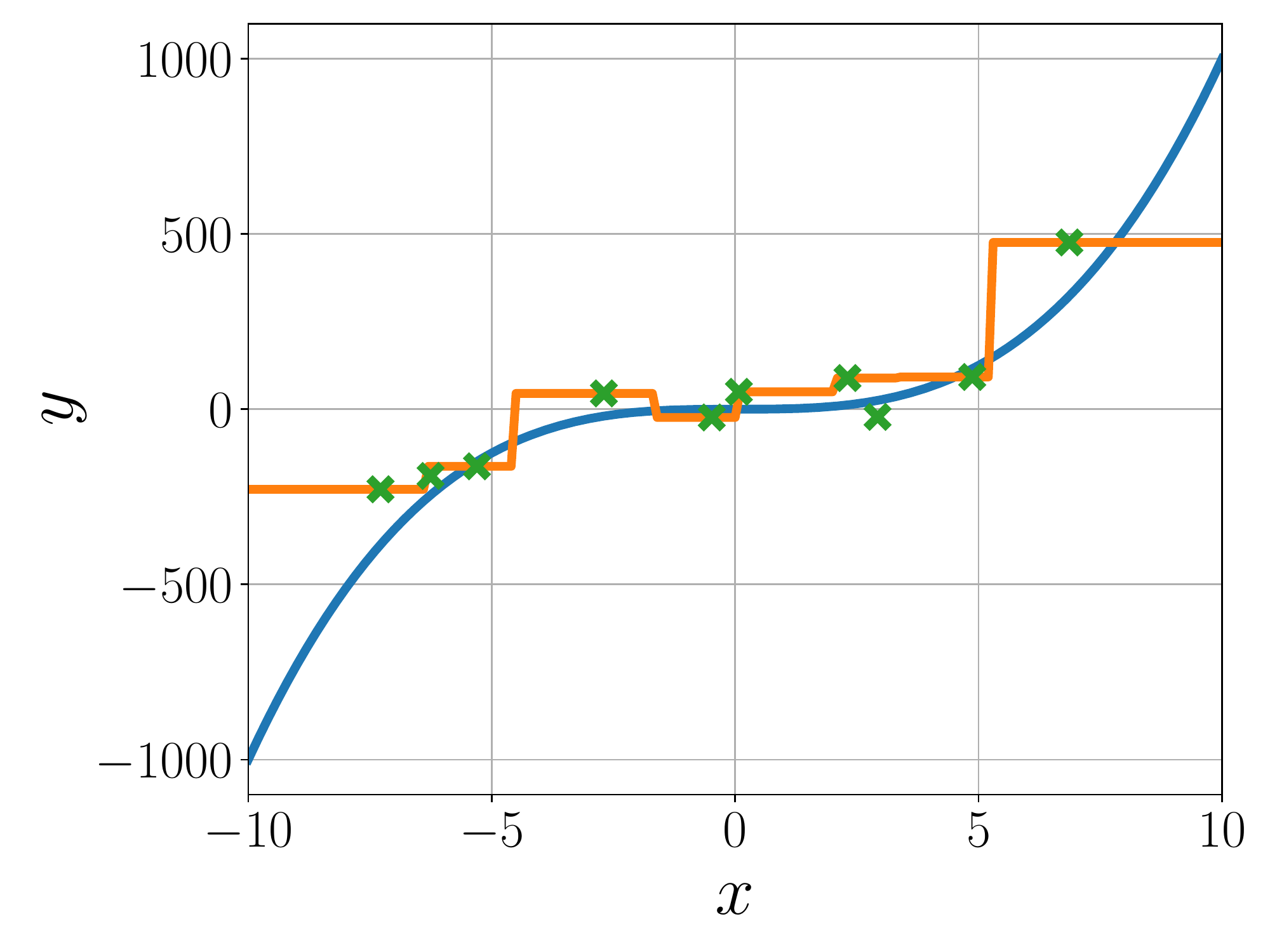}
		\label{fig:unc_1d_cubic_trees_rb}
	}
	\subfigure[R + B + O (i.e., BwO forest)]{
		\includegraphics[width=0.235\textwidth]{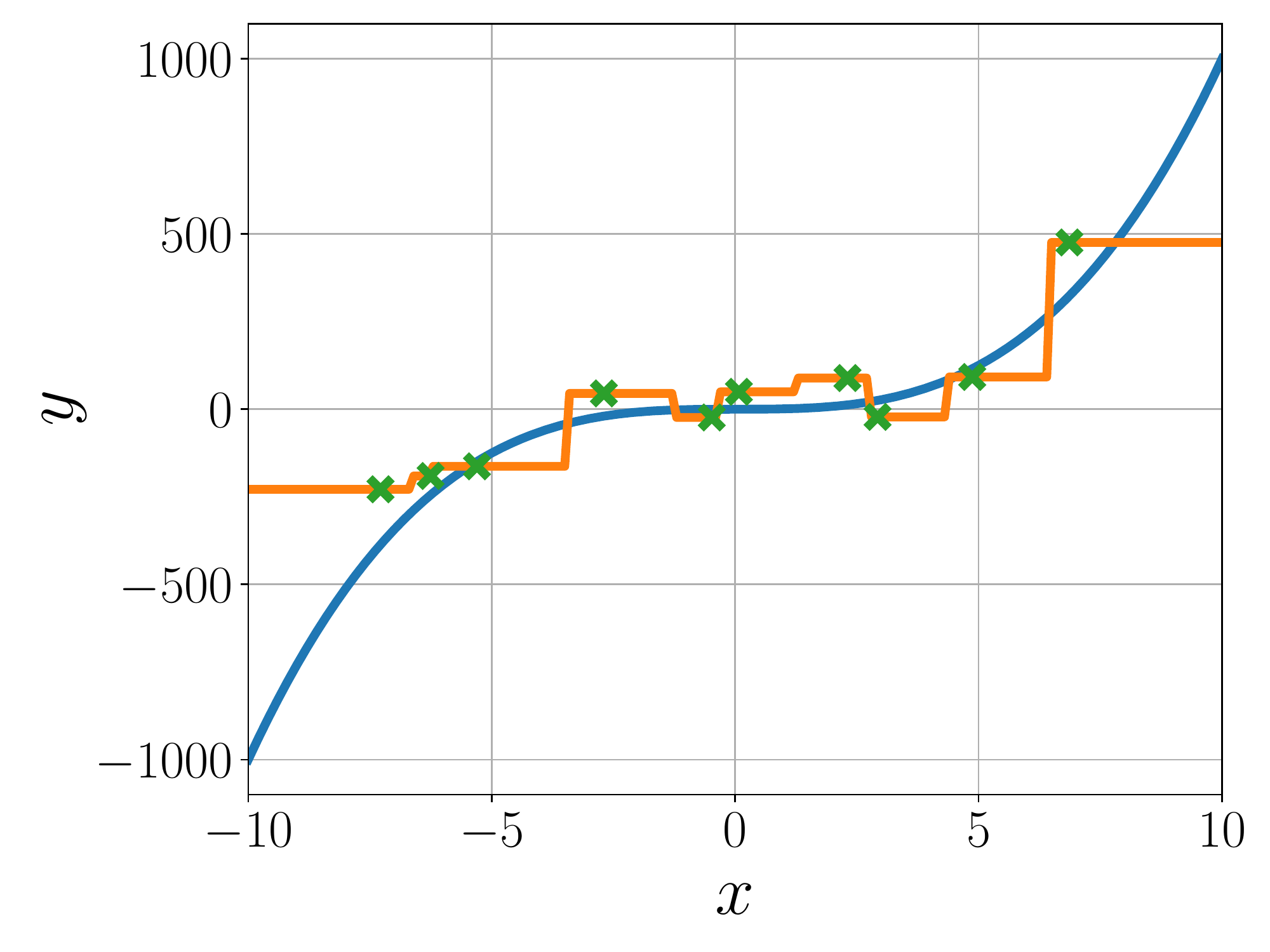}
		\includegraphics[width=0.235\textwidth]{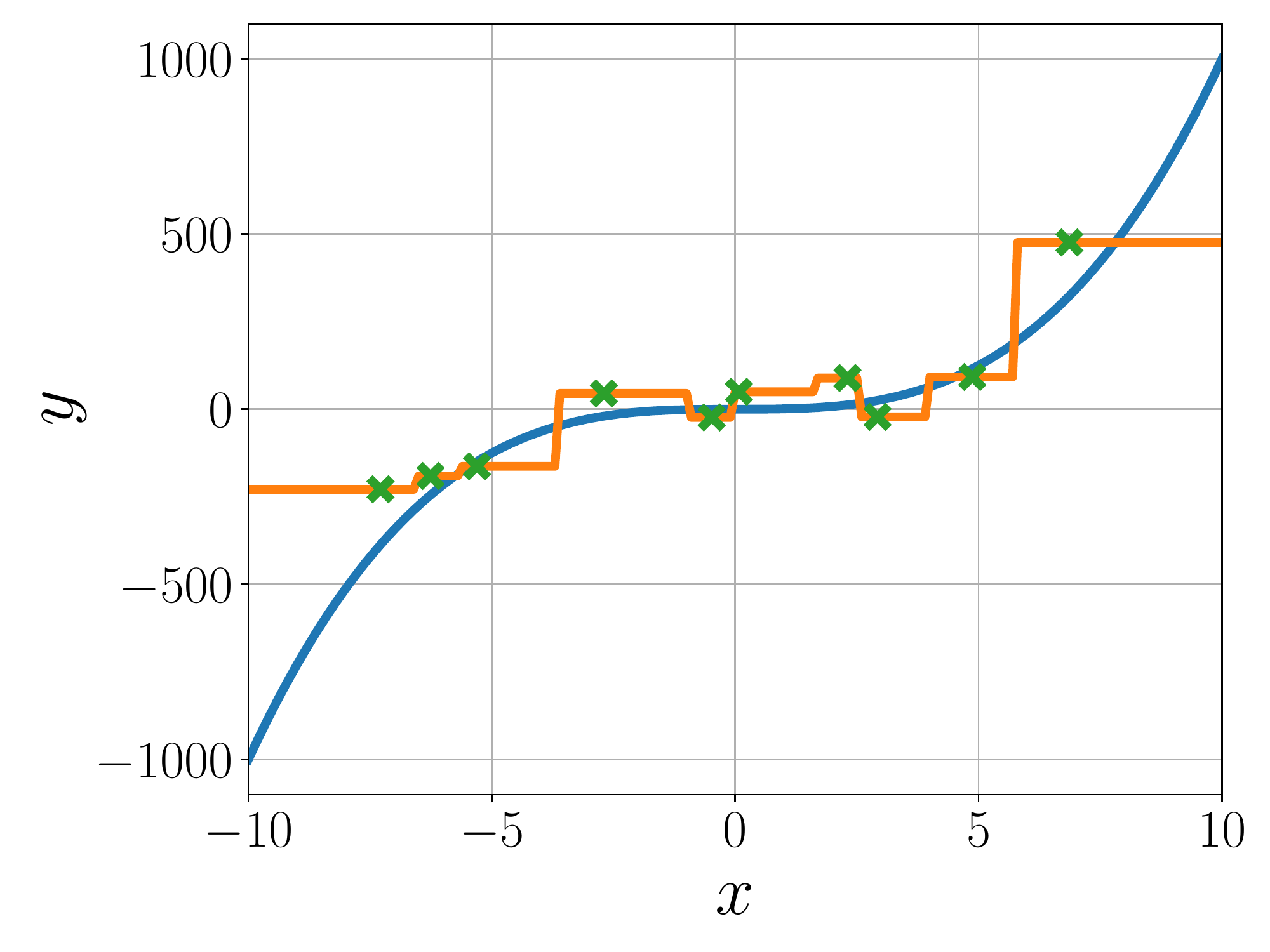}
		\includegraphics[width=0.235\textwidth]{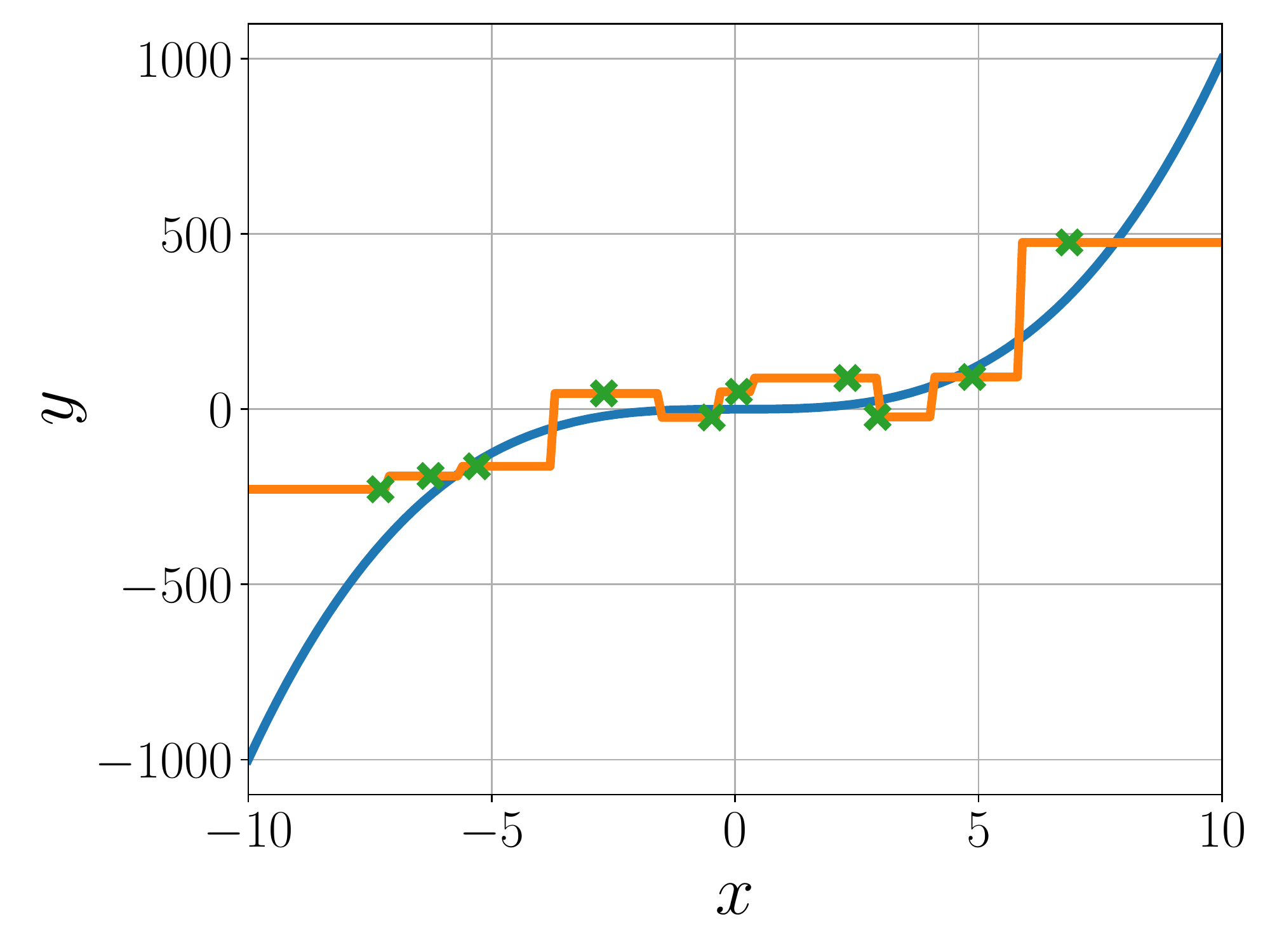}
		\includegraphics[width=0.235\textwidth]{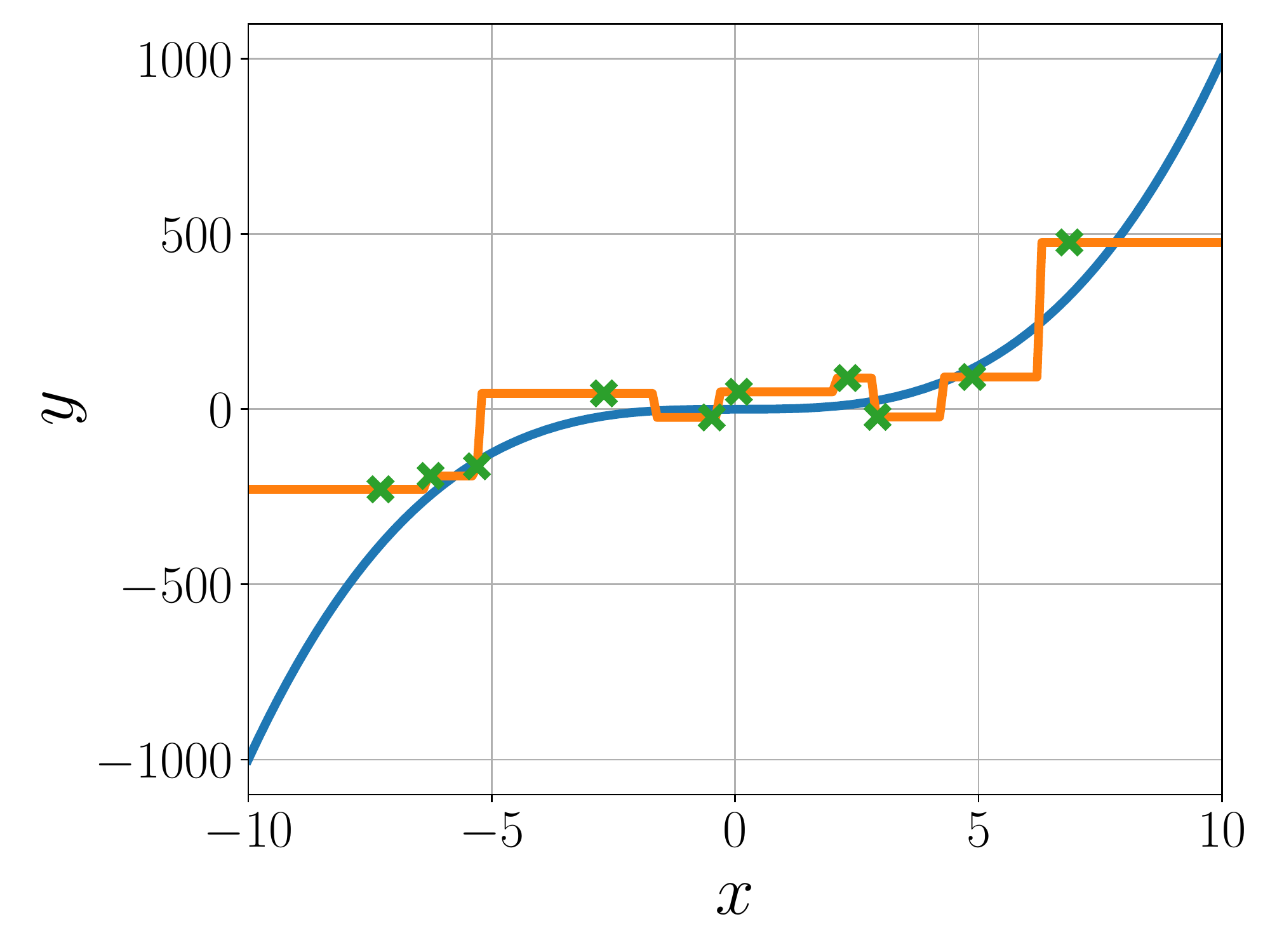}
		\label{fig:unc_1d_cubic_trees_rbo}
	}
	\caption{Results by individual trees for the case shown in \figref{fig:unc_1d_cubic}.\label{fig:unc_1d_cubic_trees}}
\end{figure}

We visualize the regression results by individual trees for the cases shown in \figref{fig:unc_1d_few},
\figref{fig:unc_1d_many}, and \figref{fig:unc_1d_cubic}.
These results presented in \figref{fig:unc_1d_few_trees}, \figref{fig:unc_1d_many_trees}, and \figref{fig:unc_1d_cubic_trees}
help us to understand the consequences by tree-based ensemble models.

\section{DETAILS OF EXPERIMENTS}

In this section, we describe the detailed setup of experiments.
As described in the main article, to implement our BwO forest,
we sufficiently duplicate an original dataset and then subsample part of duplicated datasets,
where $\alpha = 4$ and $\beta = 16$.
As part of our implementation, we utilize scikit-learn~\citep{PedregosaF2011jmlr} 
and QMCPy~\citep{ChoiSCT2021arxiv}.
For the implementation of BART, Mondrian forest, and NGBoost,
we use the following open-source projects:
\url{https://github.com/JakeColtman/bartpy},
\url{https://github.com/scikit-garden/scikit-garden},
and \url{https://github.com/stanfordmlgroup/ngboost}, respectively.

\paragraph{High-Dimensional Binary Search Spaces.}
We adopt the experimental setup suggested by~\citet{OhC2019neurips}.
Ising sparsification is to optimize the KL-divergence between two probability mass functions
with the regularization technique controlled by $\lambda$.
Contamination control in food supply chain is a problem that
optimizes food contamination with minimum prevention cost.
It is also regularized by a balancing hyperparameter $\lambda$.
See the work~\citep{OhC2019neurips} for the details of experiments.

\paragraph{Mixed Search Spaces.}
Our automated machine learning problem selects one of such classifiers:
(i) AdaBoost, (ii) GradientBoosting, (iii) Decision Tree, (iv) ExtraTrees, and (v) Random Forest.
This selection is represented by a one-hot encoding.
In addition to this categorical variable, we optimize the following hyperparameters for classifiers:
(i) the size of ensemble model, (ii) maximum depth, (iii) maximum features, and (iv) $\log$ learning rate.
It is summarized in~\tabref{tab:hyps}.
We use scikit-learn~\citep{PedregosaF2011jmlr} to implement all the algorithms.
Datasets used in this work: (i) Authorship~\citep{VanschorenJ2013sigkdden}, (ii) Breast Cancer~\citep{DuaD2019uci}, (iii) Digits~\citep{DuaD2019uci}, and (iv) Phoneme~\citep{VanschorenJ2013sigkdden},
are available at the respective references.

\end{document}